\documentclass[journal,draftcls,onecolumn,12pt]{IEEEtran}
\usepackage{cite}
\usepackage{epsfig}
\usepackage{graphicx}
\usepackage{color}
\usepackage{multirow}
\usepackage{verbatim}

\usepackage{fancyhdr}

\date{ }
\begin{document}

\pagestyle{fancy}
\lhead{T-PAMI vol. 30 no. 12, pp. 2188-2203, 2008  }
\lfoot{}
\title{Disconnected Skeleton: Shape at its Absolute Scale\thanks{Corresponding authors: Sibel Tari (stari@metu.edu.tr) and Cagri Aslan (caslan@microsoft.com)}}

\author{C.~Aslan,~A.~Erdem,~E.~Erdem~and~S.~Tari}
\maketitle

\thispagestyle{empty}

\begin{abstract}
We present a new skeletal representation along with a matching
framework to address the deformable shape recognition problem. The
disconnectedness arises as a result of excessive regularization that
we use to describe a shape at an attainably coarse scale. Our
motivation is to rely on the stable properties of the shape instead
of inaccurately measured secondary details. The new representation
does not suffer from the common instability problems of traditional
connected skeletons, and the matching process gives quite successful
results on a diverse database of 2D shapes. An important difference
of our approach from the conventional use of the skeleton is that we
replace the local coordinate frame with a global Euclidean frame
supported by additional mechanisms to handle articulations and local
boundary deformations. As a result, we can produce descriptions that
are sensitive to any combination of changes in scale, position,
orientation and articulation, as well as invariant ones.
\end{abstract}

\section*{NOTES}

The work excluding \S\ref{sec:slcf} and  \S\ref{sec:morph} has first appeared in 2005~\cite{Aslan05,Aslantez}:
\begin{itemize}
\item{Aslan, C., Tari, S.: An Axis-Based Representation for Recognition. In ICCV(2005) 1339-1346.}
\item{Aslan, C., : Disconnected Skeletons for Shape Recognition. Master’s thesis, Department of Computer Engineering, Middle East Technical University, May 2005.}
\end{itemize}

\newpage

\section{Introduction}
\label{intro} Local symmetry axis based representations, commonly
referred to as shape skeletons, have been used in generic shape
representation since the pioneering work of Blum \cite{Blum73} on
the study of form via {\sl axis morphology}. The strength of axis
based representations lies in expressing the links among shape
primitives and providing a shape centered coordinate frame.

\begin{figure}[!h]
\centering
\begin{tabular}{cccc}
\includegraphics[width=3cm]{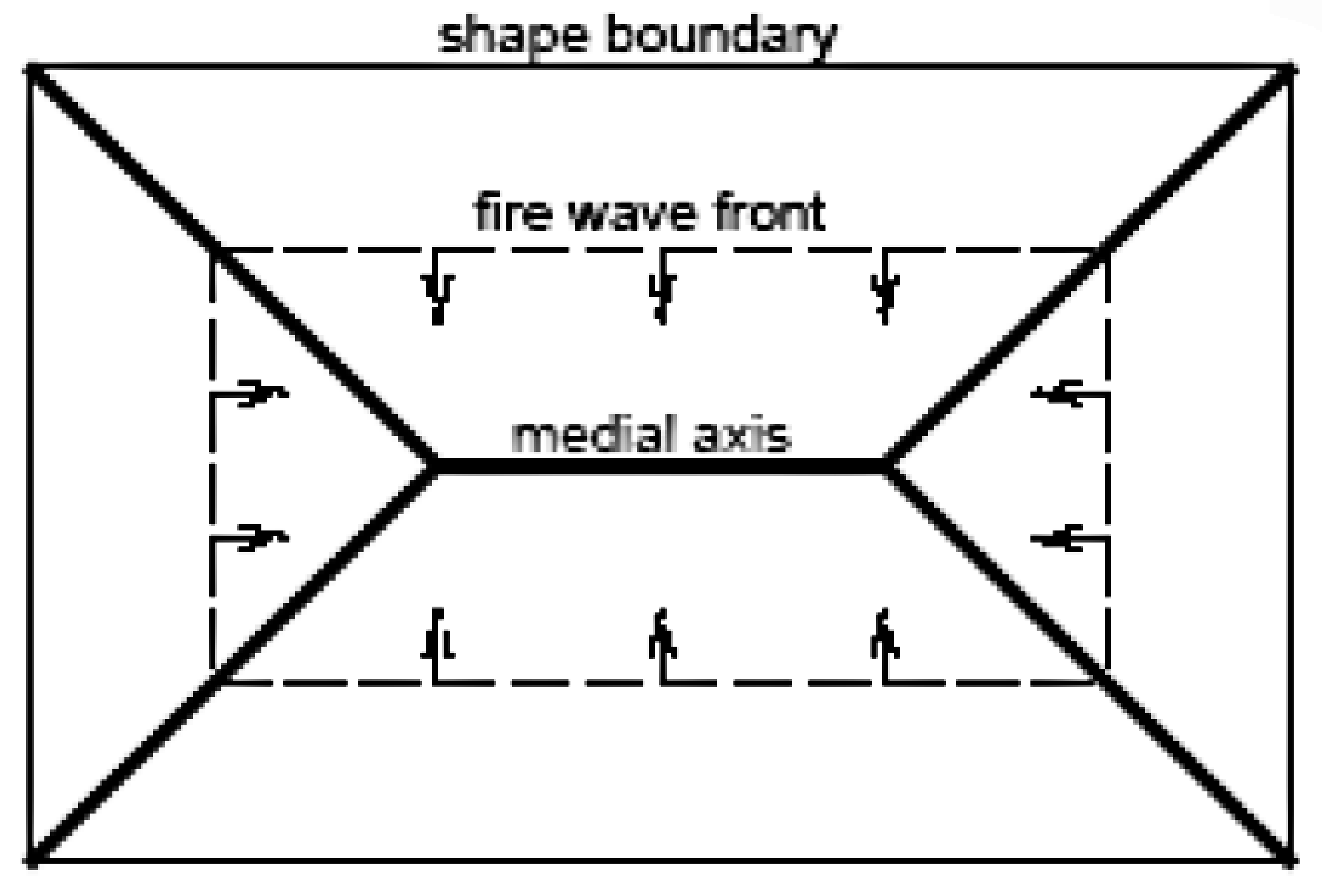} &
\includegraphics[width=4.5cm]{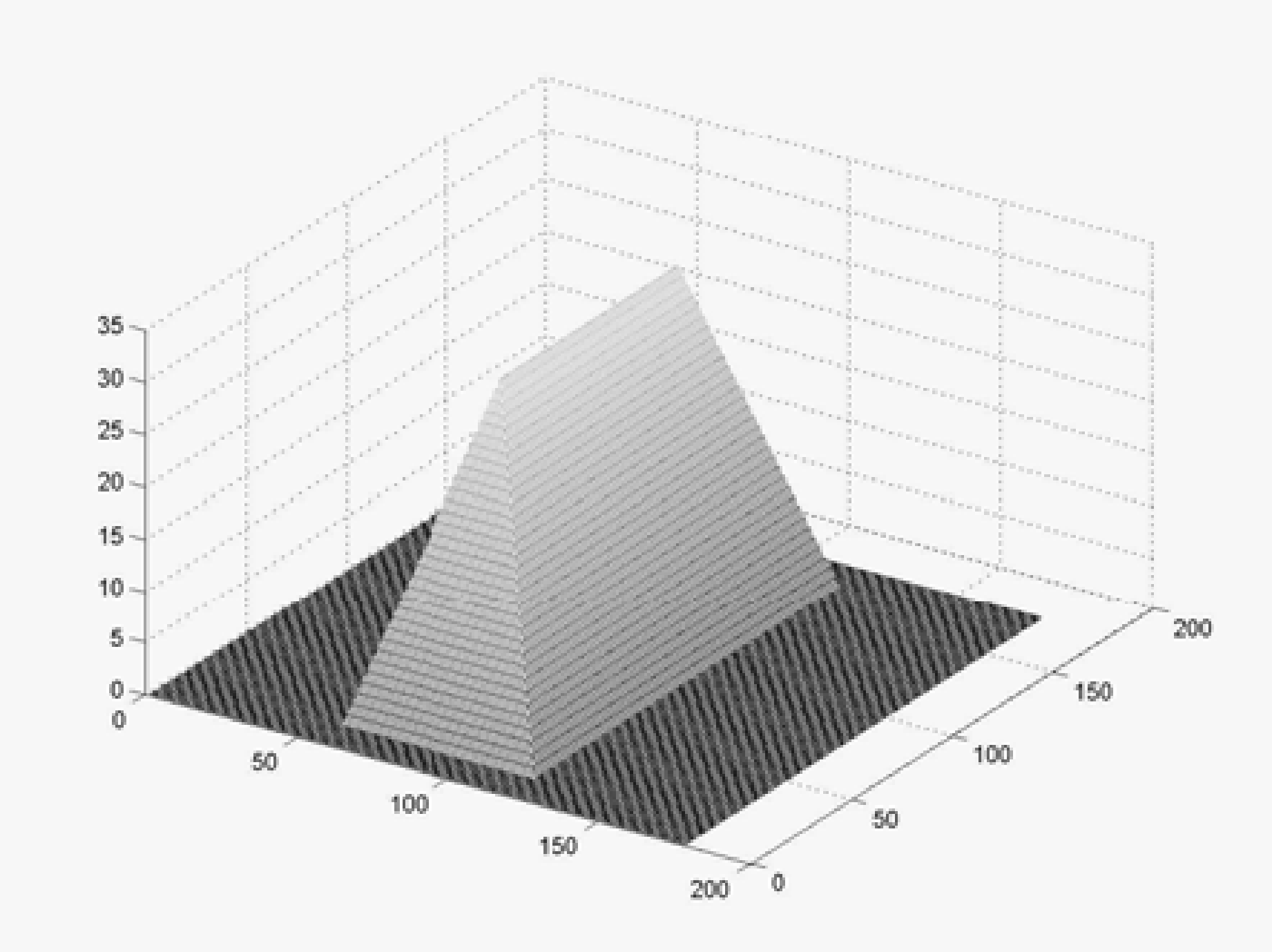} &
\includegraphics[width=3cm]{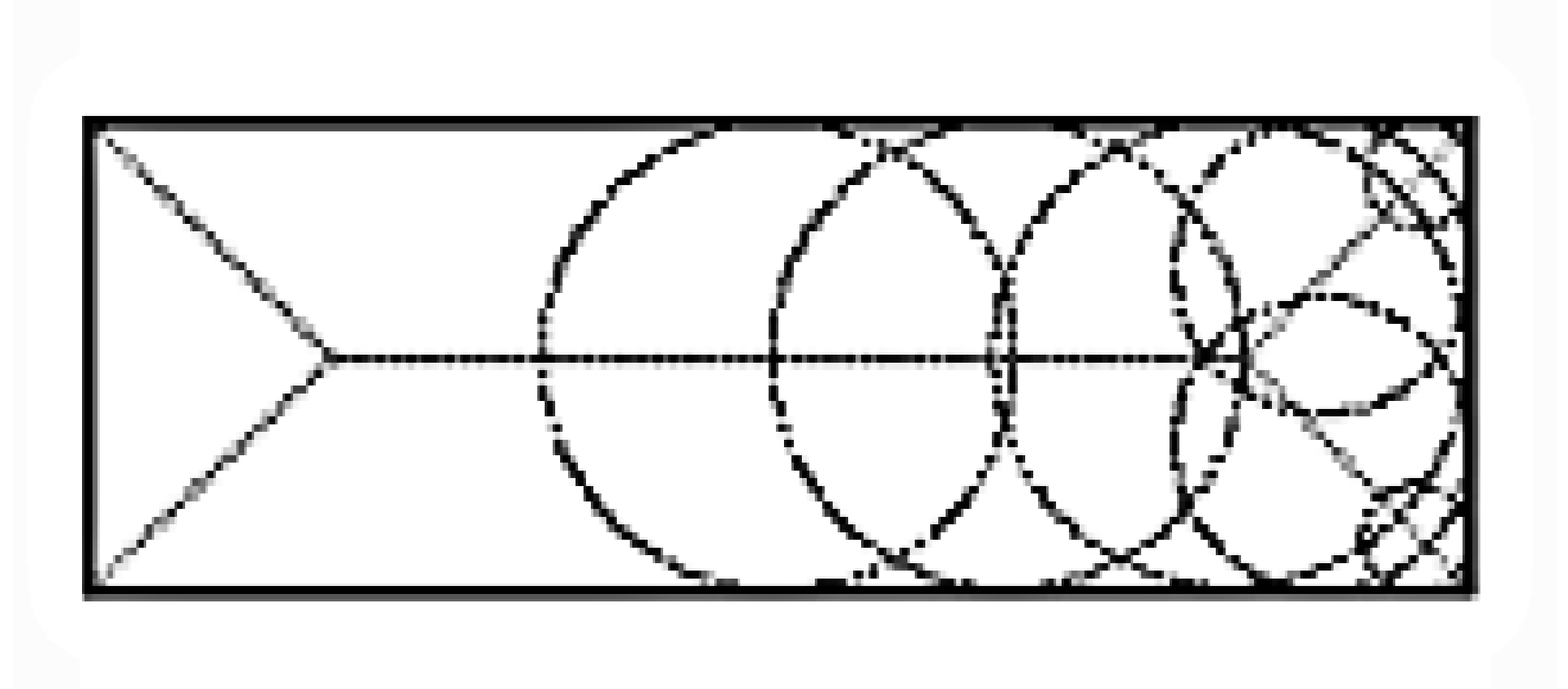} &
\includegraphics[width=4cm]{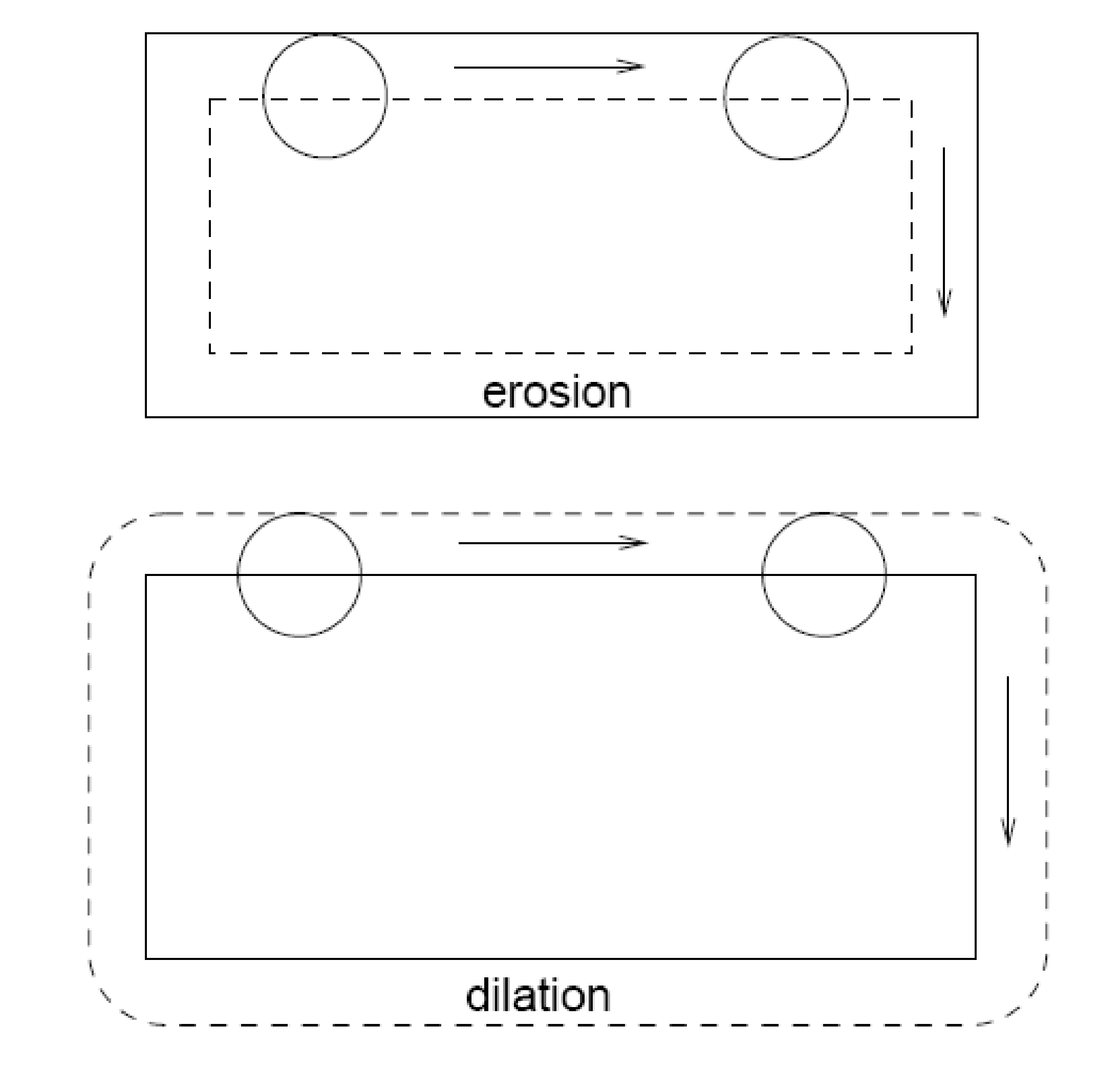} \\
(a) & (b) & (c) & (d) \\
\end{tabular}
\caption{Construction of the skeleton for a rectangle. (a) grass-fire method.
(b) distance transform. (c) maximal circles. (d) erosions and dilations with a
disc as the structuring element. Dashed lines are the transformed versions of
solid lines.} \label{fig:4methods}
\end{figure}

Blum's skeleton can be explained using three alternative
constructions. The original one is to initiate fire fronts at time
$t = 0$ along all the points of the shape boundary, and to let these
fronts propagate as wavefronts toward the center of the shape at
uniform speed (Fig.~\ref{fig:4methods}(a)). The locus of {\sl shock}
points, where these wavefronts would intersect and extinguish,
defines the skeleton. An important property of the skeleton is its
ability to reconstruct the shape boundary by propagating the
wavefronts backwards. The second construction replaces the dynamic
view of propagating fronts with a static view by embedding time
dependent wavefronts as the level curves of a surface whose value at
each point is the minimum distance of the point to the boundary
(Fig.~\ref{fig:4methods}(b)). Skeleton points are the ones which are
equidistant from at least two boundary points. They coincide with
the shock points. They are the sharp points of the level curves of
the distance function. It is possible to detect these points by
projecting the ridges of the distance function on the shape plane.
The third construction is via {\sl maximal circles} that are
inscribable inside the shape and touch the shape boundary at more
than one point. In this construction, circle radius plays the role
of time of arrival in the symmetry axis function. Considering the
envelope of maximal circles, one can reconstruct the shape boundary
(Fig.~\ref{fig:4methods}(c)).

Independent of Blum's morphological analysis, {\sl mathematical
morphology} has been developed as a set theoretic approach to
shape~\cite{Matheron75,Serra82}. Following Lantuejoul's
work~\cite{Lantuejoul78} which showed that the skeleton can be
constructed by set transformations, a significant amount of
skeletonization work developed within the mathematical morphology
community. Basic set transformations that are applied to shapes with
the help of {\sl structuring elements} are called {\sl erosions} and
{\sl dilations} (Fig.~\ref{fig:4methods}(d)). Note that when the
structuring element is chosen as a disk of diameter $d$, the eroded
shape boundary becomes equivalent to the wave front at $t=d$.
Eroding the shape first and dilating it later with the disk
structuring element removes the sharp corners which correspond to
the shock points of the wave model.
%Hence, taking a set difference
%between shape and its eroded and dilated form, gives the skeletal points.

While all of these constructions lead to the same representation,
they inspired many others having different properties and being
computed with different algorithms,
e.g.~\cite{Brady84,Bruce85,Geiger03,Leyton88,MaragosSchafer86,GoutsiasSchonfeld91,LeviMontanari70,Ogniewicz95,
Pizer03,Siddiqi02,Tari97,Zhu99,Zhu96,Kresch95,KreschMalah94}. There
is a connection between positive curvature maxima of the shape
boundary and the skeleton in the sense that each curvature maximum
gives rise to a skeletal branch~\cite{Leyton88}. This connection is
often used to extract the skeleton. When the skeleton is extended to
include branches that arise from curvature minima, a richer set
commonly referred to as \textsl{local symmetry axes} is
obtained~\cite{Brady84}. Note that the term symmetry as used in this
context is different than conventional symmetry and is a property of
points generated by the shape rather than being a property of shapes
themselves~\cite{Blum73,Tari98}.

A major reason for having alternative skeleton constructions is that
the skeleton is an unstable representation in the sense that a small
change in the shape may cause a significant change in its
description. Difficulties associated with a robust implementation
are the major source of a significant body of research dedicated to
regularization and richer definitions of local symmetry.

Even though there are interesting ideas in recent
works~\cite{Zhu99,Geiger03,Siddiqi02}, pruning of the axes has been
mostly used to simplify or regularize the skeleton. In general,
pruning methods define a saliency measure for the axis points and
discard those points whose significance is below a threshold. Axis
length, propagation velocity, maximal thickness, and the ratio of
the axis to the boundary it unfolds are the most typical
significance measures; however, they do not necessarily reflect the
perceptual prominence of parts~\cite{Shaked98}.

Interestingly, -- as also observed in
\cite{August99,Eede06,Shah05,Zhu99} -- while the accurate
computation of the skeleton branches corresponding to noise and
secondary details is difficult, the ones corresponding to
ribbon-like parts of the shape with slowly varying width can be
accurately determined with any method. This observation motivates us
to propose a new skeletal representation, by computing the local
symmetries only at the locations where they can be accurately
determined. Unlike conventional skeletons, our skeleton is {\sl
disconnected}. The disconnectedness is not a result of pruning
according to some measure as in Shah~\cite{Shah05}, but it is a
result of excessive regularization added to local symmetry
extraction. In that respect, our work is closely related to
multi-scale skeleton ideas despite the fact that we have a single
shape dependent scale which is as coarse as possible, yet  permits a
morphological analysis. We refer to this scale as the {\sl absolute
scale}.

The rest of the introduction is divided into two parts. In the first
part, multi-scale skeleton ideas from the  literature are reviewed.
In the second part, an overview of our approach and its connections
to the previous multi-scale ideas is given.

\subsection{Multi-scale Skeleton}
The multi-scale skeleton and the related {\sl pattern
spectrum}~\cite{MaragosSchafer86,Maragos89} form one of the most
interesting ideas proposed in shape analysis. In a rather simplified
view, it is a nested representation in which the skeleton is
expressed as the union of skeleton subsets detected using circles of
varying width. Excluding skeletal subsets detected with small
circles (e.g. by zeroing the corresponding component in the pattern
spectrum), provides a coarser description. In this sense, the Blum
skeleton may be interpreted as the multi-scale morphological
skeleton  at the smallest possible scale. The multi-scale skeleton
idea found interesting alternative algorithms in  recent
works~\cite{Pizer03}.

An alternative definition of multi-scale skeleton is the computation
of skeleton at various regularization levels via initial smoothing
of the shape
boundary~\cite{PizerOliverBloomberg87,DillLevineNoble87,Latecki07,Ogniewicz95}.
In this group of works, multiple skeletons are extracted forming a
{\sl skeleton scale space}. As the smoothing increases, the skeleton
becomes less detailed. Note that a coarse skeleton representation in
a skeleton scale space, e.g.~\cite{Latecki07}, is a morphological
skeleton of a smoothed version of the shape boundary.

In contrast to this view of smoothing shape boundary, one can
consider continuous smoothing of the propagating fronts (or the
level sets of the distance
function)~\cite{Kimia95,Siddiqi96,Tari97,Tari98,Gorelick06}. Kimia,
Tannenbaum and Zucker~\cite{Kimia95} considered a front propagating
with a speed which is a linear combination of a constant term
corresponding to morphology and a curvature dependent term serving
as a smoother. Even though it is one of the interesting ideas, the
shape scale space obtained via this model has not been used in shape
recognition. In the shock-loci methods,  Kimia \textsl{et al.}
switched to the use of constant speed corresponding to
morphology~\cite{Pizer03}. One reason is that the speed function
taking negative values poses computational difficulties. Another
difficulty is that due to smoothing, shock points turn into high
curvature points which are harder to detect accurately.
Computational difficulties could be alleviated by considering a
speed which is positive everywhere as demonstrated by Tari, Shah and
Pien (TSP)~\cite{Tari96,Tari97} and
 Gorelick~\textsl{et al.}~\cite{Gorelick06}. In these alternative frameworks, one obtains
an analogue of the smoothed distance surface via solving elliptic
PDEs. The solution can be related to the wave model by considering a
propagation speed which is an increasing function of medialness. In
the  TSP method, time of arrival is approximately a decaying
exponential function of distance from the shape boundary. Skeletal
points moving with a faster speed than the non-skeletal points can
be detected by the minima of the gradient which are indirectly
related to curvature maxima.

\subsection{Our Approach}
Our approach to compute the symmetry set (described in
Section~\ref{sec:dosa}) is directly related to that of
TSP~\cite{Tari96,Tari97}. Our major departure is to let the
regularization tend to infinity and dominate over morphology. To
achieve this, we propose a special surface $\phi$ taking values in
$(0,1]$ and whose $1$-~level curve corresponds to the shape boundary
and other level curves roughly mimic the evolution of the boundary
towards a circle. In a naive sense, the surface $\phi$ is the limit
of the surface in TSP when the smoothing tends to infinity. The
level curves of $\phi$ can be interpreted as multi-scale erosions
with non-homogeneous structuring elements: ellipses of varying
aspect ratio and size. The local symmetry points corresponding to
the curvature extrema of the evolving boundary (note that these are
smoothed shock points) end as soon as the evolving curve {\sl
locally} becomes a circle. For this reason, unlike the conventional
skeleton, ours is not connected.

The disconnection points have an interesting interpretation in terms
of the connected morphological skeleton. In the traditional
construction, the angle between two symmetry-shape geodesics
(shortest distance from axis point to boundary) gives the {\sl
object angle}~\cite{Blum73}. On one hand, at regions of slowly
varying width, this angle is large. On the other hand, it is small
near branch points where the skeleton computation is sensitive to
noise and secondary details. As such, the disconnected axes are
analogous to pruned skeletons using object angle or its
variants~\cite{Shah05}.

We remark that the connection of our approach to multi-scale
skeleton ideas is rather counter-intuitive. In comparison to the
definition by Maragos~\cite{MaragosSchafer86,Maragos89}, our
skeleton representation is roughly {\sl high-passed} where the
skeletal subsets detected with circles of larger radius are filtered
out. In contrast to skeleton scale space methods which smooth the
boundary first  and then compute the skeleton, we, in a dynamic way,
continuously smooth the shape boundary and record its special points
till the whole boundary becomes simple enough to admit any. Unlike
in skeleton scale space representation, we do not form a hierarchy
of skeletons. We have just a single skeleton computed at the
coarsest possible scale in which a skeletal description is still
possible.

The disconnected axes corresponding to ribbon-like sections (or
equivalently isolated simple symmetry branches) form the primitives
of our representation. Each primitive is described by its
disconnection location and its length. As such, the proposed
representation is an unlabeled attributed point set and forms a
trade-off between unstructured point sets,
e.g.~\cite{Belongie02,Burl96,Chui03} and skeletal graphs,
e.g.~\cite{Geiger03,Sebastian04,Siddiqi99,Zhu96}.

\begin{figure}[!h]
\centering
\begin{tabular}{cccc}
\includegraphics[height=2.5cm]{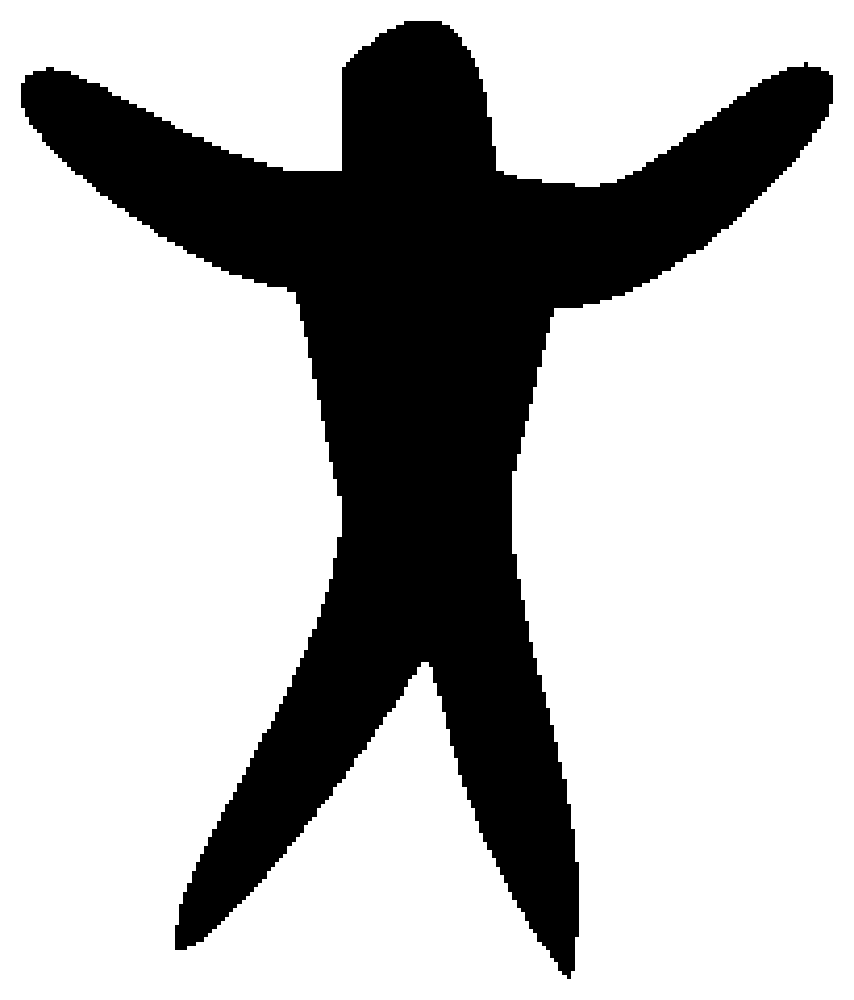} &
\includegraphics[height=2.5cm]{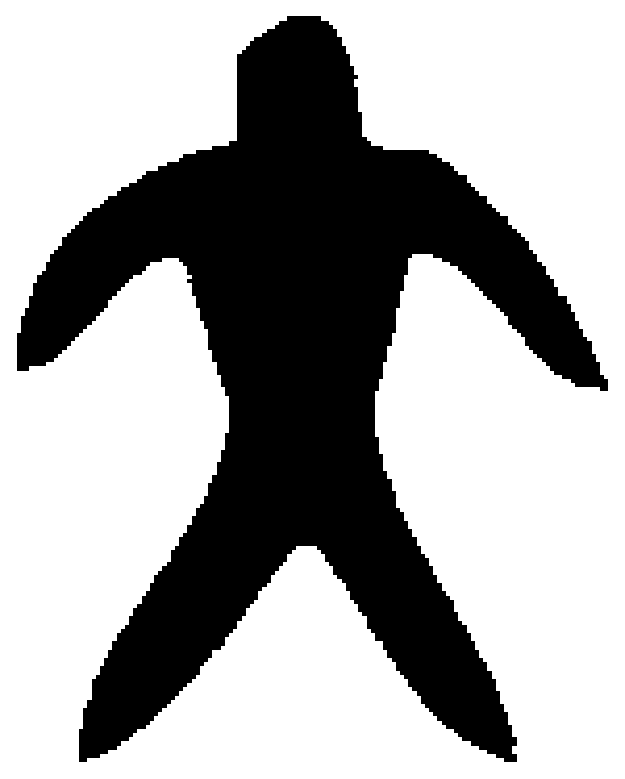} &
\includegraphics[height=2.7cm]{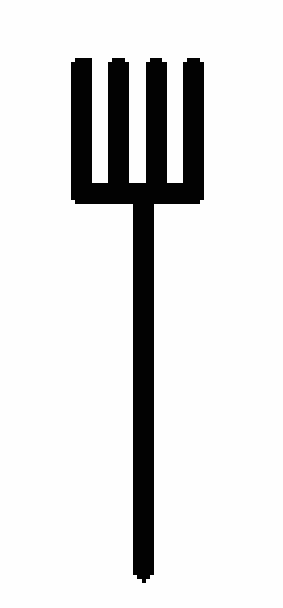} &
\includegraphics[height=2.7cm]{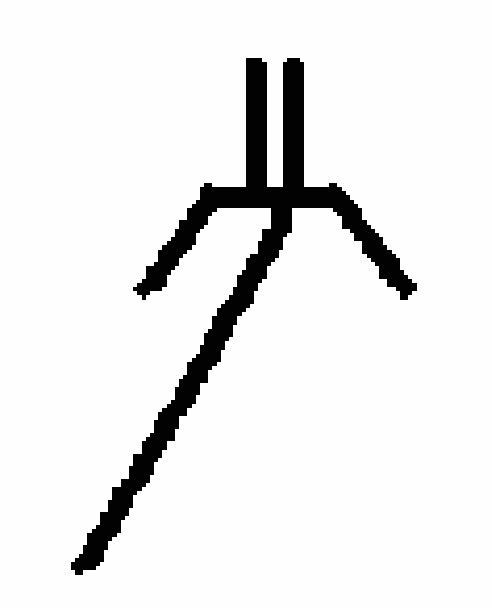} \\
(a) & (b) & (c) & (d)
\end{tabular}
\caption{On one hand, certain context require articulation
invariance such as being able to  say that (a) and (b) are the same
shape. On the other hand, it is less natural to make the same claim
for (c) and (d). Such a distinction may require previous encounter
to the shapes in consideration. Generic shape representation should
be able to handle selected sensitivity to articulations depending on
the context.} \label{fig:goal}
\end{figure}

{A  strong motivation for computing a skeleton is to define a shape
centered coordinate frame in order to cope with visual
transformations. Blum~\cite{Blum73} defined this frame as a local
affine frame defined by three parameters: symmetry axis curvature,
width and width angle. Such a local frame is  utilized in  recent
appearance based recognition work~\cite{Ozcanli06}. However, when a
skeleton becomes disconnected due to excessive smoothing, the local
coordinate frame in a large central area is lost. As a solution, we
propose to  separate visual changes into three groups; as rigid
transformations, piecewise rigid transformations (articulations) and
infinitesimal transformations (deformations). We develop a global
Euclidean frame (Section~\ref{sec:gcf}) and demonstrate that the
coarse spatial structure imposed by this frame  is sufficient for
shape comparison in a way which is robust to articulations as well
as Euclidean transformations (Section~\ref{sec:smas}). Even though
invariance to visual transformations is desirable, there are
situations in which transformation variant descriptors must be used,
e.g. discriminating `6' from `9' or discriminating a likely
articulation from an unlikely one (Fig. \ref{fig:goal}). To
introduce sensitivity to  articulations we propose a part-centered
coordinate frame (Section~\ref{sec:slcf}). Finally, to handle
deformations we resort to the techniques developed for {\sl landmark
point} based representations~\cite{Bookstein91,Kendall99}
(Section~\ref{sec:morph}). By separating visual transformations, we
can produce descriptions that are sensitive to any combination of
changes in scale, location, orientation, and articulations in
addition to descriptions that are invariant to these changes. The
hybrid (axis vs. point) nature of our local symmetry representation
makes the necessary constructions and computations almost trivial.}

A preliminary conference version of this work appeared in
\cite{Aslan05}.  Implementation details are provided in the
supplementary material~\cite{Appendix}.

\section{Detection of Symmetry Axes}
\label{sec:dosa} For a given shape boundary $\Gamma$, we define the
distance surface $\phi$  as the solution of the following linear
diffusion equation at a special scale $\sigma^*$:
\begin{eqnarray}
\label{eq:eqn1}
\frac{\partial}{\partial\sigma}\phi(x,y,\sigma)=\left(\frac{\partial^2}{\partial
x^2}+\frac{\partial^2}{\partial y^2}\right)\phi(x,y,\sigma) \\
\phi(x,y,\sigma)|_{(x,y)\in\Gamma}=1 \nonumber
\end{eqnarray}

The scale $\sigma^*$ is not directly computed and depends
specifically on the shape itself; hence we interpret it as the
\textsl{absolute scale} of the shape. It is related to the amount of
maximum allowable regularization that  still permits morphological
analysis. An algorithm for the solution of (\ref{eq:eqn1}) is
provided in the Appendix~\cite{Appendix} along with other
implementation details.

In order to demonstrate the motivation for the surface
 $\phi$ and explain the selection of
$\sigma^*$, we relate the construction process to that of the edge
strength function  of TSP \cite{Tari96,Tari97,Tari98}.
\subsection{The Edge Strength Function of TSP}
\label{ssec:esf} The edge strength function $v$ is the solution of the
following PDE:

\begin{eqnarray}
\label{eq:eqn2} \left(\frac{\partial^2}{\partial
x^2}+\frac{\partial^2}{\partial
y^2}\right)v(x,y) - \frac{v(x,y)}{\rho^2}=0 \\
v(x,y)|_{(x,y)\in\Gamma}=1 \nonumber
\end{eqnarray}

The function $v$ attains its highest value of $1$ at the shape
boundary $\Gamma$, and decays monotonically as a function of the
distance from the boundary. As such, it is a smoothed analogue of
the distance function. Solution of (\ref{eq:eqn2}) can also be
expressed as the steady-state solution of a linear diffusion
equation with a bias term:

\begin{eqnarray}
\frac{\partial}{\partial\sigma}v(x,y,\sigma)=\left(\frac{\partial^2}{\partial
x^2}+\frac{\partial^2}{\partial y^2}\right)v(x,y,\sigma) - \frac{v(x,y,\sigma)}{\rho^2} \\
v(x,y)|_{(x,y)\in\Gamma}=1 \nonumber
\end{eqnarray}

\begin{figure}[!h]
\centering
\begin{tabular}{cc}
\includegraphics[height=4cm]{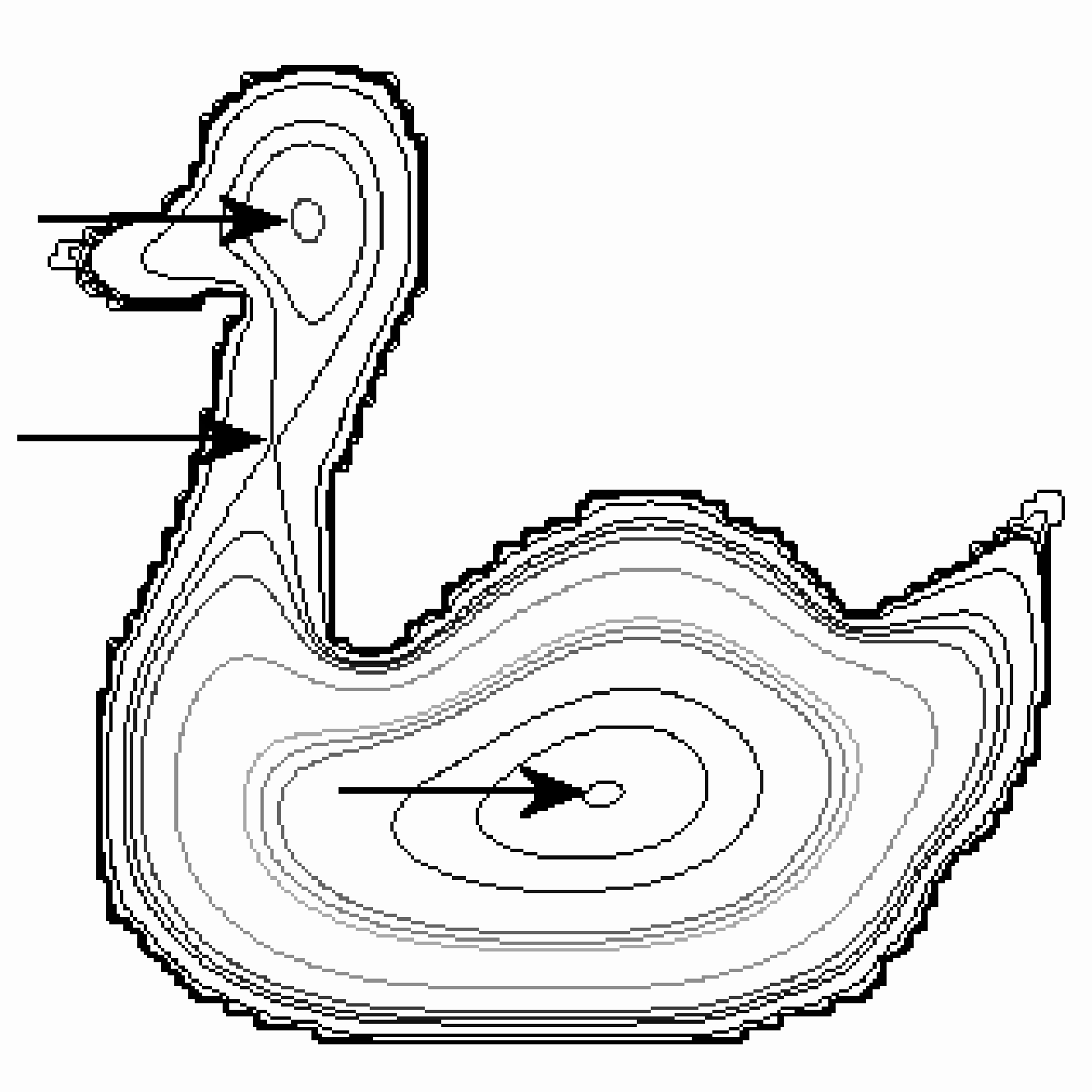} &
\includegraphics[height=4cm]{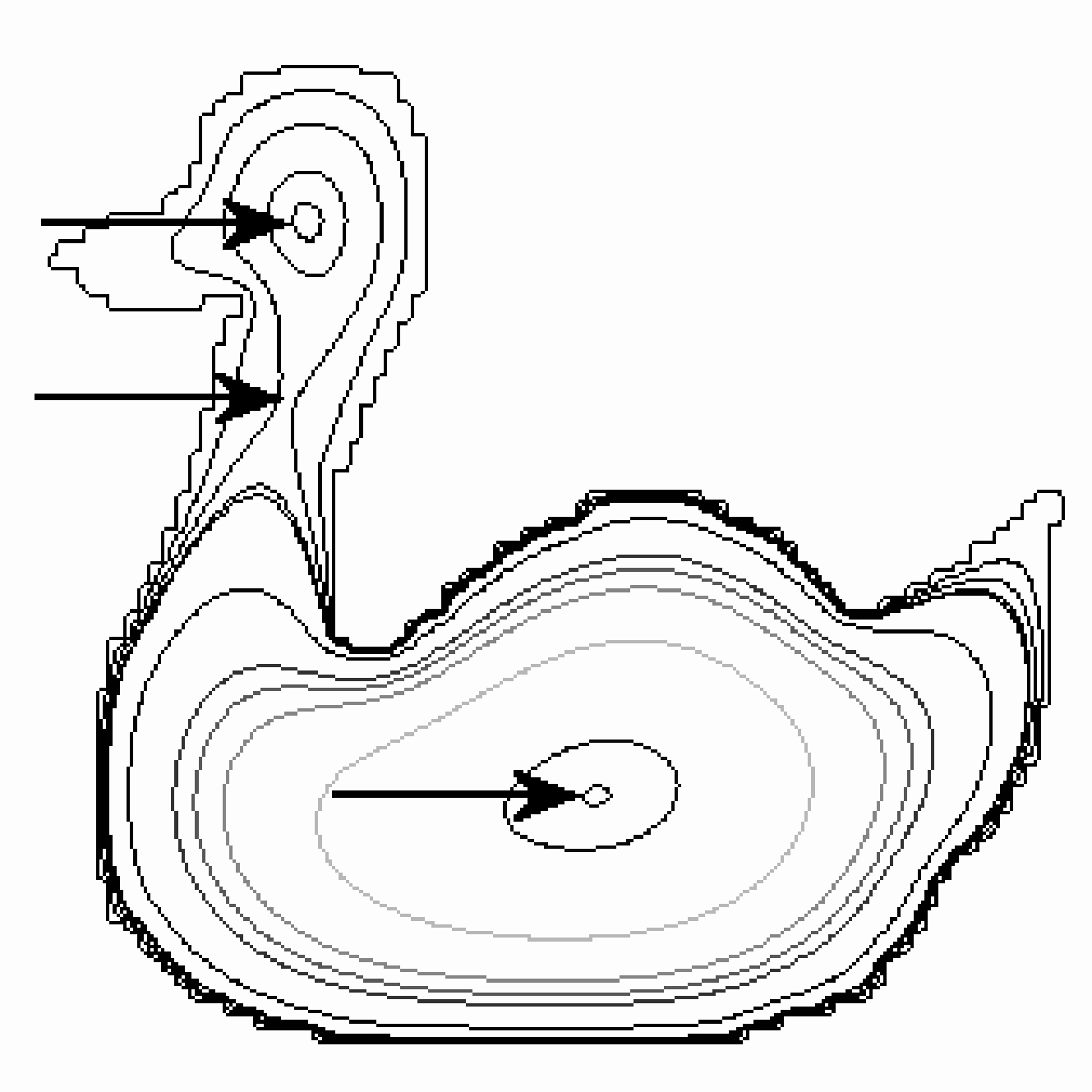}
\end{tabular}
\caption{The TSP method: level curves of $v$ for two different
$\rho$ values, $\rho=32$ (left) and $\rho=128$ (right). Arrows
indicate the extremum points of the surface $v$. There are two
minima ({\sl centers}) and one saddle point ({\sl neck}).}
\label{fig:tsp}
\end{figure}

As the only parameter $\rho$ approaches to zero, $v$ function
becomes an approximation of the discontinuity locus of the
Mumford-Shah segmentation model
\cite{Mumford89,Ambrosio90,Shah92,Shah96}. For larger $\rho$ values
(e.g. 4, 8, 16), the edge strength function acts like a level set
function whose level curves mimic the propagation of the shape
boundary in the inward normal direction with a curvature dependent
velocity \cite{Tari96,Tari97,Tari98}, as in the {\sl curve
evolution} of Kimia, Tannenbaum and Zucker \cite{Kimia95}. Sample
$v$ functions computed for a duck shape using two different $\rho$
values are illustrated in Fig.~\ref{fig:tsp}.

\begin{figure}[!h]
\centering
\begin{tabular}{ccc}
\includegraphics[height=3cm]{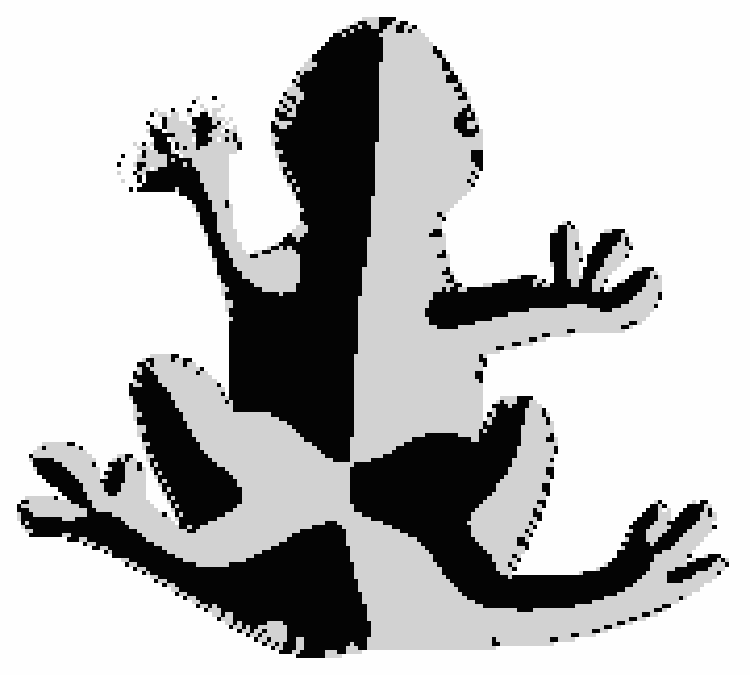} &
\includegraphics[height=3cm]{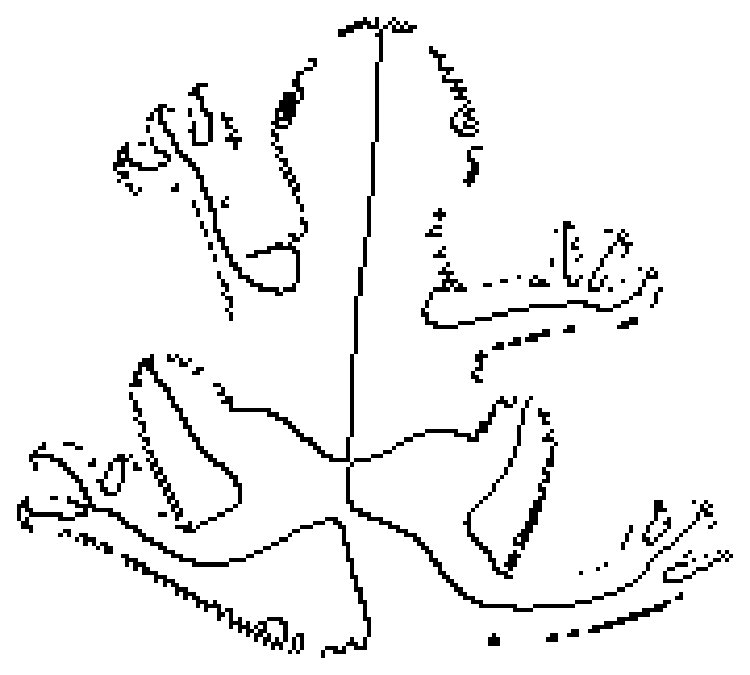} &
\includegraphics[height=3cm]{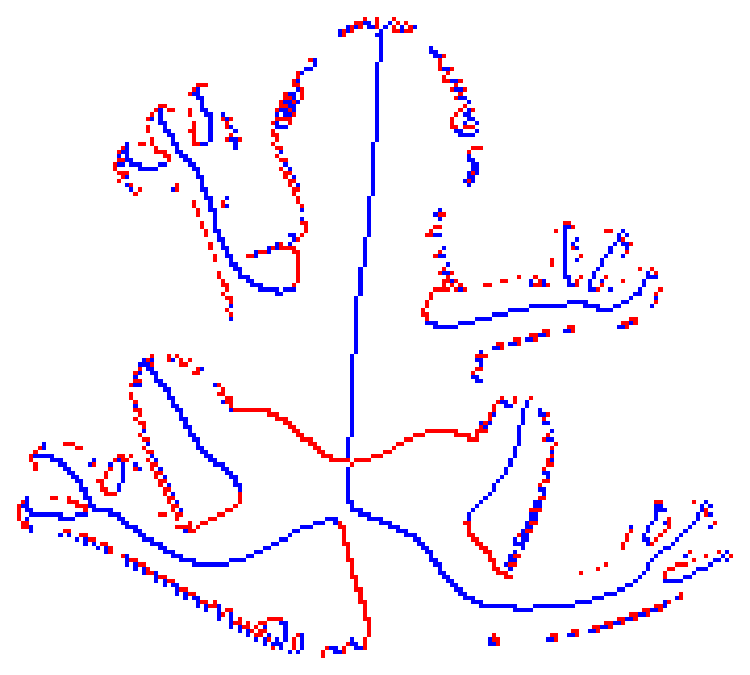} \\
(a) & (b) & (c)\\
\end{tabular}
\caption{The TSP method: (a) sign($d|\nabla v|/ds$). (b) Zero
crossings of $d|\nabla v|/ds$. (c) Positive (shown in blue) and
negative (shown in red) axes (see Algorithm 2 in the
Appendix\cite{Appendix}).} \label{fig:sym_points}
\end{figure}

Tari, Shah and Pien \cite{Tari96,Tari97}, demonstrated the
connection between the differential properties of the edge strength
function and the curvature extrema of the evolving shape boundary
$\Gamma$ and proposed to detect the local symmetry axes as the locus
of points where $\frac{d|\nabla v|}{ds}$ ($s$ is the arclength in
the direction of the level curves) given by

\begin{equation}
\frac{d|\nabla v|}{ds} =
\frac{((v_y^2-v_x^2)v_{xy}-v_xv_y(v_{yy}-v_{xx}))}{|\nabla v|^2}
\label{eq:vns}
\end{equation}
\noindent vanishes.  One can distinguish between the axes that track
the curvature maxima (\textsl{positive axes}) and the curvature
minima (\textsl{negative axes}) using {a} second derivative
\cite{Tari97} or a sign test \cite{Shah05} (Fig.
\ref{fig:sym_points}). {In the TSP skeletons unintuitive branches
appear in the vicinity of necks, due to a major pathology, which we
 refer to as  {\sl saddle point instability.} As discussed
below, saddle point instability is related to {\sl insufficient
diffusion}. }

\subsection{Saddle Point Instability}
\label{ssec:spi} 
\begin{figure}[!h]
\centering
\begin{tabular}{cc}
\includegraphics[height=3cm]{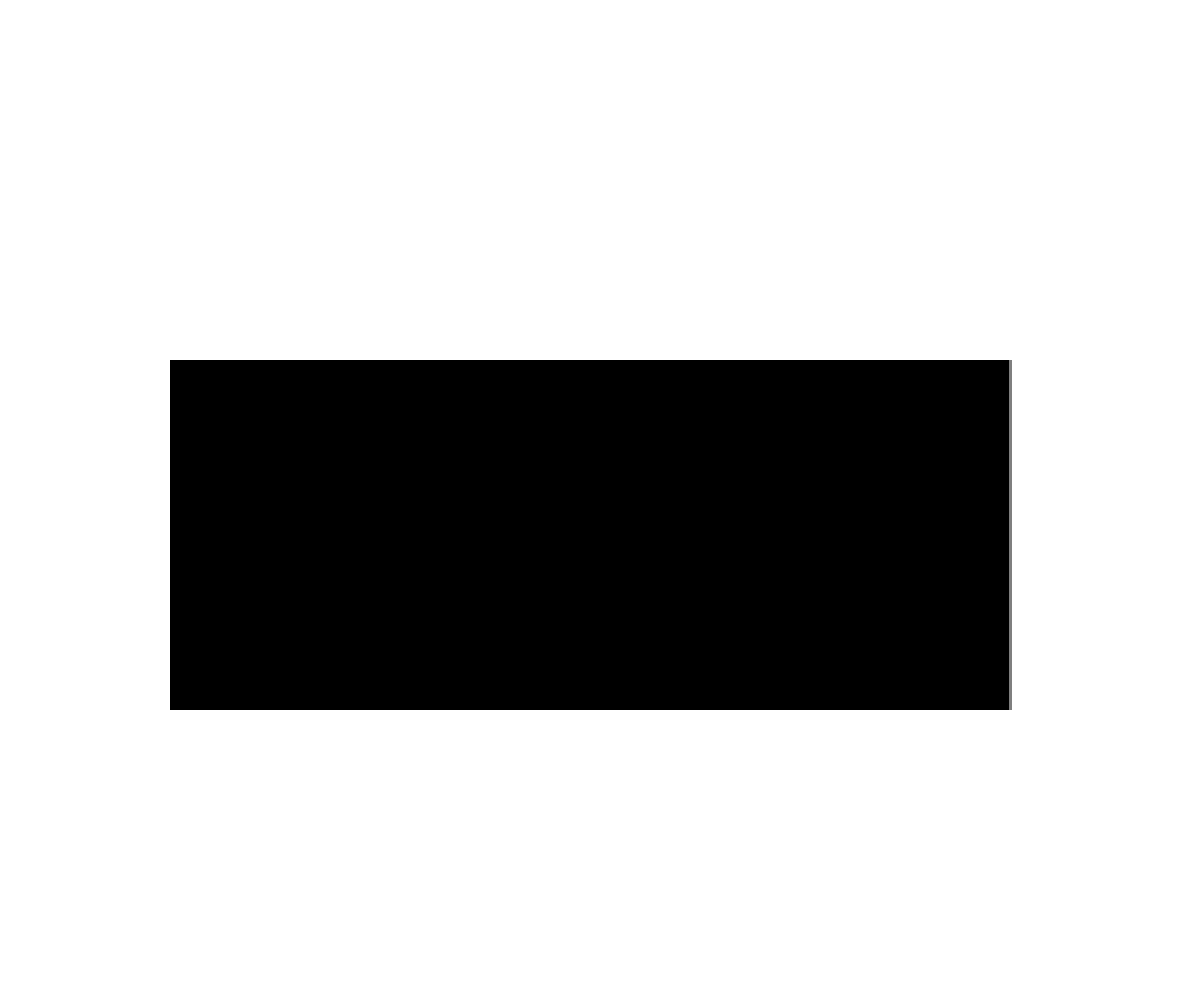} &
\includegraphics[height=4cm]{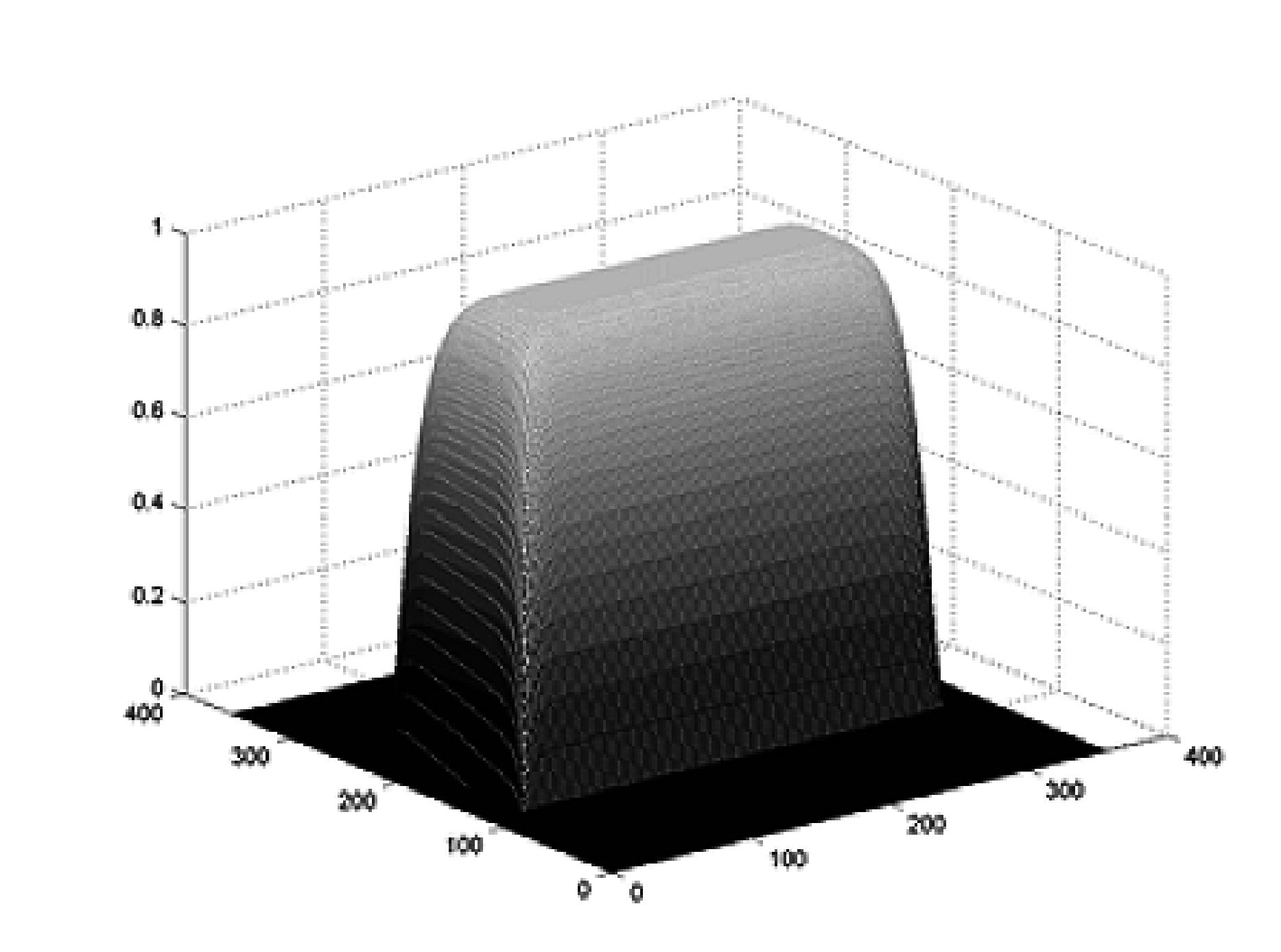} \\
(a) & (b) \\
\end{tabular}
\caption{Insufficient diffusion in the TSP method. (a) A rectangle
shape. (b) Surface $1-v$ ($\rho=8$).} \label{fig:tsp_rect}
\end{figure}

Consider the rectangle and the surface $1-v$
(computed from it) {which are} shown in Fig. \ref{fig:tsp_rect}. The
surface is computed using $\rho=8$, which is a reasonable value for
many shapes of the same size. It can be seen that most of the
interior of the rectangle has remained intact (surface has the
initial value of zero at these points). Because of the insufficient
diffusion, the differential properties of the surface cannot be
computed accurately and further shape analysis cannot be carried
out.

Insufficient diffusion occurs because the speed of a point on the
curve depends on the interaction between nearby points. For the
relatively thin parts of a shape, the opposite boundaries affect
each other sooner than the broad regions; hence the speed of the
level curves in these thin parts increases. On the contrary, the
opposite boundaries in the broad regions start to affect each other
much later. If the smoothing parameter is negligible compared to the
width of the broad regions, the diffusion equation reaches  the
steady state without affecting the inner parts of the shape enough.
 It can be suggested that some sort of scaling and
resizing may be performed before computing the surface $v$. This
would likely  fail because the amount of smoothing required depends
on the thickness of the limbs of a shape, rather than its overall
size.

\begin{figure}
\centering
\begin{tabular}{ccc}
\includegraphics[height=8cm]{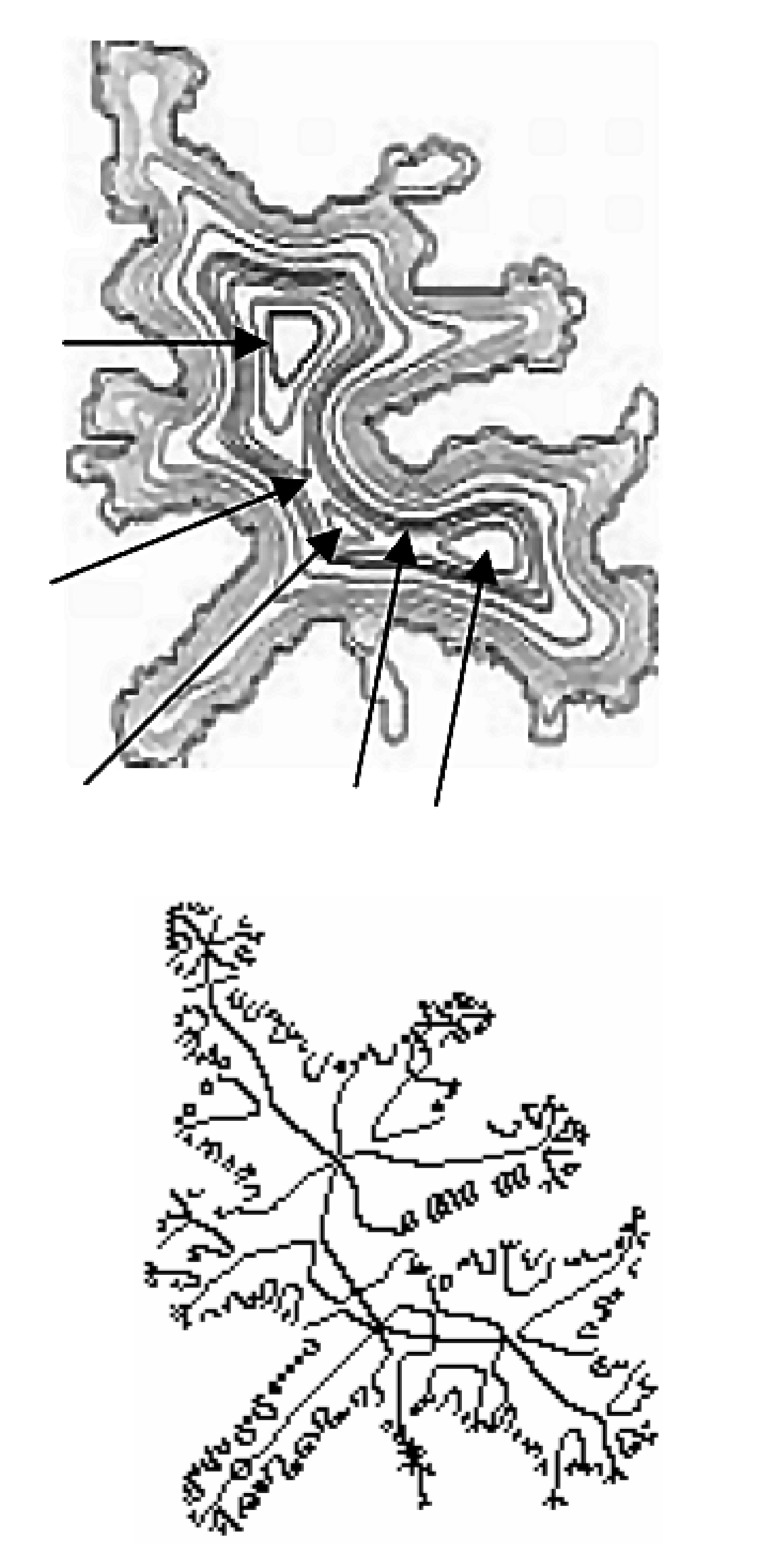} &
\includegraphics[height=8cm]{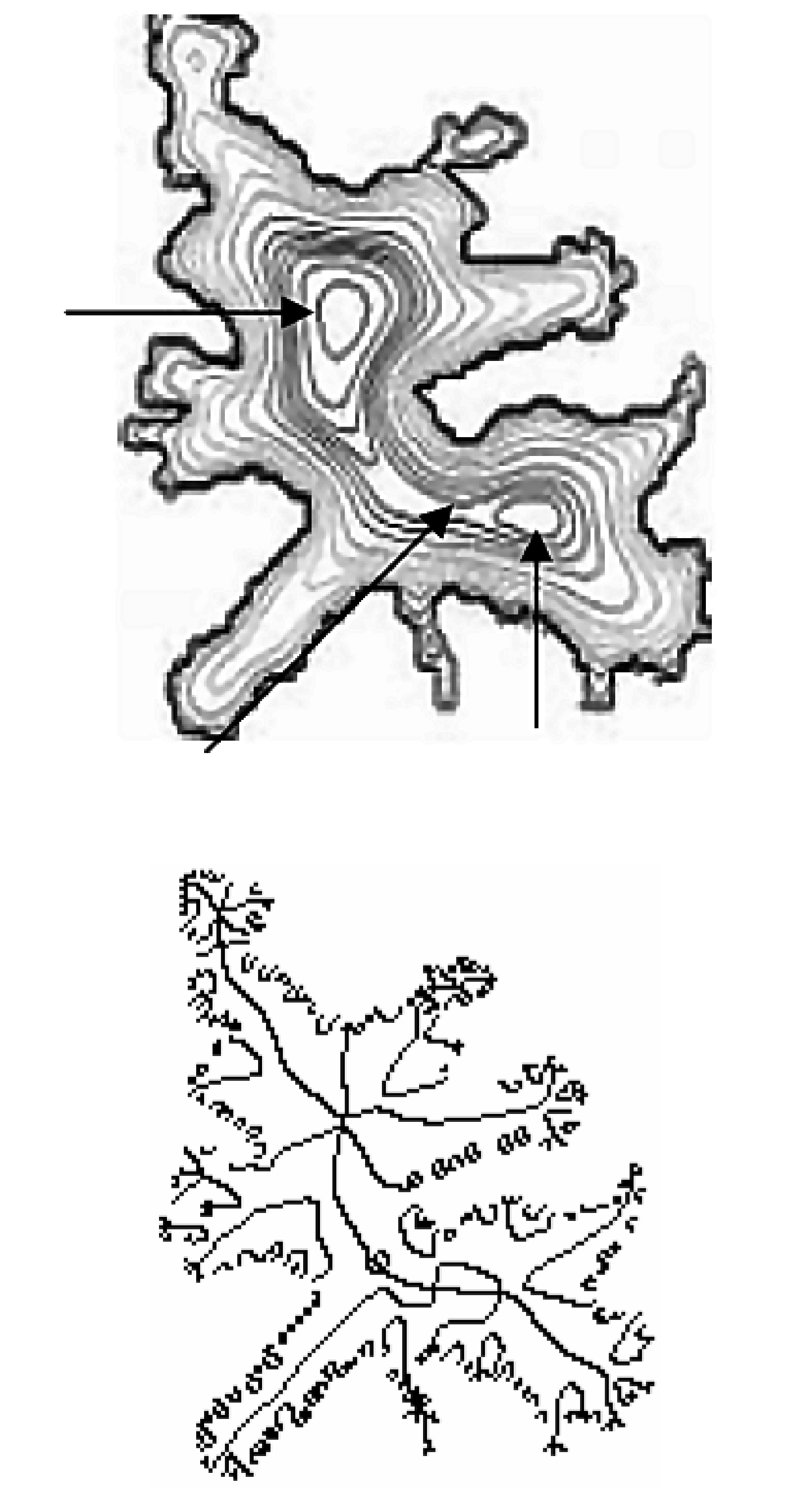} &
\includegraphics[height=8cm]{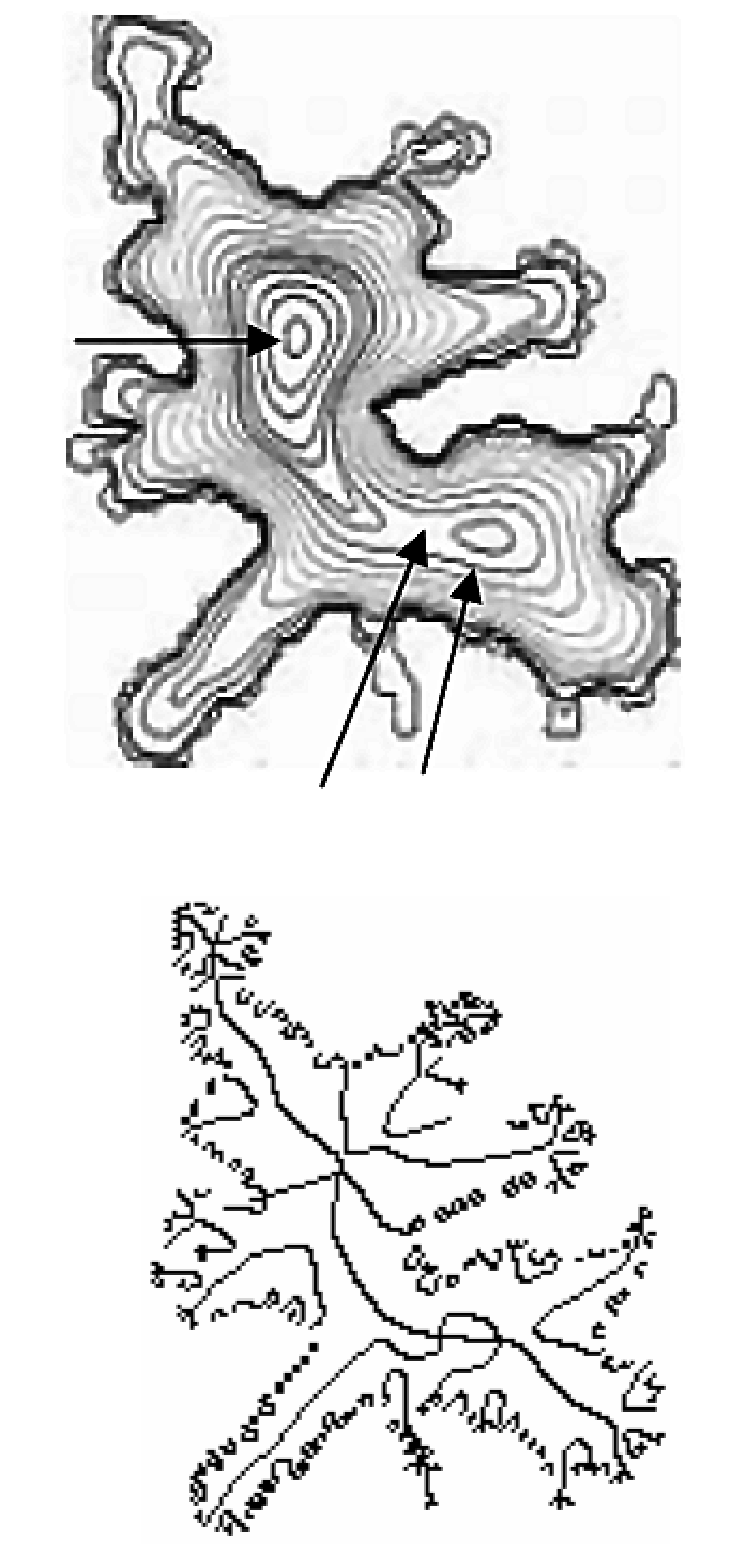} \\
(a) & (b) & (c)\\
\end{tabular}
\caption{Symmetry point detection using the TSP method on a
segmented MRI image using 3 different $\rho$ values. (a) $\rho=4$
(b) $\rho=8$ (c) $\rho=16$. {The top row (level curves) illustrates
the topological interpretation at three different scales by varying
$\rho$. The arrows indicate the minima and the saddle points. The
bottom row displays the branches computed from the respective
surfaces.}} \label{fig:tsp_mri1}
\end{figure}

\begin{figure}
\centering
\begin{tabular}{cc}
\includegraphics[height=4cm]{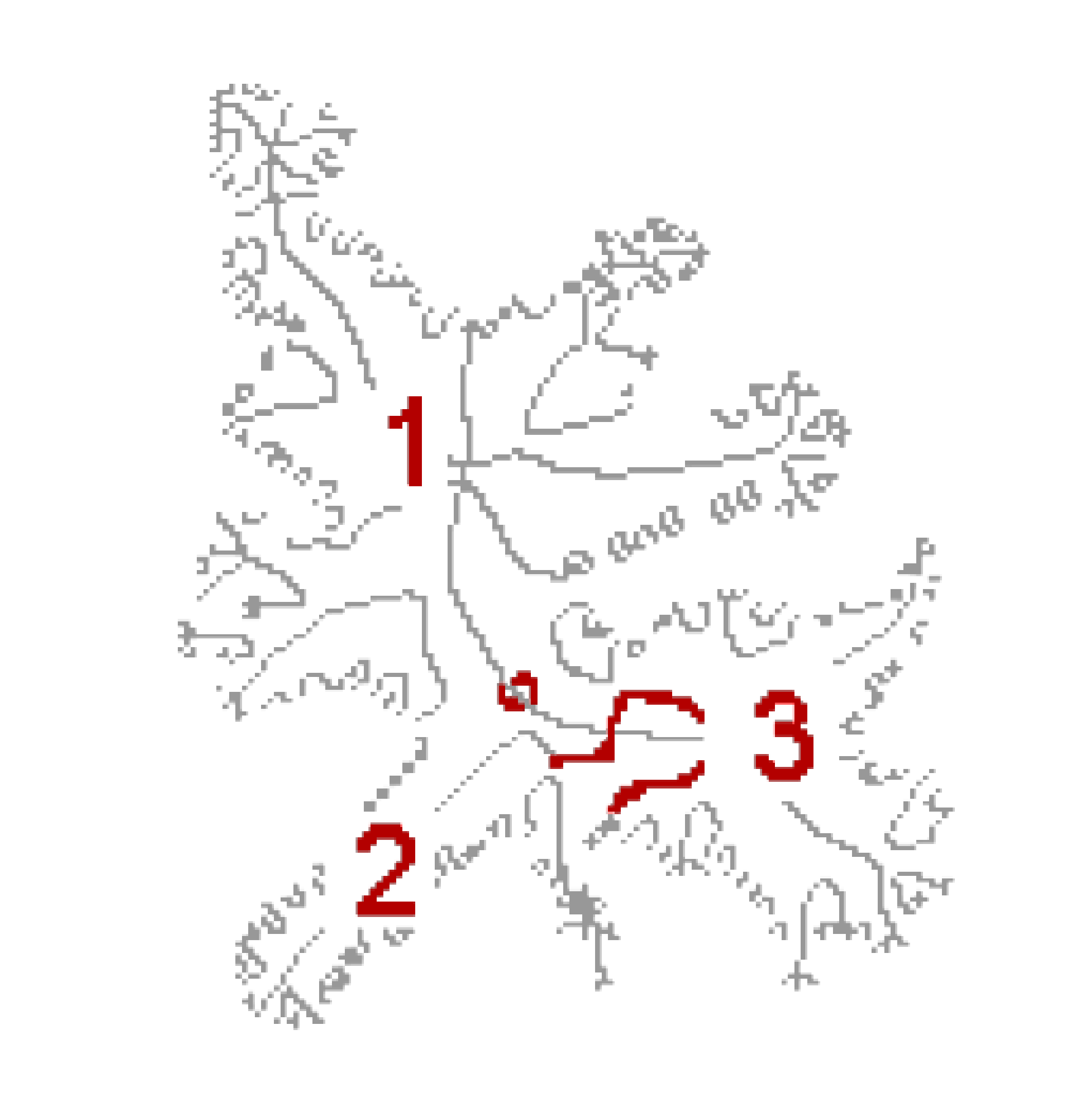} &
\includegraphics[height=4cm]{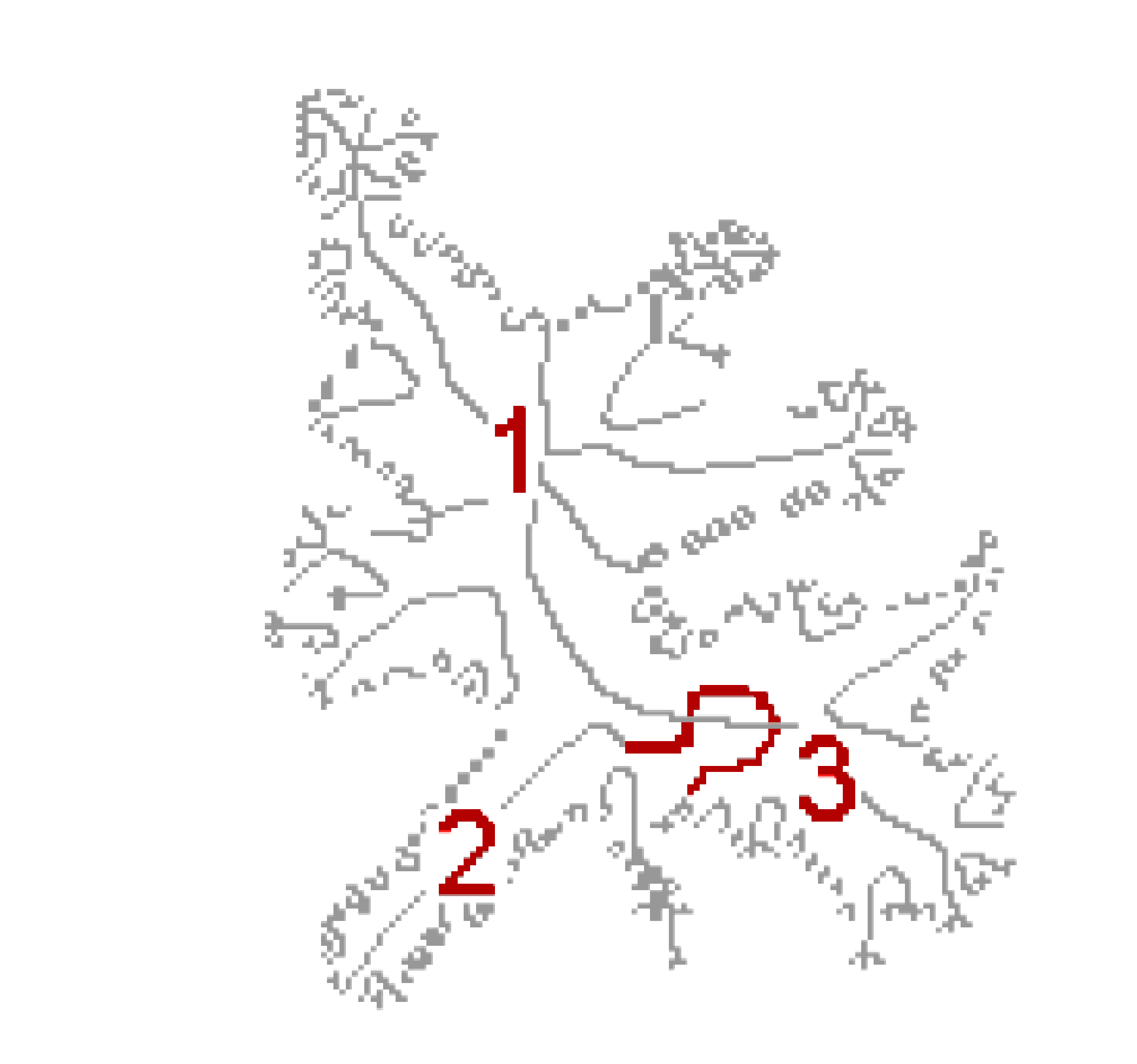} \\
(a) & (b) \\
\end{tabular}
\caption{Saddle point instability in TSP. The symmetry points of the
segmented MRI image when topological changes are not complete. (a)
$\rho=8$ (b) $\rho=16$. In  \ref{fig:tsp_mri2}(a), the computation
stopped while the shape was transforming from a shape with three
major blobs to a shape with two major blobs. The circular symmetry
branch colored with red is due to the interaction of the center of
parts two  and the neck between parts one and two. As shown in
\ref{fig:tsp_mri2}(b), increasing the amount of diffusion makes this
branch  disappear since the topological change is complete. This
time, the shape is between the state with two blobs (parts one and
two together and part three) and the state with one blob. The red
branch is due to the interaction of the center point of part
 three and the neck  between parts two and three.}
\label{fig:tsp_mri2}
\end{figure}

The effect of $\rho$ on the symmetry axis and on the level curves of
surface $v$ in the TSP formulation is visible in Fig.
\ref{fig:tsp_mri1}. The level curve plots for an MRI image are shown
in the top row. In the case of $\rho=4$, the evolving shape boundary
splits into three curves, therefore it shrinks into three distinct
minima separated by two saddle points. For $\rho=8$ and $\rho=16$
there are two minima. The symmetry points computed {from the
respective surfaces are shown in the bottom row. The ones computed}
with $\rho=8$ and $\rho=16$ exhibit a pathological situation that
frequently occurs in the TSP framework. Notice that these images
contain branches that do not correspond to any protrusion or
indentation (hence the local symmetry) of the shape.
The reason for these computation artifacts is that the diffusion is
stopped at such a critical time when the shape is between two
different interpretations that differ in topological structure (Fig.
\ref{fig:tsp_mri2}).

\subsection{Motivation for the New Surface}
\begin{figure}
\centering
\begin{tabular}{ccc}
\includegraphics[height=4cm]{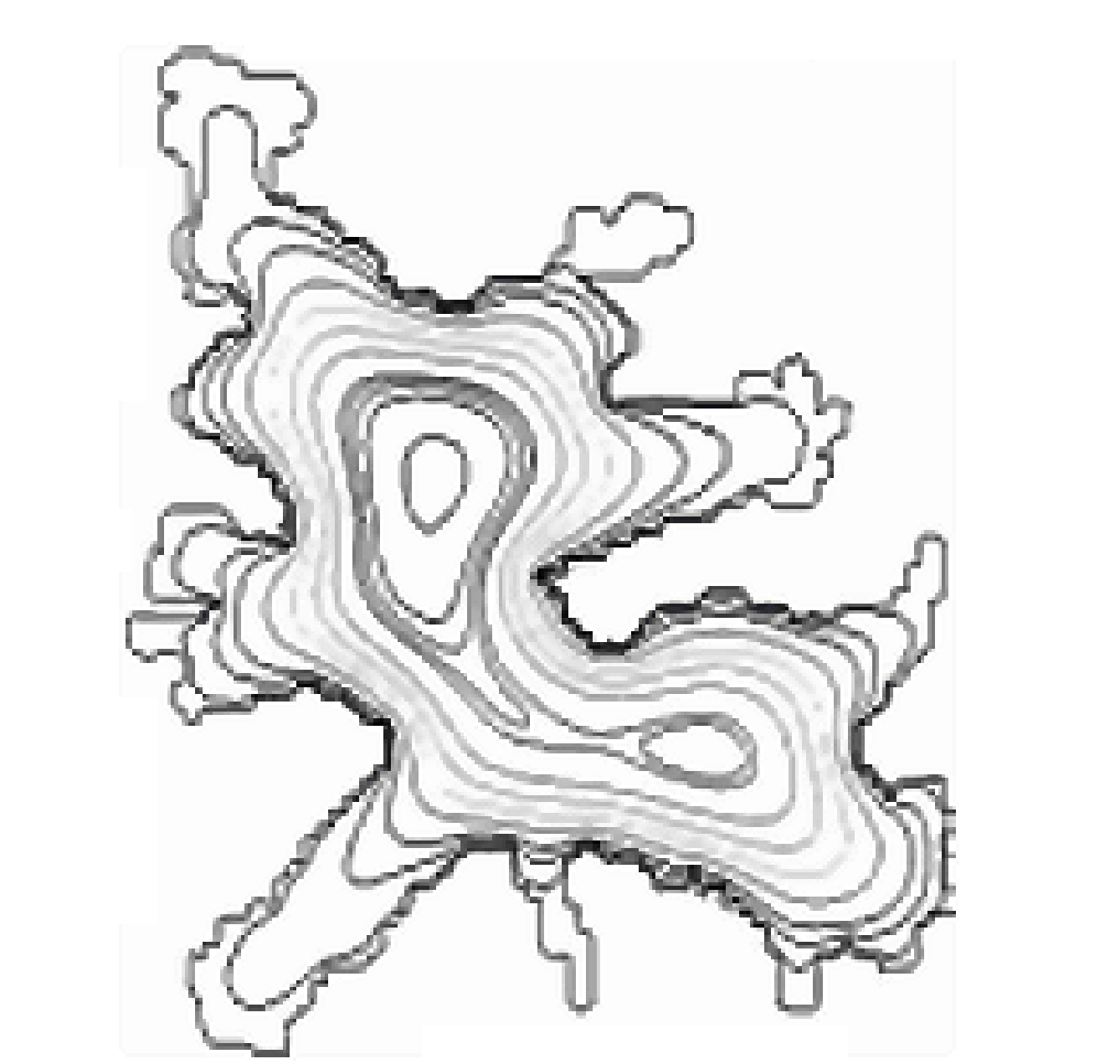} &
\includegraphics[height=4cm]{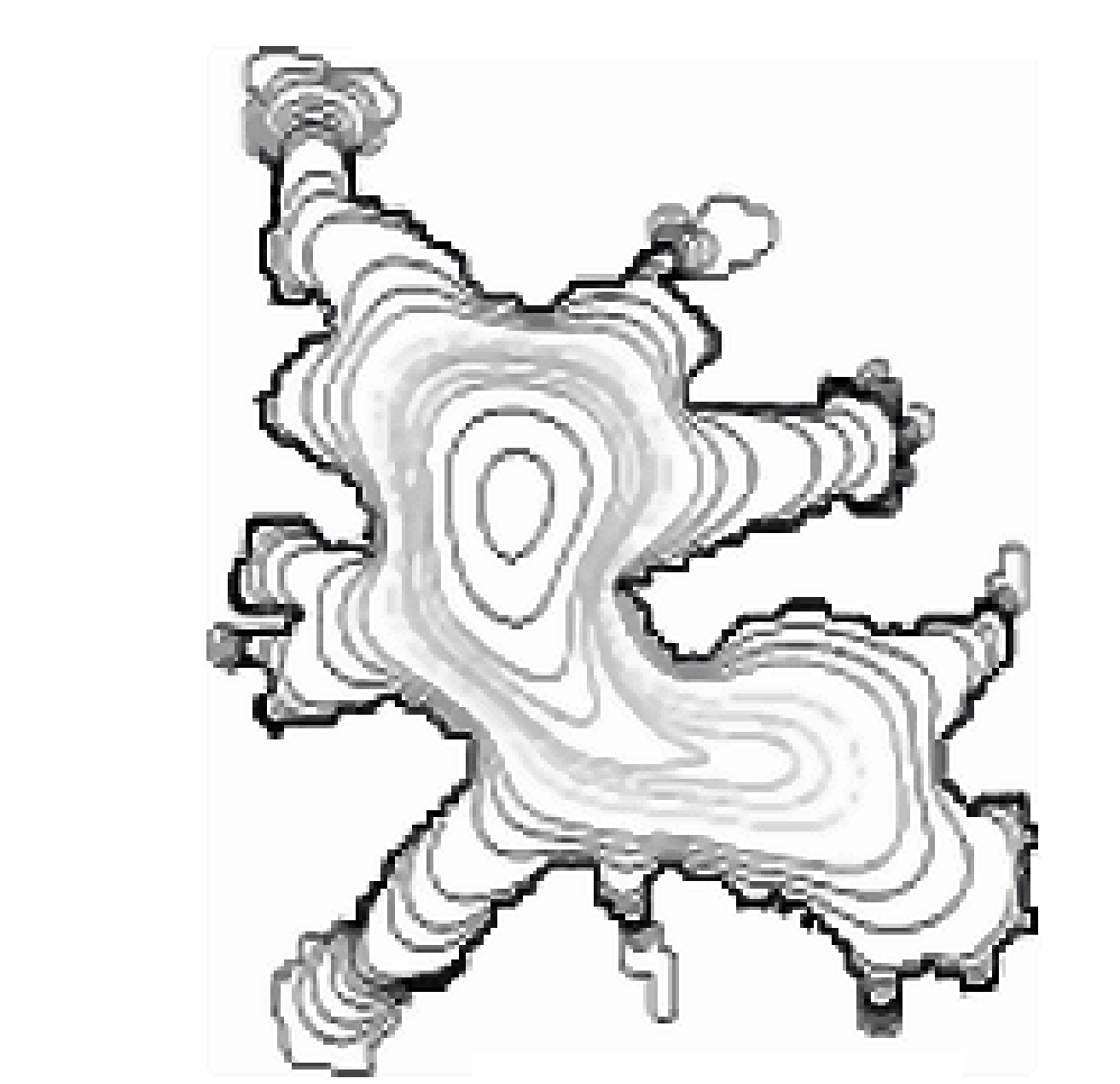} &
\includegraphics[height=4cm]{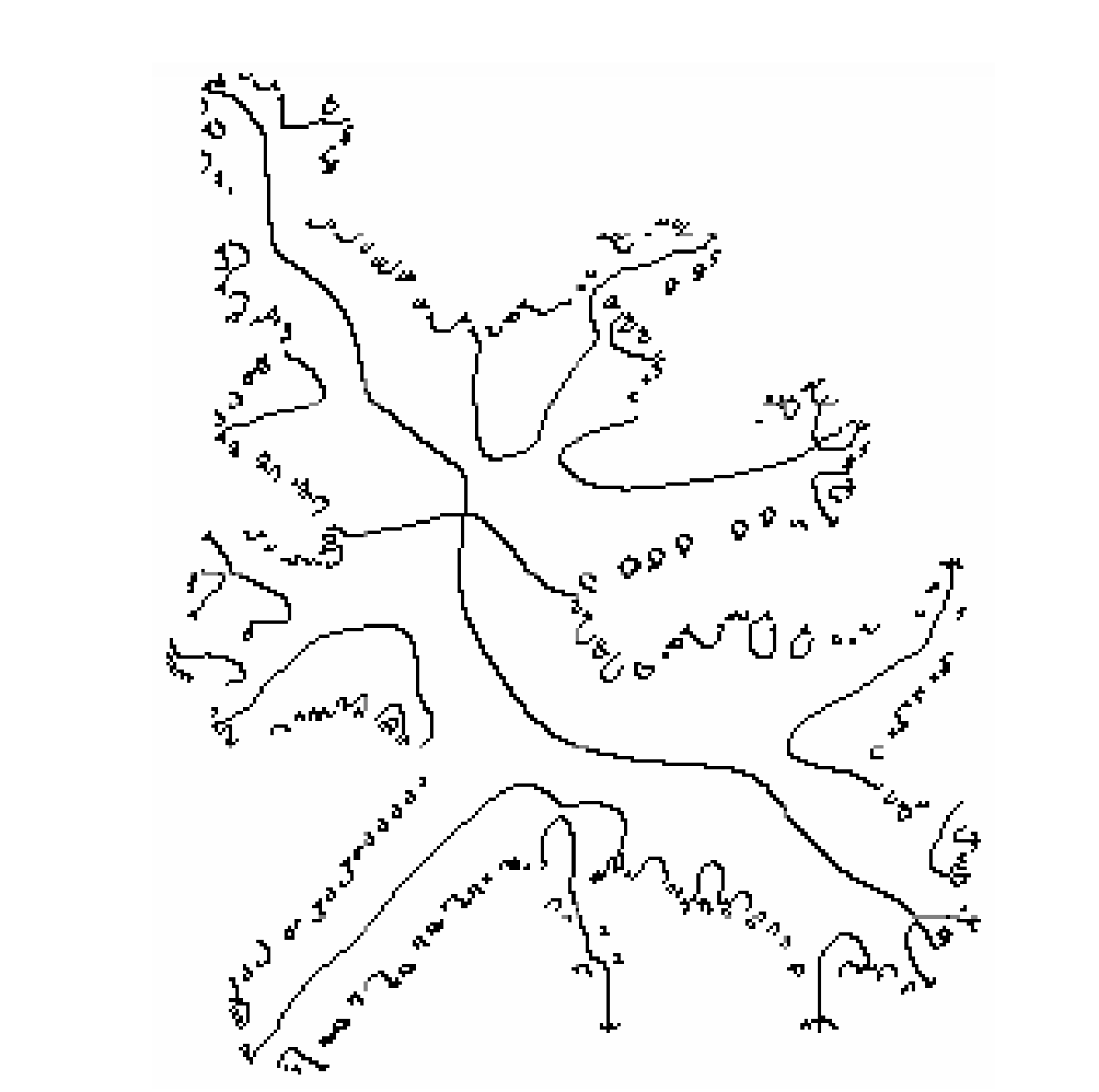} \\
(a) & (b) & (c)\\
\end{tabular}
\caption{New surface. (a) The level curves after initial diffusion.
(b) The level curves after final diffusion. (c) Symmetry points.}
\label{fig:aslan_mri}
\end{figure}

The diffusion of the surface until a single extremum is obtained is
expected to prevent  pathological situations of the TSP formulation
since there is no chance that the computation will be stopped in the
process of a topological change.  Fig. \ref{fig:aslan_mri}(a) shows
the level curves of the MRI image after some initial diffusion,
which is used to determine the approximate shape topology. The level
curves break at the neck point between parts two and  three. After
the final diffusion (Fig. \ref{fig:aslan_mri}(b)), there is only one
center point and the local symmetry set contains no computation
artifacts or unintuitive symmetry branches (Fig.
\ref{fig:aslan_mri}(c)).

Notice in Figures~\ref{fig:tsp_mri1} and \ref{fig:aslan_mri} that as
$\rho$ gets larger, the points of high curvature on the level curves
move faster, the protrusions are smoothed out earlier, less
important symmetry branches shrink, and  the length of a branch
becomes an accurate measure of its importance. In morphological
evolution, a curvature extremum, and therefore the symmetry branch
tracking it, survive until the end of the evolution (ignoring
bifurcations). {\sl This is why the length of a branch in a
morphological skeleton does not reflect its prominence}.

\begin{figure}
\centering
\begin{tabular}{cccc}
\includegraphics[height=3.1cm]{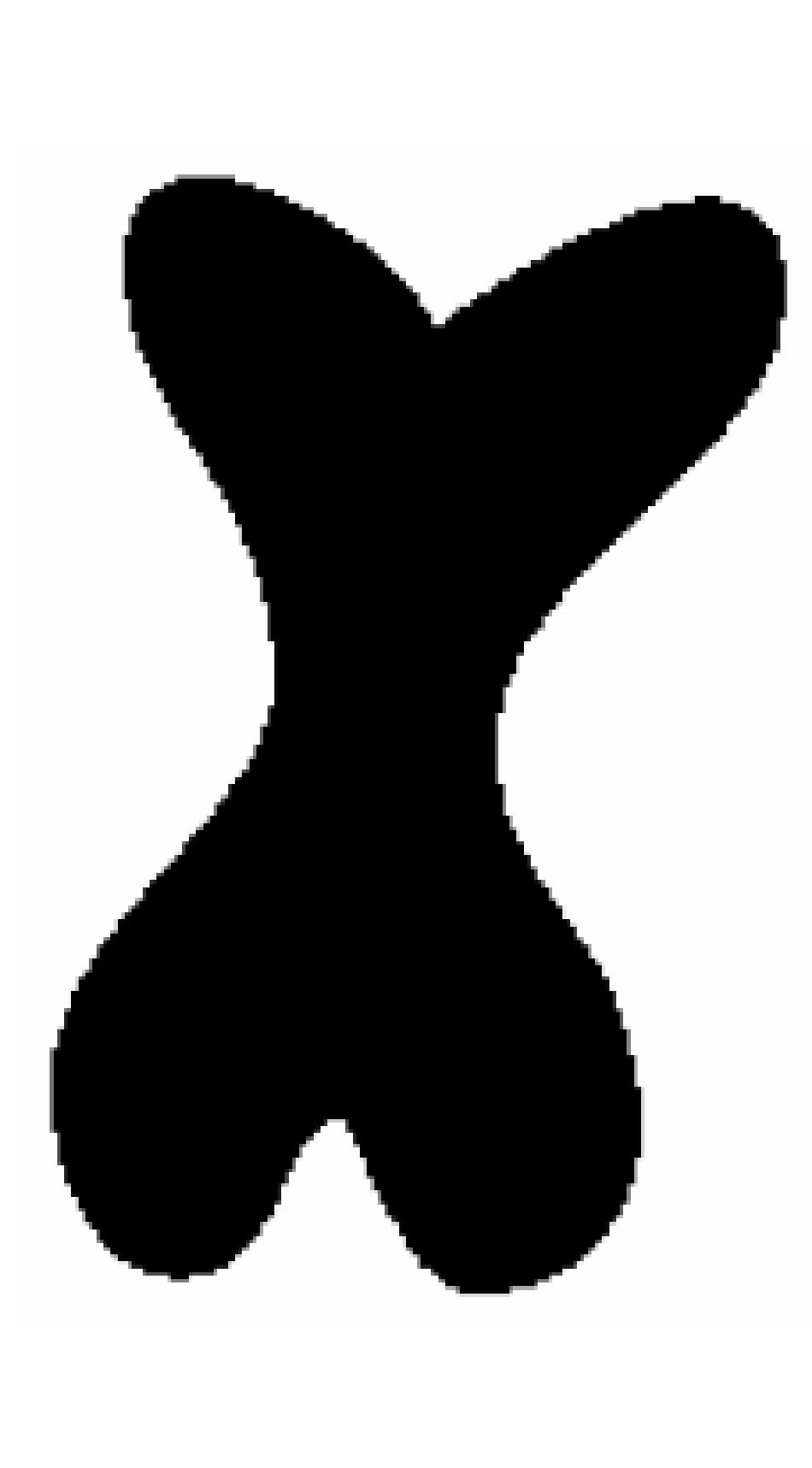} &
\includegraphics[height=3.1cm]{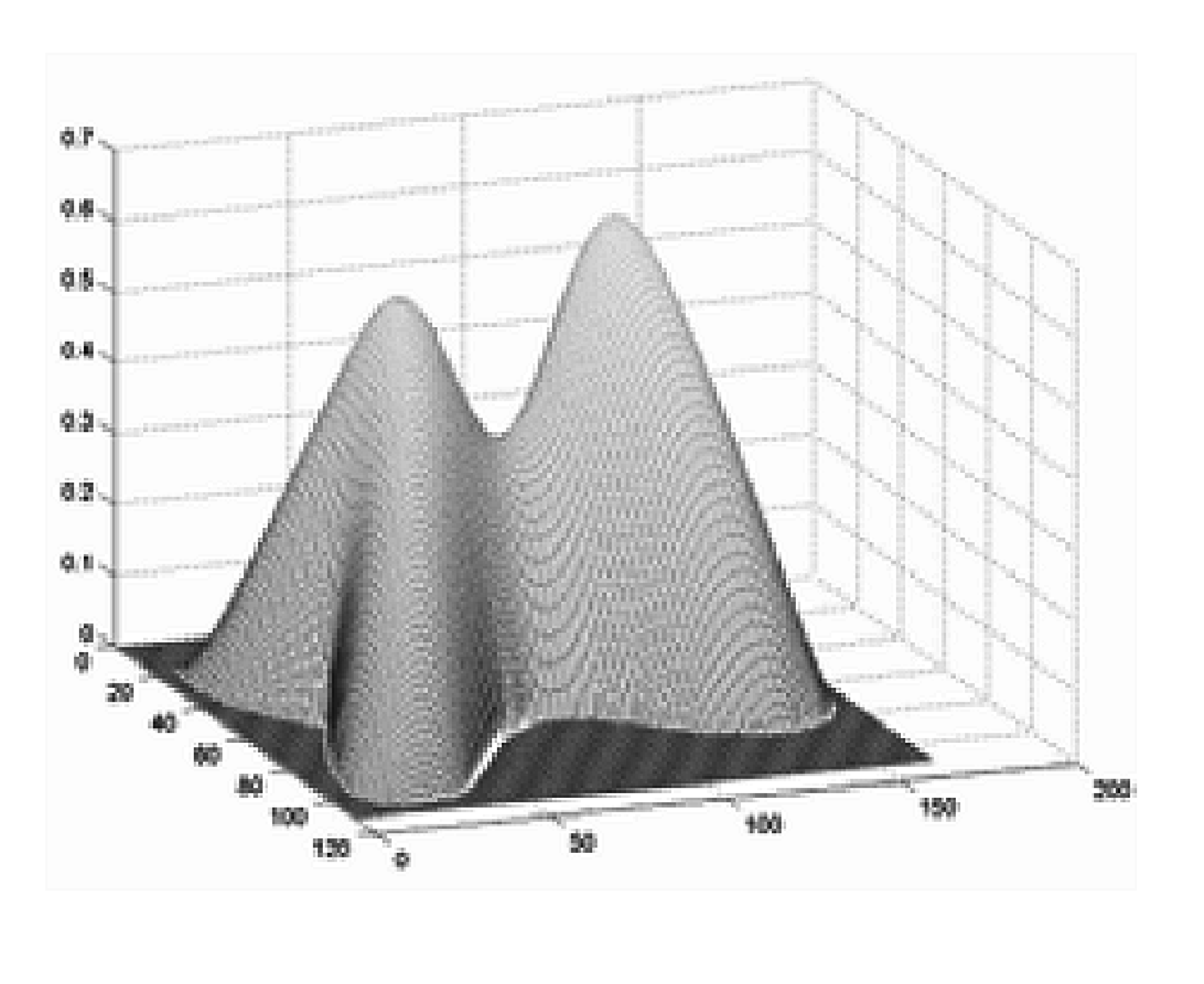} &
\includegraphics[height=3.1cm]{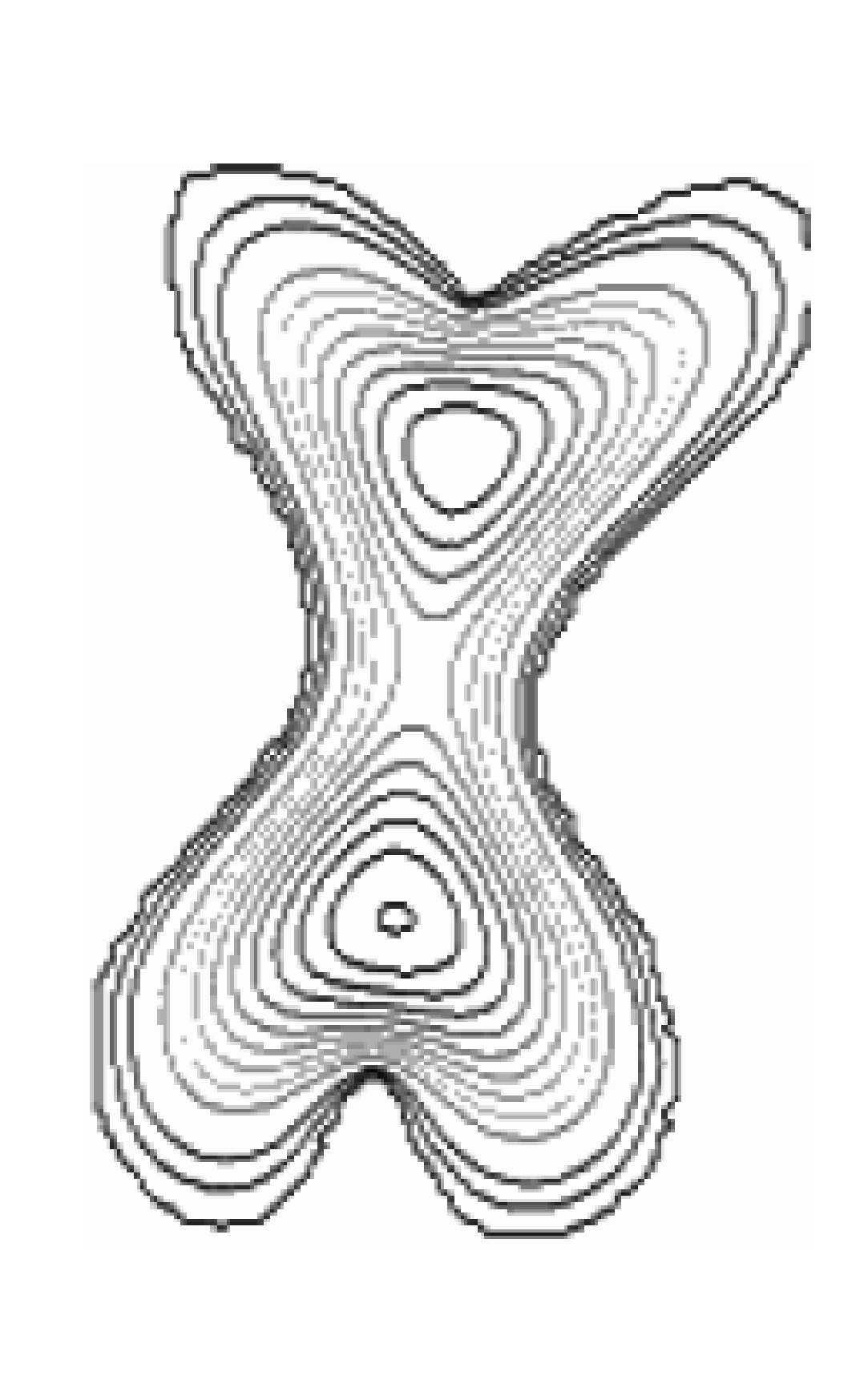} &
\includegraphics[height=3.1cm]{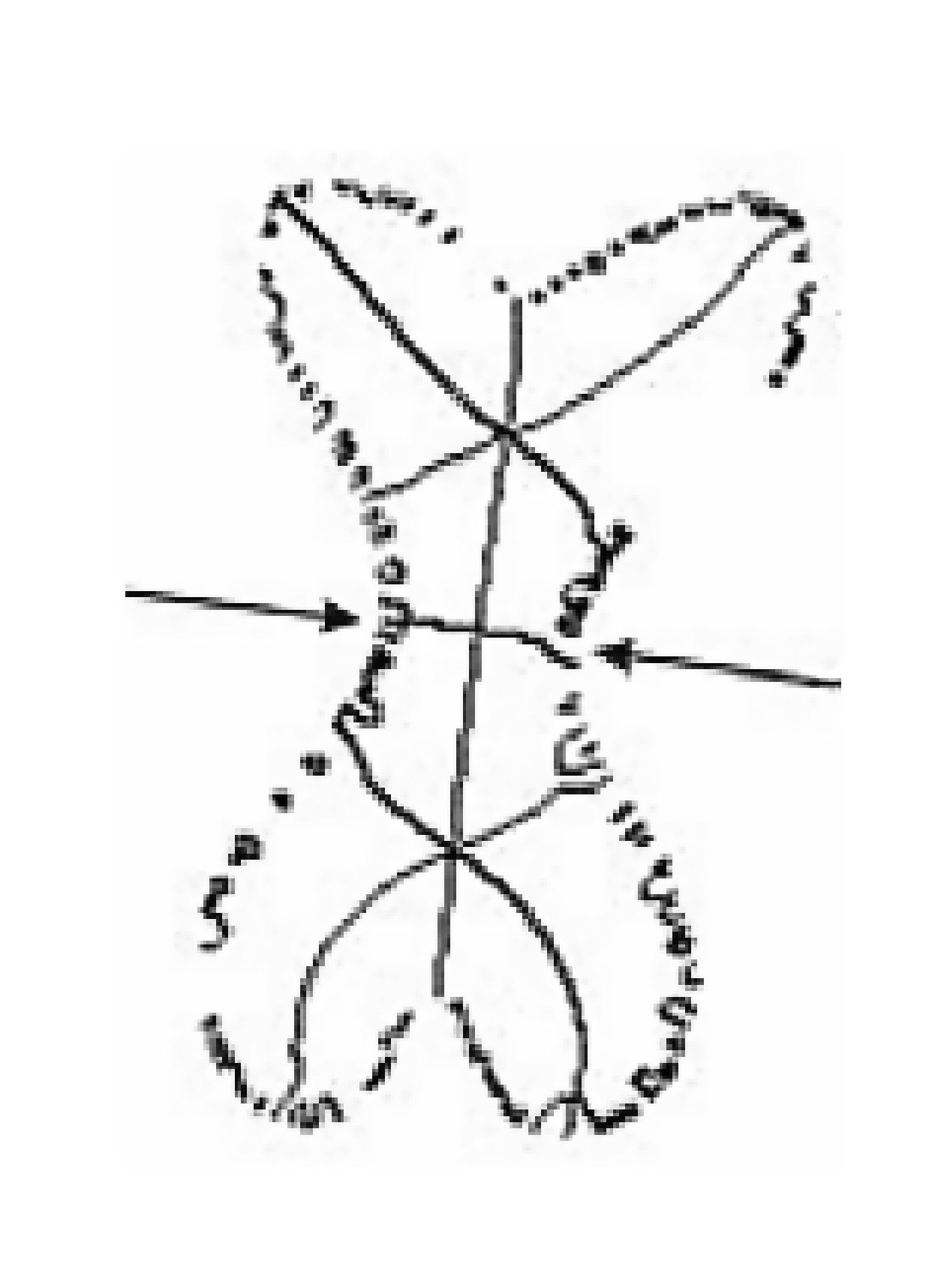} \\
(a) & (b) & (c) & (d)\\
\end{tabular}
\caption{(a) A dog-bone shape. (b) The surface $1-v$ after the
initial diffusion. (c) The level curves of the surface. (d) Symmetry
points. Arrows indicate symmetry branches that capture the neck of
the shape.} \label{fig:aslan_neck}
\end{figure}

There is a practical consideration associated with some dog-bone or
dumbbell-like shapes where the two main parts of the shape have
nearly the same prominence. It takes a significant amount of
computation time to reduce these kinds of shapes to a single point.
Therefore, it is logical to retain their dumbbell-like topology in
the final description. Having two types of descriptions may lead to
instability when some shapes that are between these two types are
encountered. However, the problem can be eliminated either by
forcing the diffusion further or by retaining two alternative
descriptions (i.e. at two different scales) for such shapes. This is
a trade-off between computational efficiency and accuracy (handling
ambiguities using alternative descriptions is discussed in
Section~\ref{sec:amb}). An example of surface computation of a
dog-bone shape is shown in Fig. \ref{fig:aslan_neck}. After the
initial diffusion step, the topology of the shape is determined to
be dumbbell-like as the two parts of the shape have nearly the same
prominence.
Further details can be found in~\cite{Aslantez}.
\subsection{From $\phi$ to Symmetry Axes}
Once $\phi$ is computed, local symmetry points can be detected and
classified as positive and negative using the TSP method. Grouping
the symmetry points into primitives (symmetry branches) is trivial
and merely based on connectivity. Recall that the symmetry points
are detected as zero-crossings; hence they are always connected for
each branch until the branch gets disconnected. Each primitive is a
local symmetry branch starting at a local curvature extremum of the
boundary and ending at a disconnection point. Very small branches
near the boundary are  discretization artifacts and can be
eliminated.

\begin{figure}[!h]
\centering
\begin{tabular}{ccc}
\includegraphics[height=2.8cm]{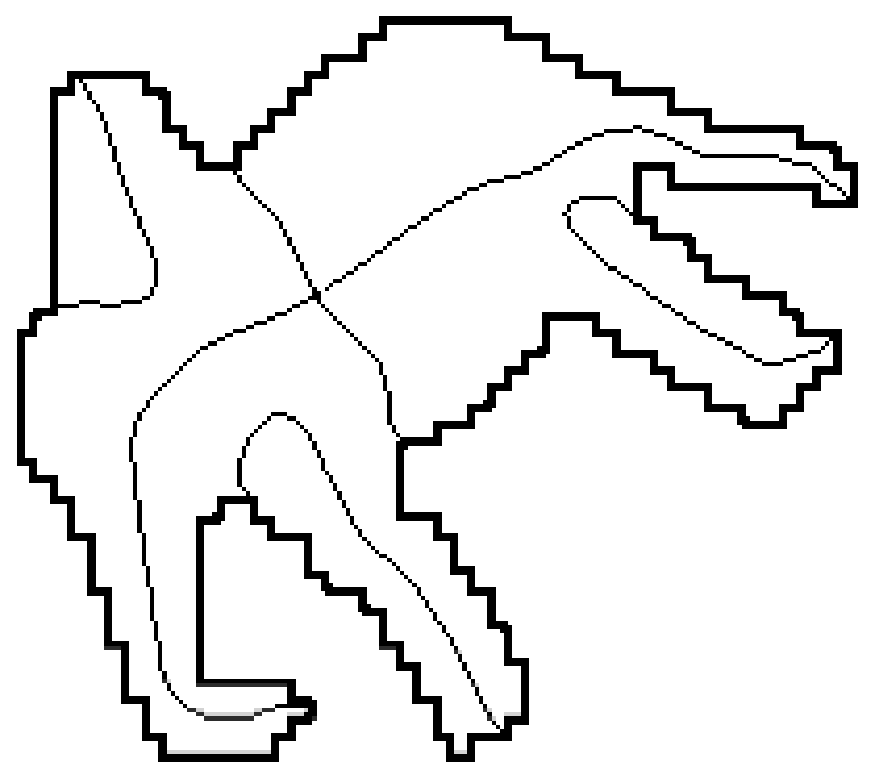} &
\includegraphics[height=2.8cm]{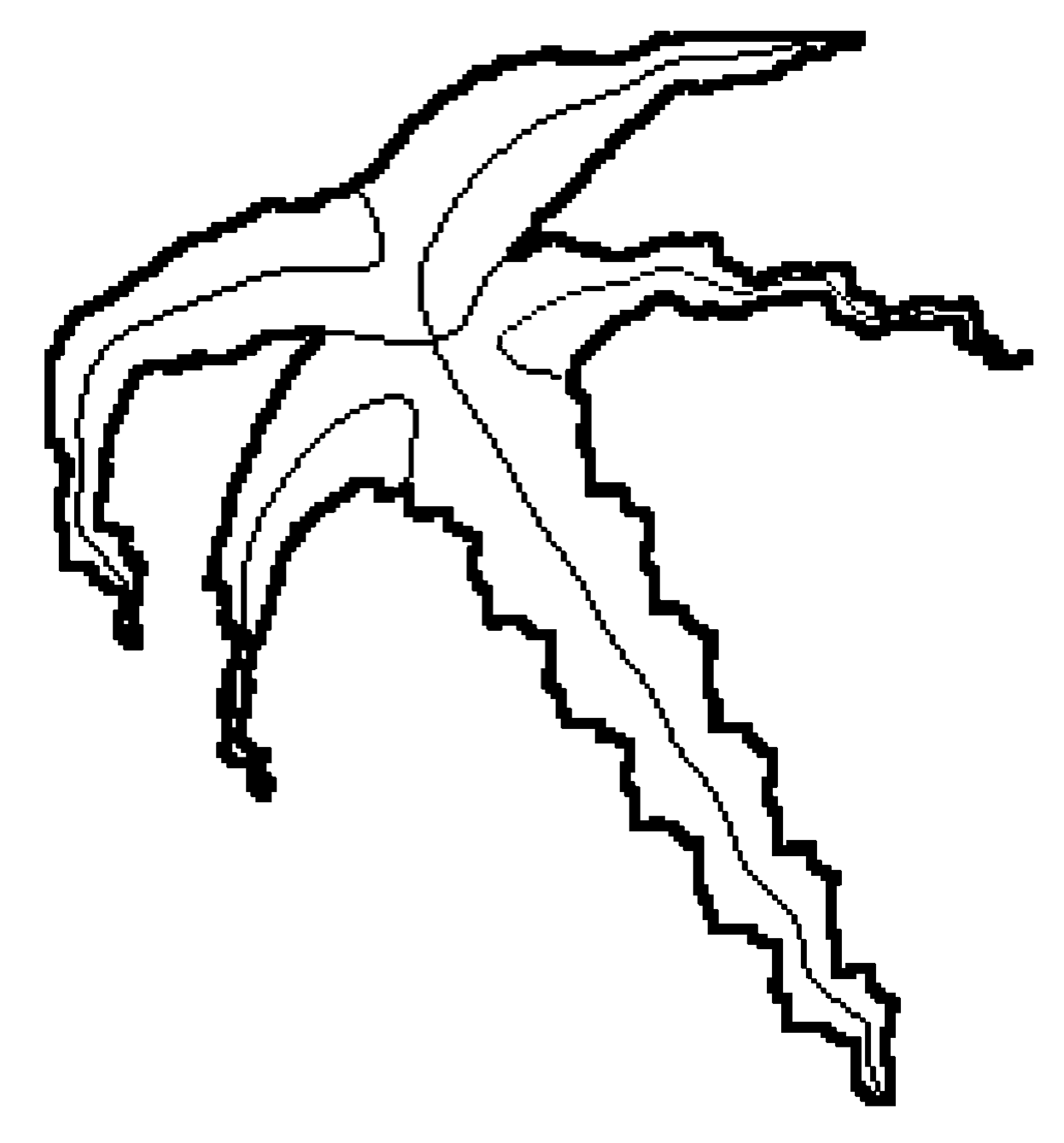} &
\includegraphics[height=2.8cm]{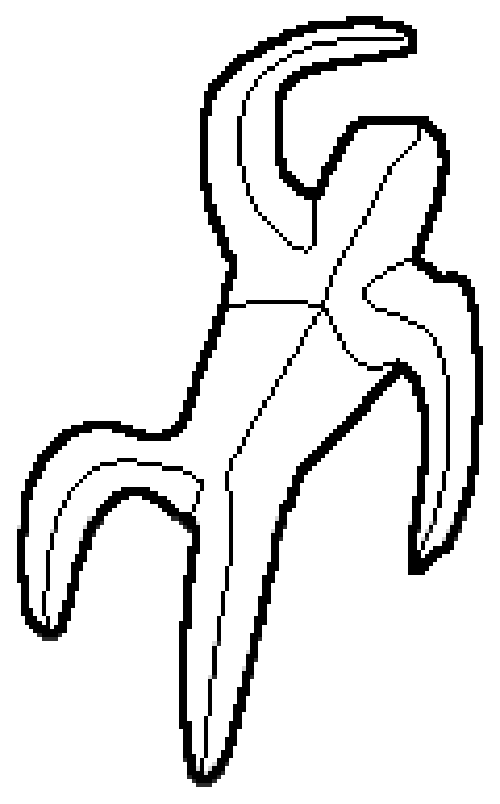} \\
\includegraphics[height=2.2cm]{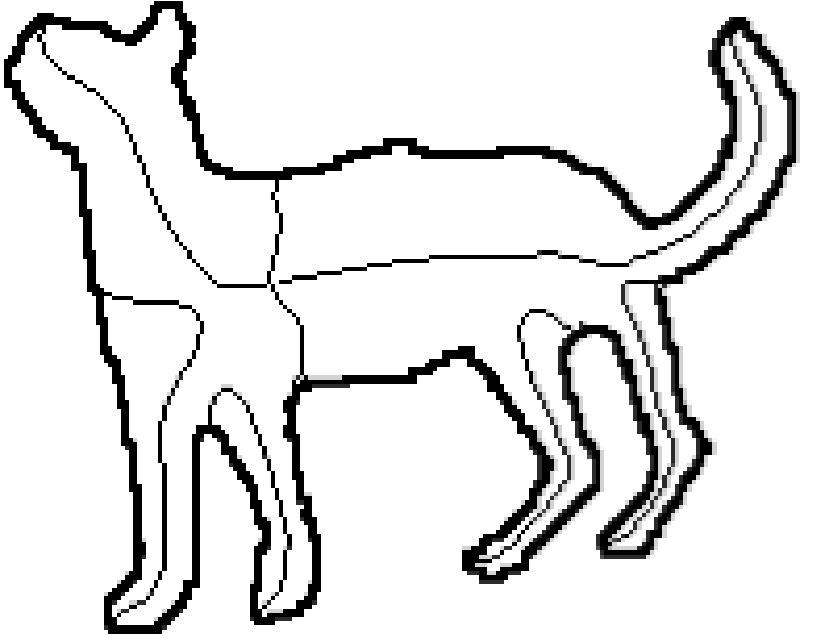} &
\includegraphics[height=2.8cm]{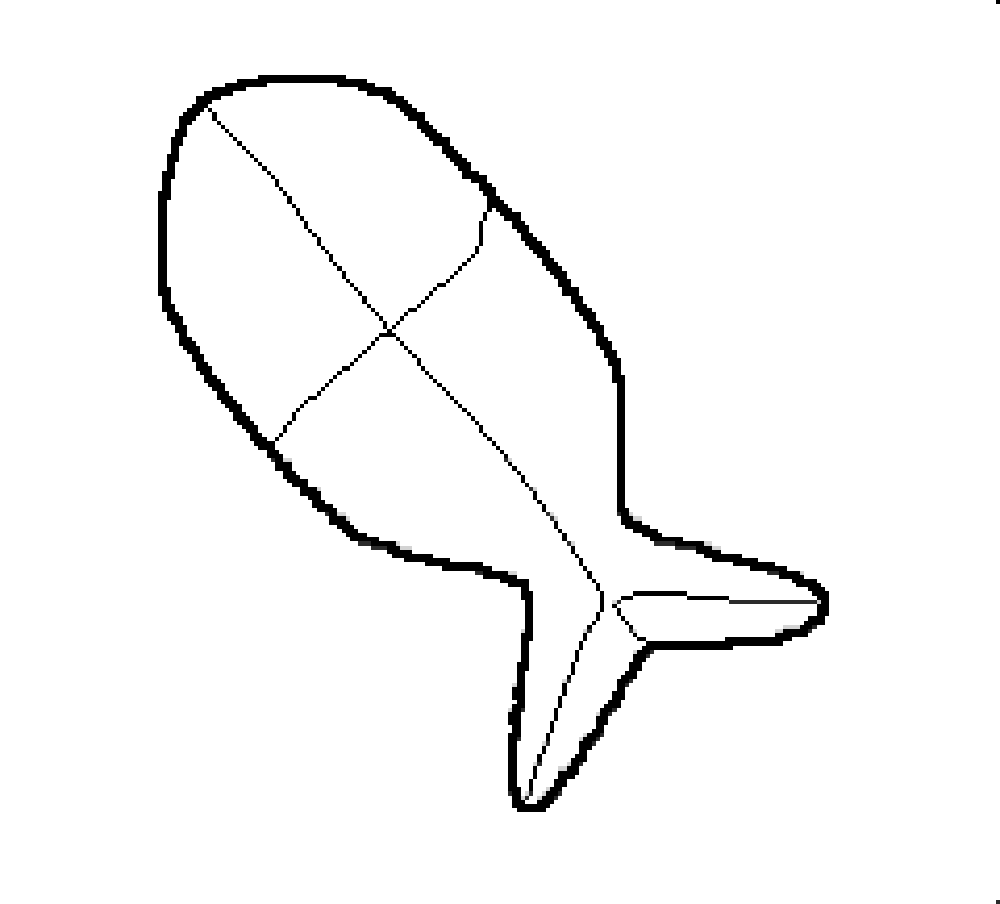} &
\includegraphics[height=2.8cm]{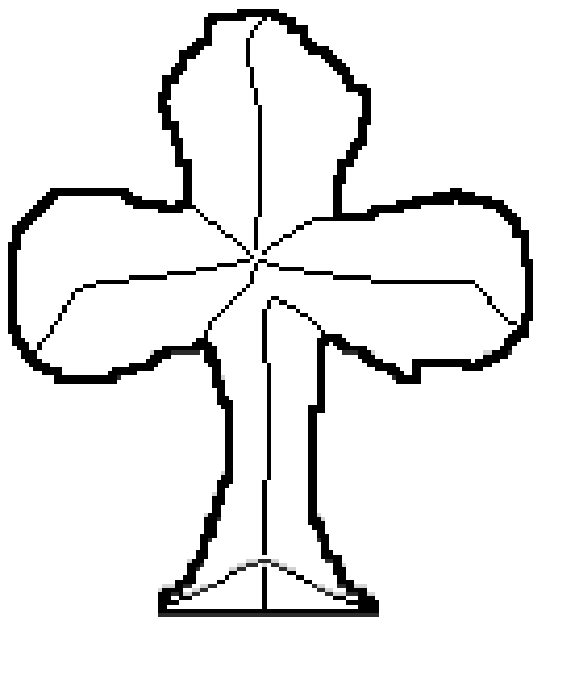} \\
\end{tabular}
\caption{Symmetry axes for sample shapes after prunning.}
\label{fig:skel_examples}
\end{figure}

In Fig. \ref{fig:skel_examples}, detected symmetry branches for some
shapes are shown. Notice that, with the exception of the bottom part
of the spades shape (the rightmost shape in the second row), all
branches are simple and indicate one level hierarchy. The spades
case is discussed and resolved in Section~\ref{sec:amb}. Further
notice that some symmetry branches survive without getting
disconnected and come to rest at a minimum or a saddle point of the
surface $\phi$. In the next section, we use these properties to
define a global coordinate frame and represent each simple branch.

\section{Global Coordinate Frame and the Disconnected Skeleton}
\label{sec:gcf} 
\begin{figure}[!h]
\centering
\begin{tabular}{cccc}
{\includegraphics[height=4cm]{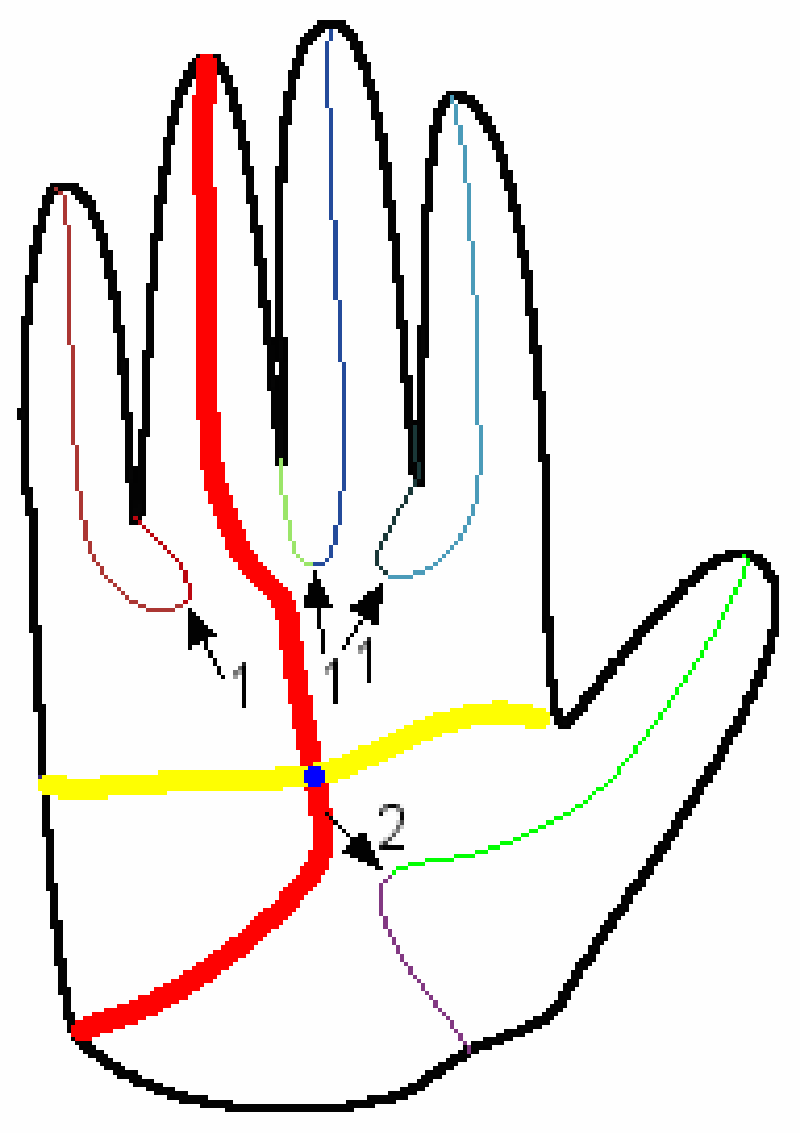} \hspace{1.5cm}} &
\includegraphics[height=4cm]{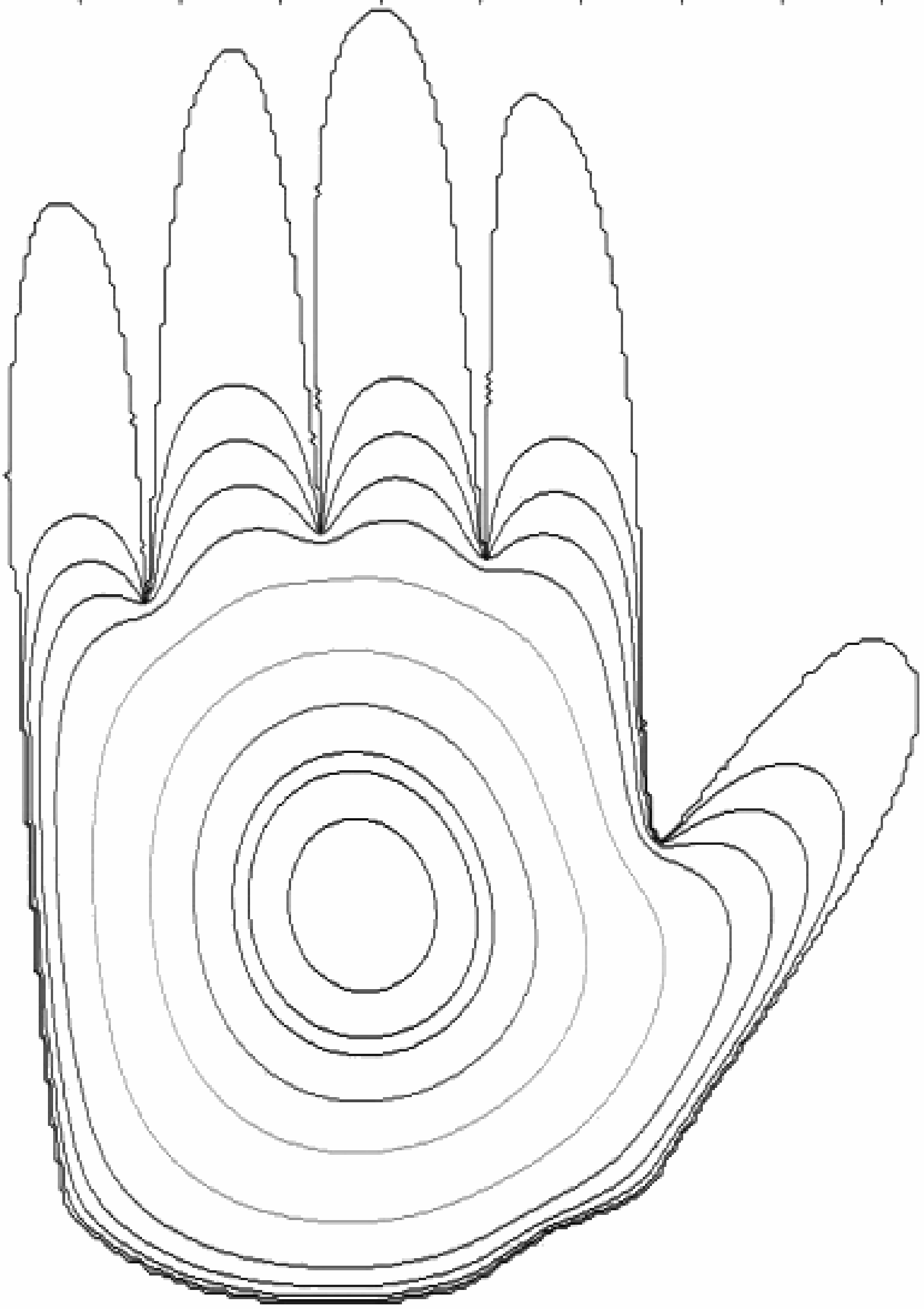} &
\includegraphics[height=4cm]{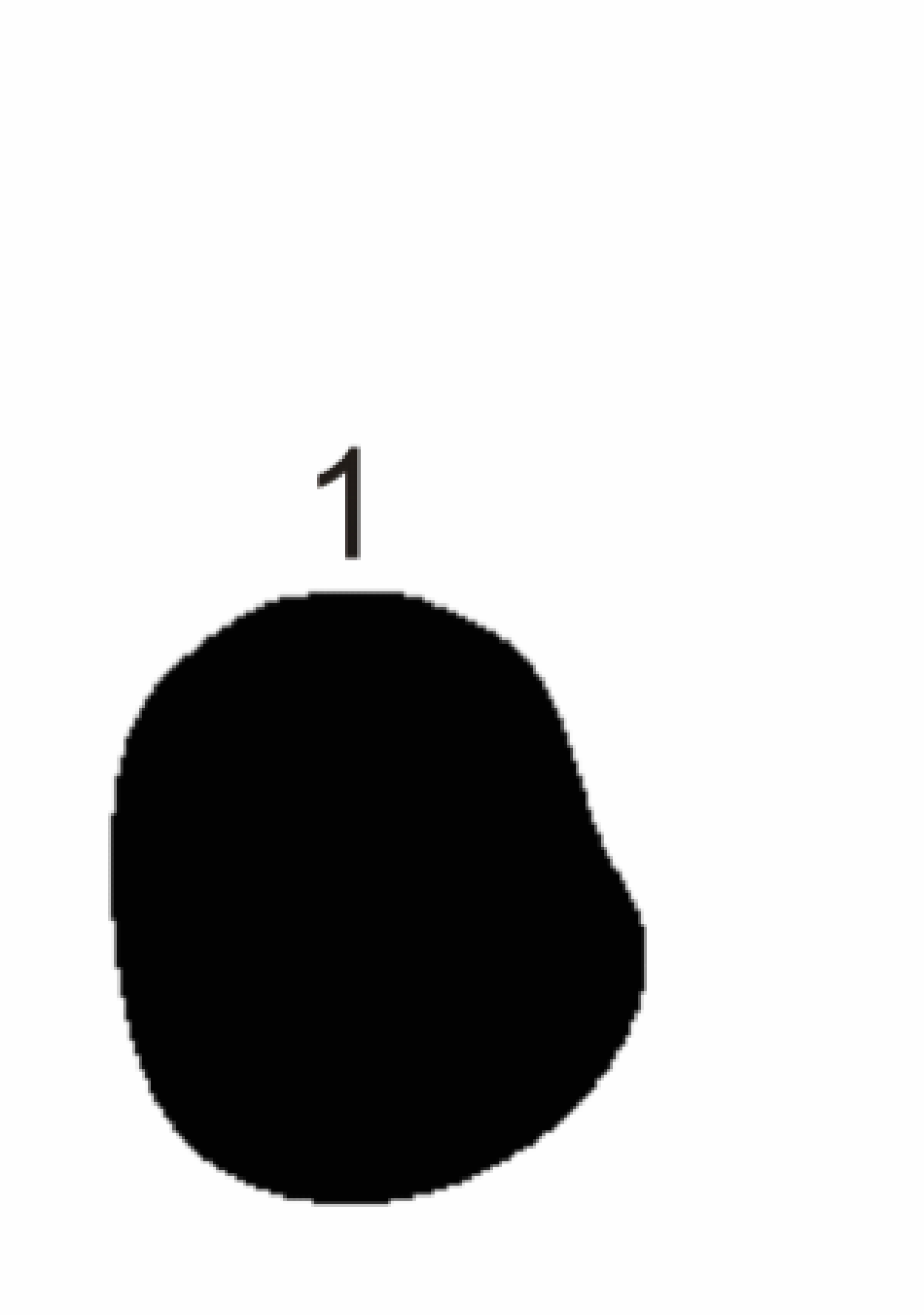} &
\includegraphics[height=4cm]{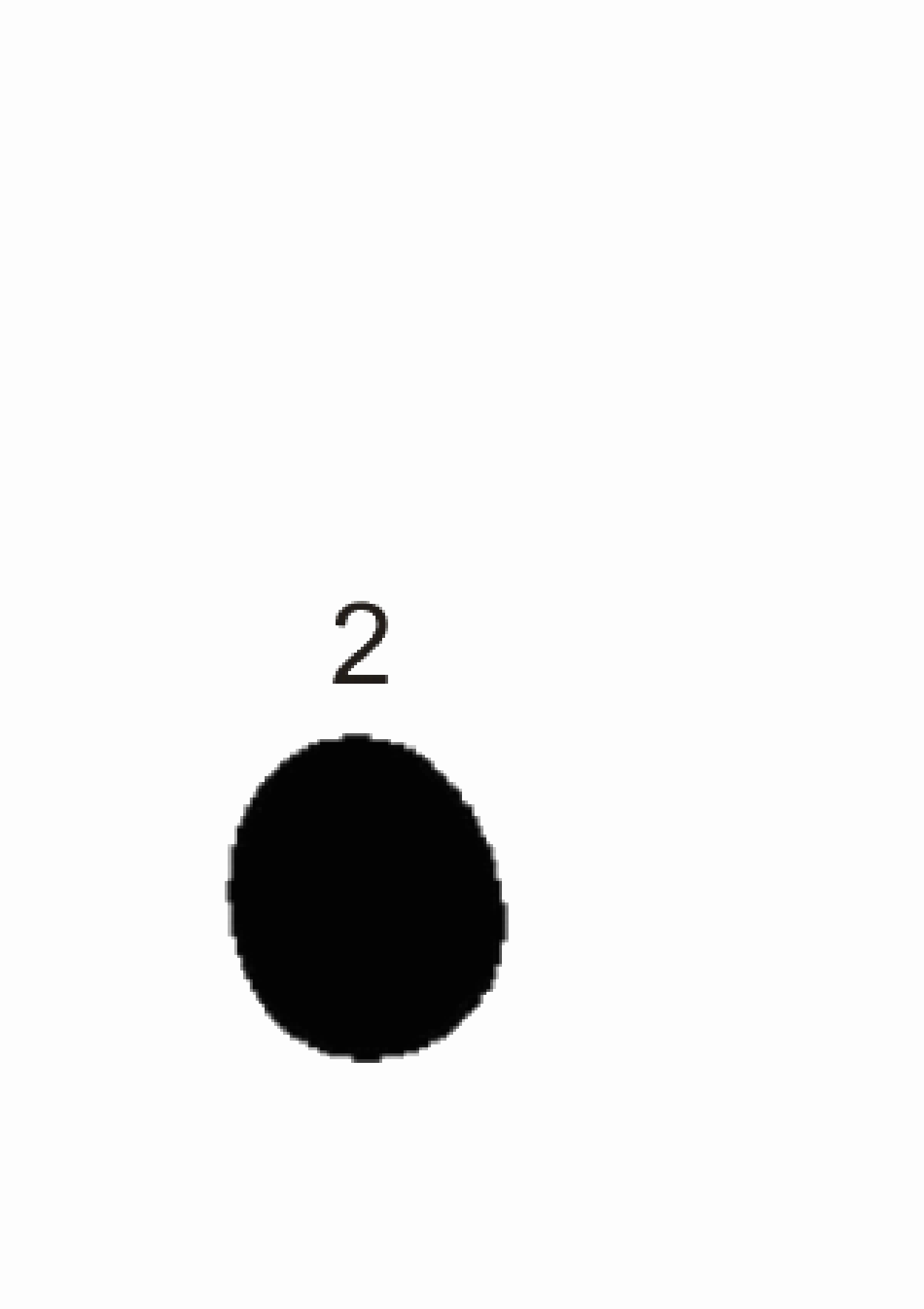} \\
(a) & (b) & (c) & (d)\\
\end{tabular}
\caption{(a) The symmetry branches of a hand shape (positive major
symmetry branches are shown in bold red and negative major symmetry
branches are shown in bold yellow). (b) The level curves of the
evolving shape boundary. (c) The state of the hand shape after three
branches corresponding to three fingers terminate. (d) The state of
the hand shape when all the branches except the major ones
terminate.} \label{fig:sym_hand}
\end{figure}

{A shape centered coordinate frame allows one to
describe the {\sl organization} of the primitives in a way that is
robust to changes in scale, rotation and articulation. This
coordinate frame can be formed by designating the prominent branches
of the shape as the reference axes.} Symmetry branches that survive
long enough to reach a minimum or a saddle point of the surface
$\phi$ represent the most prominent features of a shape and they are
called {\sl major branches}. There are always at least two positive
and two negative major branches \cite{Tari97}. An illustration using
the hand shape is given in Fig. \ref{fig:sym_hand}. Fig.
\ref{fig:sym_hand}(a) shows the symmetry branches of a hand shape.
The shape center is indicated with a blue dot. The branches marked
with red are the major positive branches whereas the ones marked
with yellow are the major negative branches. During the evolution of
the boundary curve of the hand shape, some positive symmetry
branches merge with the negative ones and terminate. The branch
disconnection points are marked with arrows. The level curves of the
evolving shape boundary and states of the shape after these branches
terminate are shown in Fig. \ref{fig:sym_hand}(b)-(d) respectively.
The blob in Fig. \ref{fig:sym_hand}(c) represents the state of the
hand shape after three branches corresponding to three fingers
terminate. Notice that the protrusion corresponding to the thumb is
still present. The branch corresponding to the thumb is more
prominent, and it terminates later in the process. The ellipse-like
blob in Fig. \ref{fig:sym_hand}(d) represents the state of the shape
when all the branches except the major ones terminate. The resulting
shape includes only the most significant branches, therefore, it can
be considered as the coarsest description of the original shape.
From that point, only the two positive branches and the two negative
branches continue towards the shape center and survive until the end
of the evolution.

When a  shape undergoes  non-rigid transformations such as
articulation and boundary perturbations, the coarsest structure
remains roughly the same.

\begin{figure}[!h]
\centering
\includegraphics[height=4cm]{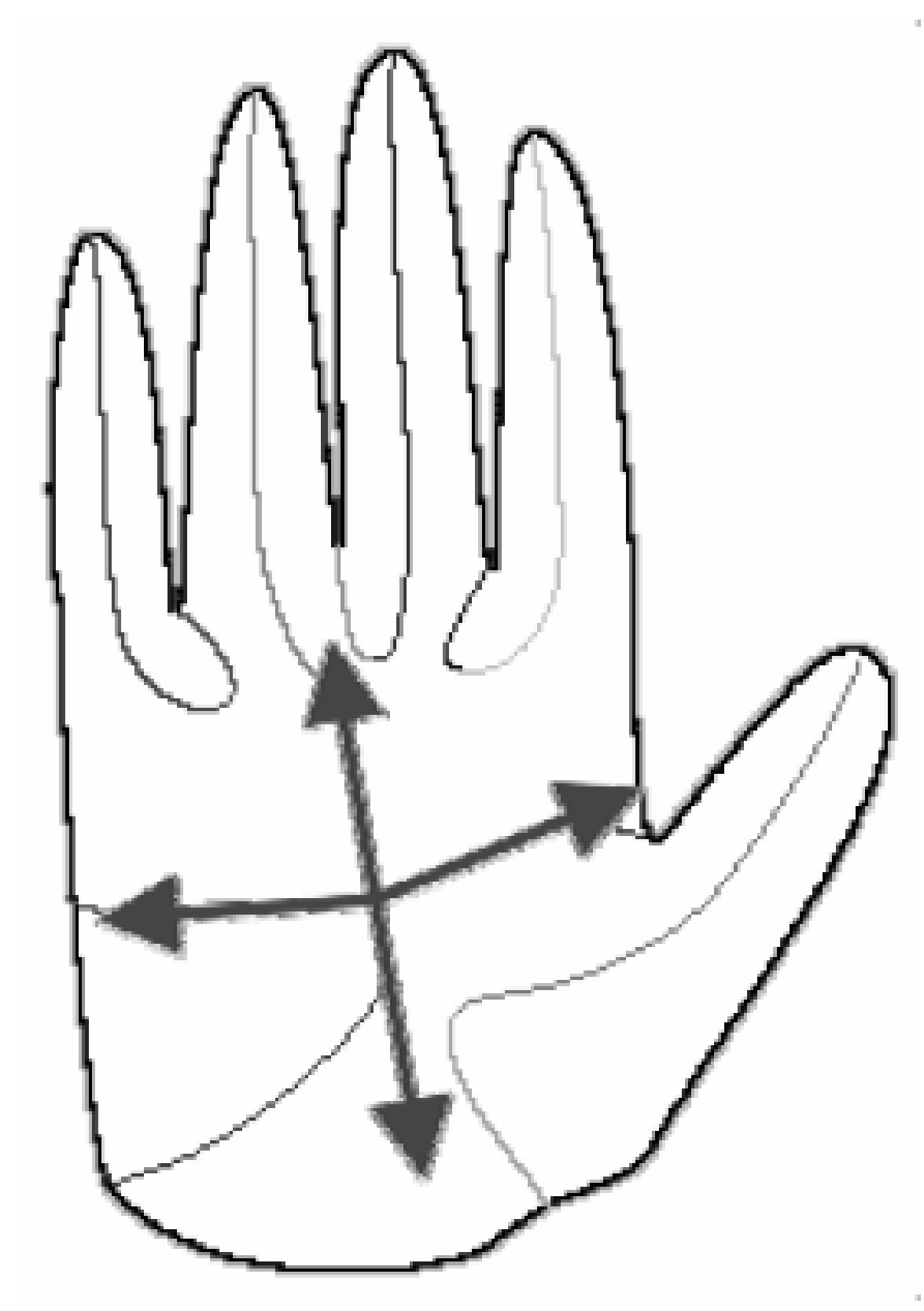}
\caption{Four possible reference axes of the hand shape.}
\label{fig:ref_hand}
\end{figure}

\begin{figure}[!h]
\centering
\begin{tabular}{ccc}
\includegraphics[height=2cm]{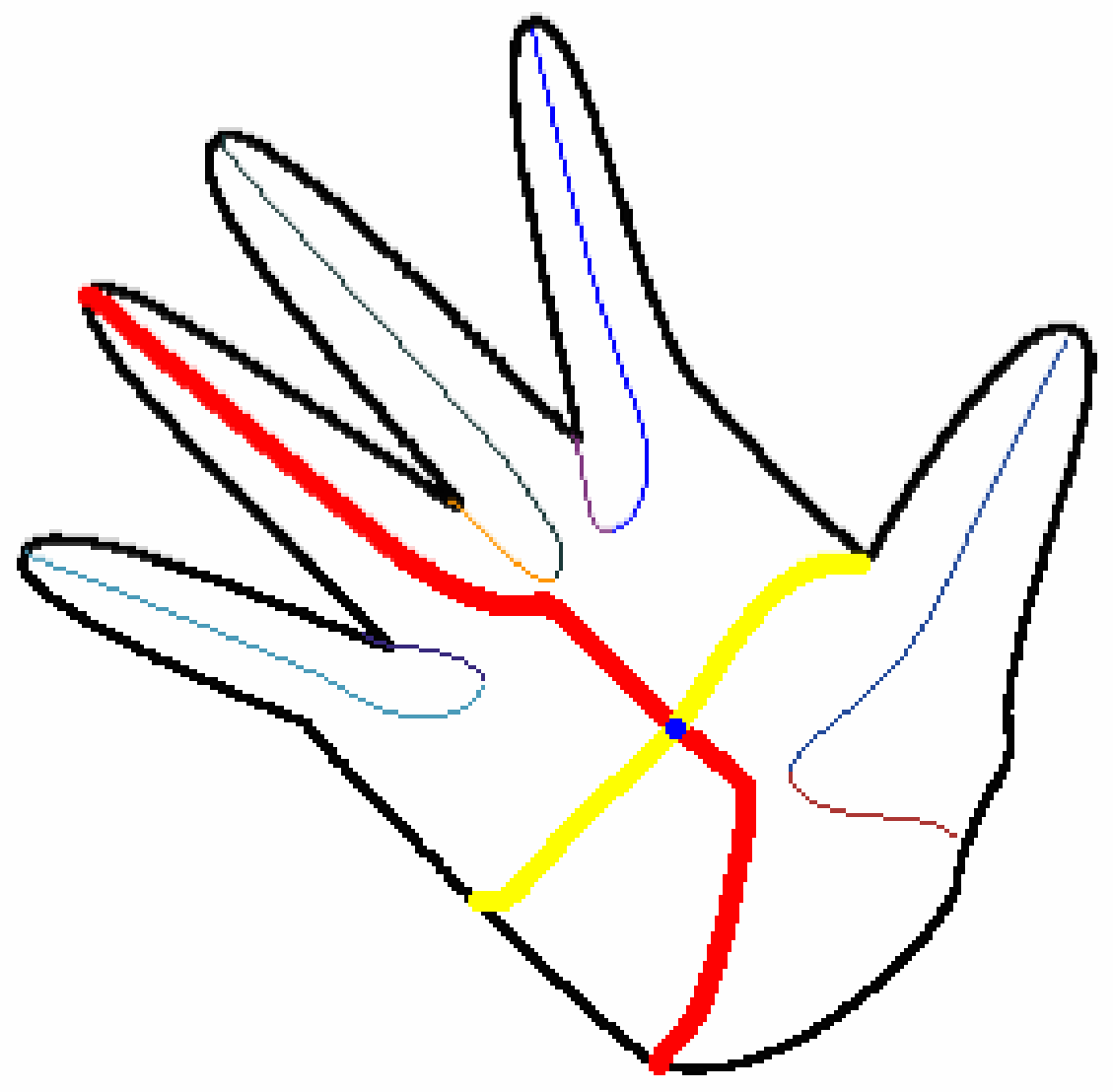} &
\includegraphics[height=1.75cm]{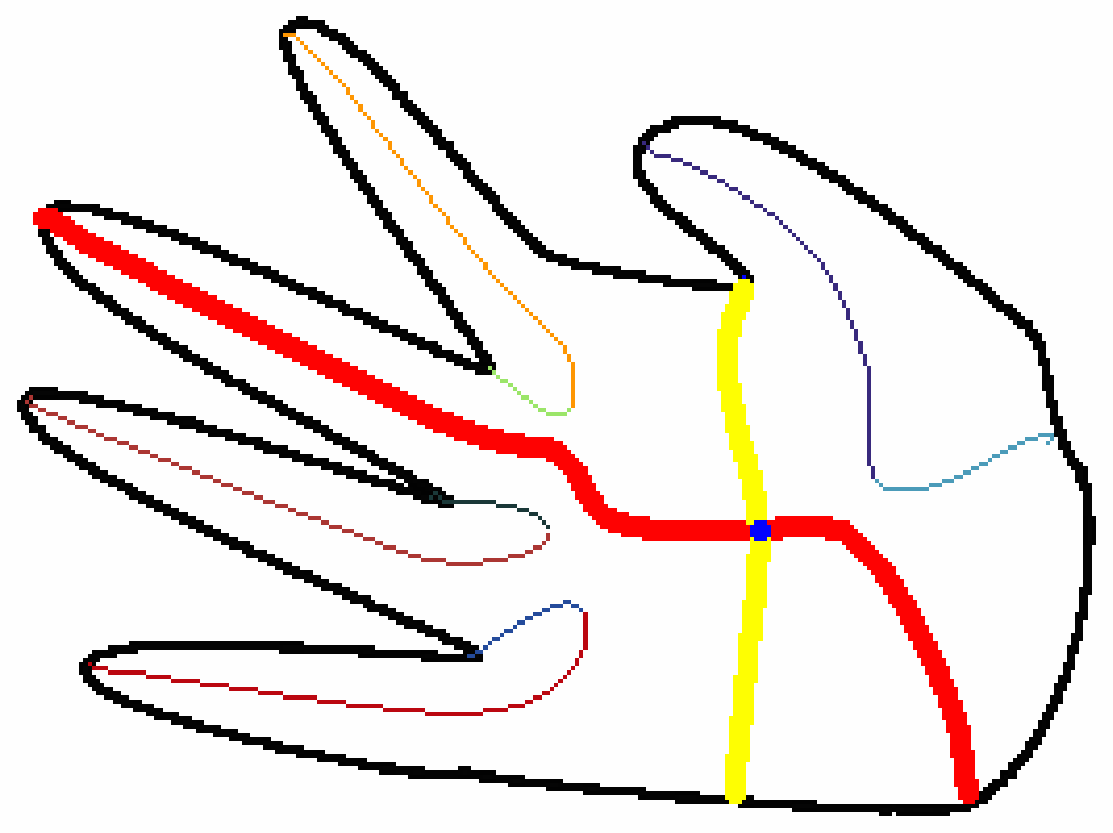} &
\includegraphics[height=2cm]{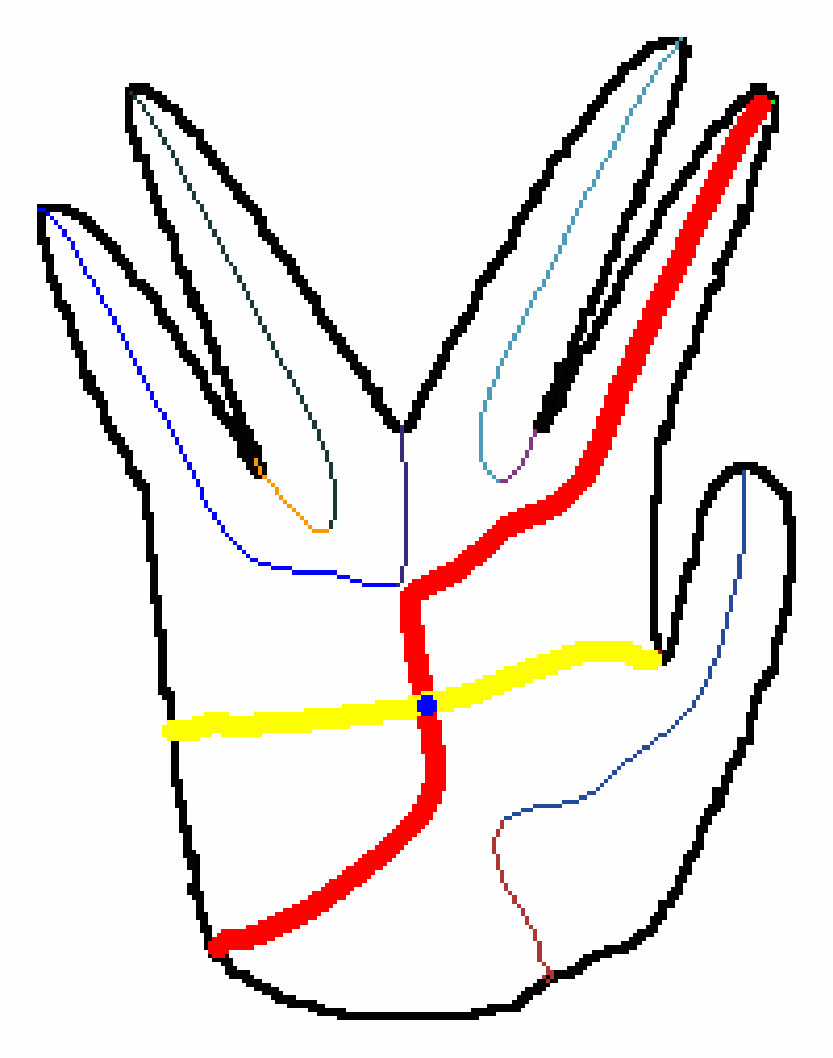} \\
\includegraphics[height=2cm]{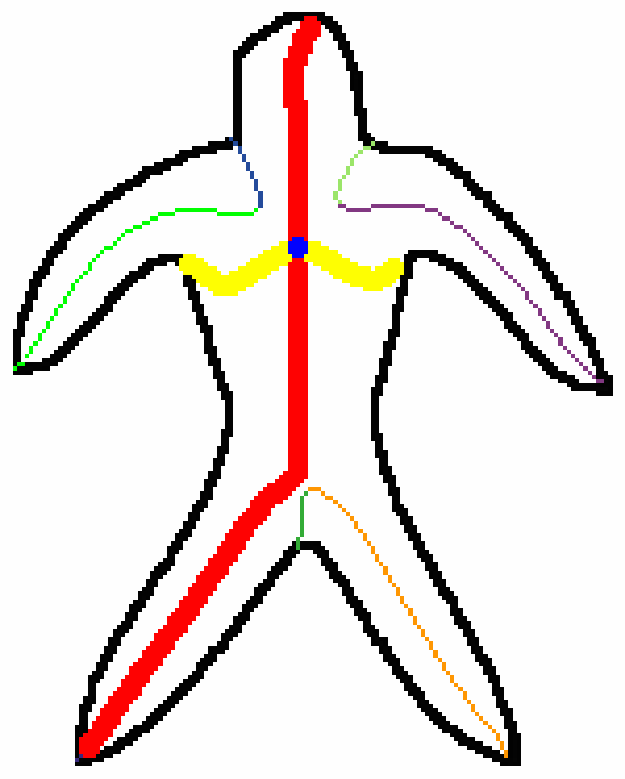} &
\includegraphics[height=2cm]{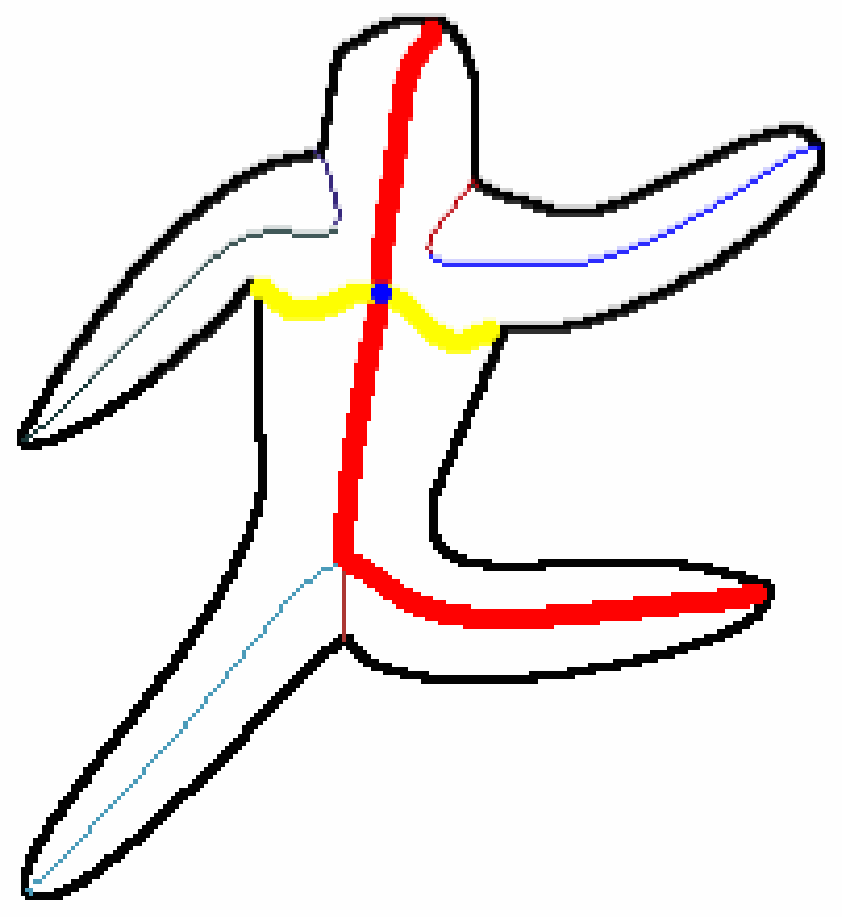} &
\includegraphics[height=2cm]{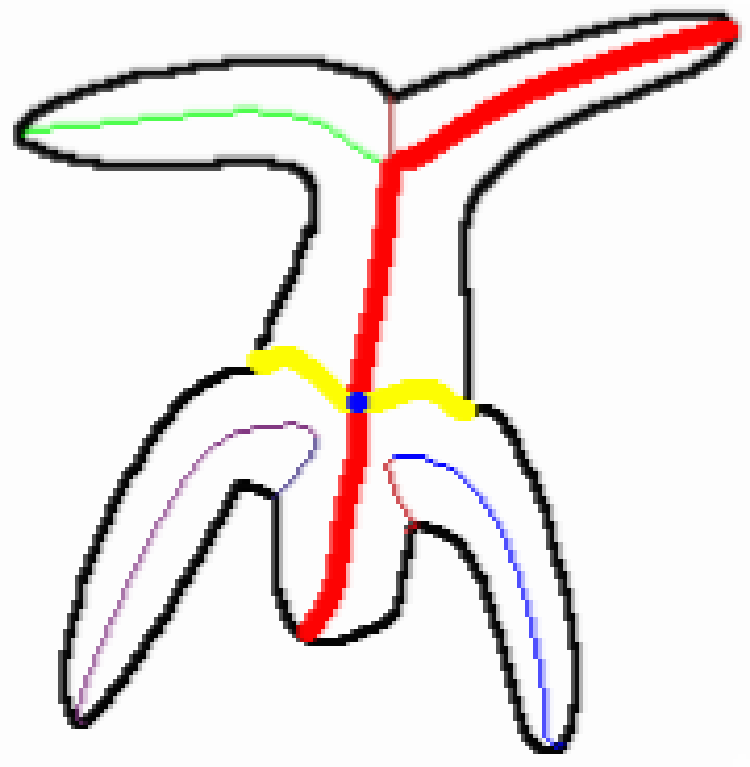}
\end{tabular}
\caption{The major branches of the hand and human shapes. Positive
major symmetry branches are shown in bold red and negative major
symmetry branches are shown in bold yellow.}
\label{fig:axes_hand_human}
\end{figure}

The center point and one of the major branches are used  to set up a
canonical coordinate frame (Fig. \ref{fig:ref_hand}). Any one of the
major branches may be selected. The line that connects the origin to
a nearby point on the selected major branch defines the reference
axis. This point on the major branch should be chosen within the
ellipse-like blob (e.g. Fig. \ref{fig:sym_hand}(d)), representing
the coarsest form, because the major branches may bend as one moves
away from this region (Fig. \ref{fig:axes_hand_human}).

No matter which major branch is chosen as a reference axis, the same
axis must be chosen for similar shapes. Since there are two major
axes of the same type, there is an ambiguity in the process. If the
descriptions of two similar shapes depend on different coordinate
frames, the matching algorithm will be unable to determine the
similarities of shapes. This situation may necessitate creating at
least two descriptions.

To overcome this problem, we use each negative major symmetry branch
as a reference axis for one half of the shape. Hence, the shape is
described as two halves with each half having its own coordinate
frame (Section \ref{ssec:org}). This approach decreases the
computation time of the matching algorithm drastically.

For a dumbbell-like shape, any one of the three surface extrema may
be chosen as the origin. The fact that each hyperbolic point of the
surface has at least two positive symmetry axes with negative
curvature \cite{Tari97} removes this ambiguity (Fig.
\ref{fig:pos_ref}(a)).
\subsection{Handling Ambiguities}
\label{sec:amb} 
\begin{figure}[!h]
\centering
\begin{tabular}{cccc}
\includegraphics[height=1.75cm]{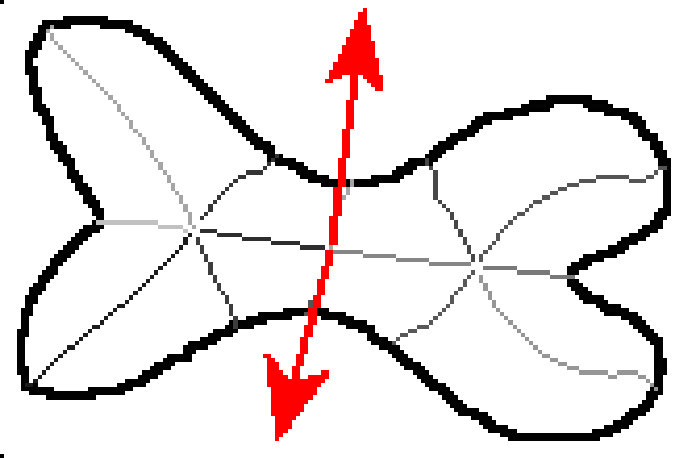} &
\includegraphics[height=2.5cm]{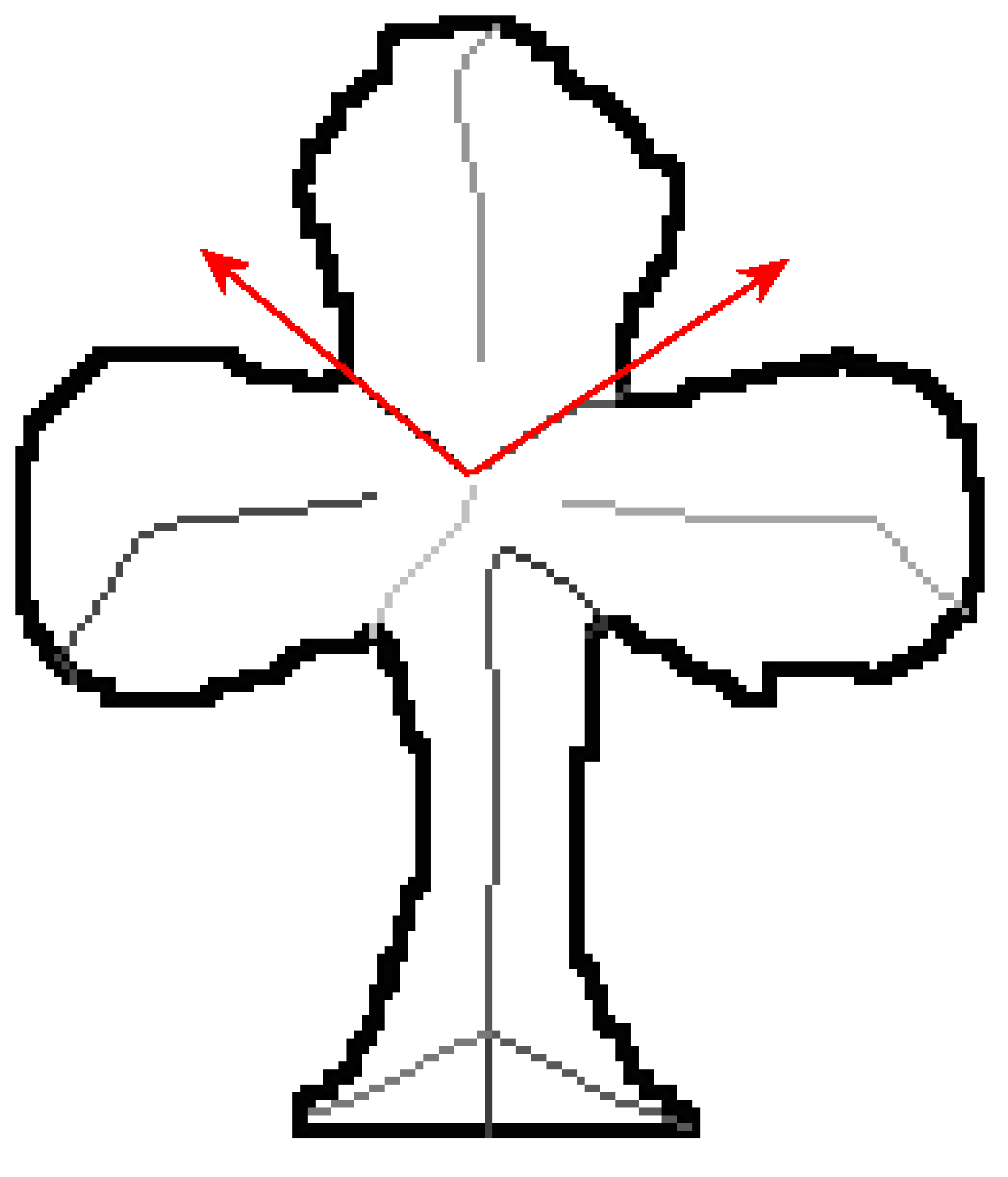} &
\includegraphics[height=2.5cm]{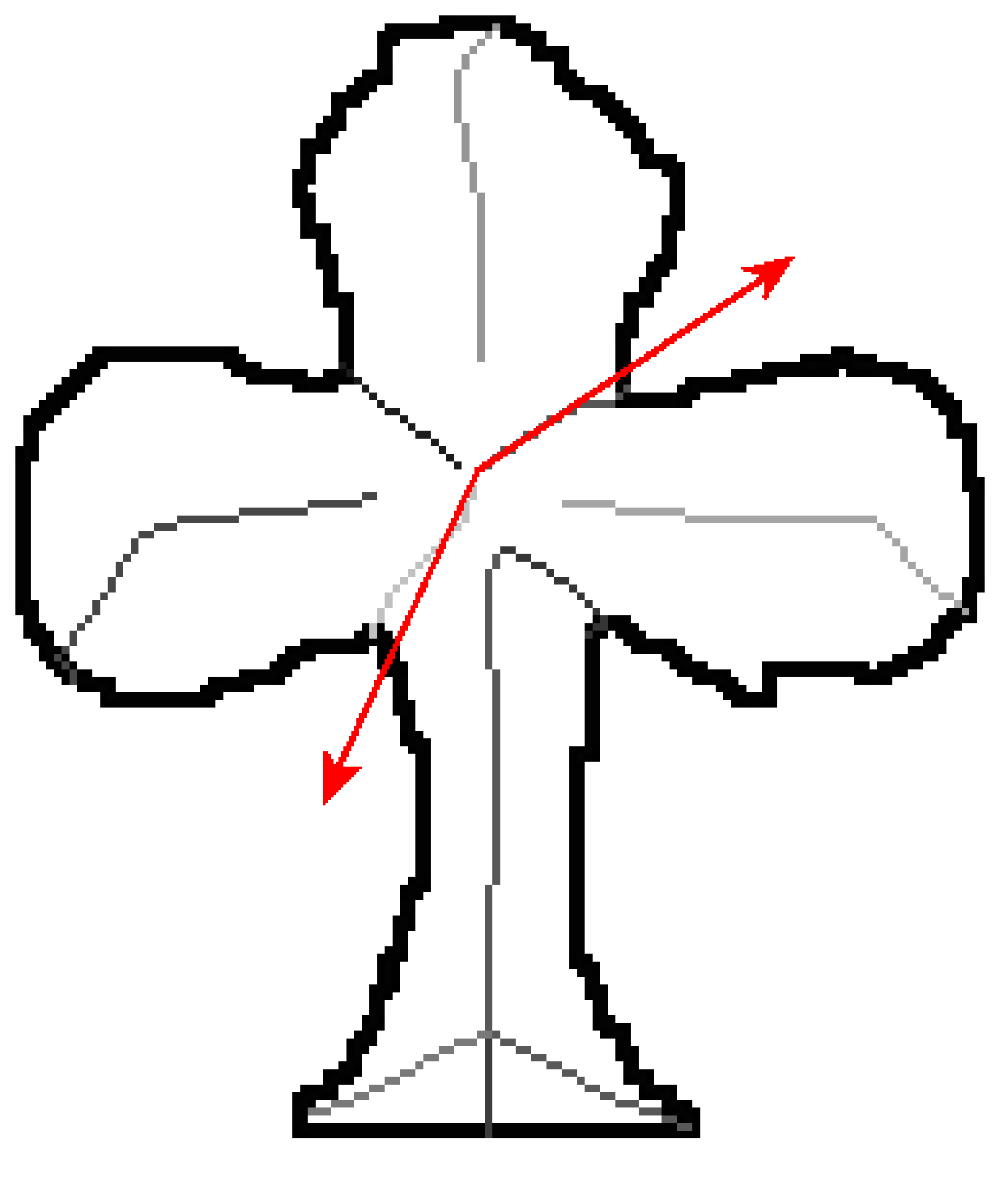} &
\includegraphics[height=2.5cm]{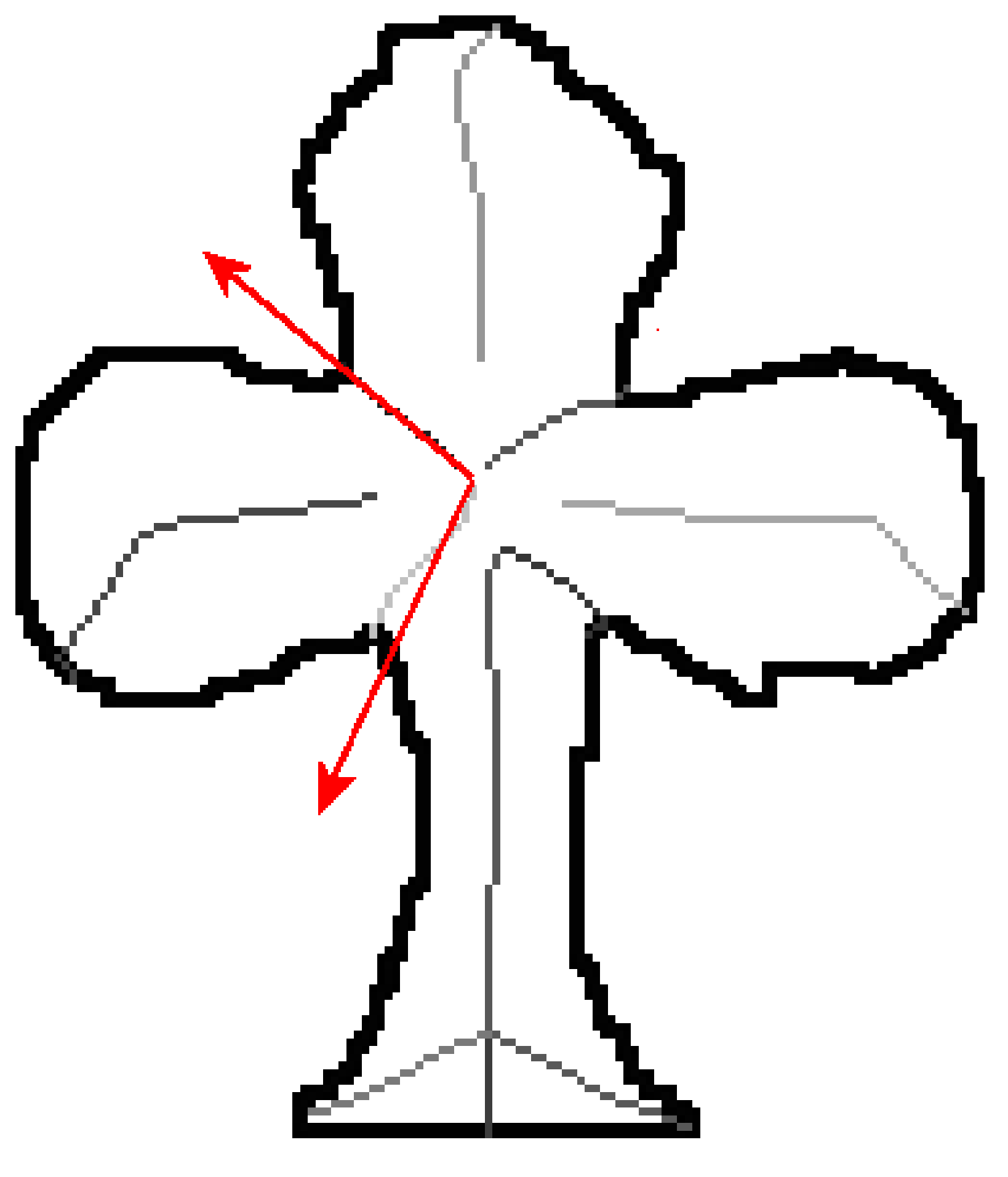} \\
(a) & & (b) &
\end{tabular}
\caption{(a) Reference axes for a dog-bone shape. (b) Possible
reference axes of a shape in which three negative branches meet at
the shape center. The three possible descriptions of this shape are
used when the shape is matched against the other shapes.}
\label{fig:pos_ref}
\end{figure}

\begin{figure} [!h]
\centering
\begin {tabular}{ccccc}
\epsfig{figure=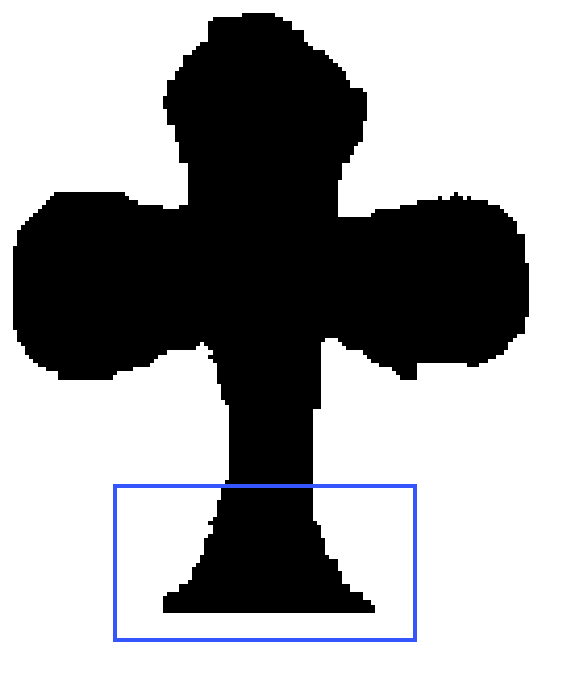, width=2cm} &
\epsfig{figure=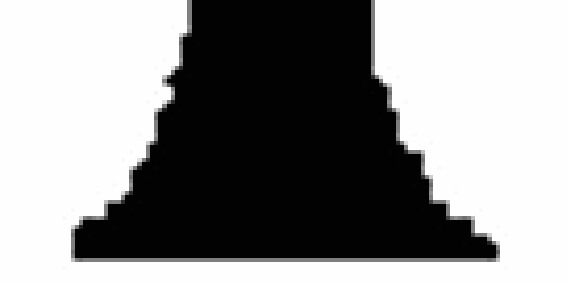, width=1.9cm} &
\epsfig{figure=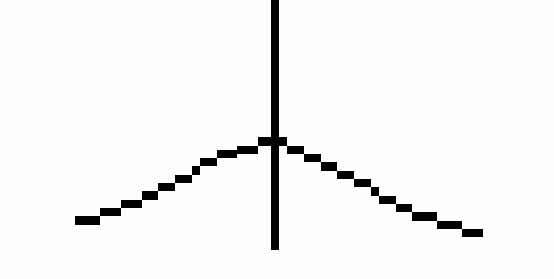, width=1.9cm} &
\epsfig{figure=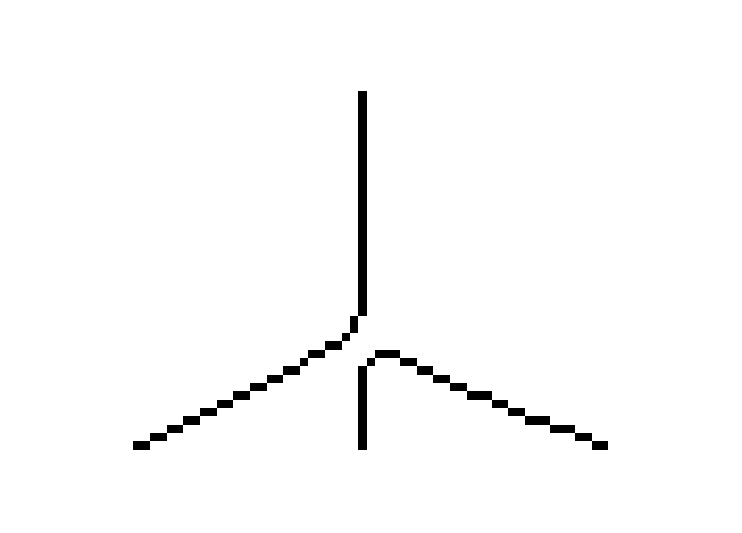, width=2cm} &
\epsfig{figure=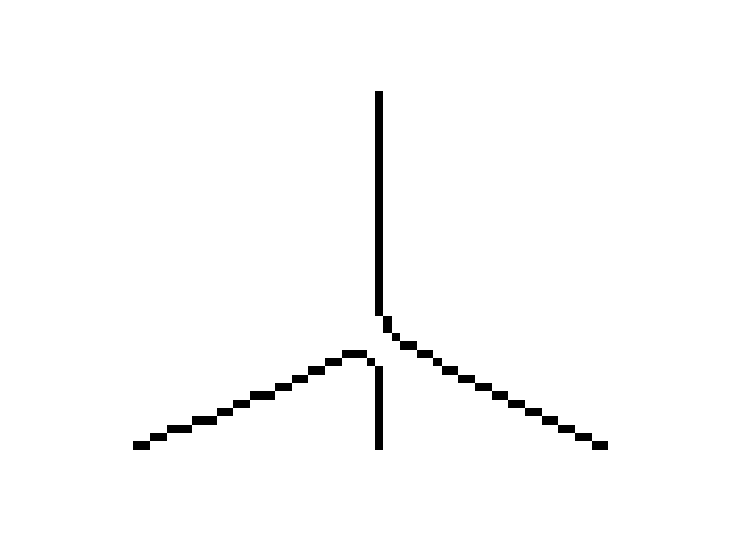, width=2cm}\\
(a) & (b) & (c) & (d) & (e)\\
\end{tabular}
\caption{Triple junction example. (a) Shape with mirror symmetry.
(b) Part of the shape. (c) Part of the symmetry axis: Intersecting
branches at junction point. (d)-(e) Skeletal configurations which
arise as a result of slight deviation from mirror symmetry.}
\label{fig:junct_1}
\end{figure}

When the symmetry of a shape is more than two-fold,
obtaining the description of the shape is more complicated because
there are more than two negative branches reaching the shape center.
We have to guarantee that the same coordinate frame is formed for
those shapes. A simple solution is to interpret these situations as
the ambiguities of the representation and to generate a number of
possible descriptions. If there are $n$ major branches that reach
the shape center, we select all the two permutations of those major
branches to generate possible descriptions (Fig.
\ref{fig:pos_ref}(b)). This redundancy of descriptions may not incur
high computational penalties in the matching process because it
appears to us that high order perfect symmetries (e.g. four-fold)
are encountered rarely.
{ One final remark} is that triple junctions rarely occur due to
discretization issues. However, these occasional cases obviously
violate the simplicity of the primitives. Therefore, they must be
handled accordingly. Our approach is to transform the triple
junction configuration into two stable alternative forms and keep
both forms as alternative descriptions of the same shape (Fig.
\ref{fig:junct_1}).

\subsection{Spatial Organization of the Symmetry Branches}
\label{ssec:org}Once the coordinate frame is set up, it is easy to
describe the relative placement of the symmetry branches. Each
symmetry branch is represented by a vector from the origin to the
disconnection point. We emphasize that the disconnection points are
used because these are the points where the limbs connect to the
body, and they are robust with {respect to an articulated motion}.

\begin{figure}[!h]
\centering
\includegraphics[height=4cm]{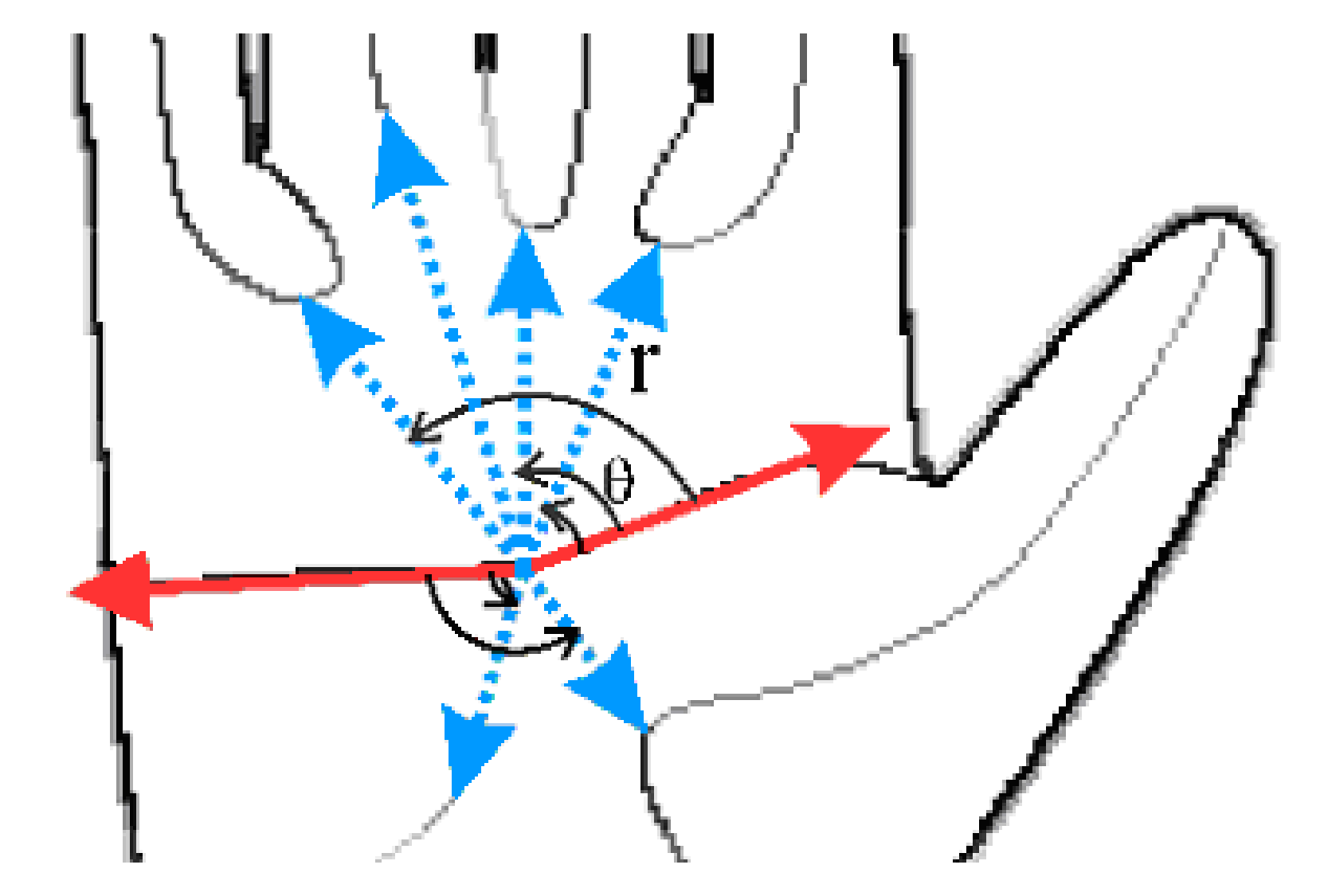}
\caption{The reference axes (red) and the position vectors (blue) of
the symmetry axes of the hand shape.} \label{fig:ref_axes}
\end{figure}

\begin{figure}[h!]
\centering
\begin{tabular}{ccc}
\includegraphics[height=3cm]{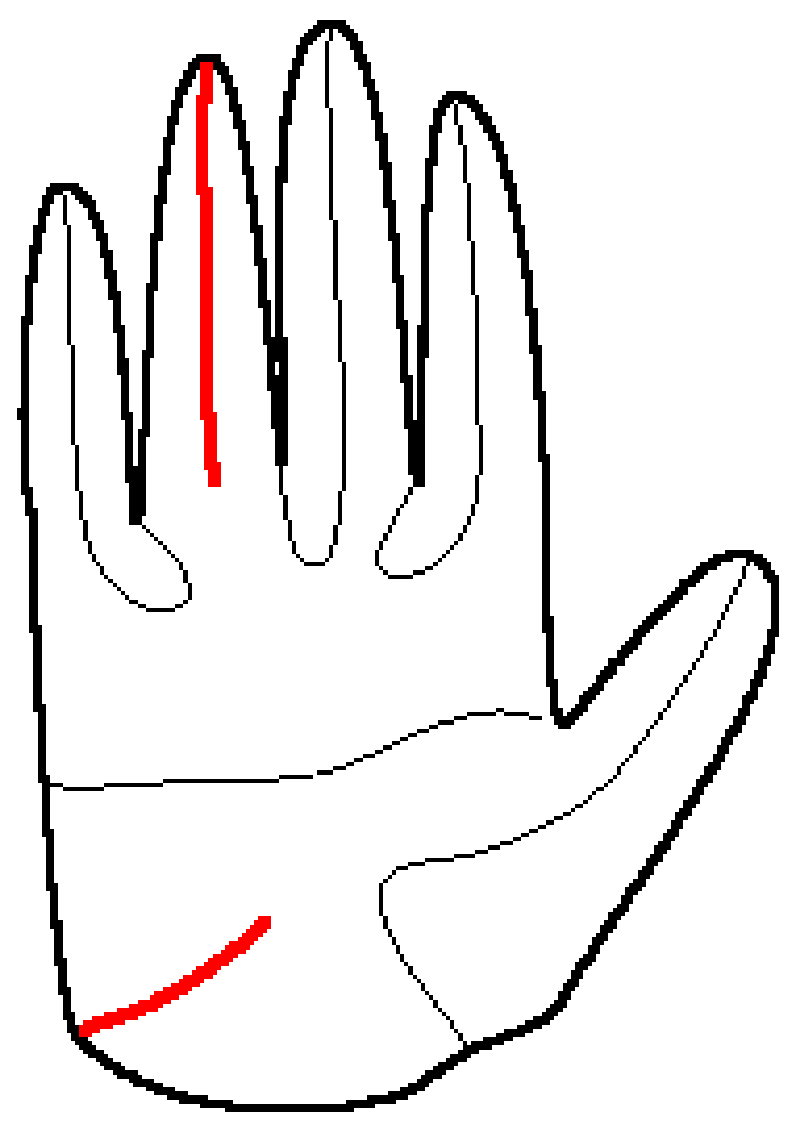} &
\includegraphics[height=3cm]{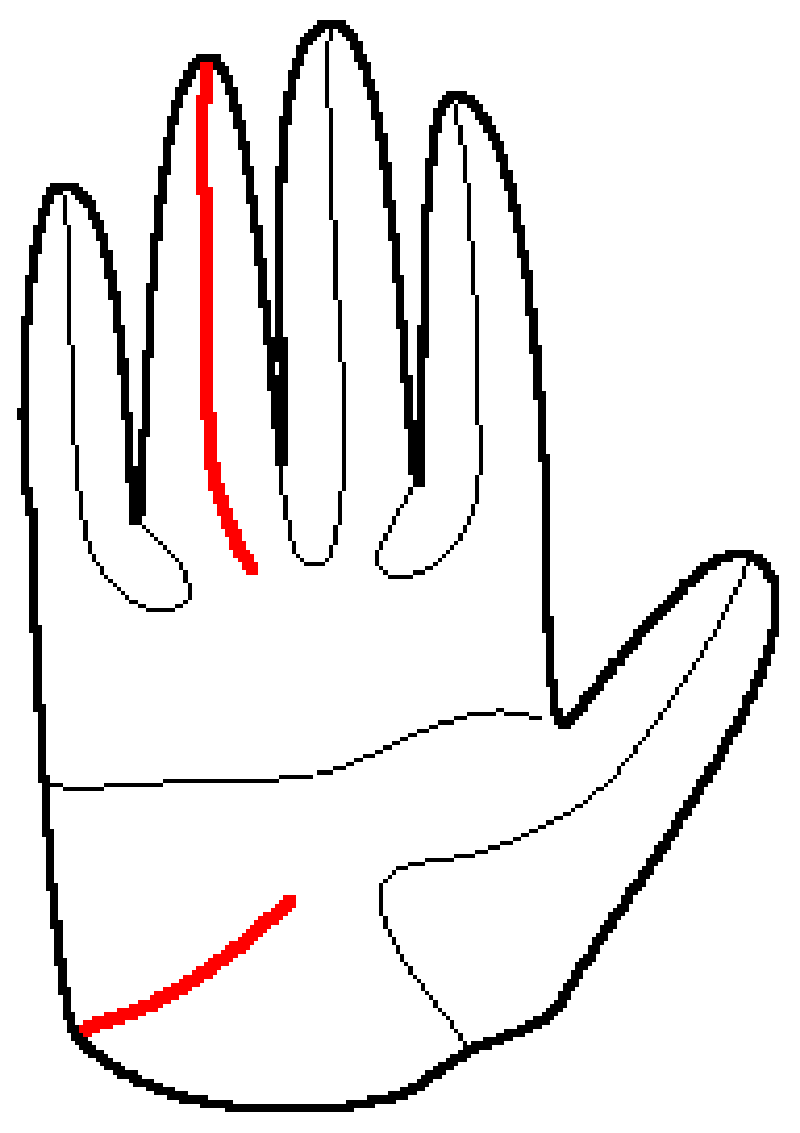} &
\includegraphics[height=2.25cm]{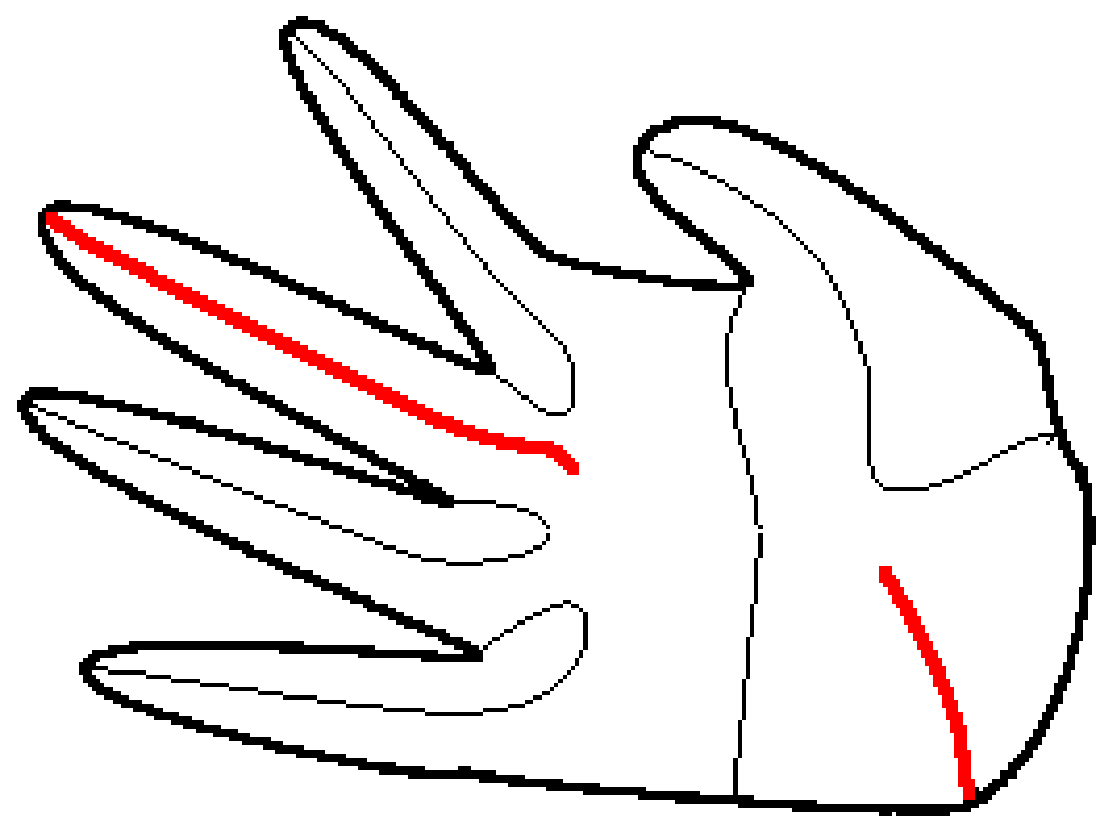} \\
(a) & (b) & (c)
\end{tabular}
\caption{Artificially defined disconnection points for major
positive axes. Compare the representations in (a)-(b) to the one in
Fig. \ref{fig:sym_hand}(a).} \label{fig:mcut}
\end{figure}

\begin{figure}[h!]
\centering
\begin{tabular}{ccc}
\includegraphics[height=3cm]{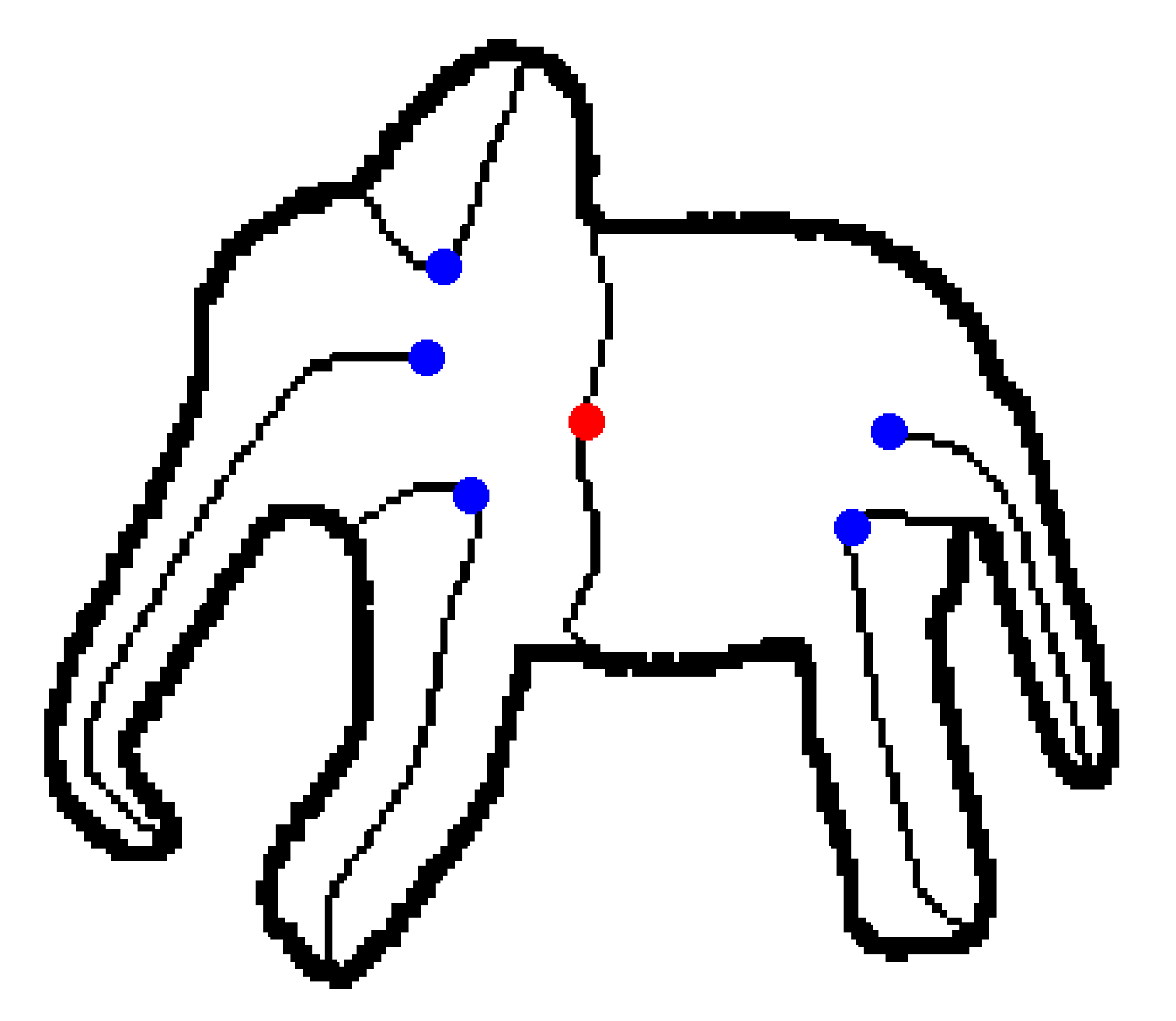} &
\includegraphics[height=3cm]{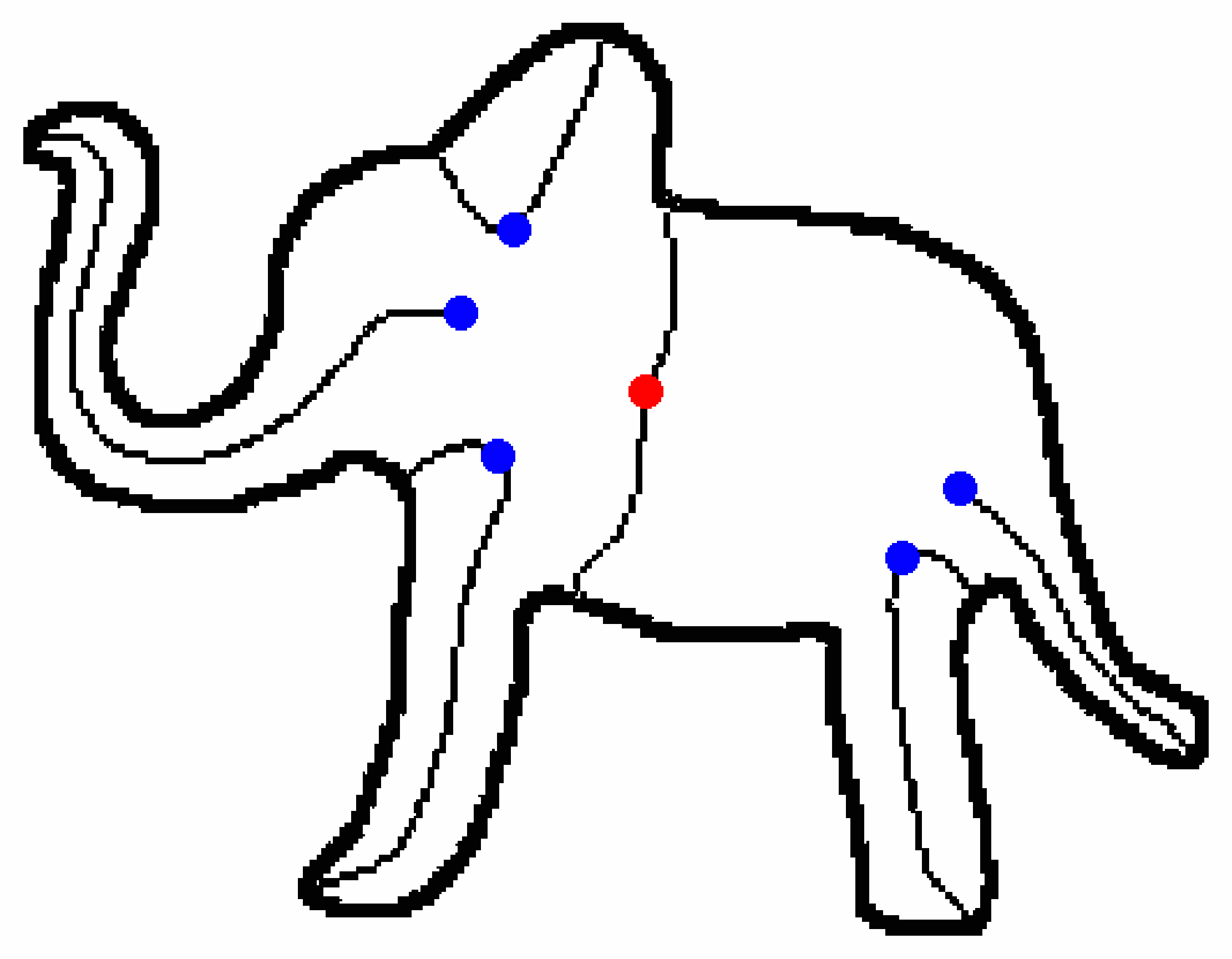} &
\includegraphics[height=3cm]{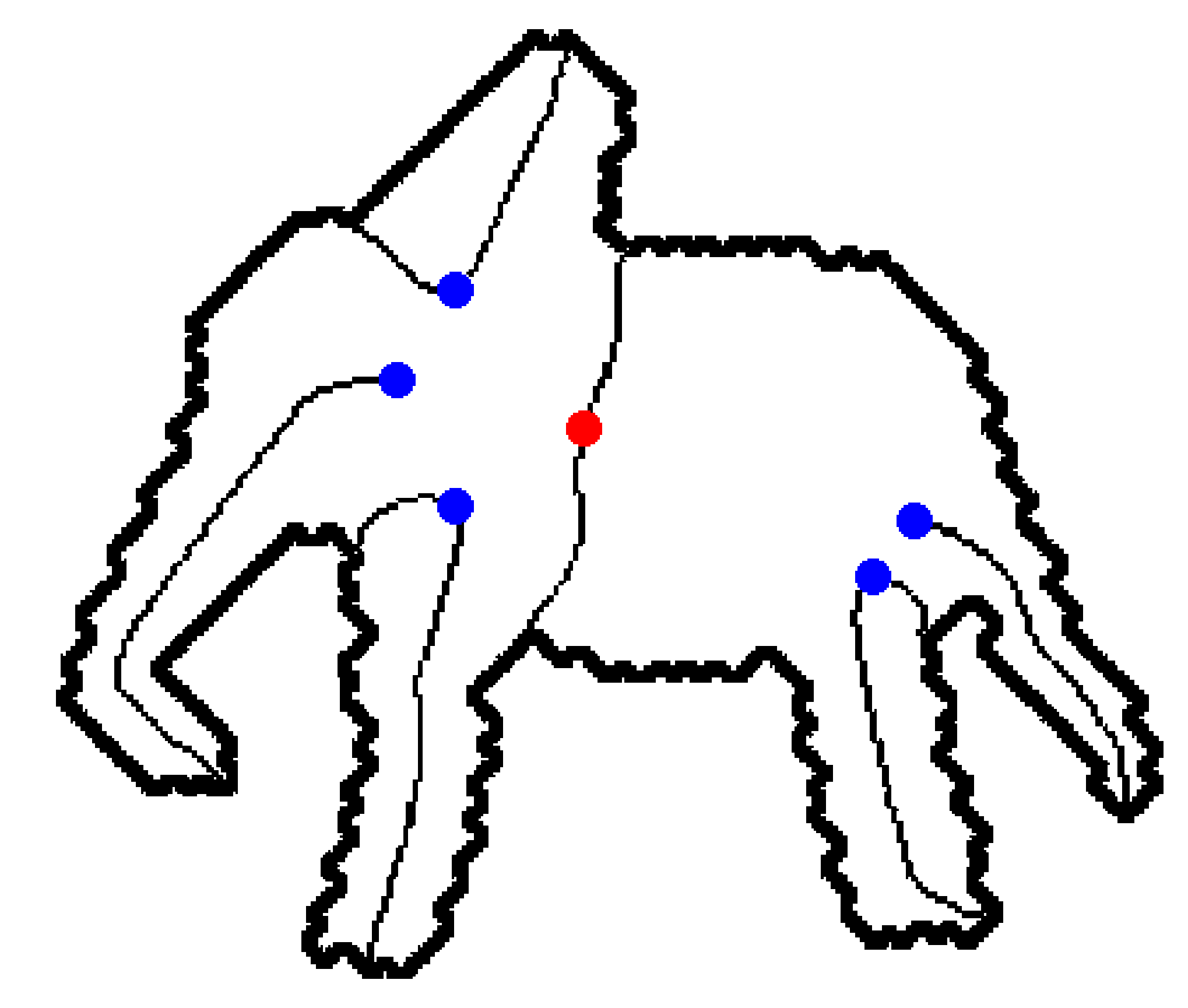} \\
(a) & (b) & (c)
\end{tabular}
\caption{The disconnected skeleton for some elephant shapes. Note
that a quadruple is associated with each disconnection point
indicating the sign, location, and length in the global frame.}
\label{fig:elephantdis}
\end{figure}

The construction is depicted in Fig. \ref{fig:ref_axes}. The dotted
arrows denote the disconnection points.  The solid arrows are the
reference axes. The length of the vector defines $r$; the angle
between the vector and a reference axis defines $\theta$, providing
a polar representation. There are two reference axes. Considering a
counter-clockwise ordering, the upper half  is represented in the
coordinate frame formed by the arrow pointing towards the thumb. The
lower half is represented in the coordinate frame formed by the
other arrow. The representation allows us to infer the order of
symmetry branches along the shape boundary and to use this as a
constraint in the matching process.
A critical practical issue is that the termination points of the
major positive branches are the origin. Hence, one should
artificially define a disconnection point for the positive major
branches. One alternative is to cut the branch by intersecting it
with a line connecting the two nearest indentations indicated by the
start point of the negative symmetry branches (Fig.
\ref{fig:mcut}(a)). Another alternative is to cut the branch at the
location where the $\phi$ value is equivalent to the $\phi$ values
of the disconnection points of the neighboring branches (Fig.
\ref{fig:mcut}(b)). The second alternative is {more robust in the
sense that} if a major branch jumps from one finger to another, the
disconnected skeleton representation does not change (Fig.
\ref{fig:mcut}(c)).
Disconnected skeletons for elephant shapes are shown in Fig.
\ref{fig:elephantdis}. Each disconnection location indicated by a
dot is associated with a quadruple indicating the branch length $l$,
the location $(r,\theta)$ measured in the global frame and the sign
of the branch.

The information stored in the  shape descriptions is listed in
Table~\ref{table:sm}.  Recall that  in a certain context, invariance
may not be desired.  Therefore, we store the extrinsic coordinates
of the center point (for translation), the total length of symmetry
axes (for scale) and the orientation of the reference axes in the 2D
image plane (for rotation).
\begin{table}  \caption{Information stored in the descriptions}
\label{Table1}
 { \begin{center}
\begin{tabular}{ll}
\hline
Description element & Information stored \\
\hline
Shape & Center point ($x_0,y_0$)\\
& Total length of the axes\\
& Orientation of the reference axes ($m_0$, $m_1$)\\
& \\
Local Symmetry Branch & Type (Positive, Negative) \\
& Location ($r$, $\theta$)\\
& Normalized Length\\
& Reference Axis (Yes, No)\\
& Next Symmetry Axis\\
& Previous Symmetry Axis\\
\hline
\end{tabular}
\end{center} \label{table:sm}}
\end{table}
\section{Shape Matching and Similarity}
\label{sec:smas} Shape matching is the stage where the best
correspondence of the primitives forming two shapes is determined.
 The similarity scores between a shape and a selected set of shapes
 may be used as identifying features~\cite{Basri93,Edelman98}. When
skeletons are used for shape matching and recognition, the common
paradigm is to convert the skeletal description to a graph or a tree
and reduce the problem to the matching of these structures. Existing
methods mainly differ from each other by the distance measures they
use to compute the similarities between representation primitives
and by the graph (or tree) matching algorithms they employ, e.g.
\cite{Geiger03,Pelillo99,Sebastian04,Siddiqi99,Zhu96}. An
interesting idea in Zhu and Yuille \cite{Zhu96} is the generation of
more than one possible  graph for the input shape to overcome the
unreliability of the skeleton description. Even though the
approaches based on connected skeletons are successful to some
extent, the instabilities of the representations lower their
performance. In addition, the complexity of the descriptions  leads
to computationally expensive matching and recognition algorithms.
Rich descriptions may be suitable for reconstructing a shape, but
may not be necessary for recognition. In fact, weakly informative
secondary details with numerical inaccuracies may hamper recognition
or  classification performance.

{Recall that we describe each primitive using the set of attributes
given in Table~\ref{table:sm}. The differences between these
attributes are used to compute the dissimilarity between two
primitives.}
 A normalized similarity scale that
varies between $0$ and $1$ is used, with $1$ indicating that the two
branches are identical. Let $(l_1,r_1,\theta_1,type_1)$ be the
attributes of the first branch and let $(l_2,r_2,\theta_2,type_2)$
be the attributes of the second branch. If $type_1$ and $type_2$ are
identical, then the similarity between these two branches is
computed by using
\[ sim(l_1,r_1,\theta_1,type_1;l_2,r_2,\theta_2,type_2)=e^{-0.5[l_1-l_2, r_1-r_2,
\theta_1-\theta_2] \Sigma^{-1} [l_1-l_2, r_1-r_2,
\theta_1-\theta_2]^T} \] where $\Sigma$ is a diagonal weight matrix
storing the importance of each attribute.
For simplicity, we compare each matching pair independently and
obtain a similarity score based on the averaging of the similarity
scores of the matched pairs. The order of the branches along the
shape boundary is also stored in the description. It is used to sort
out impossible correspondences in the matching process. This reduces
computation time and leads to perceptually more accurate matchings.

In a recognition process, where a query shape is compared with all
the shapes in a catalogue, determining the most similar shape is not
enough. To make a distinction between known shapes and shapes that
are encountered for the first time, the question of whether the two
 compared shapes are similar must be answered. For that purpose, the
similarity scores determined by the matching process must be {\sl
absolute}. The total similarity of two shapes is determined by the
weighted sum of matched branch pairs where the weights are the
normalized lengths of the branches. As argued in Section
\ref{ssec:spi}, the relative importance of protrusions and
indentations is reflected by the normalized branch lengths
accurately as a result of excessive smoothing. Hence, an accurate
probabilistic measure of similarity is obtained. This measure is
 asymmetric. In order to retain symmetry, the total similarity value
is calculated by using both the weights of the first shape's
primitives and the second shape's primitives. The lower one of these
two similarity values is selected.

The matching process is a branch and bound algorithm that searches
over all possible matchings of two shapes. The worst case complexity
of this type of algorithm is high; however, in practice, the
matching process is very fast. The number of shape primitives is
small and additional measures are employed to reduce the number of
permutations that need to be tested. Those matchings that would
violate the order constraint are not tested. The generation of a
permutation is stopped when it is determined that the current branch
of computation will not be able to produce a higher similarity value
than the current maximum. {Finally,} the representation of the shape
as two halves in two different coordinate frames makes it possible
to reduce the problem into two half problems providing a drastic
decrease in computation time.

\subsection{Examples}
\label{examples} 
\begin{figure}[!h]
\centering
\begin{tabular}{cc}
\includegraphics[height=2.5cm]{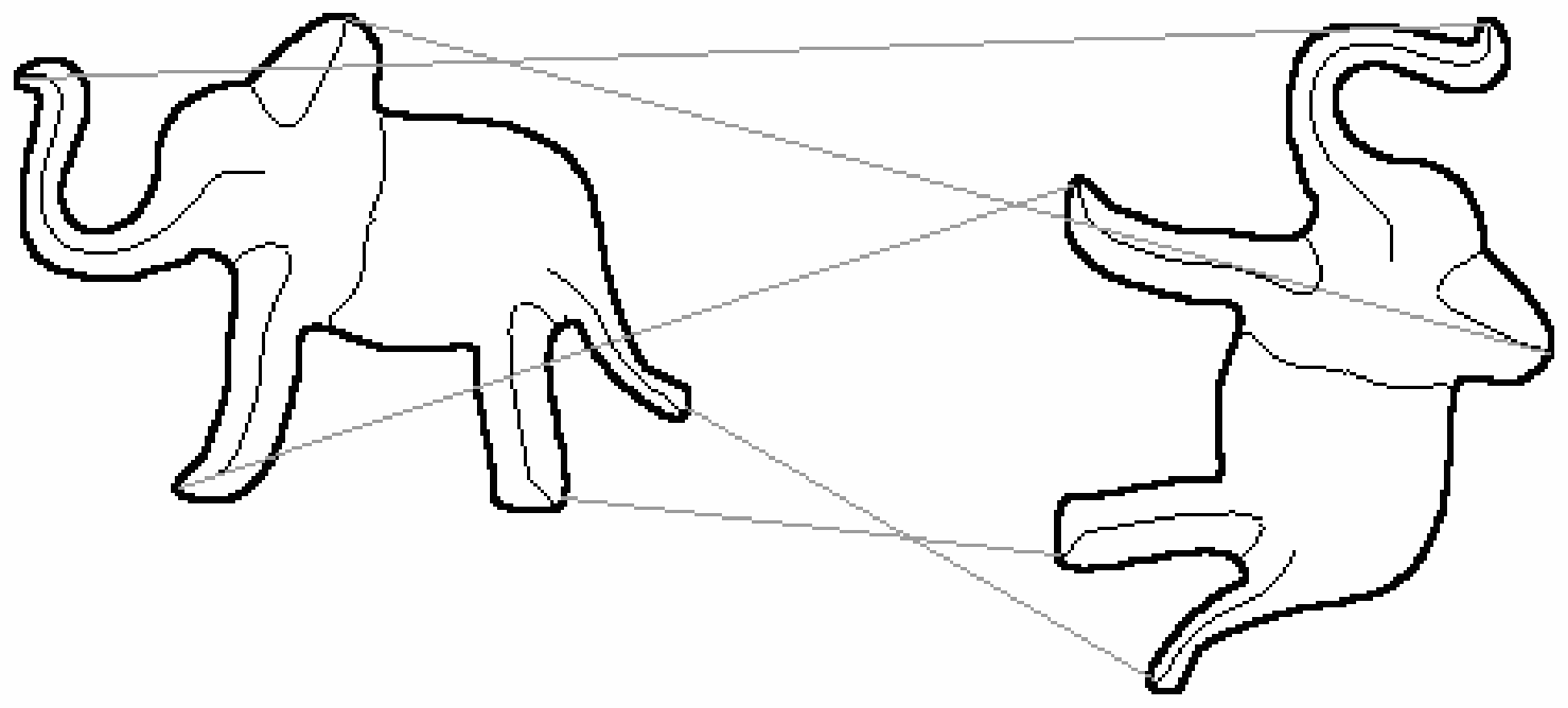} &
\includegraphics[height=2.5cm]{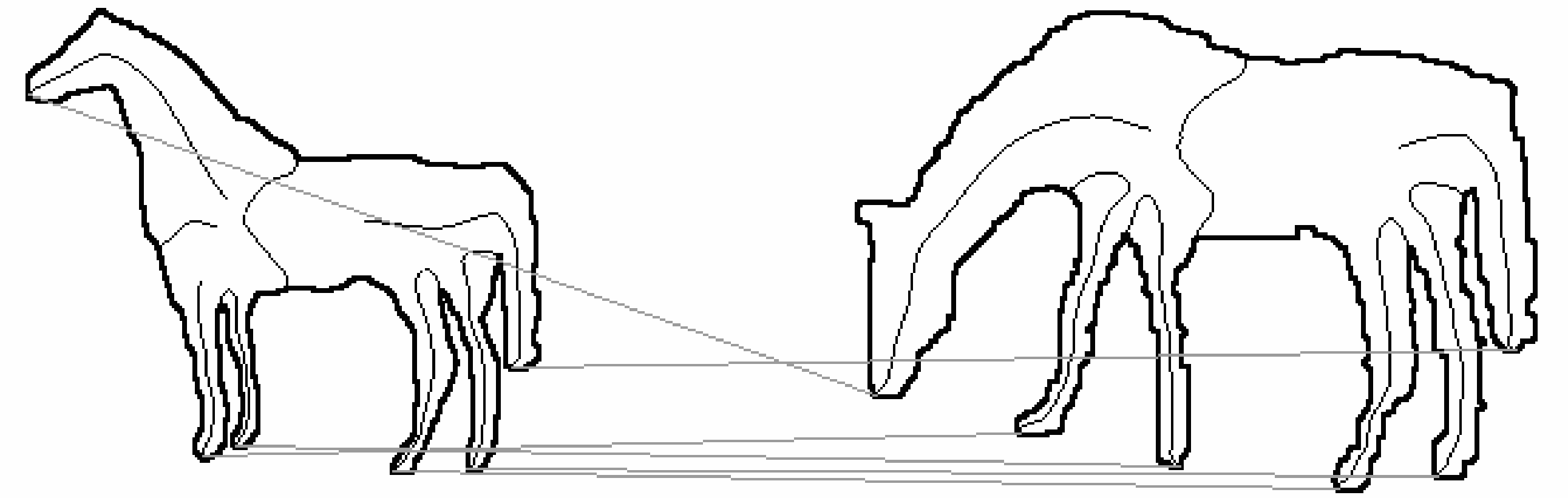} \\
(a) & (b) \\
\includegraphics[height=2.5cm]{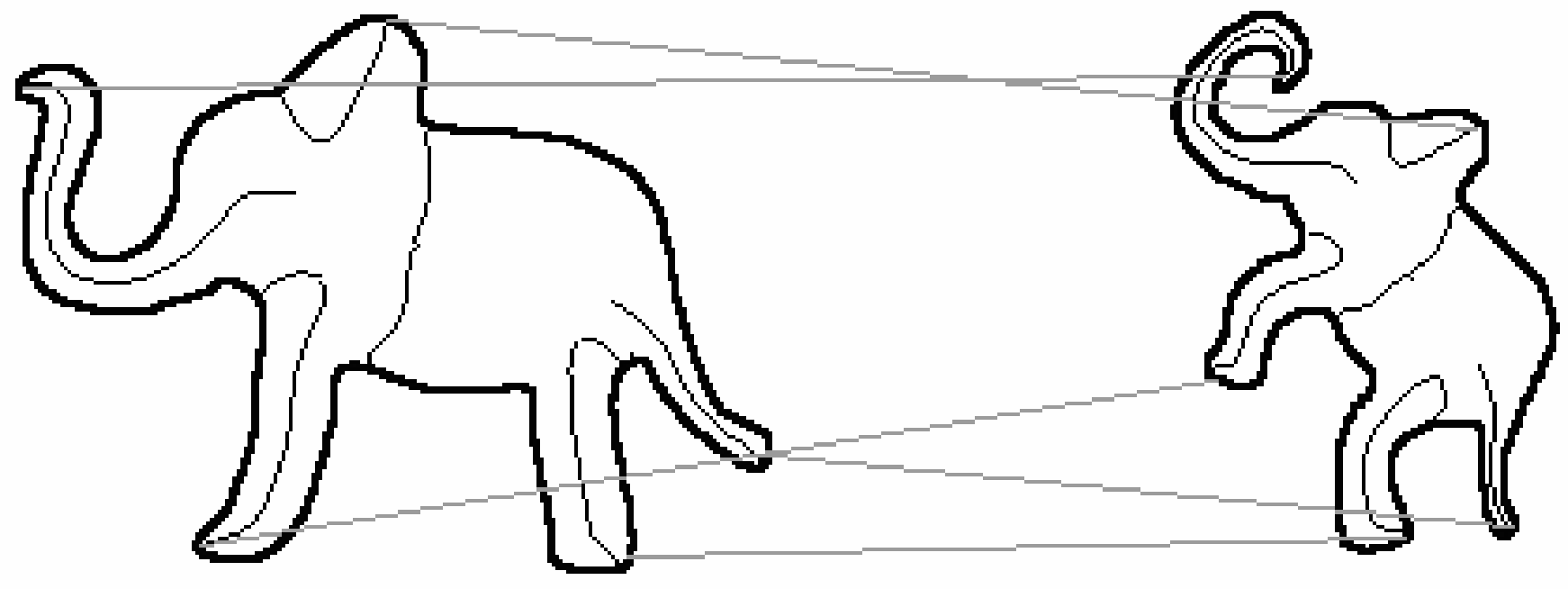} &
\includegraphics[height=2.5cm]{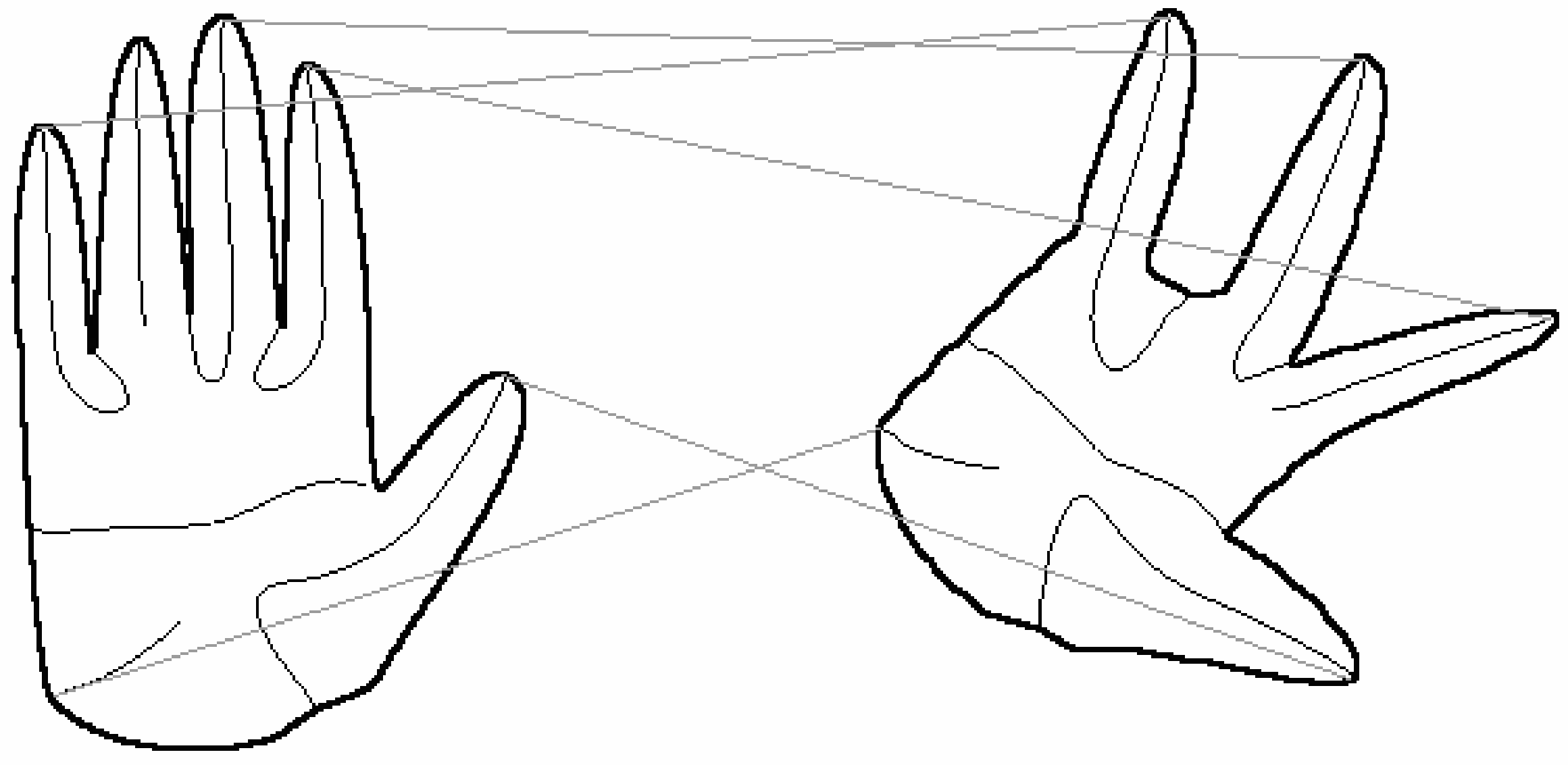} \\
(c) & (d) \\
\includegraphics[height=2.5cm]{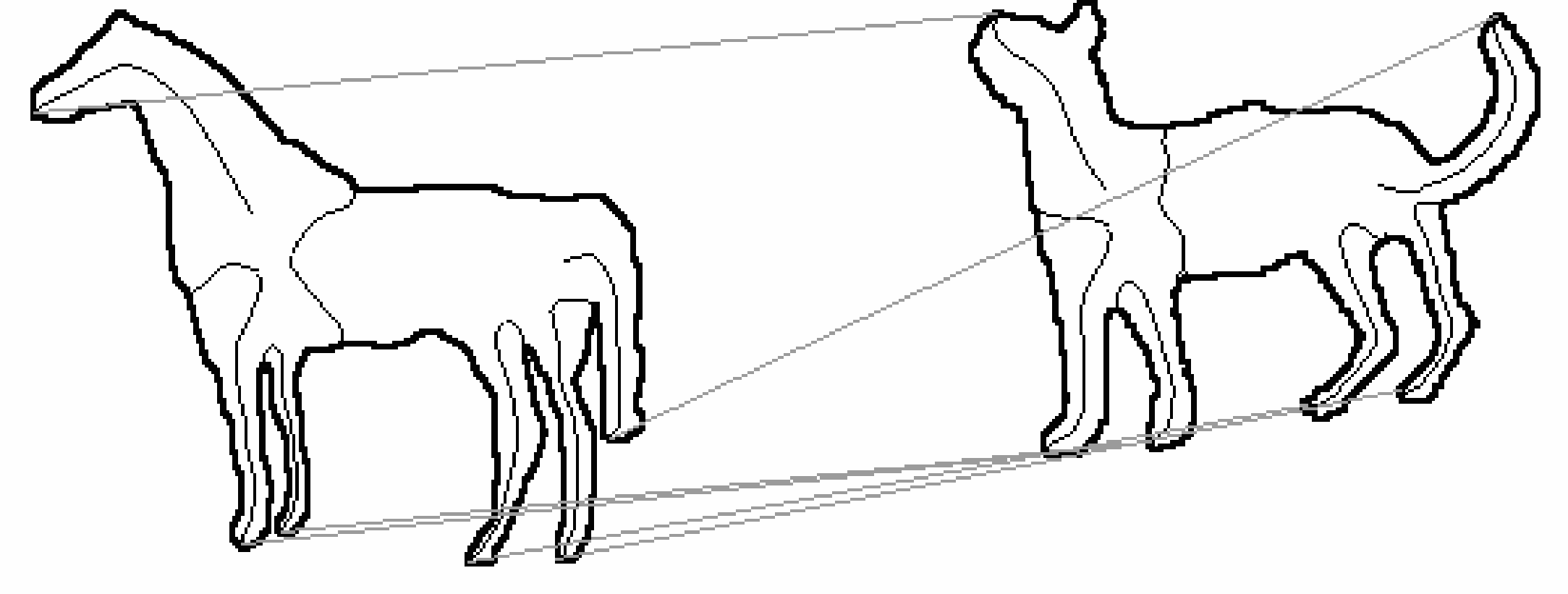} &
\includegraphics[height=2.5cm]{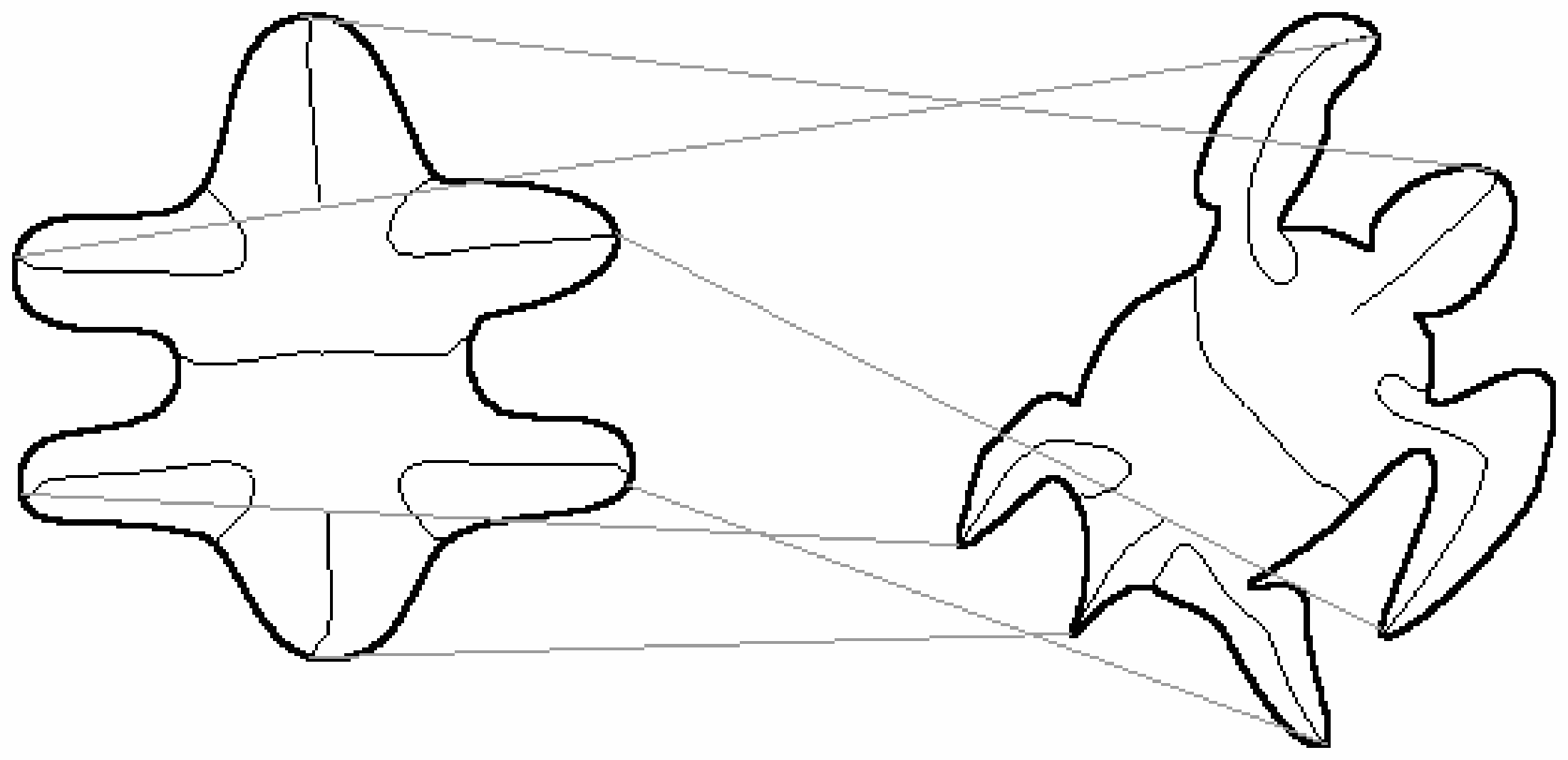} \\
(e) & (f) \\
\end{tabular}
\caption{Matching scores are 0.992, 0.708, 0.886, 0.652, 0.714, and
0.832 respectively. Note that the correct correspondences are found
even when a feature (finger) is missing. The missing finger
significantly lowers the matching score. This is a limitation of the
 matching scheme used, not the limitation of the representation.}
\label{fig:fig17}
\end{figure}

\begin{figure}[!h]
\centering \epsfig{figure=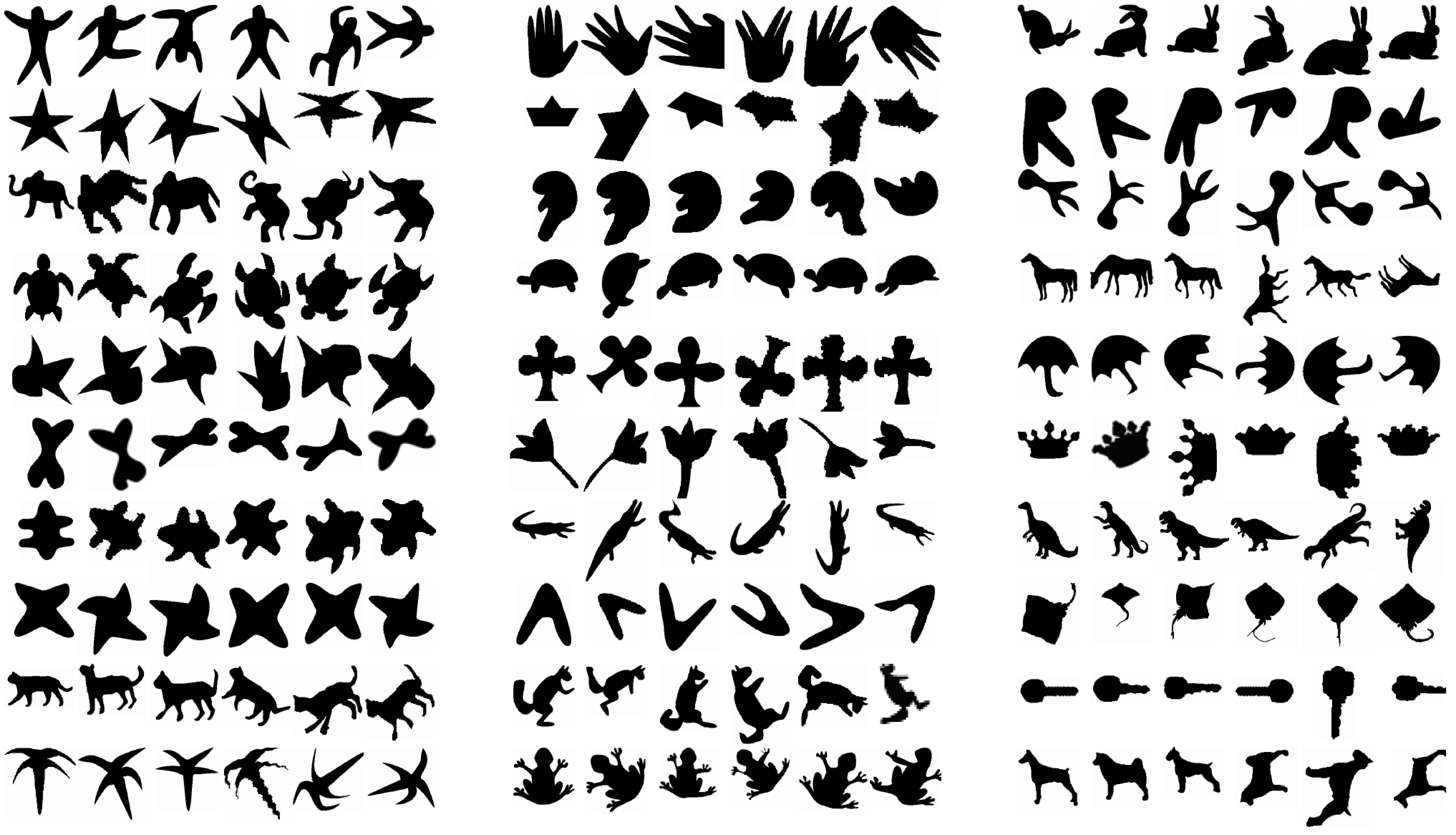,height=8cm}
\caption{The shape database used in the experiments.} \label{fig:db}
\end{figure}

\begin{figure}[!h]
\centering
\begin{tabular}{cc}
\includegraphics[height=5.5cm]{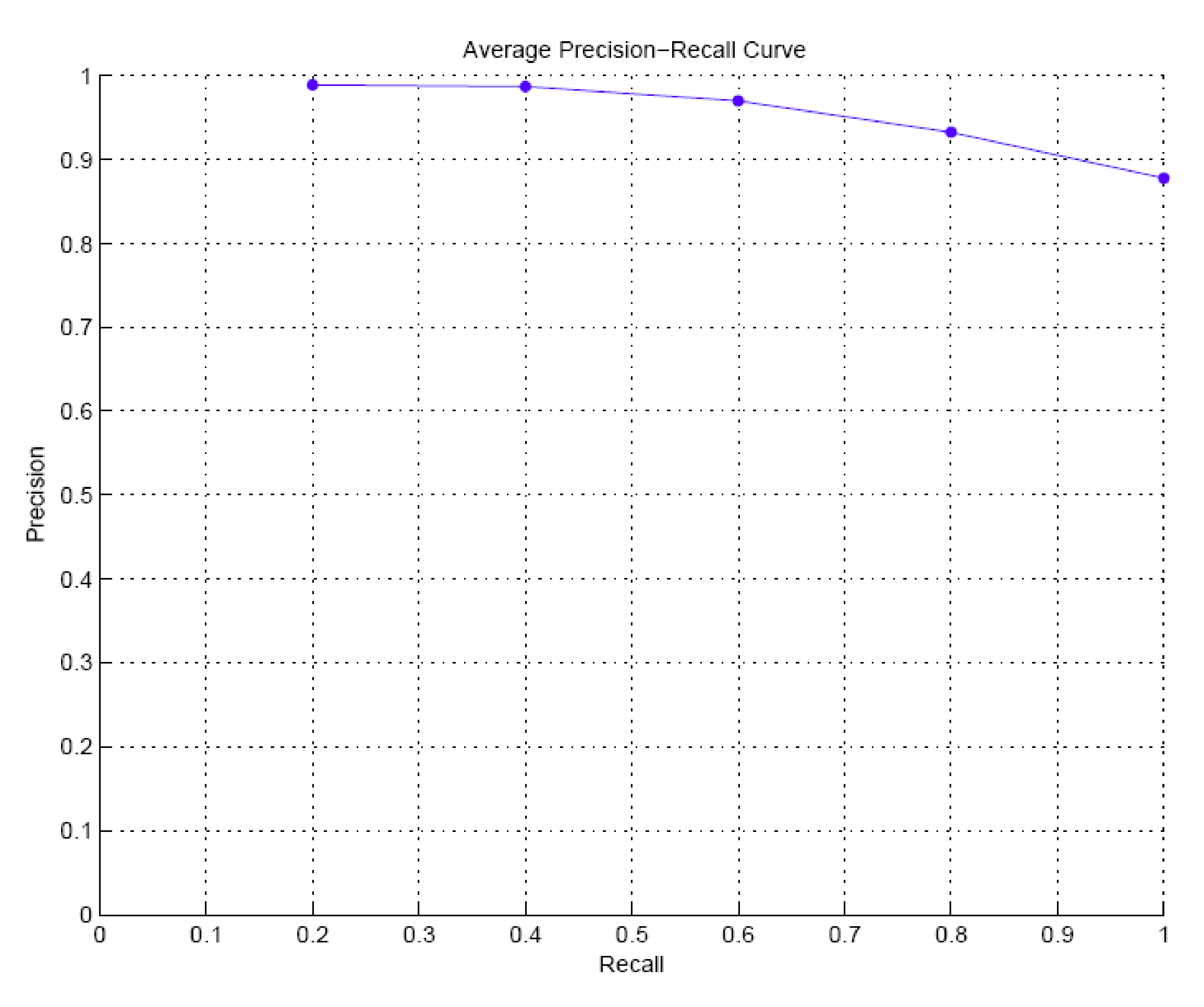} &
\includegraphics[height=5.5cm]{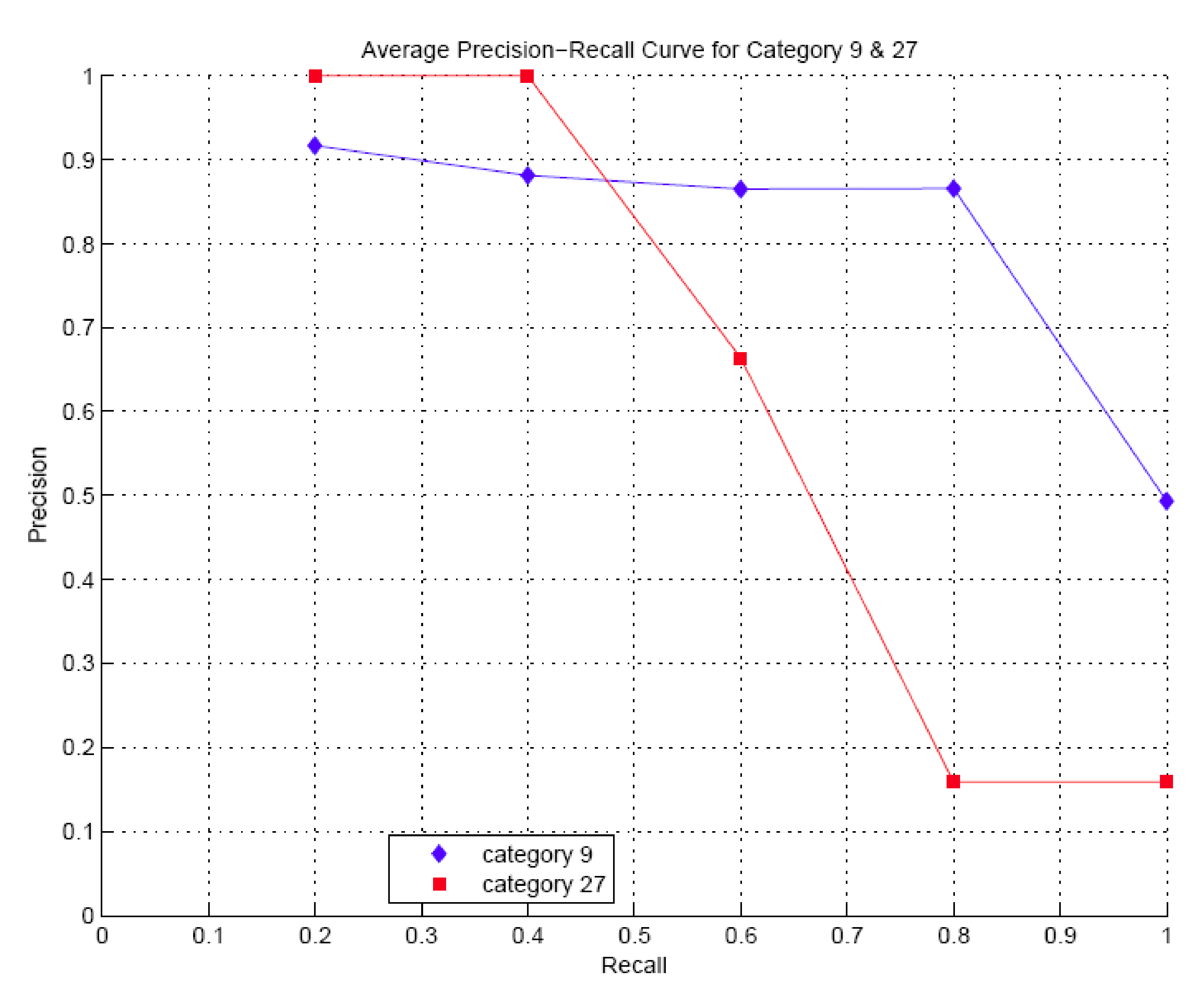} \\
(a) & (b) \\
\end{tabular}
\caption{Average precision-recall graphs.} \label{fig:precision_recall}
\end{figure}

We demonstrate the correspondence matching results
on a few illustrative shape pairs. As shown in Fig.
\ref{fig:fig17}(a)-(c), the matching process is able to find the
correspondences when the shape undergoes rotation and articulation.
In a case of missing parts (Fig. \ref{fig:fig17}(d)), the
perceptually correct correspondences are found since the spatial
organization of the symmetry branches is stored in the descriptions.
The unmatched finger lowers the total similarity score
significantly. The correspondences between shapes from different
categories are shown in Fig. \ref{fig:fig17}(e),(f). High similarity
values are detected due to the coarse descriptions generated by the
representation scheme.  In a sense, the compared shapes may be
considered similar since they have the same number of limbs at
similar locations.
{The recognition capability} of our system is tested  on a diverse
shape database. As shown in Fig. \ref{fig:db}, the database includes
30 categories with 6 shapes in each category. Among the shapes
within the same category, there are differences in orientation,
scale, articulation, and small boundary details. This is mainly to
evaluate the performance of the matching process under visual
transformations. Each shape in the database is used as a query shape
and the most similar shapes are retrieved. (The nearest neighbors of
some query shapes are shown in
Tables~\ref{table:results}-\ref{table:results2}).
Since each category consists of 6 shapes, the first five query
results (hits) must be of the same category with the query shape for
perfect performance. The Bull's eye~\cite{Ling05,Latecki00} test
score is 98\%. However, in retrieval problems, precision-recall
graphs are more informative than the Bull's eye test. The average
precision-recall curve for all shapes is displayed in Fig.
\ref{fig:precision_recall}(a). For all the recall levels, {high
precision values are obtained}. The precision is around 88\% even
when the recall is 100\%. This means that our framework returns the
most relevant shapes in the top query results.

We also present the average precision-recall curves for category 9
(cat) and category 27 (dinosaur) in Fig.
\ref{fig:precision_recall}(b). These two categories have the worst
retrieval performance. Category 27 consists of two sub categories:
dinosaurs with two arms and with one arm (Fig. \ref{fig:db}). The
additional symmetry branch indirectly introduces a cost by means of
the total branch length that is used in the normalization. For
category 9, one of the cat shapes turns out to be different than the
others and, in fact, that shape is more similar to  category 24
(horse) (Table~\ref{table:results} - Query 9).

\clearpage
\begin{table}[!htb]
\caption{Some query results.}
\centering
\begin {tabular} {|c|c|c|c|c|c||c|c|c|c|c|c|}
\hline \epsfig{figure=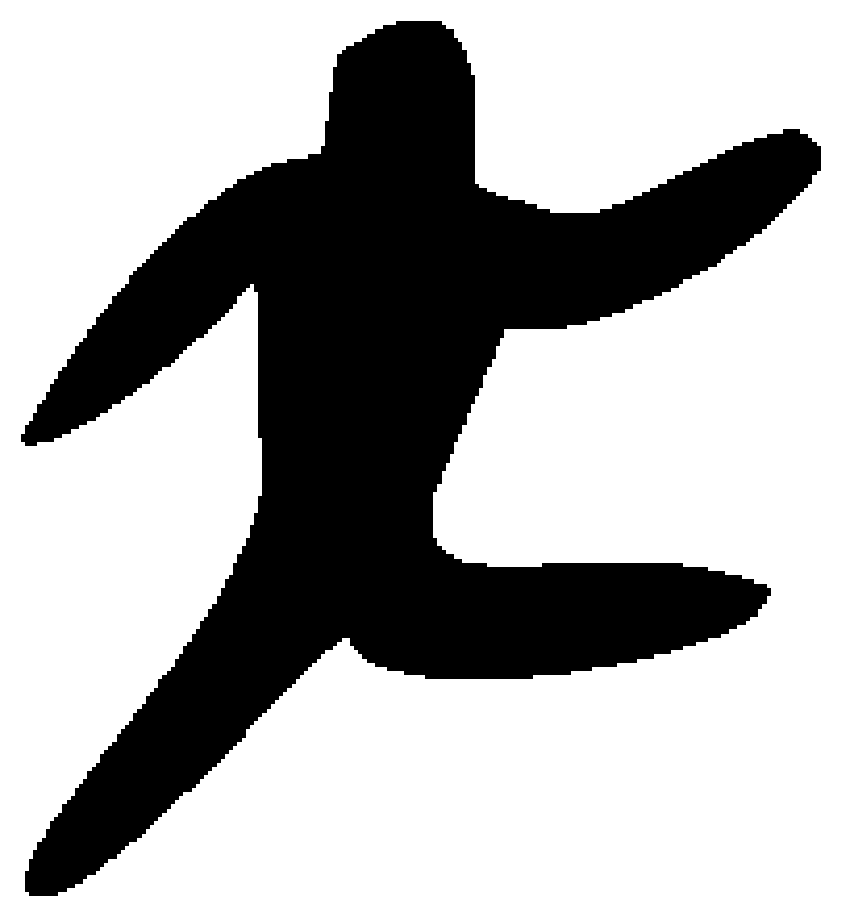,height=0.700000cm}&
\epsfig{figure=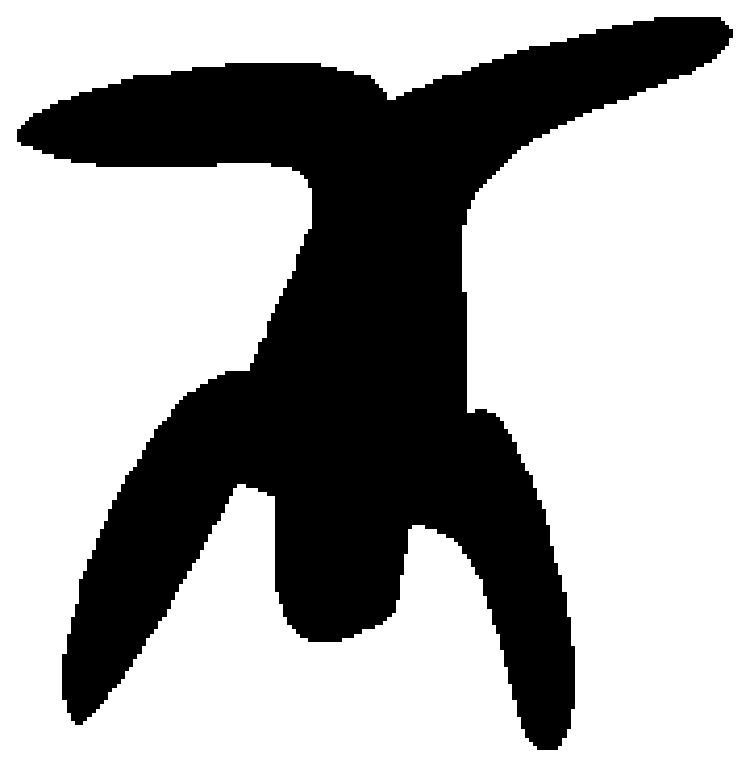,height=0.700000cm}&
\epsfig{figure=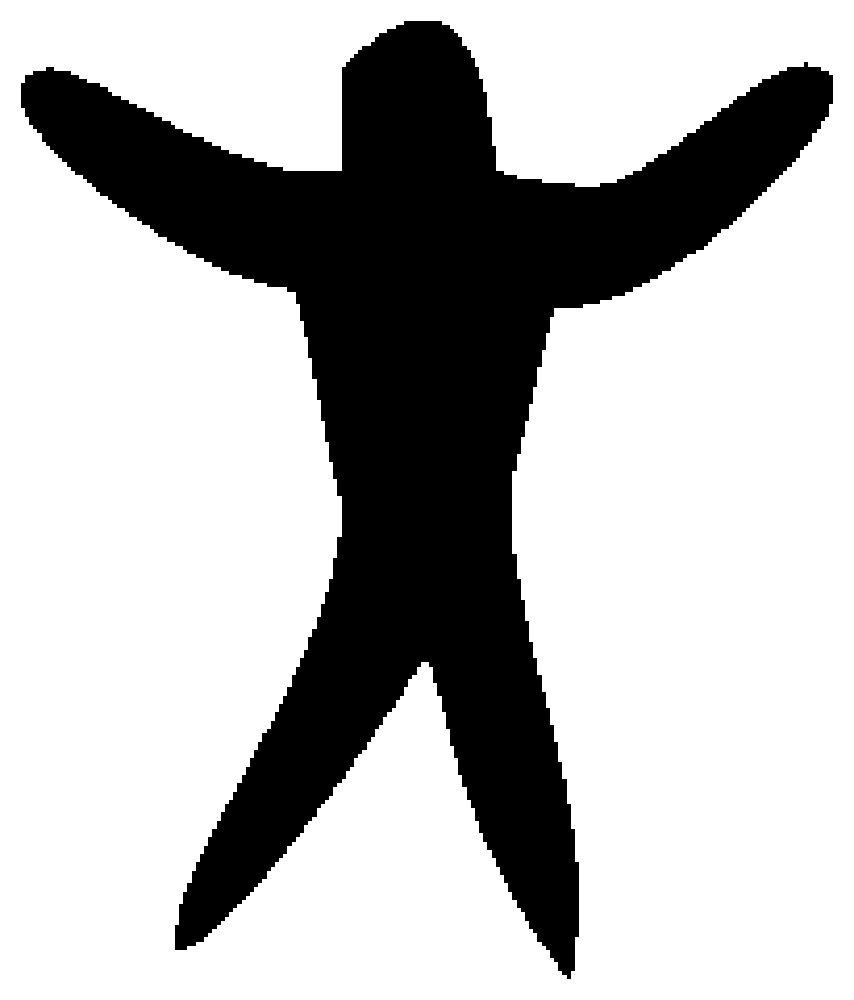,height=0.700000cm}&
\epsfig{figure=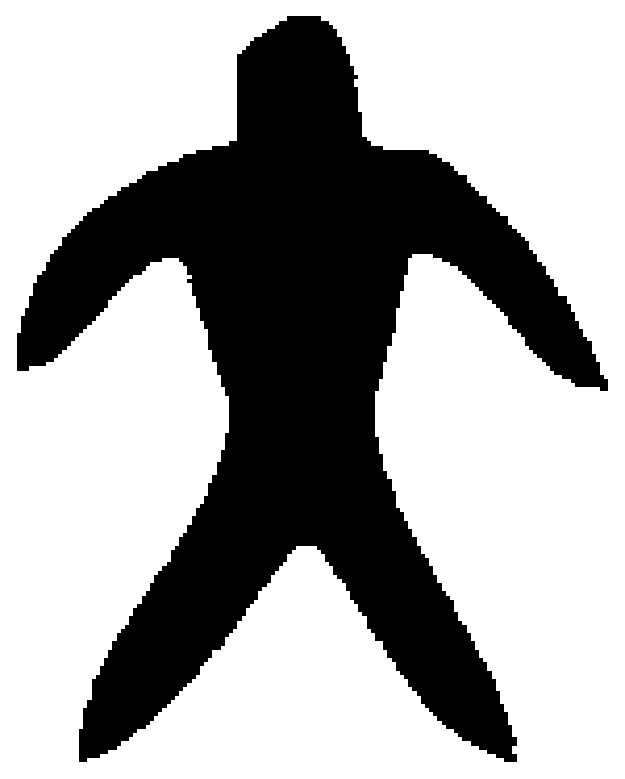,height=0.700000cm}&
\epsfig{figure=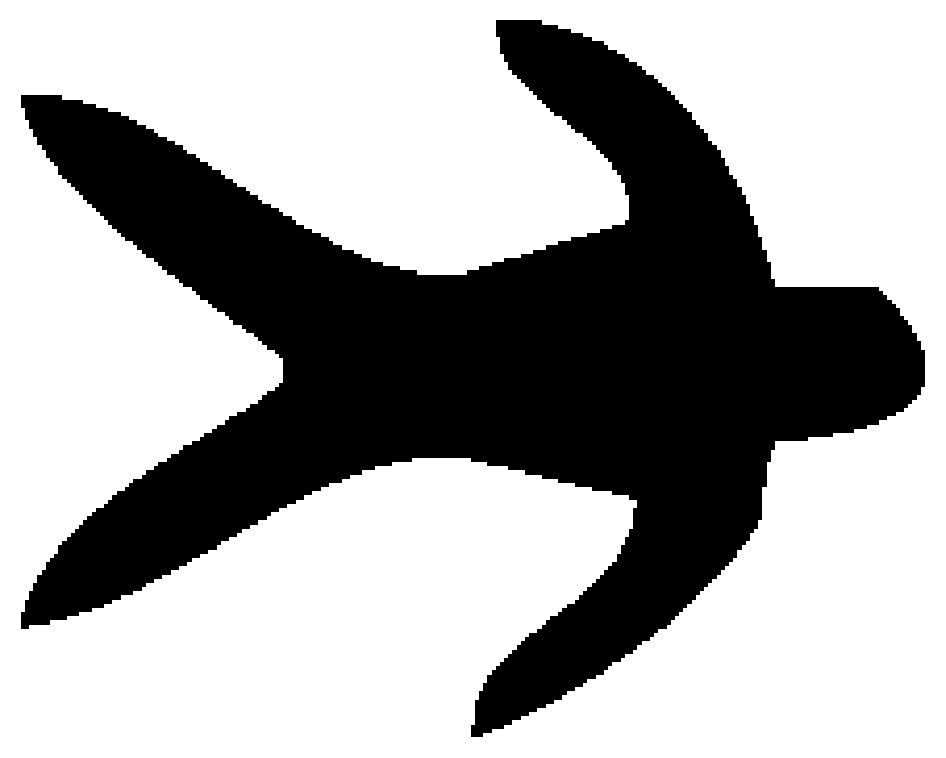,width=0.700000cm}&
\epsfig{figure=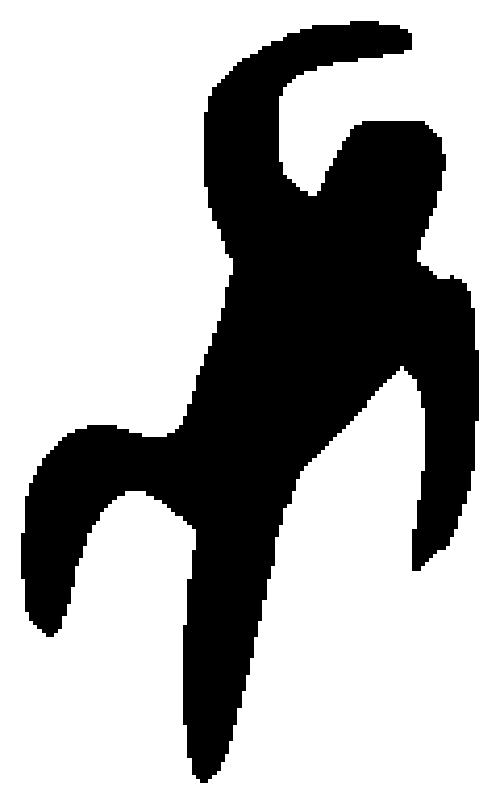,height=0.700000cm}&
\epsfig{figure=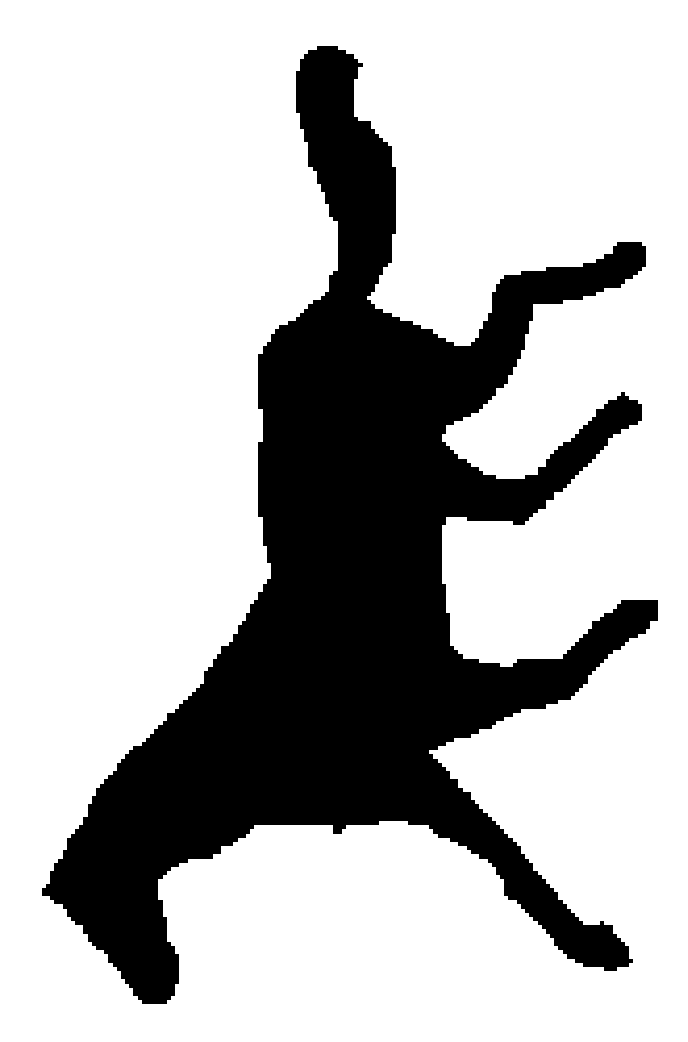,height=0.700000cm}&
\epsfig{figure=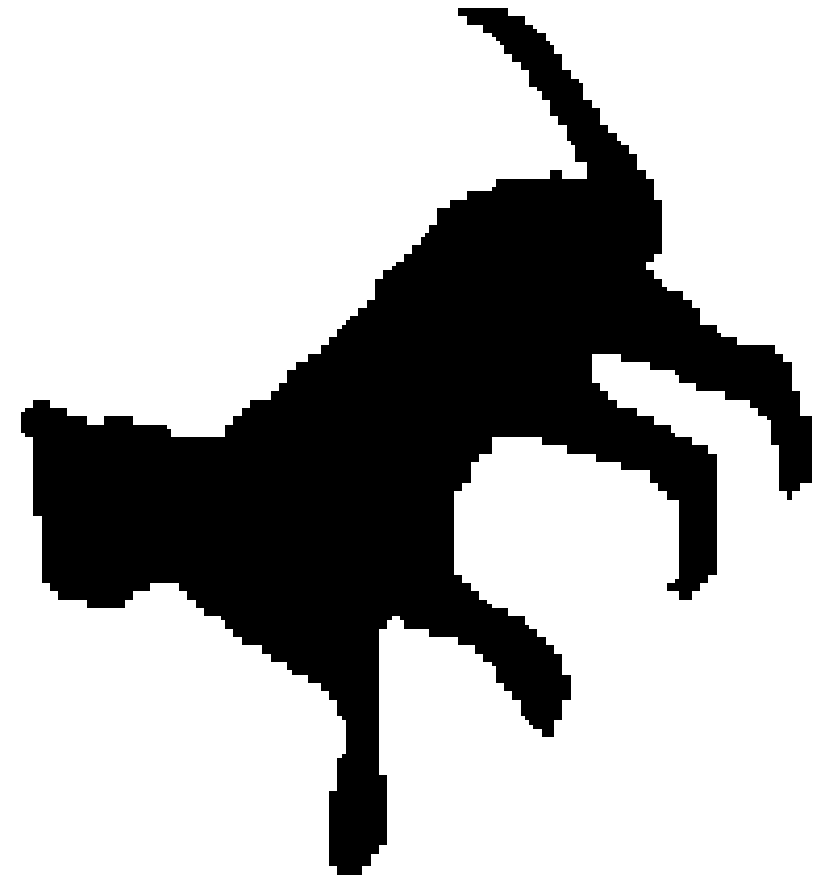,height=0.700000cm}&
\epsfig{figure=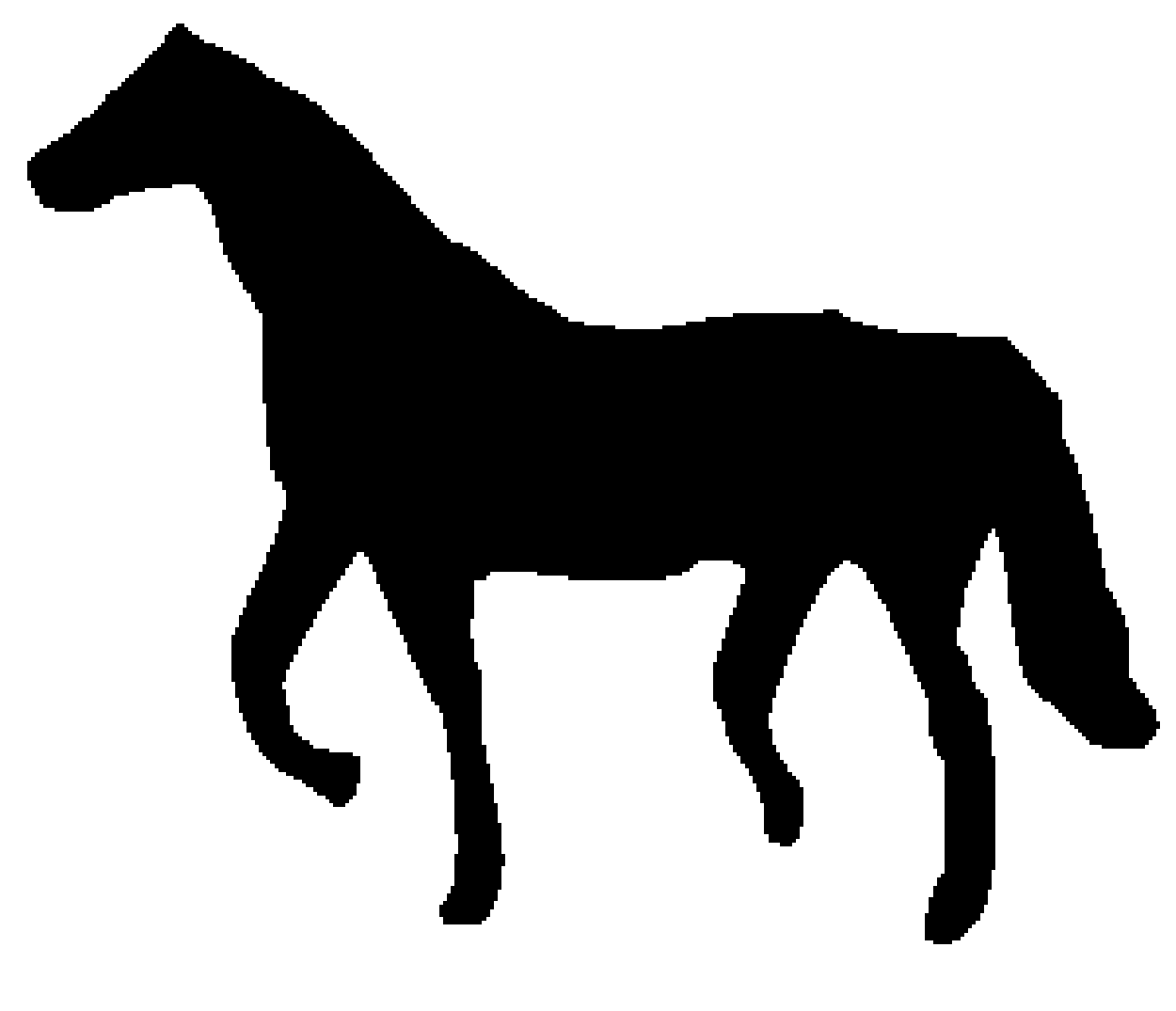,width=0.700000cm}&
\epsfig{figure=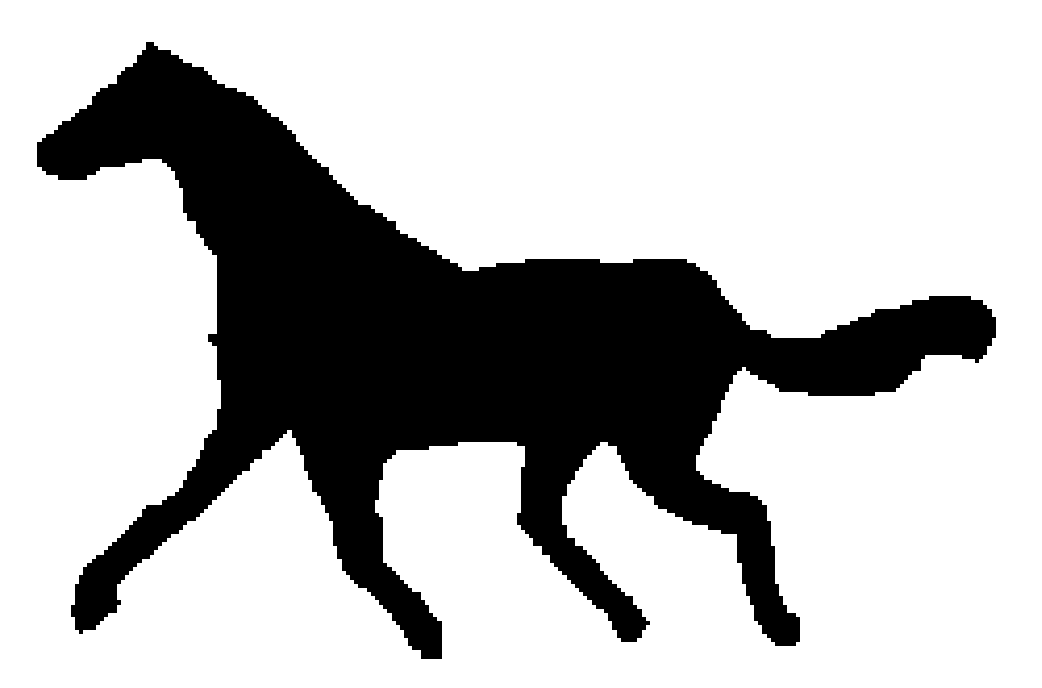,width=0.700000cm}&
\epsfig{figure=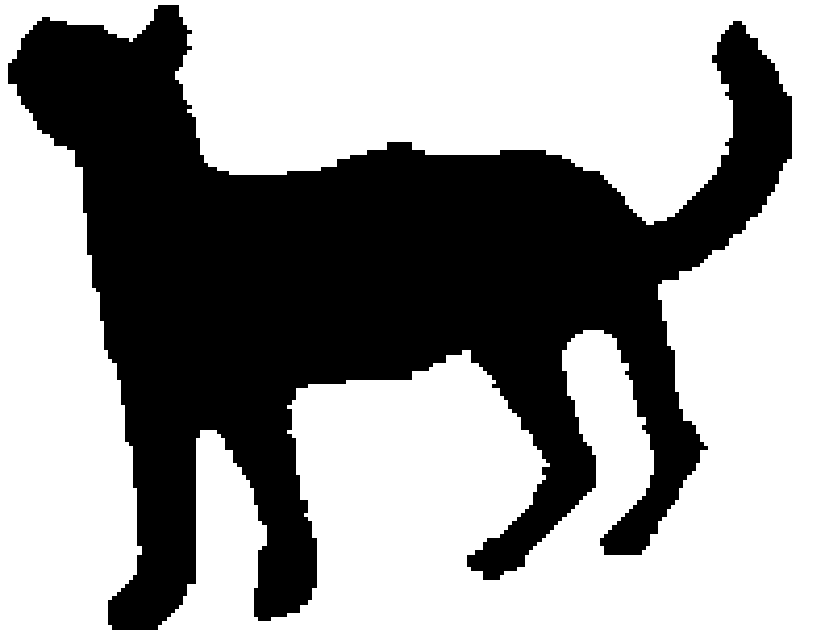,width=0.700000cm}&
\epsfig{figure=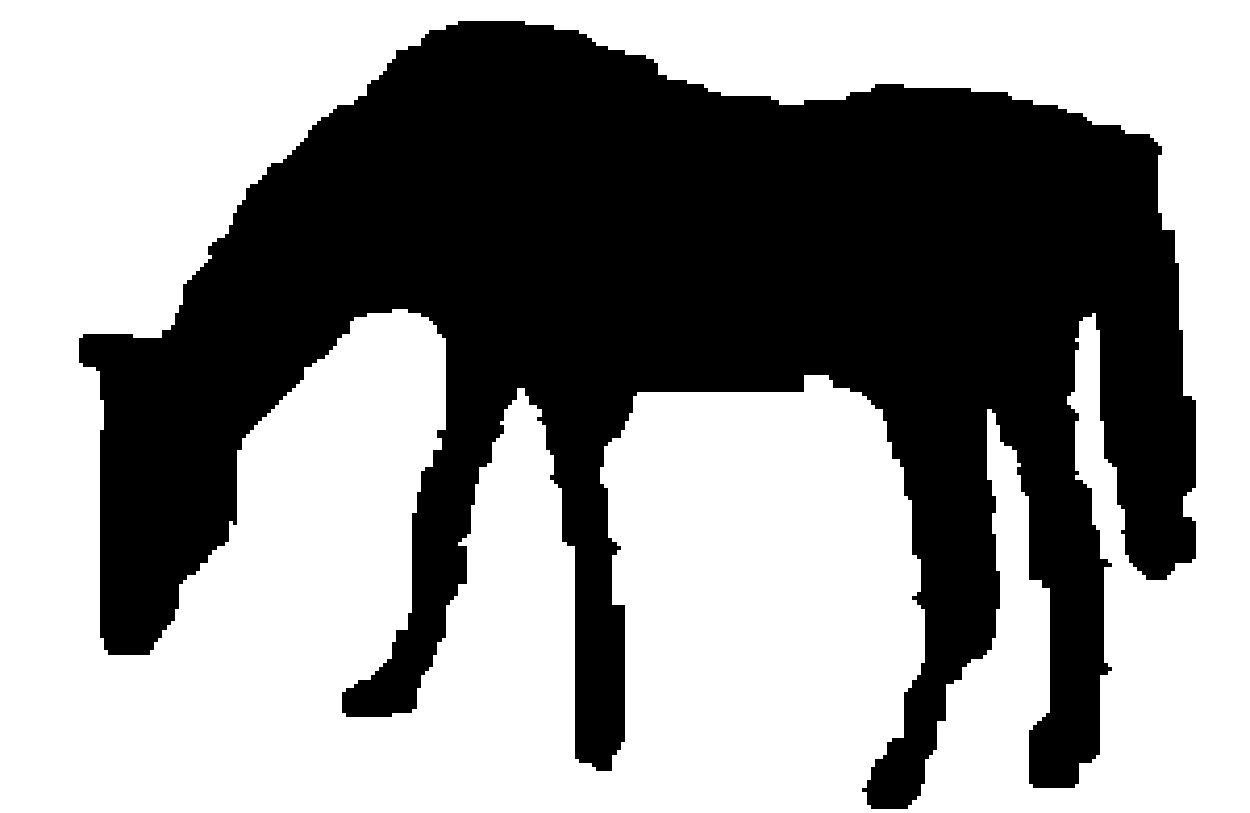,width=0.700000cm}\\
\cline{2-12}
  & \bf{0.965} & \bf{0.949} & \bf{0.941} & \bf{0.898} & \bf{0.819} & 0.578 & 0.557 & 0.550 & 0.548 & 0.541 & 0.532 \\
\hline \epsfig{figure=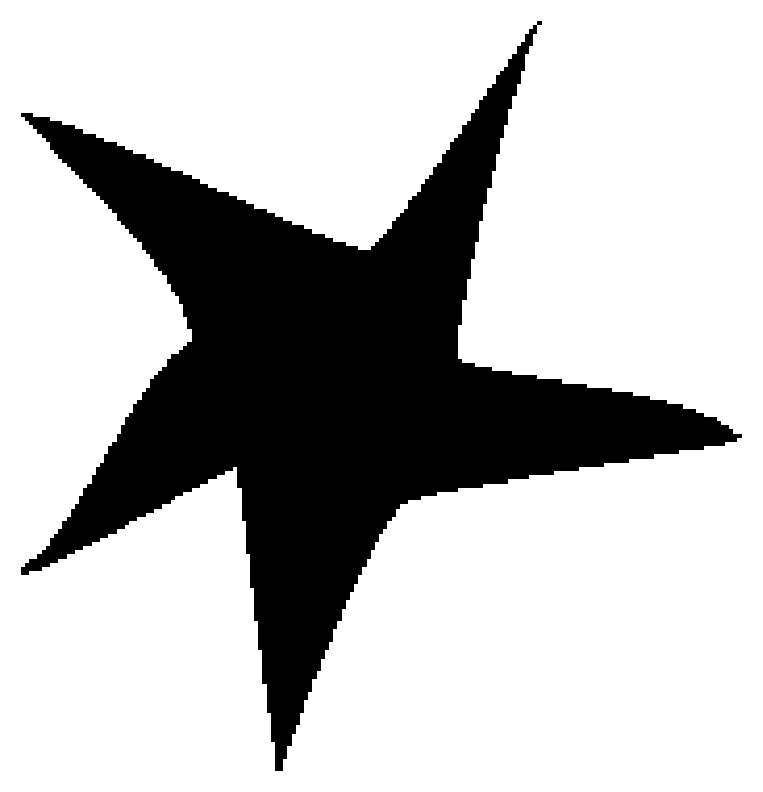,height=0.700000cm}&
\epsfig{figure=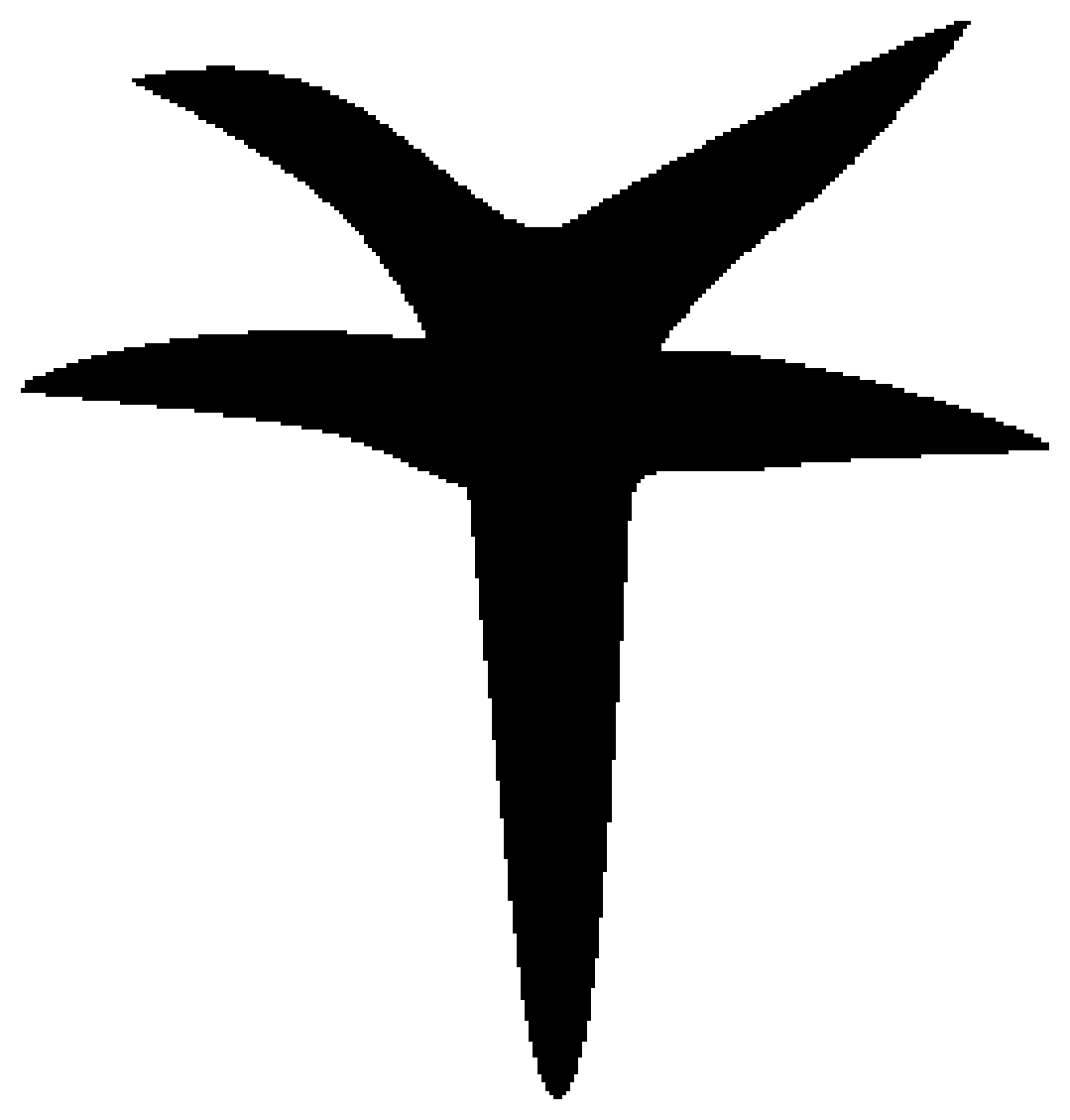,height=0.700000cm}&
\epsfig{figure=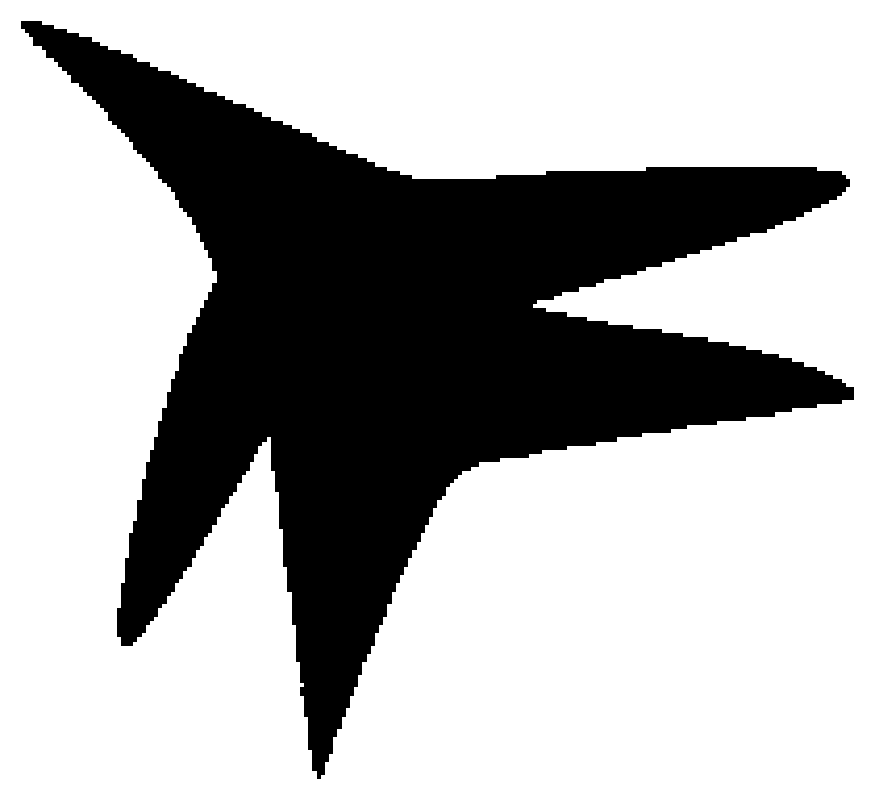,width=0.700000cm}&
\epsfig{figure=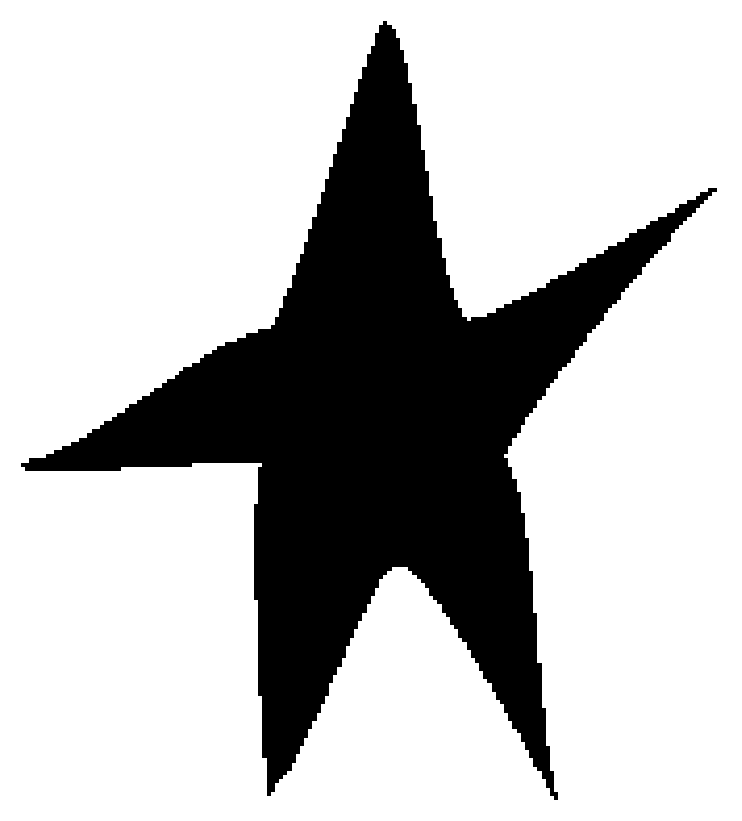,height=0.700000cm}&
\epsfig{figure=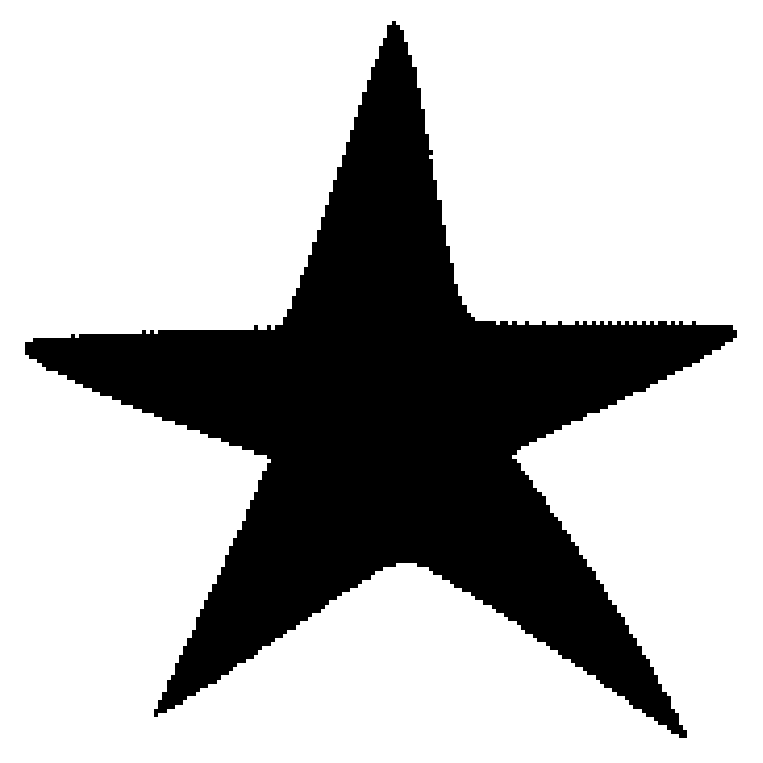,height=0.700000cm}&
\epsfig{figure=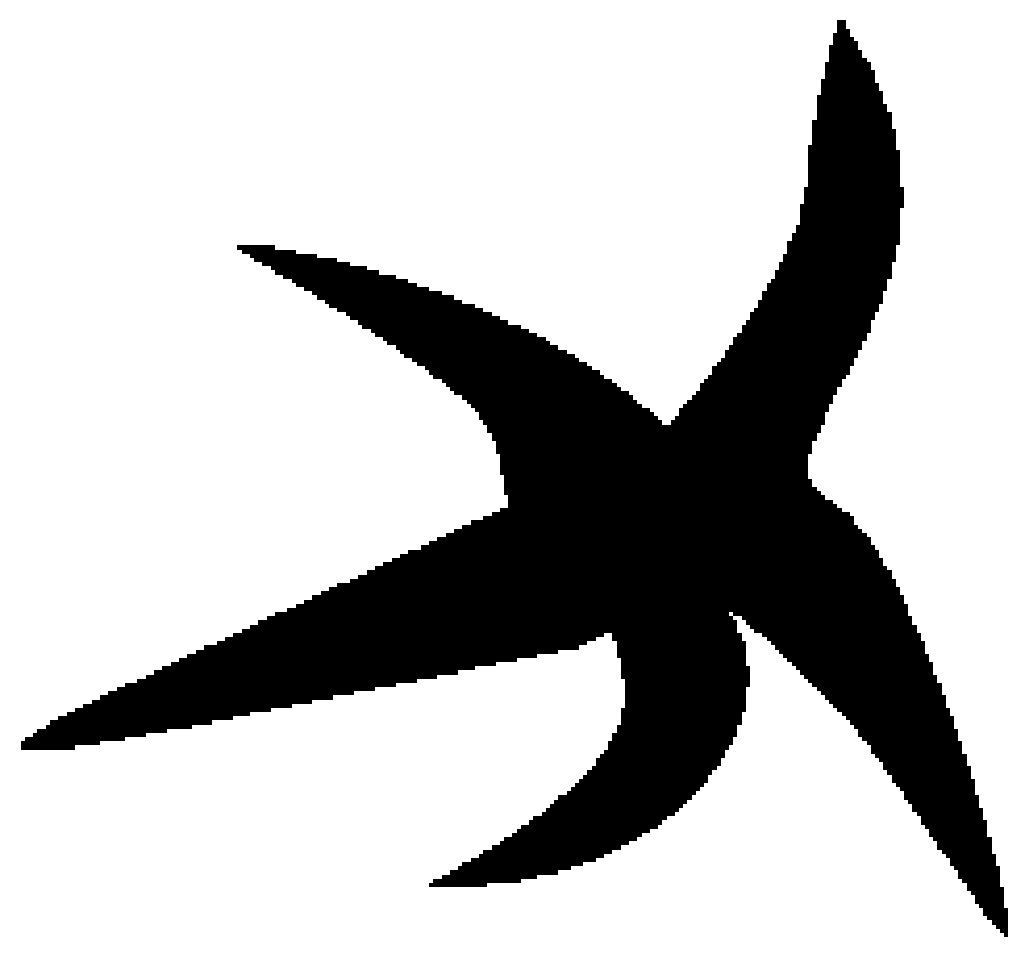,width=0.700000cm}&
\epsfig{figure=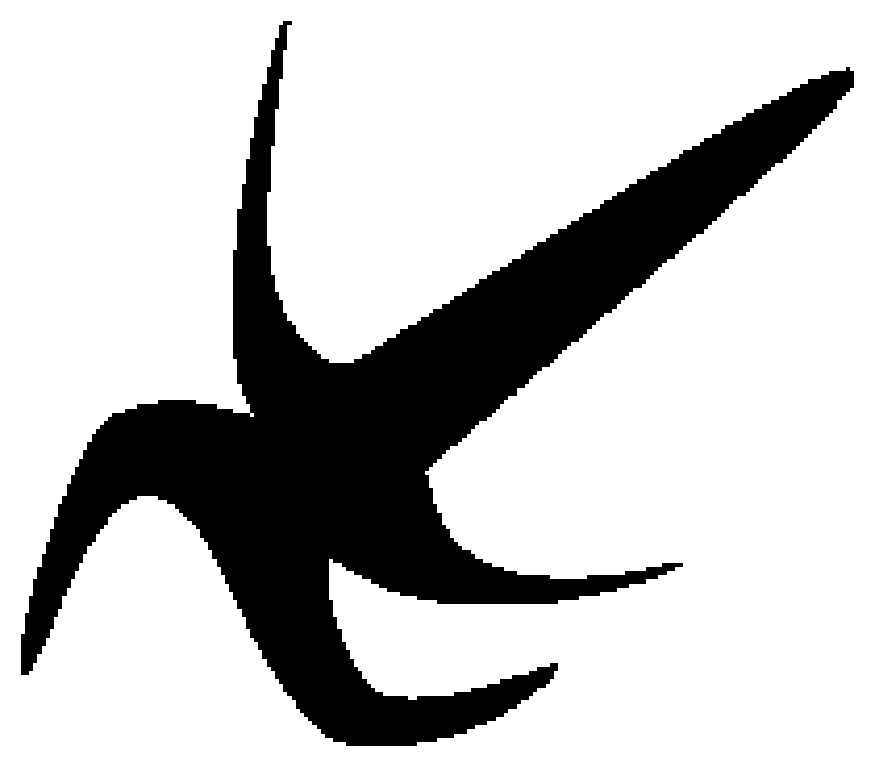,width=0.700000cm}&
\epsfig{figure=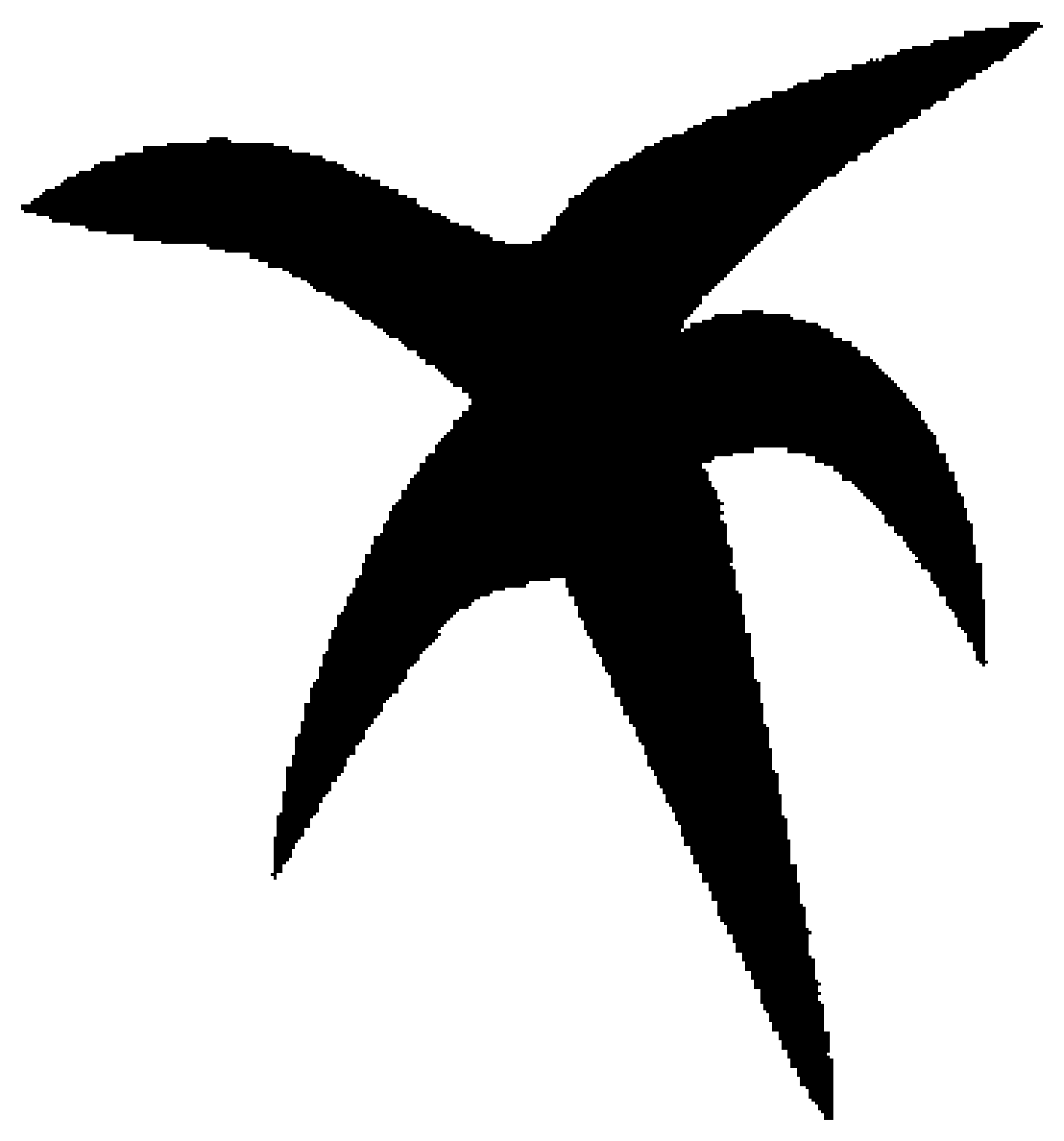,height=0.700000cm}&
\epsfig{figure=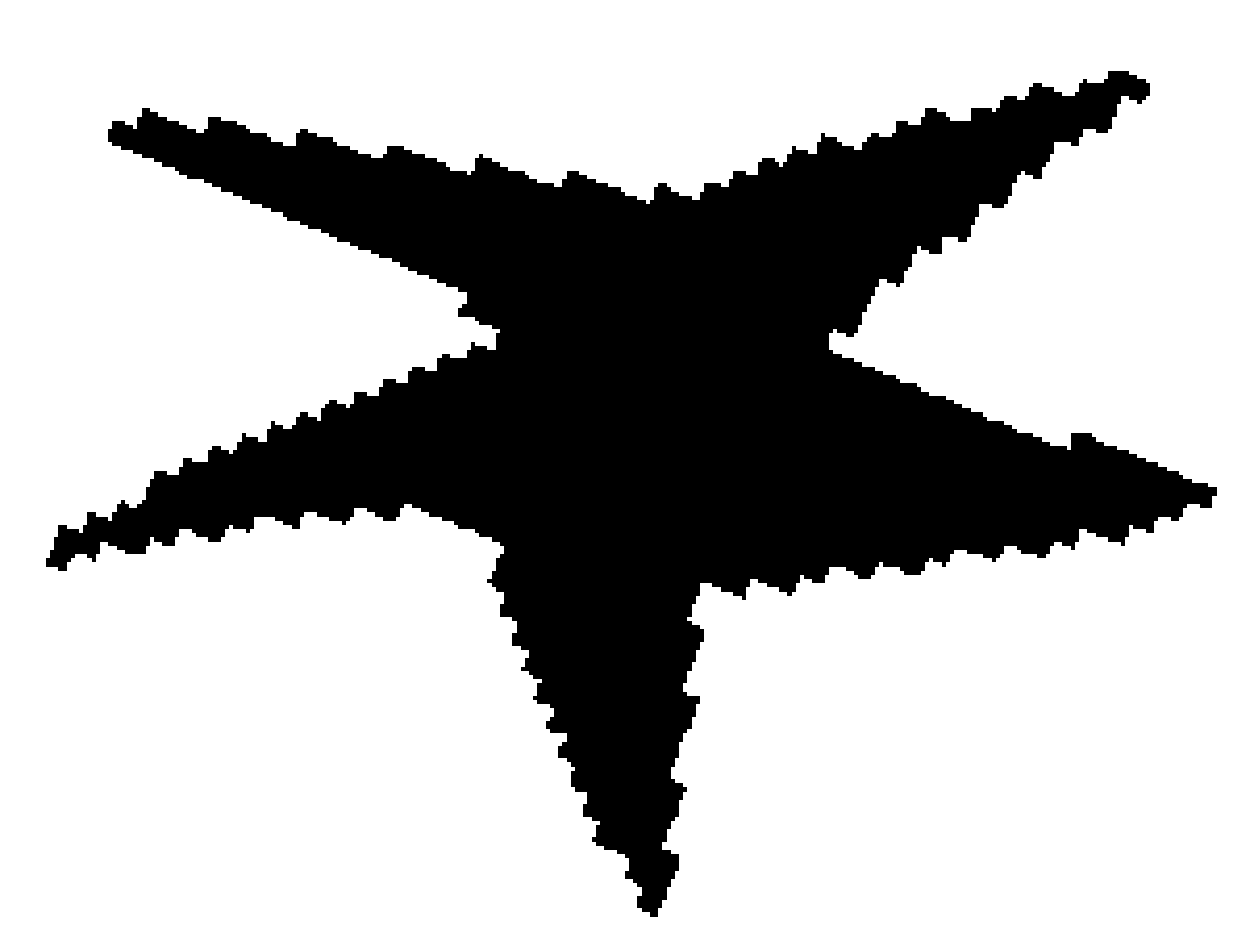,width=0.700000cm}&
\epsfig{figure=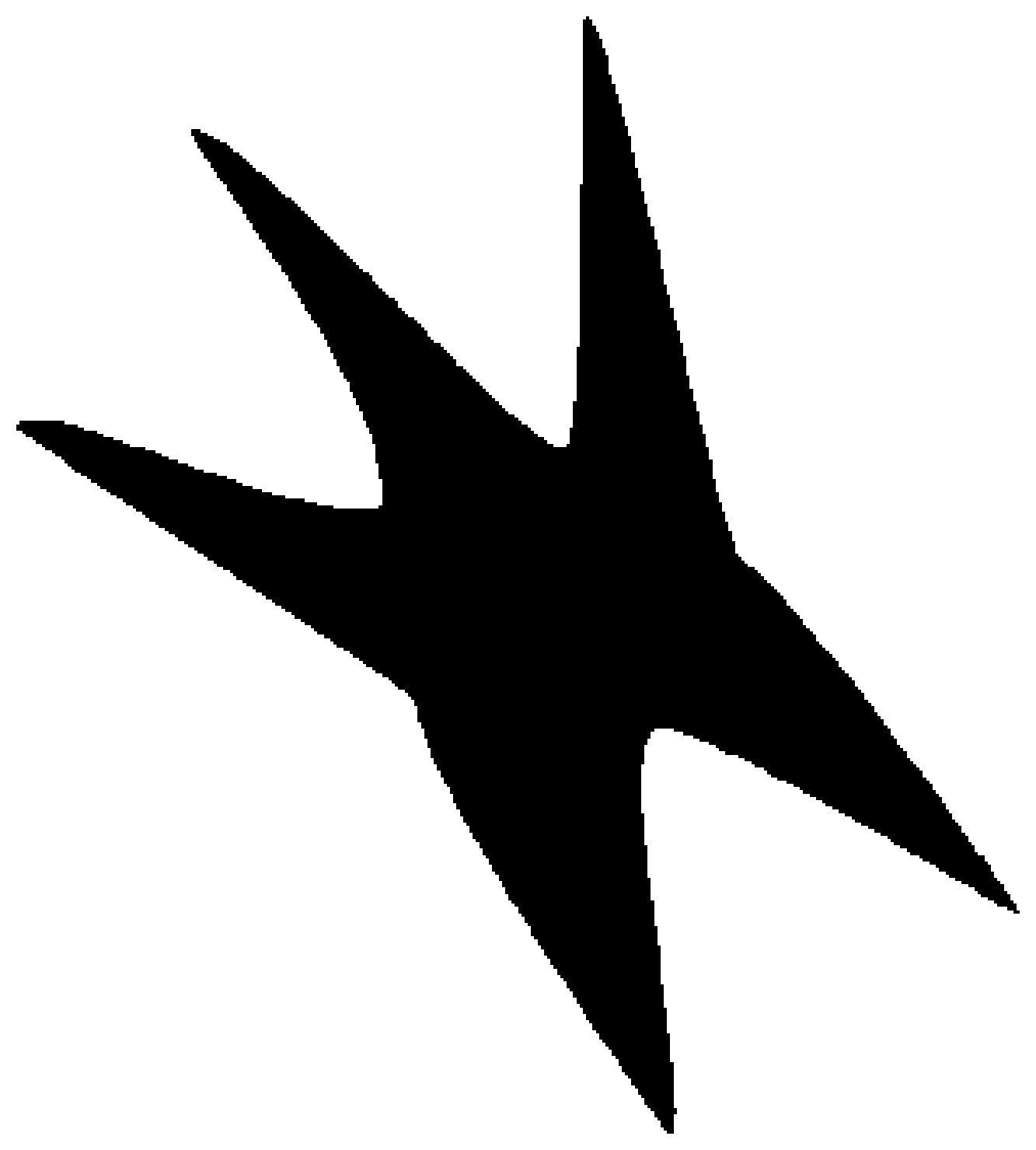,height=0.700000cm}&
\epsfig{figure=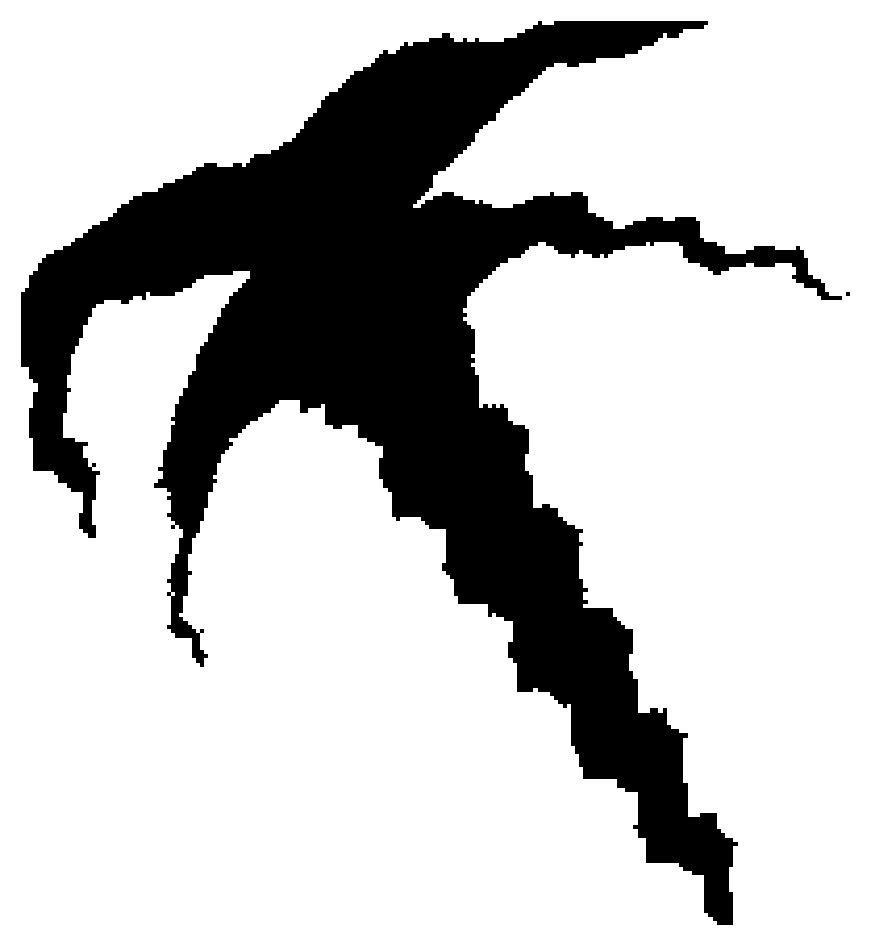,height=0.700000cm}&
\epsfig{figure=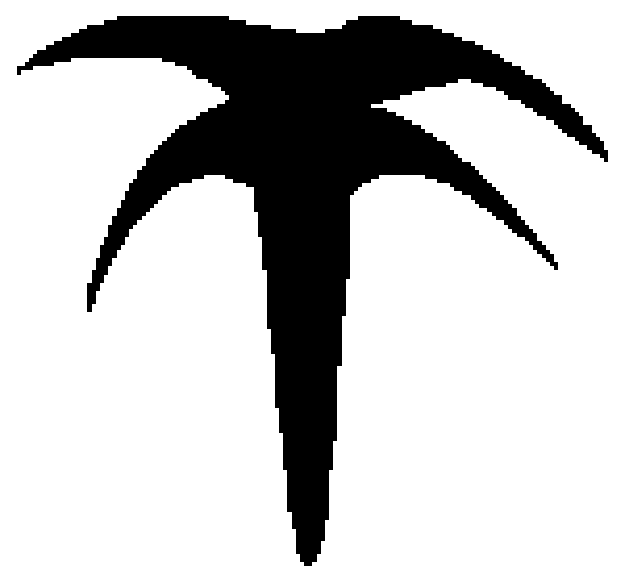,width=0.700000cm}\\
\cline{2-12}
  & 0.869 & \bf{0.844} & \bf{0.840} & \bf{0.819} & 0.811 & 0.800 & 0.784 & \bf{0.774} & \bf{0.721} & 0.718 & 0.708 \\
\hline \epsfig{figure=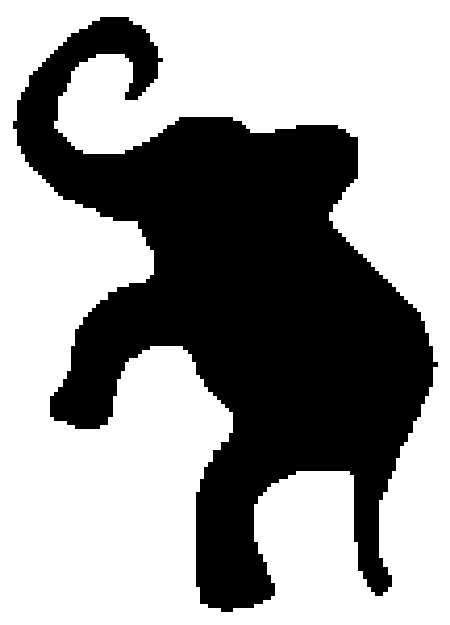,height=0.700000cm}&
\epsfig{figure=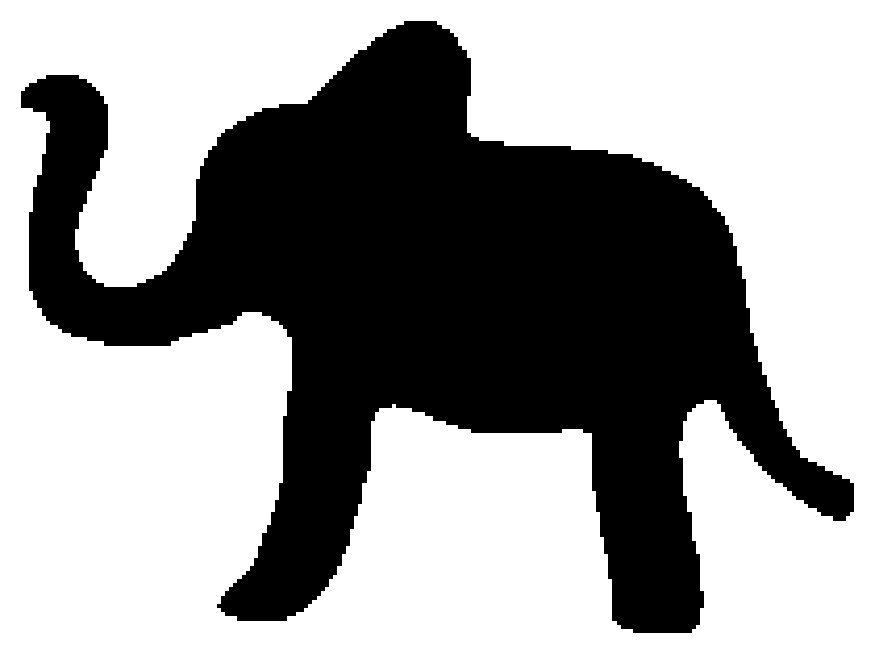,width=0.700000cm}&
\epsfig{figure=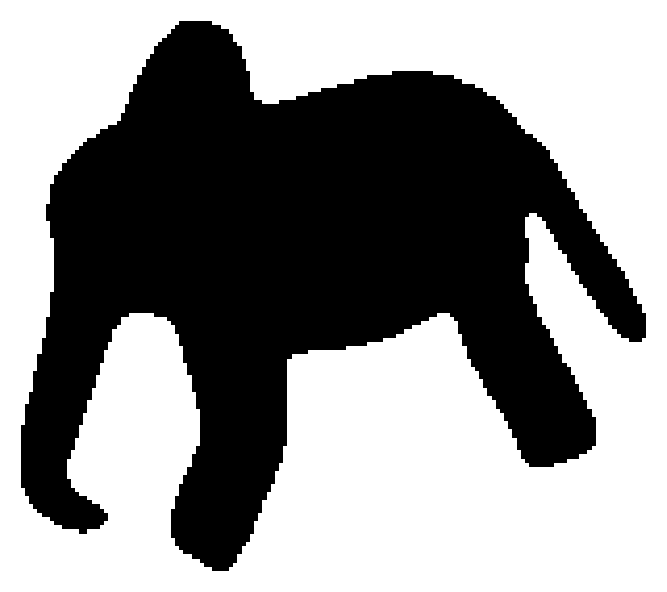,width=0.700000cm}&
\epsfig{figure=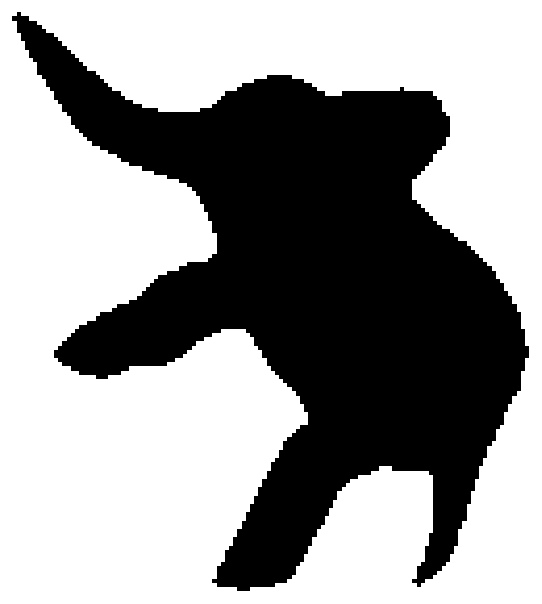,height=0.700000cm}&
\epsfig{figure=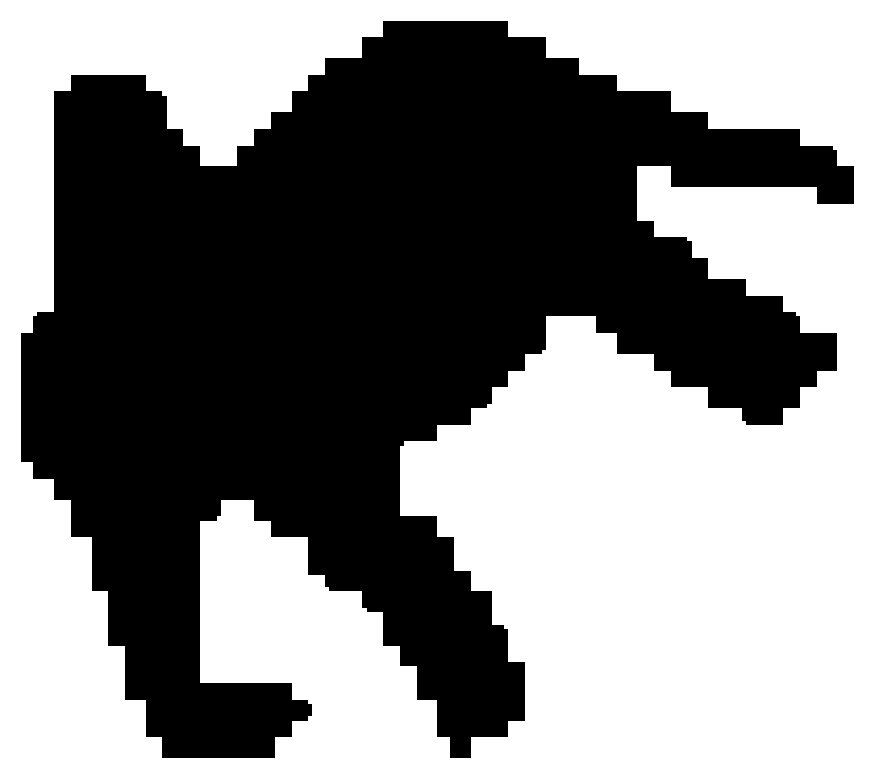,width=0.700000cm}&
\epsfig{figure=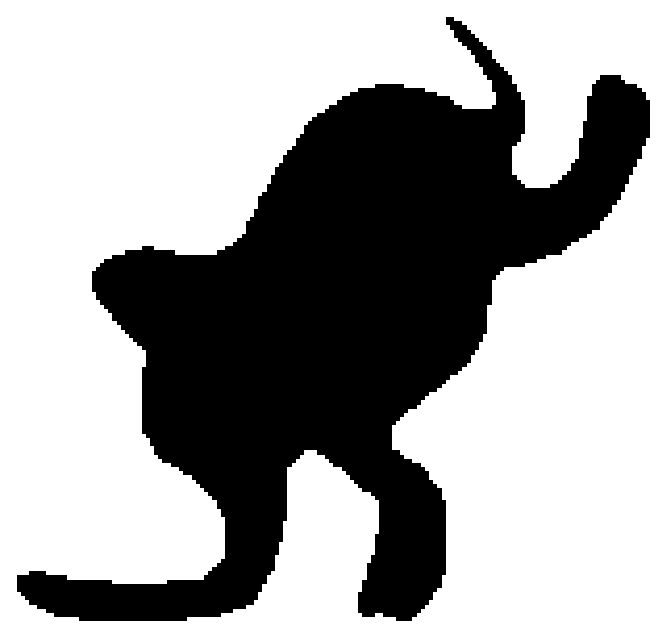,width=0.700000cm}&
\epsfig{figure=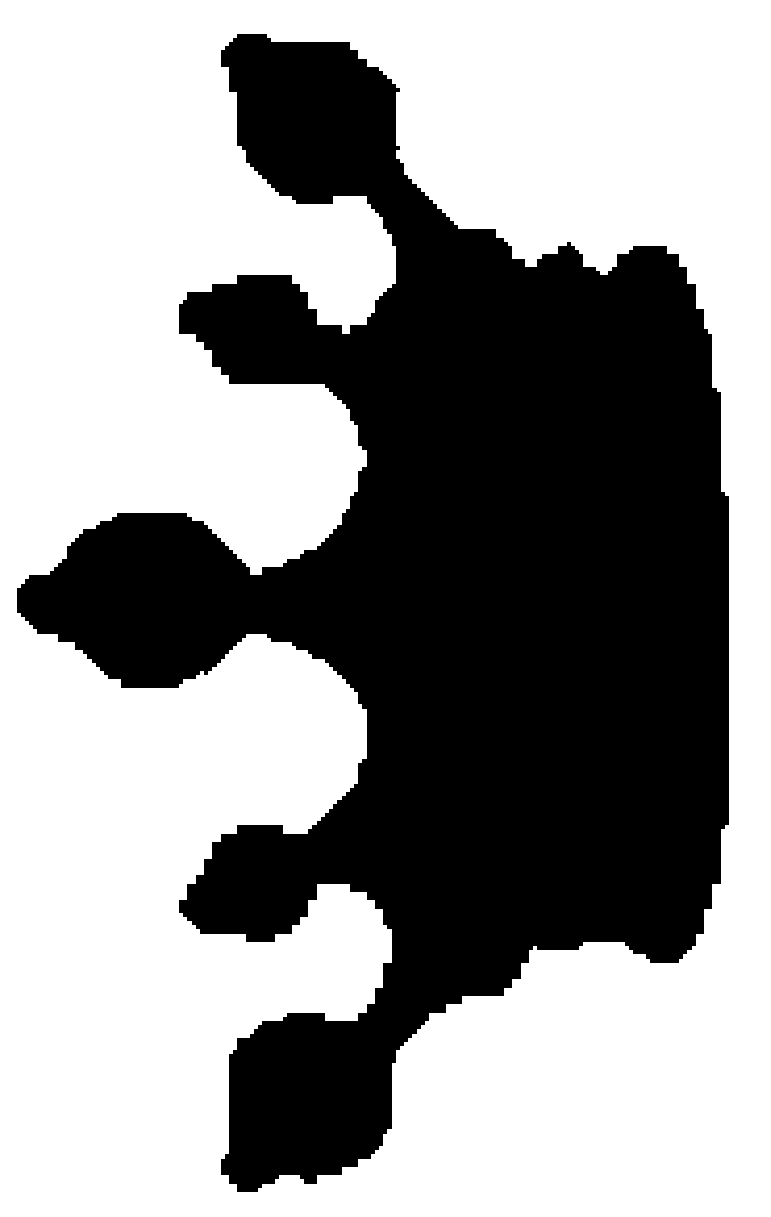,height=0.700000cm}&
\epsfig{figure=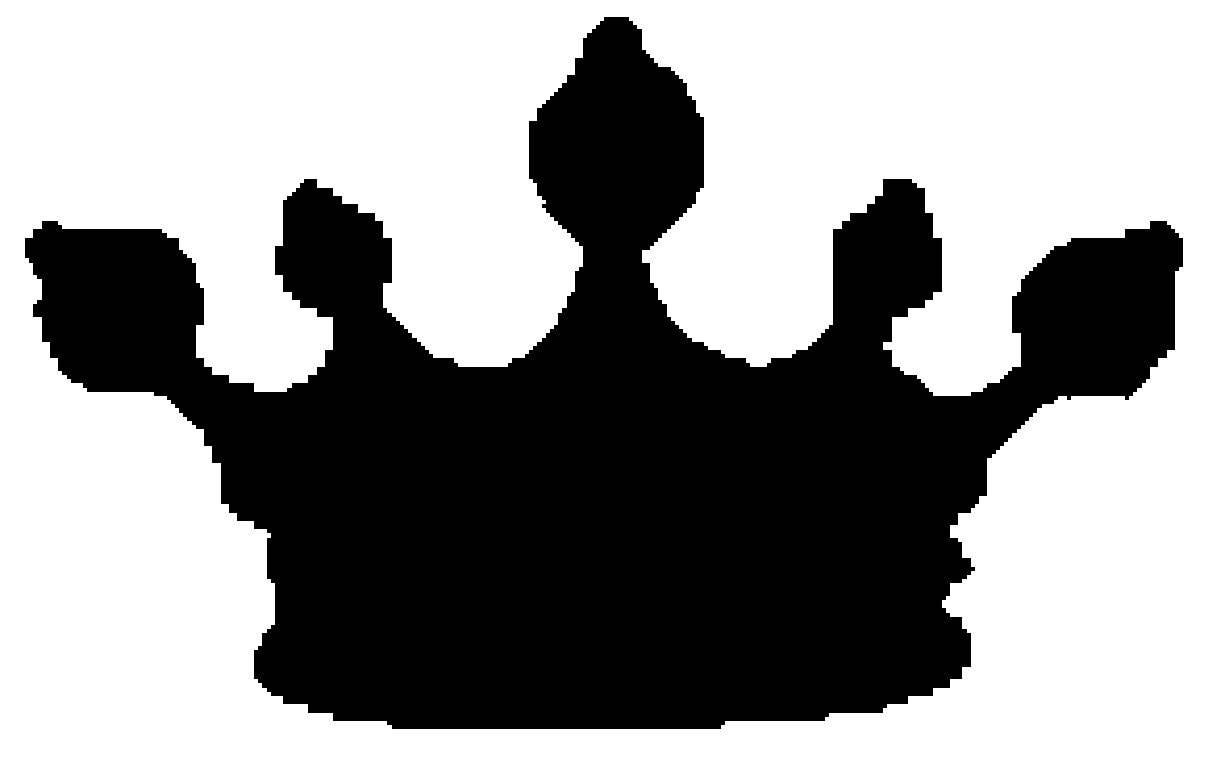,width=0.700000cm}&
\epsfig{figure=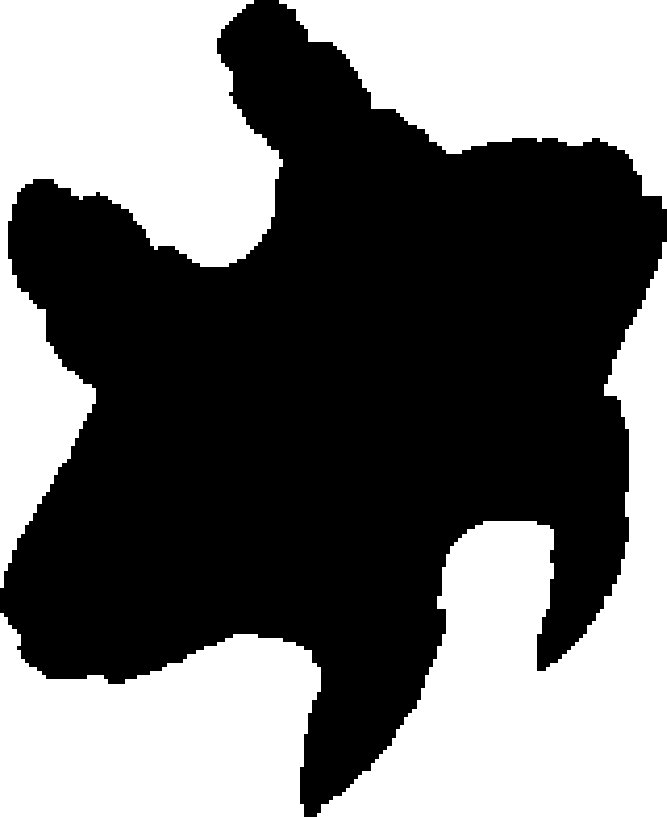,height=0.700000cm}&
\epsfig{figure=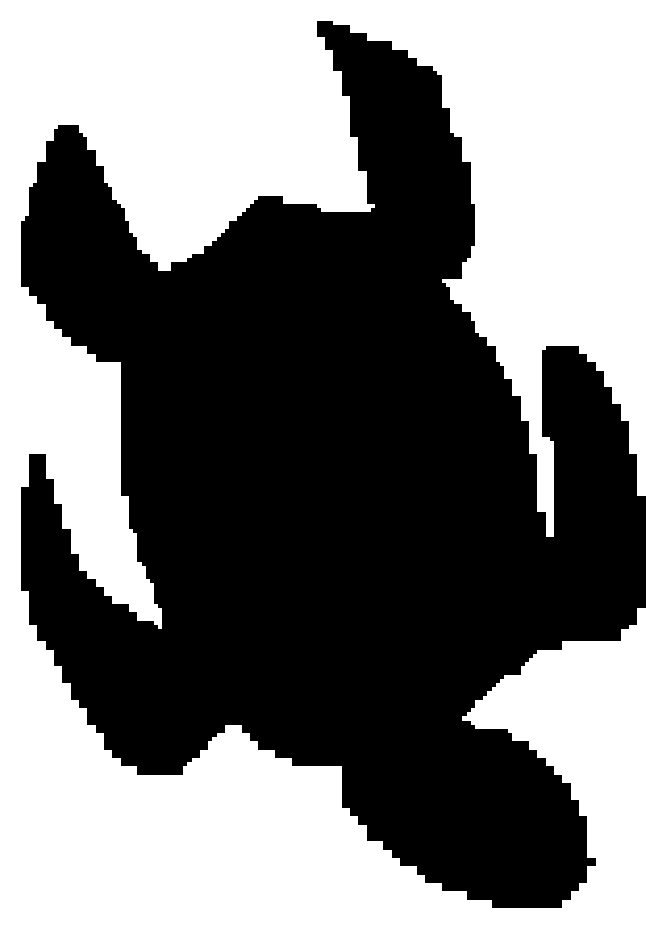,height=0.700000cm}&
\epsfig{figure=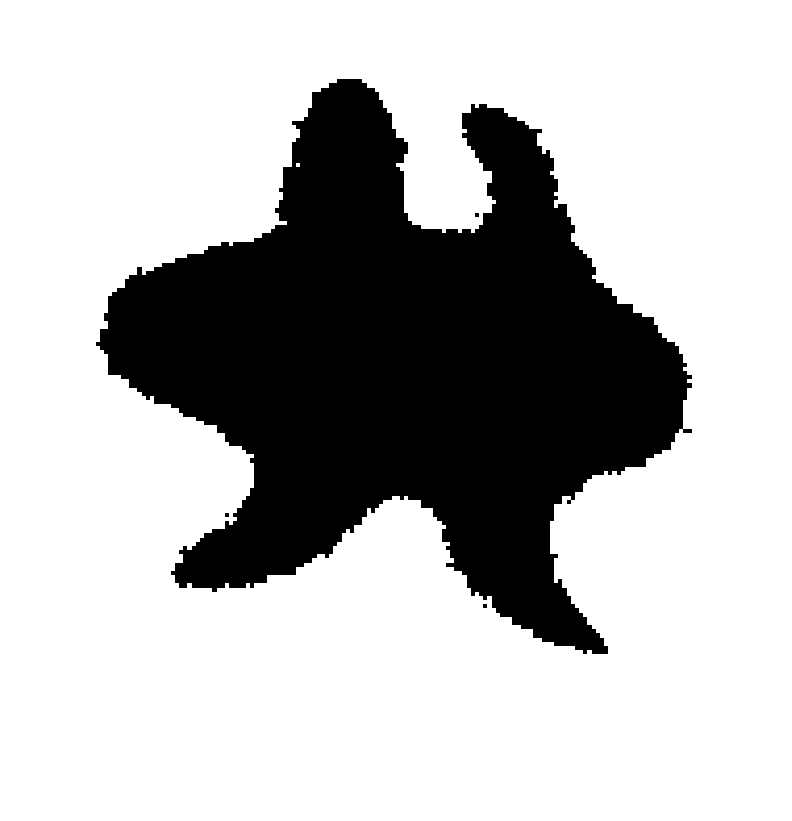,height=0.700000cm}&
\epsfig{figure=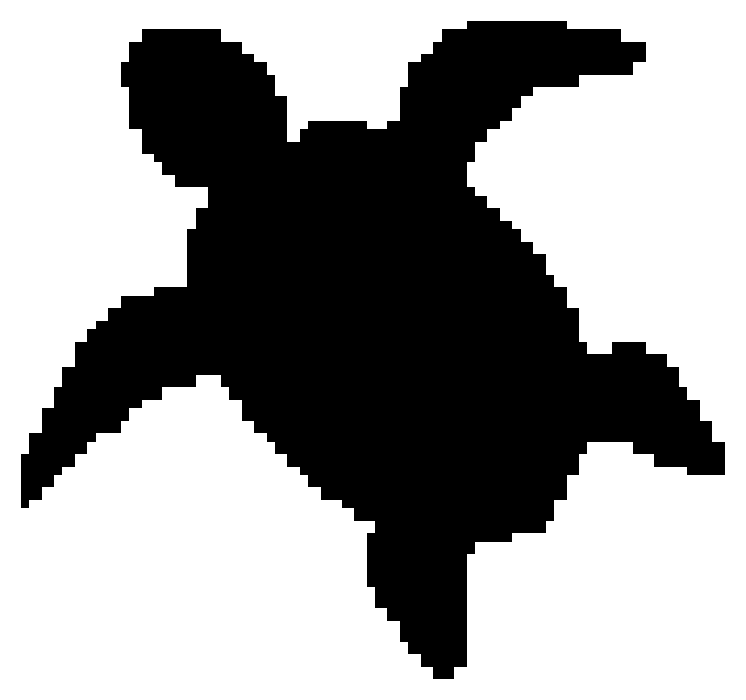,width=0.700000cm}\\
\cline{2-12}
  & \bf{0.882} & \bf{0.863} & \bf{0.823} & \bf{0.762} & \bf{0.738} & 0.569 & 0.566 & 0.530 & 0.524 & 0.516 & 0.515 \\
\hline \epsfig{figure=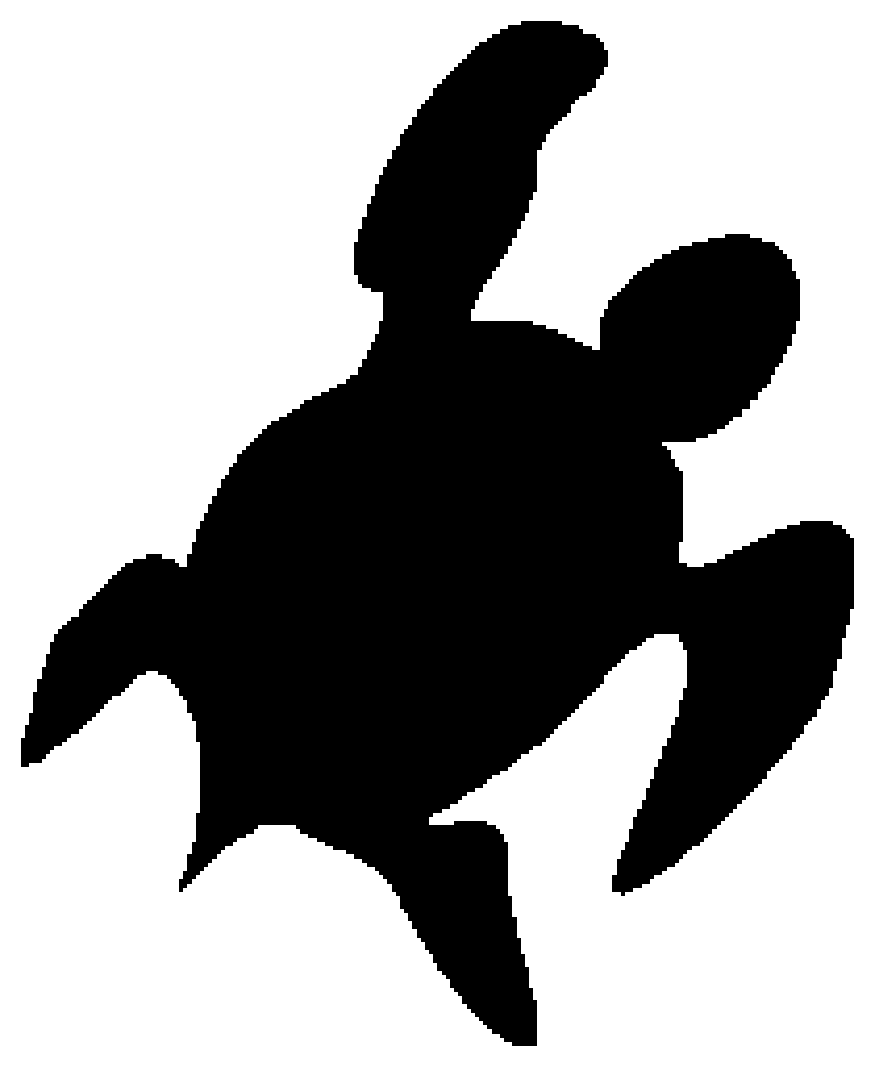,height=0.700000cm}&
\epsfig{figure=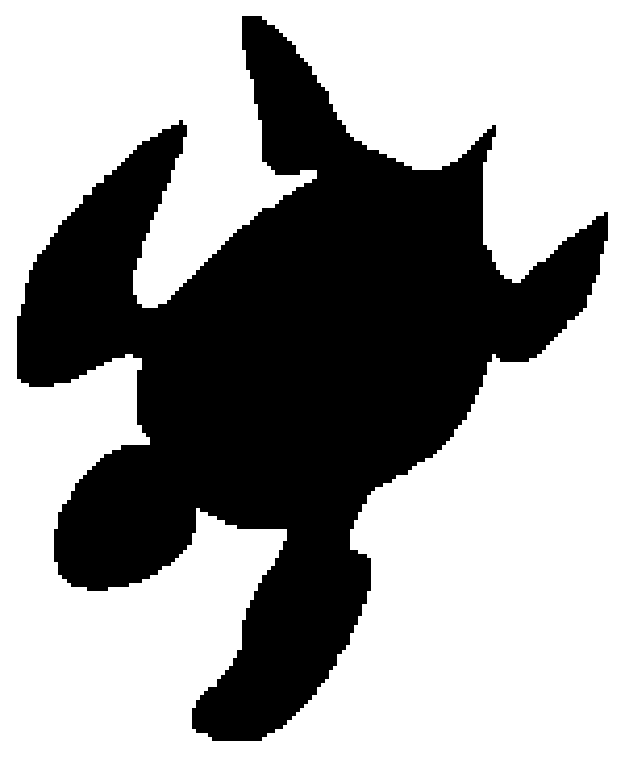,height=0.700000cm}&
\epsfig{figure=PAMI08_png/obj4_4.png,height=0.700000cm}&
\epsfig{figure=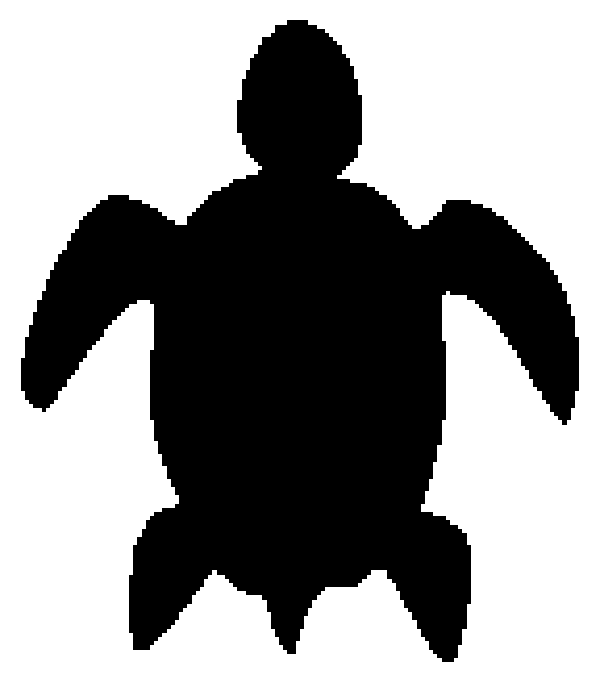,height=0.700000cm}&
\epsfig{figure=PAMI08_png/obj4_2.png,width=0.700000cm}&
\epsfig{figure=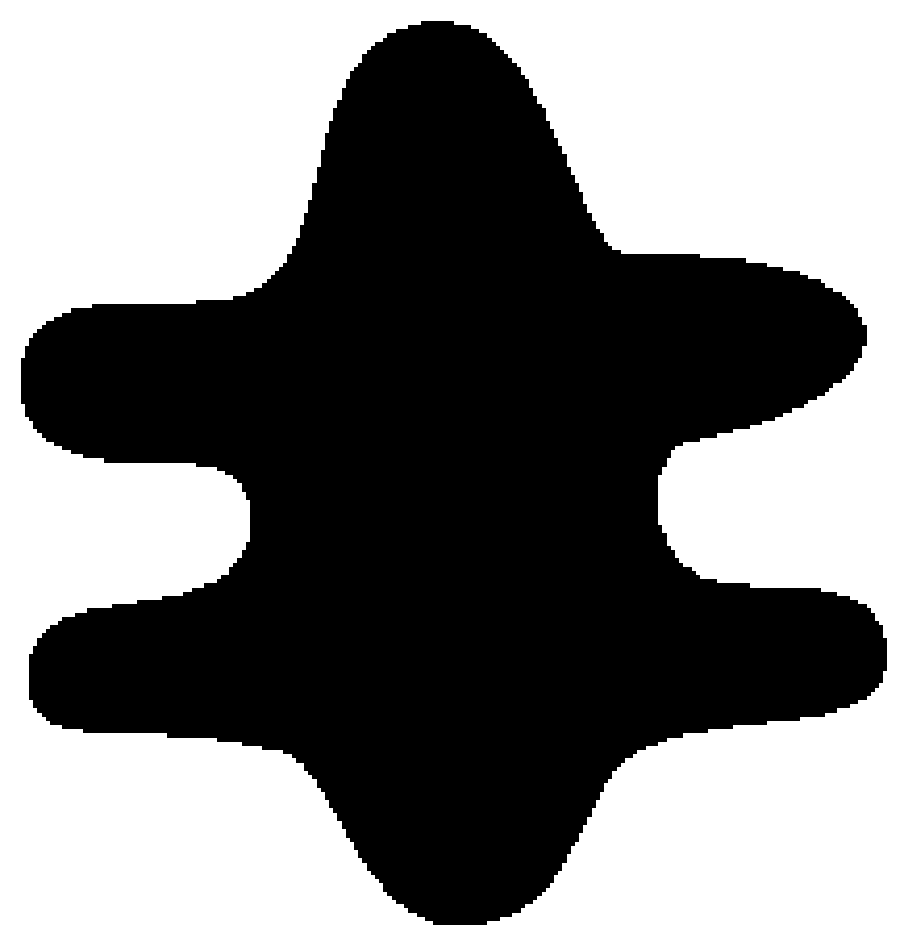,height=0.700000cm}&
\epsfig{figure=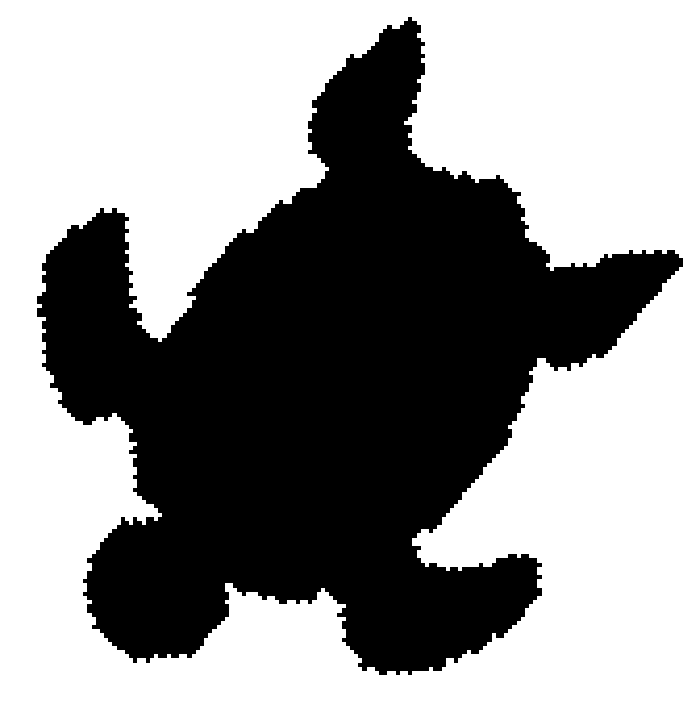,height=0.700000cm}&
\epsfig{figure=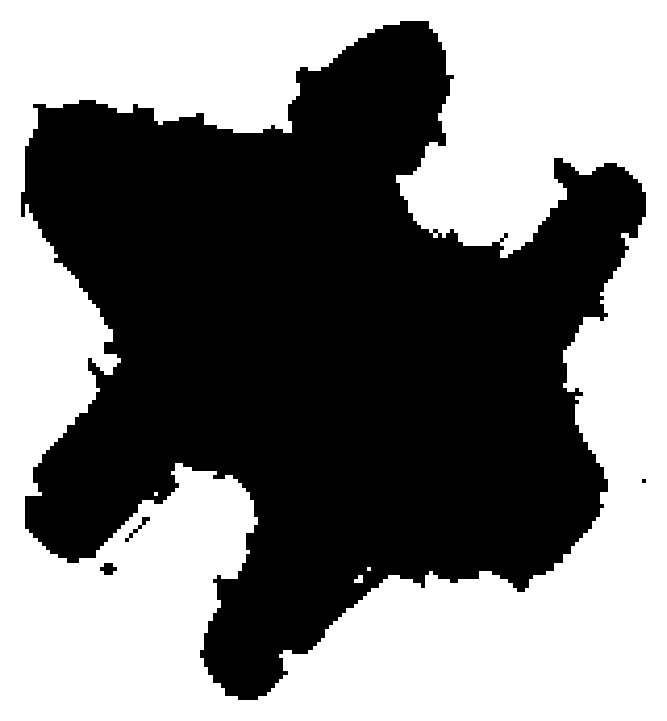,height=0.700000cm}&
\epsfig{figure=PAMI08_png/obj7_3.png,height=0.700000cm}&
\epsfig{figure=PAMI08_png/obj7_5.png,height=0.700000cm}&
\epsfig{figure=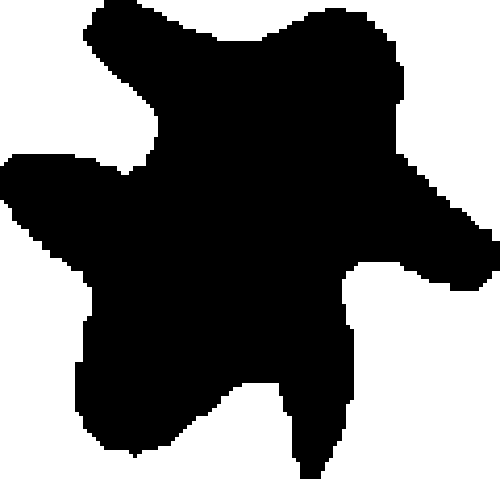,width=0.700000cm}&
\epsfig{figure=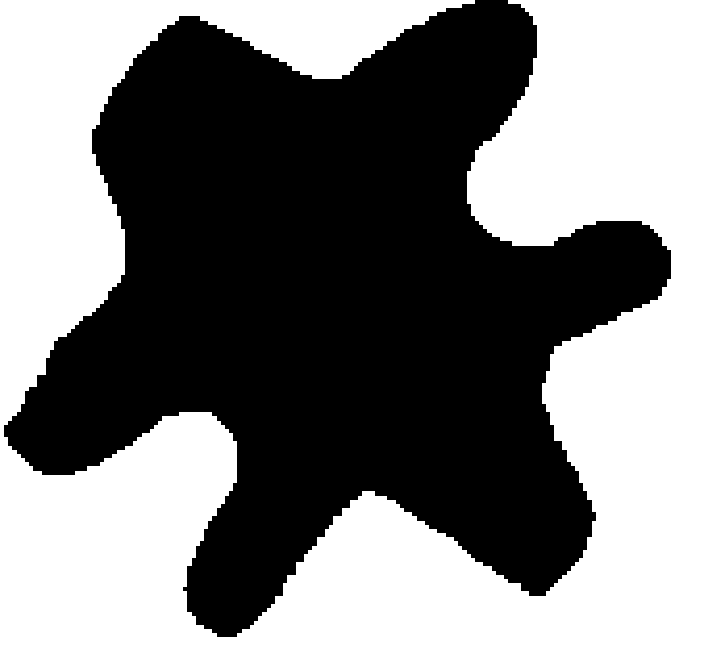,width=0.700000cm}\\
\cline{2-12}
  & \bf{0.985} & \bf{0.887} & \bf{0.866} & \bf{0.829} & 0.812 & \bf{0.775} & 0.754 & 0.750 & 0.748 & 0.706 & 0.670 \\
\hline \epsfig{figure=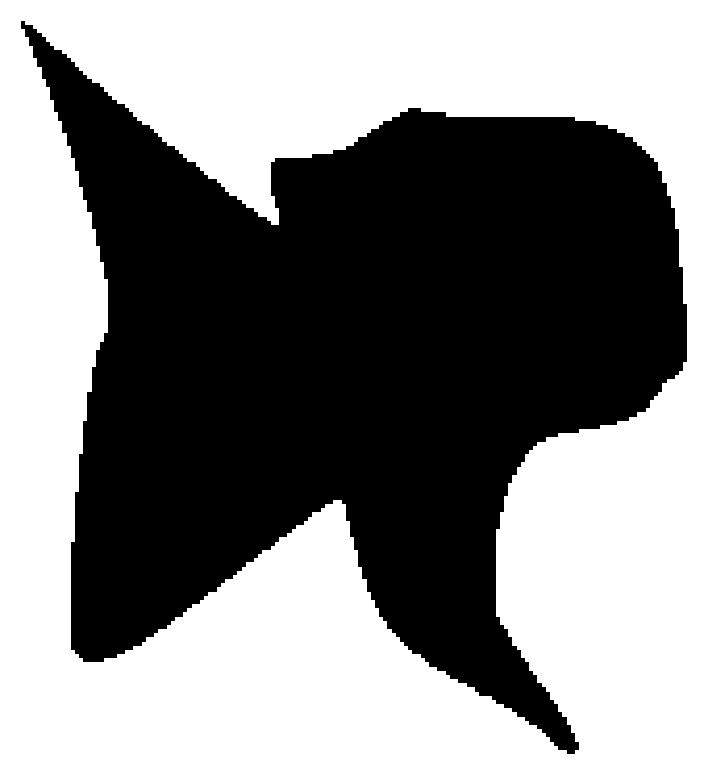,height=0.700000cm}&
\epsfig{figure=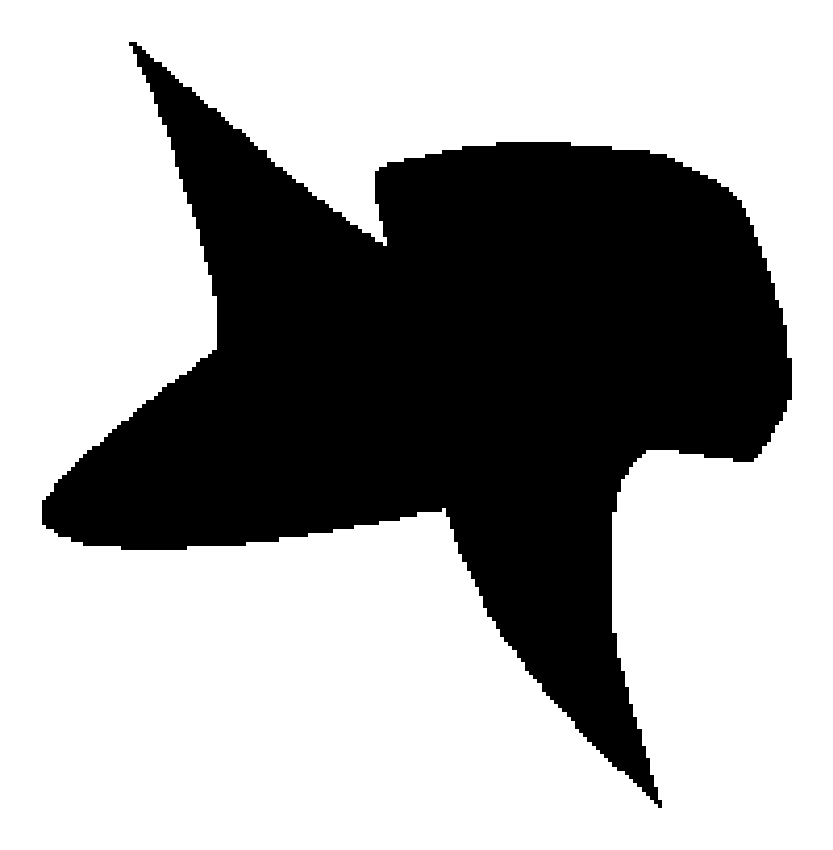,height=0.700000cm}&
\epsfig{figure=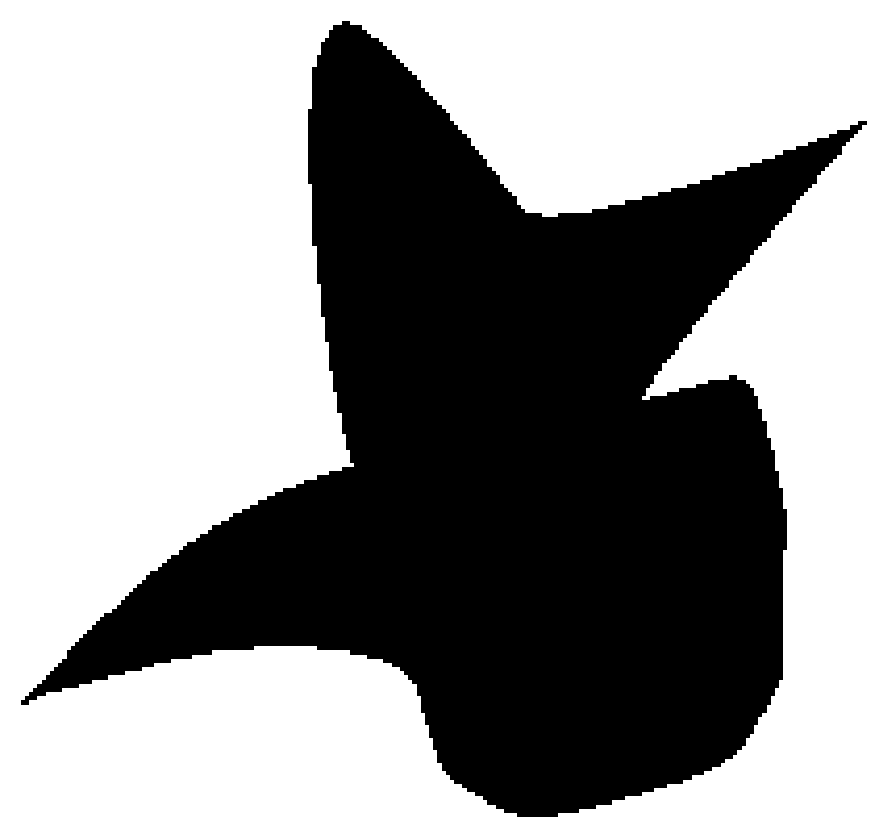,width=0.700000cm}&
\epsfig{figure=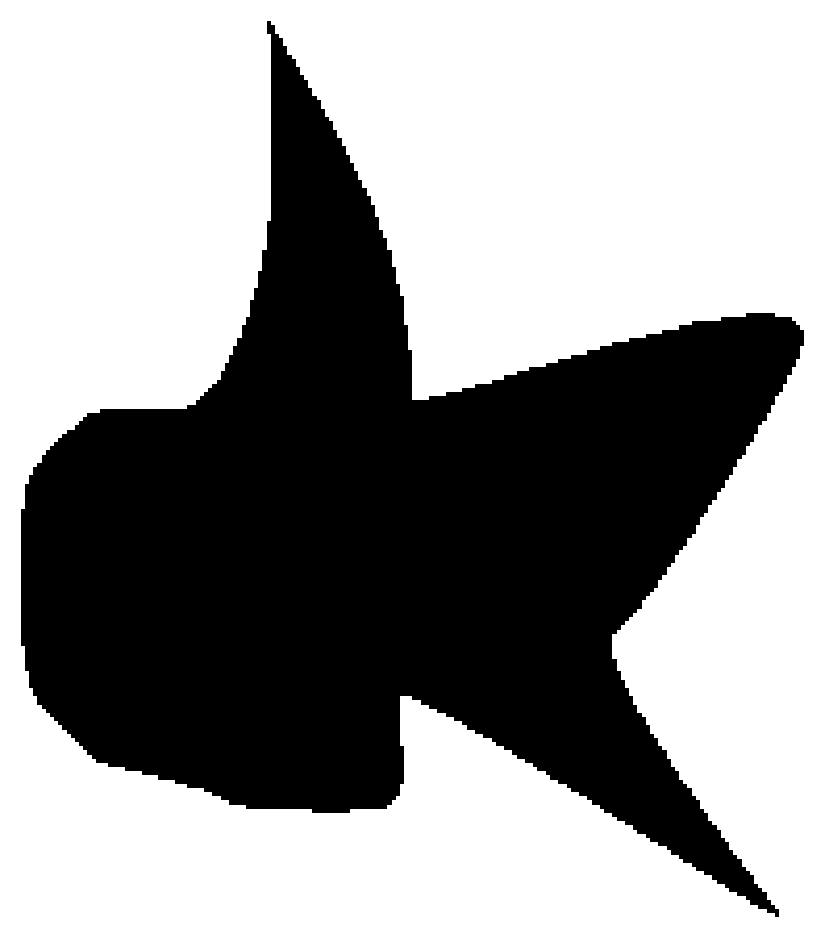,height=0.700000cm}&
\epsfig{figure=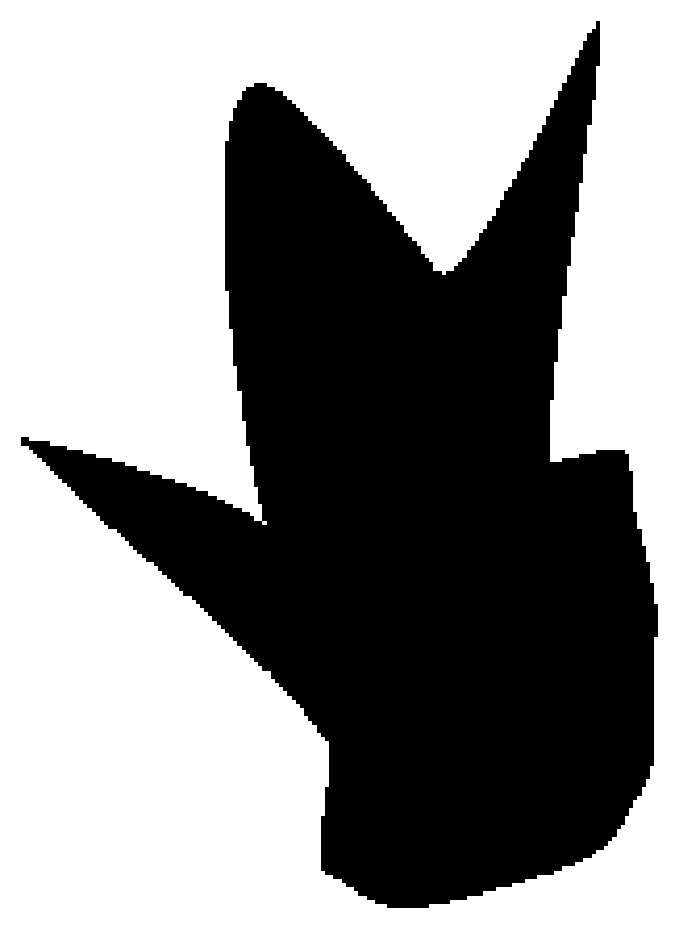,height=0.700000cm}&
\epsfig{figure=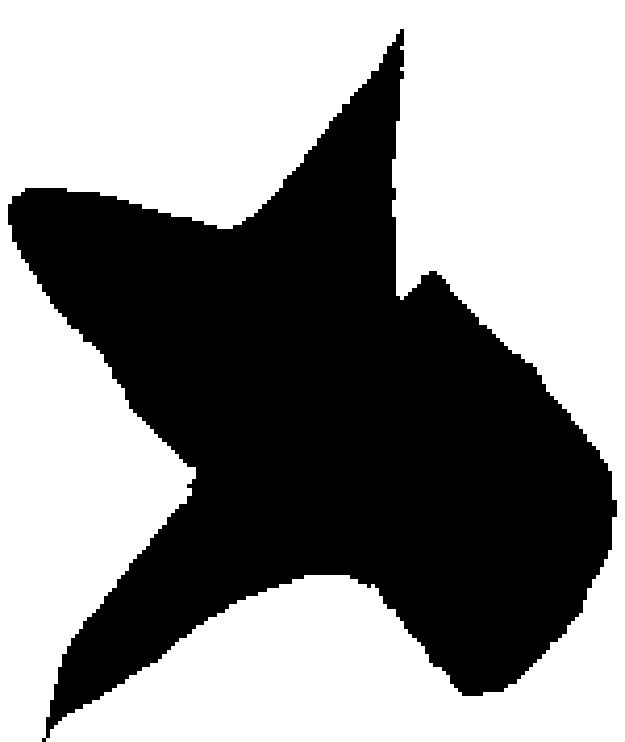,height=0.700000cm}&
\epsfig{figure=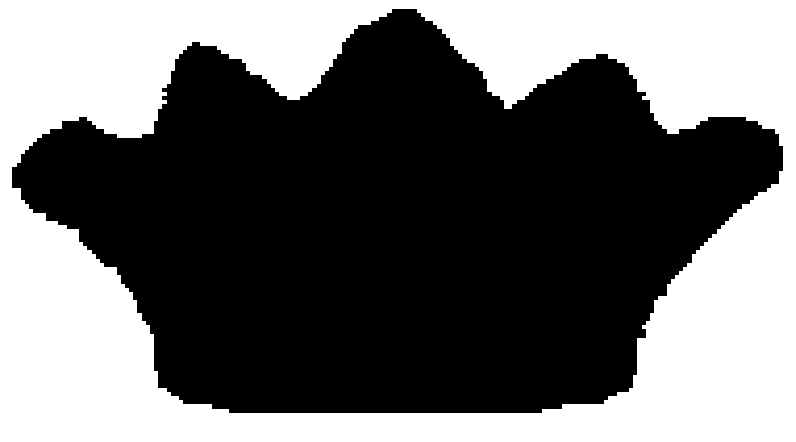,width=0.700000cm}&
\epsfig{figure=PAMI08_png/obj26_3.png,height=0.700000cm}&
\epsfig{figure=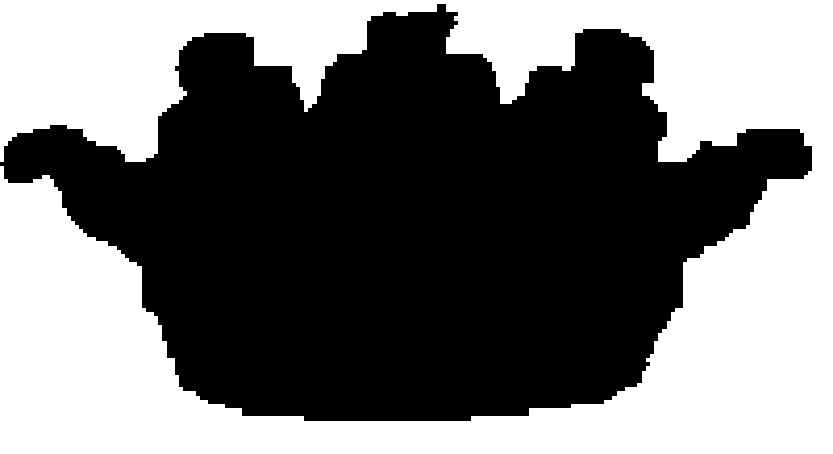,width=0.700000cm}&
\epsfig{figure=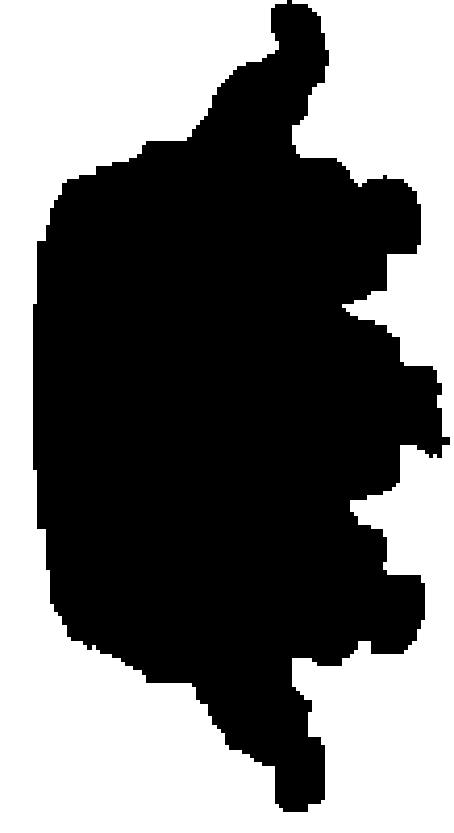,height=0.700000cm}&
\epsfig{figure=PAMI08_png/obj26_1.png,width=0.700000cm}&
\epsfig{figure=PAMI08_png/obj4_1.png,height=0.700000cm}\\
\cline{2-12}
  & \bf{0.930} & \bf{0.908} & \bf{0.907} & \bf{0.876} & \bf{0.812} & 0.582 & 0.568 & 0.568 & 0.567 & 0.561 & 0.521 \\
\hline \epsfig{figure=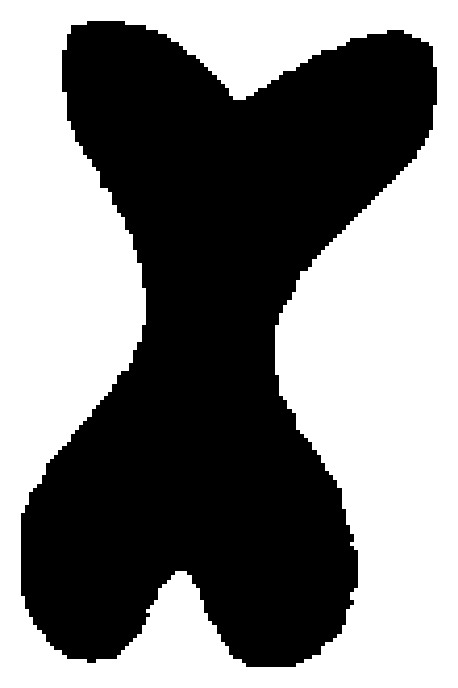,height=0.700000cm}&
\epsfig{figure=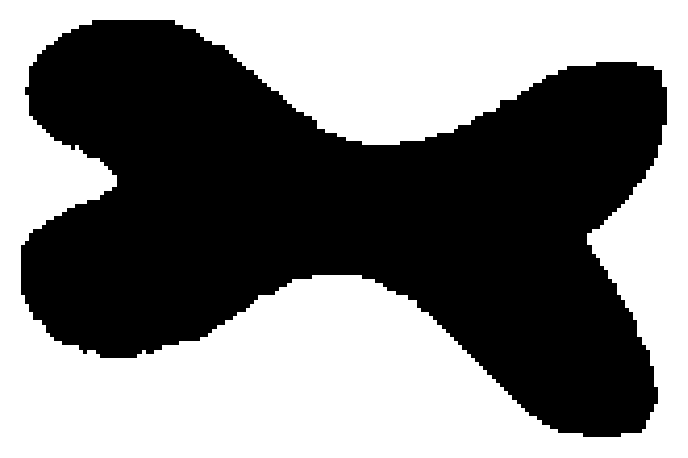,width=0.700000cm}&
\epsfig{figure=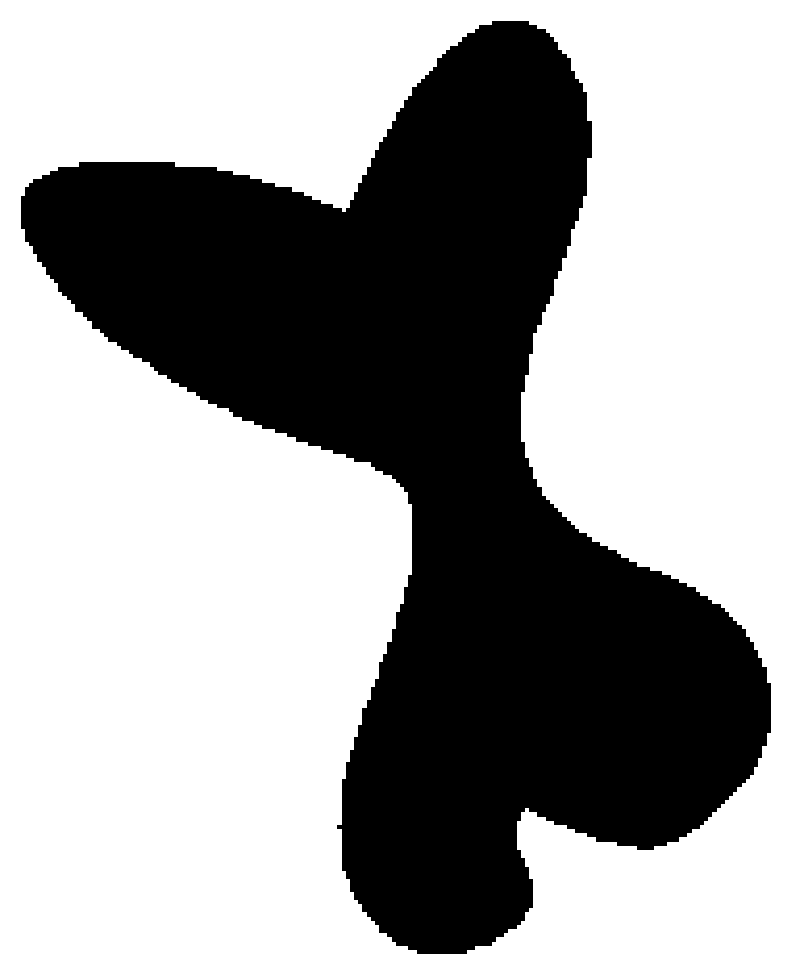,height=0.700000cm}&
\epsfig{figure=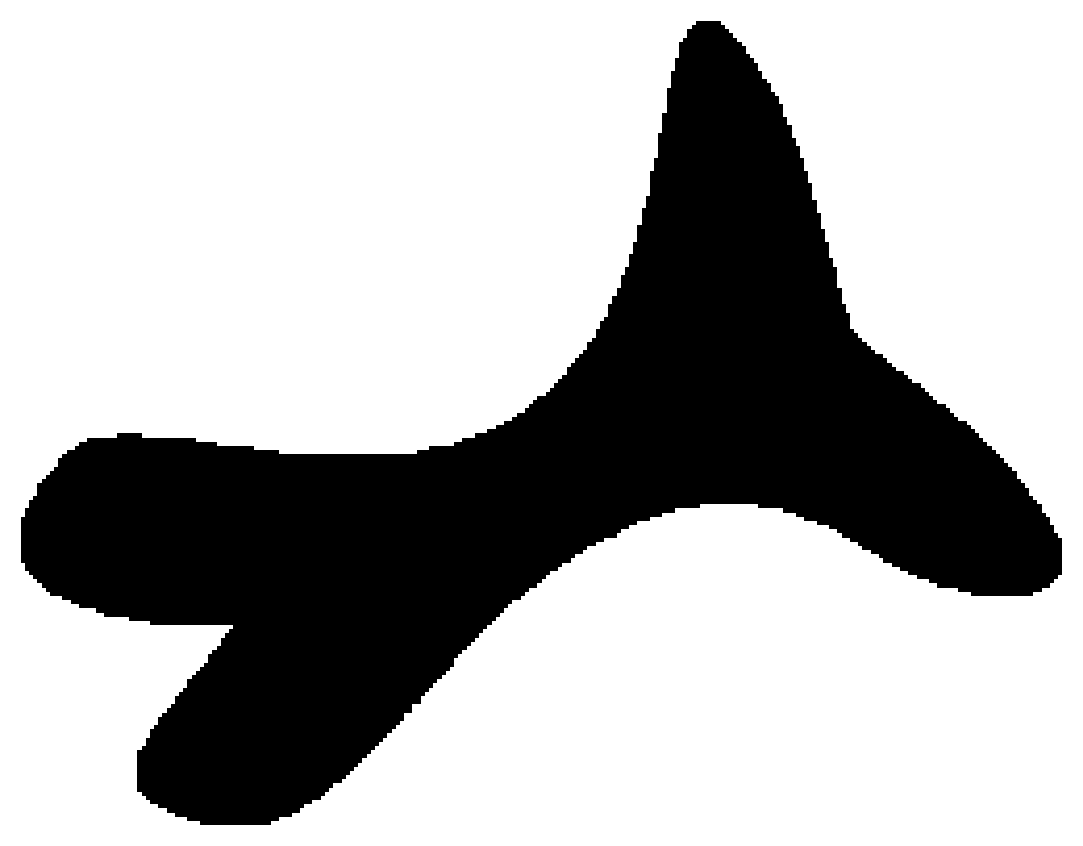,width=0.700000cm}&
\epsfig{figure=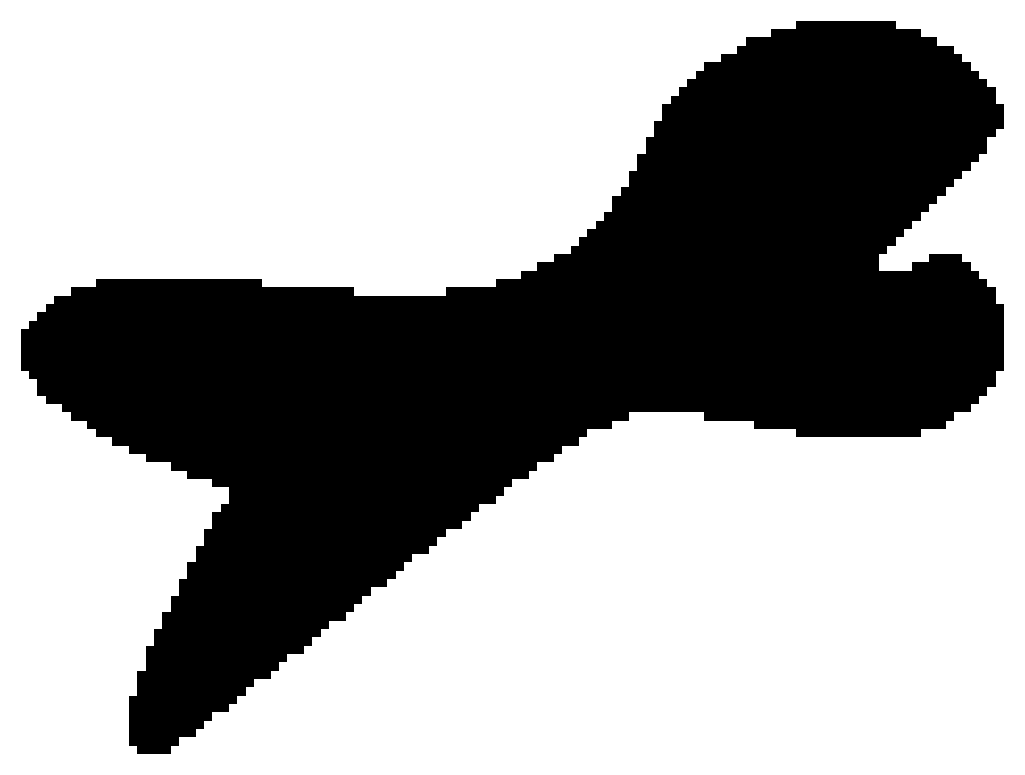,width=0.700000cm}&
\epsfig{figure=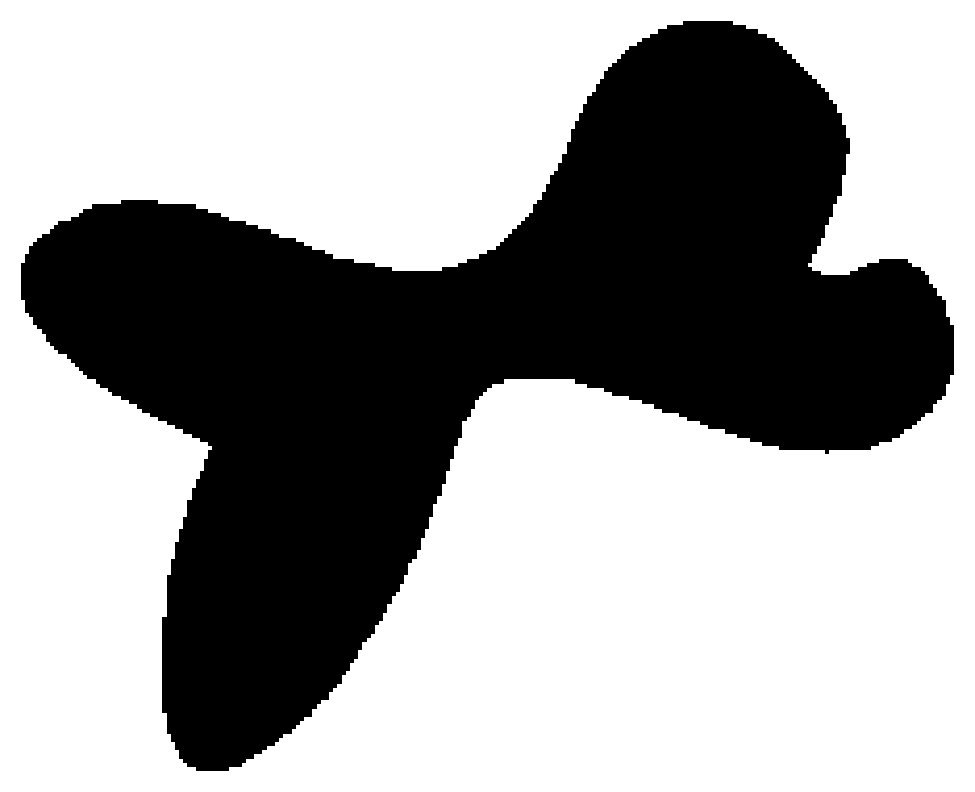,width=0.700000cm}&
\epsfig{figure=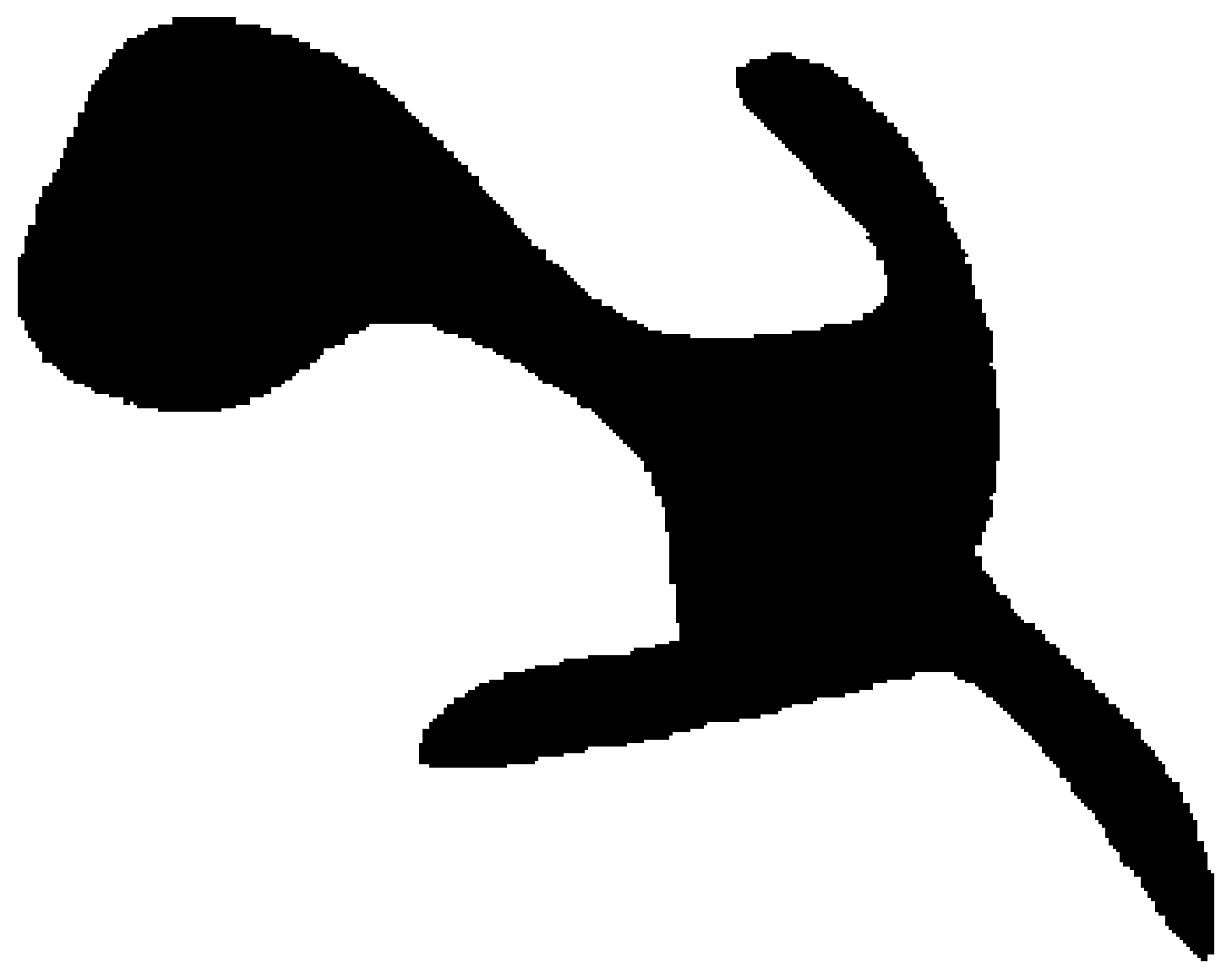,width=0.700000cm}&
\epsfig{figure=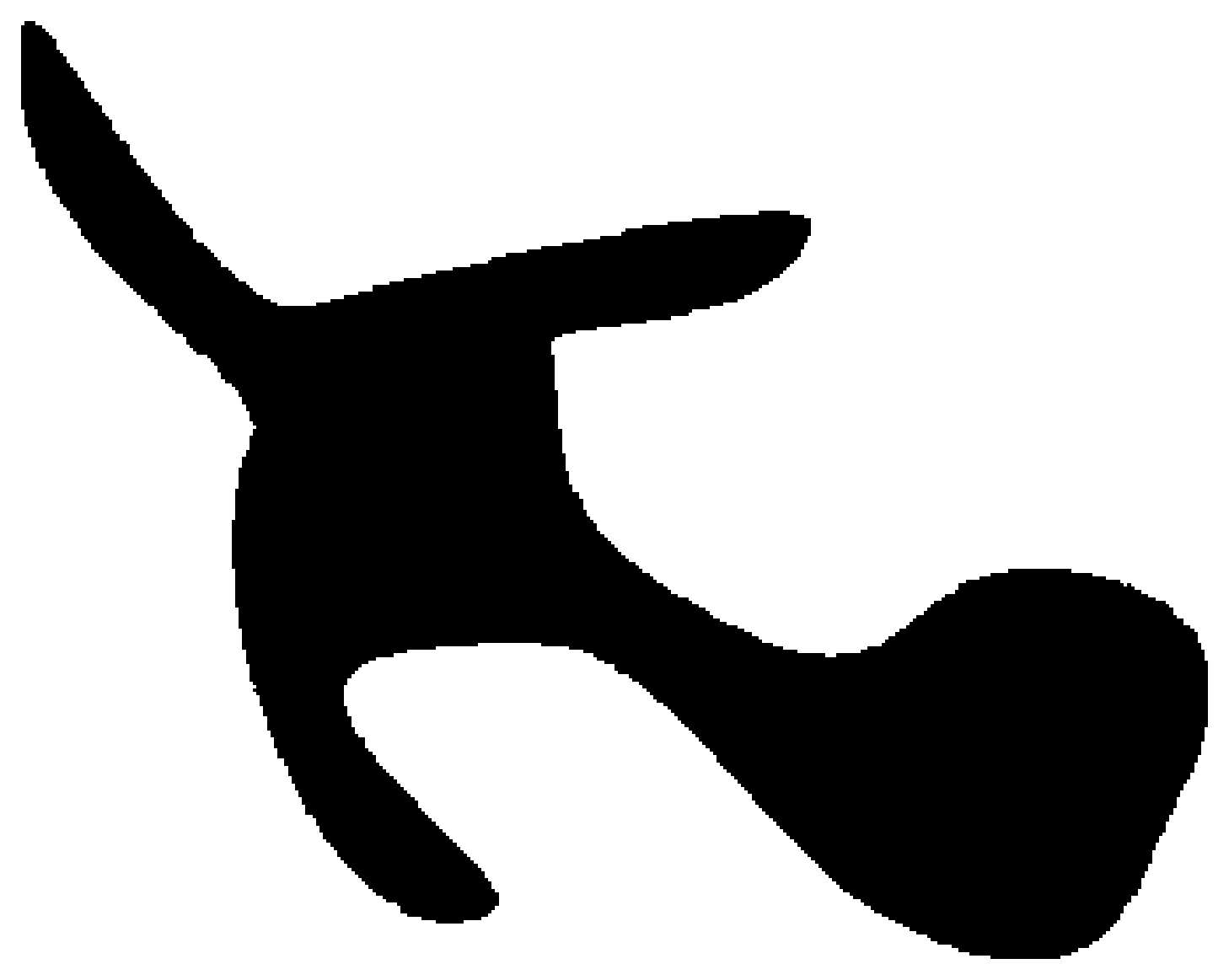,width=0.700000cm}&
\epsfig{figure=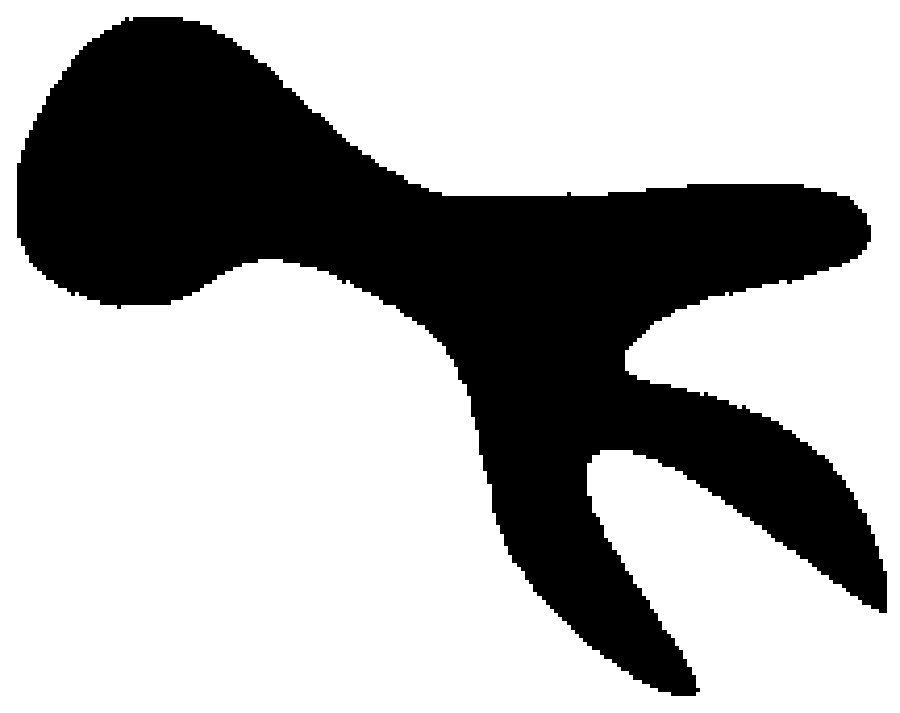,width=0.700000cm}&
\epsfig{figure=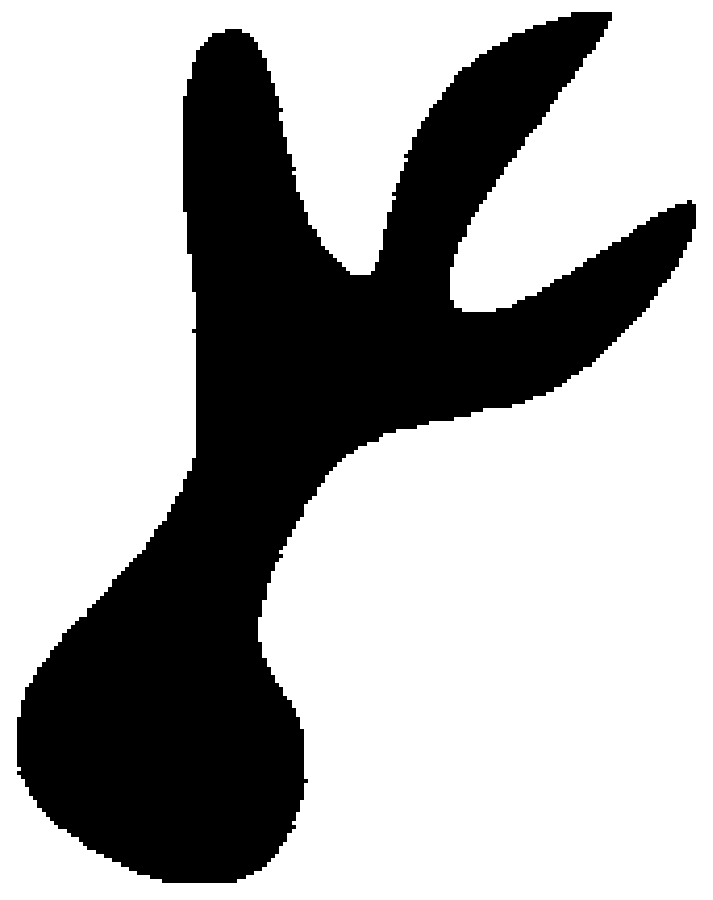,height=0.700000cm}&
\epsfig{figure=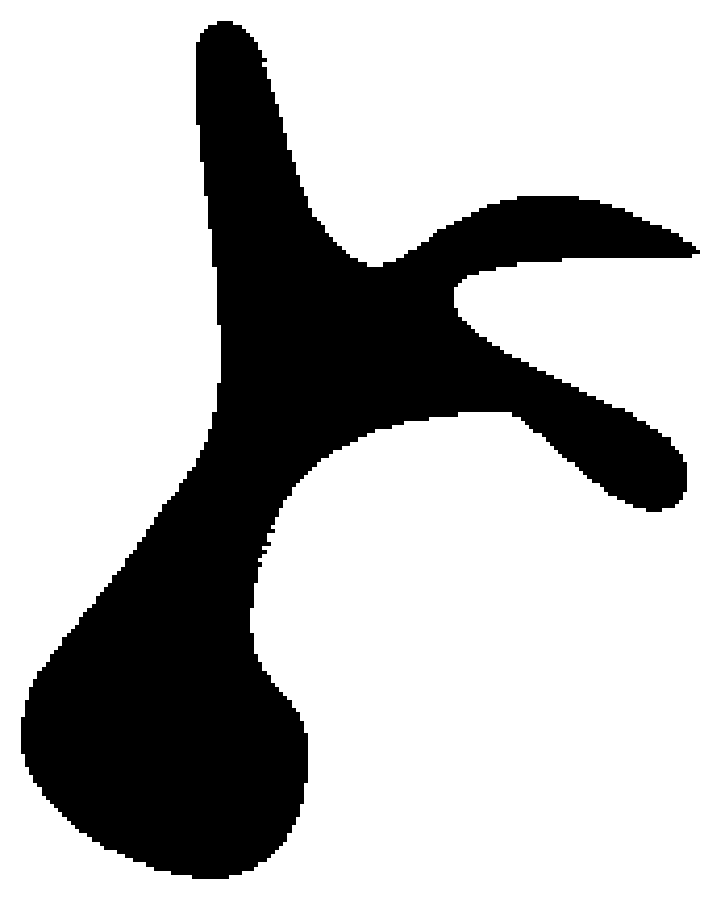,height=0.700000cm}&
\epsfig{figure=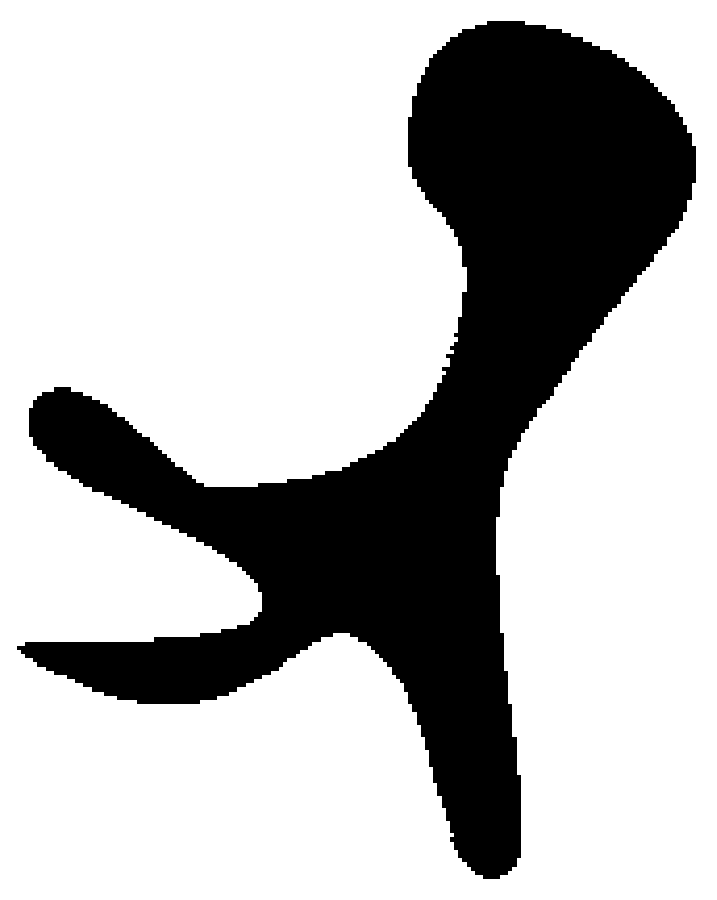,height=0.700000cm}\\
\cline{2-12}
  & \bf{0.969} & \bf{0.881} & \bf{0.881} & \bf{0.844} & \bf{0.833} & 0.628 & 0.607 & 0.528 & 0.506 & 0.499 & 0.491 \\
\hline \epsfig{figure=PAMI08_png/obj7_2.png,height=0.700000cm}&
\epsfig{figure=PAMI08_png/obj7_3.png,height=0.700000cm}&
\epsfig{figure=PAMI08_png/obj7_5.png,height=0.700000cm}&
\epsfig{figure=PAMI08_png/obj7_1.png,height=0.700000cm}&
\epsfig{figure=PAMI08_png/obj7_6.png,width=0.700000cm}&
\epsfig{figure=PAMI08_png/obj4_5.png,height=0.700000cm}&
\epsfig{figure=PAMI08_png/obj7_4.png,width=0.700000cm}&
\epsfig{figure=PAMI08_png/obj4_2.png,width=0.700000cm}&
\epsfig{figure=PAMI08_png/obj4_1.png,height=0.700000cm}&
\epsfig{figure=PAMI08_png/obj4_4.png,height=0.700000cm}&
\epsfig{figure=PAMI08_png/obj4_3.png,height=0.700000cm}&
\epsfig{figure=PAMI08_png/obj4_6.png,height=0.700000cm}\\
\cline{2-12}
  & \bf{0.967} & \bf{0.965} & \bf{0.951} & \bf{0.945} & 0.874 & \bf{0.817} & 0.807 & 0.788 & 0.773 & 0.754 & 0.740 \\
\hline \epsfig{figure=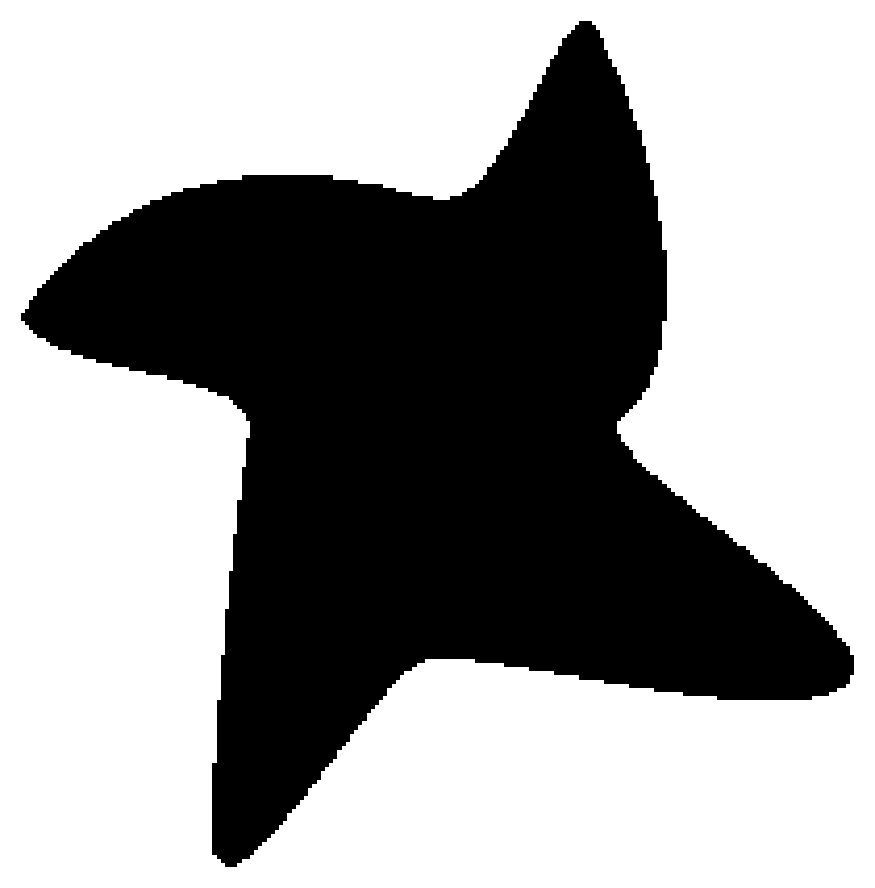,height=0.700000cm}&
\epsfig{figure=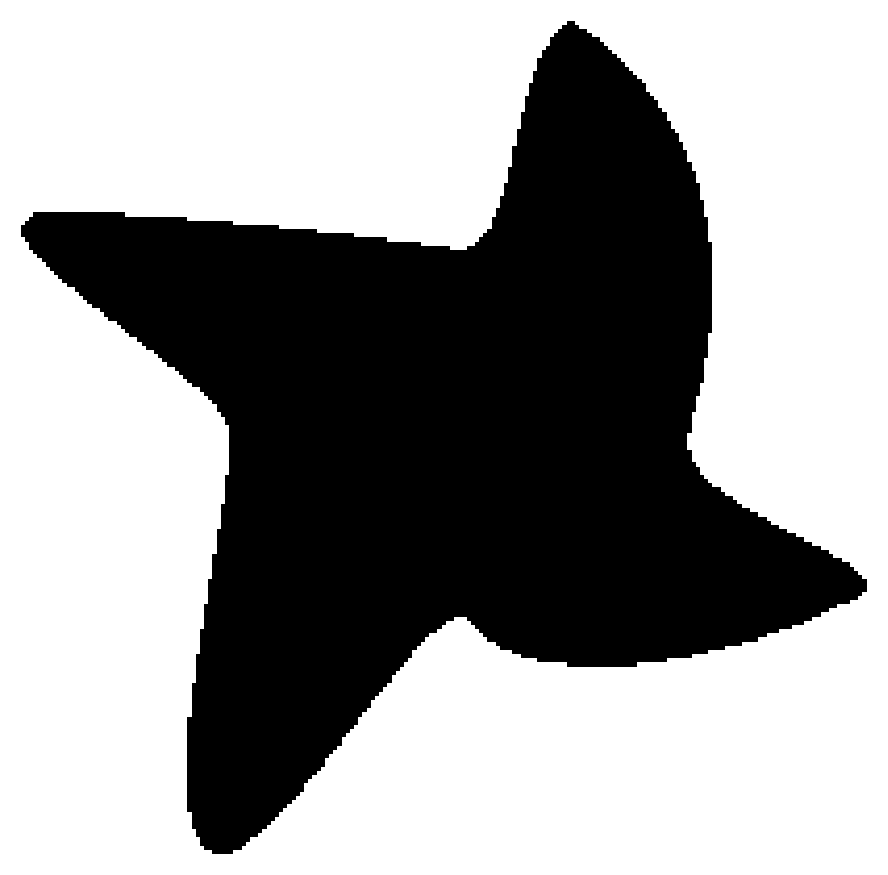,width=0.700000cm}&
\epsfig{figure=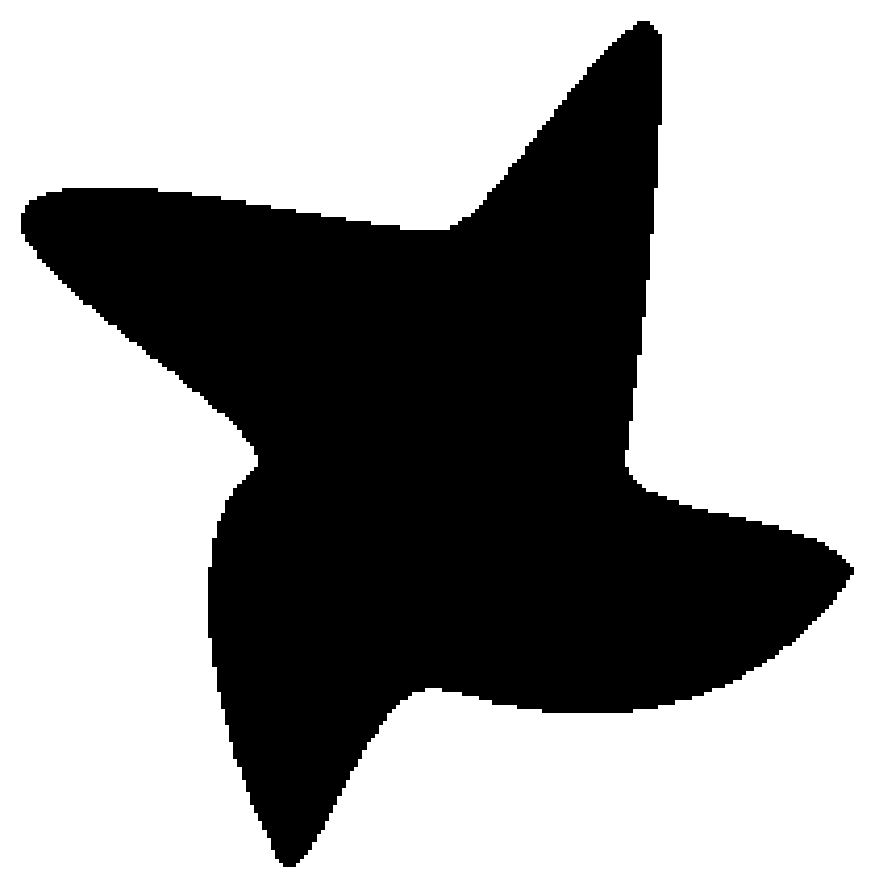,height=0.700000cm}&
\epsfig{figure=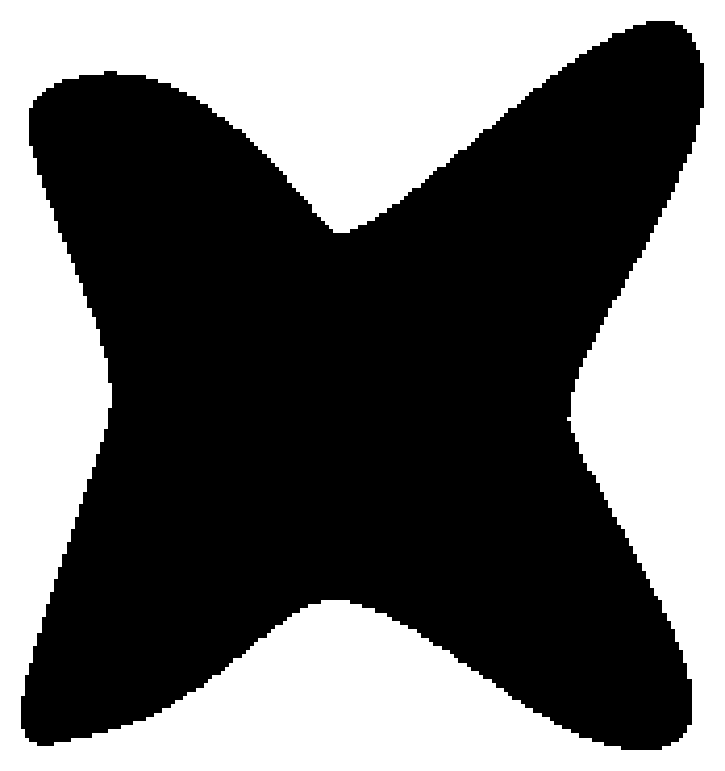,height=0.700000cm}&
\epsfig{figure=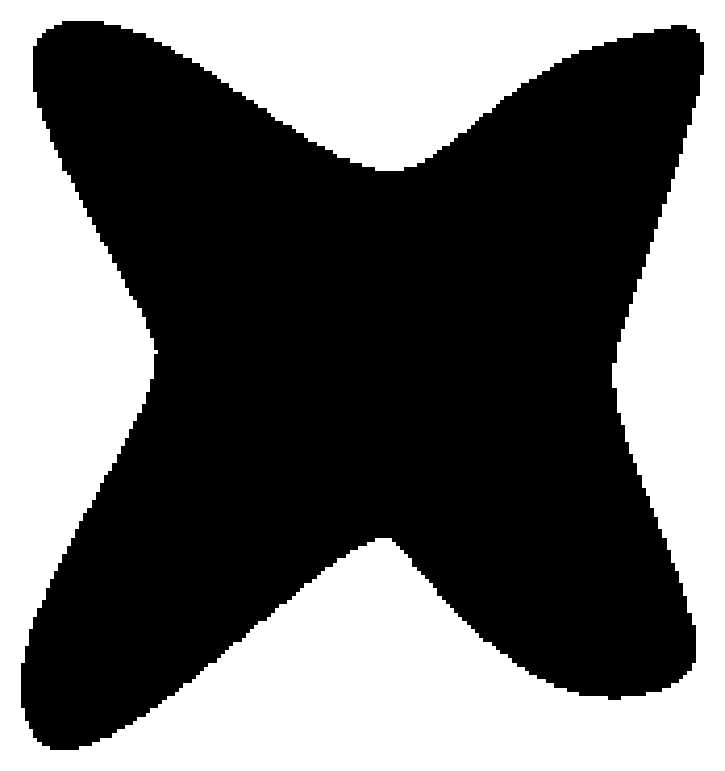,height=0.700000cm}&
\epsfig{figure=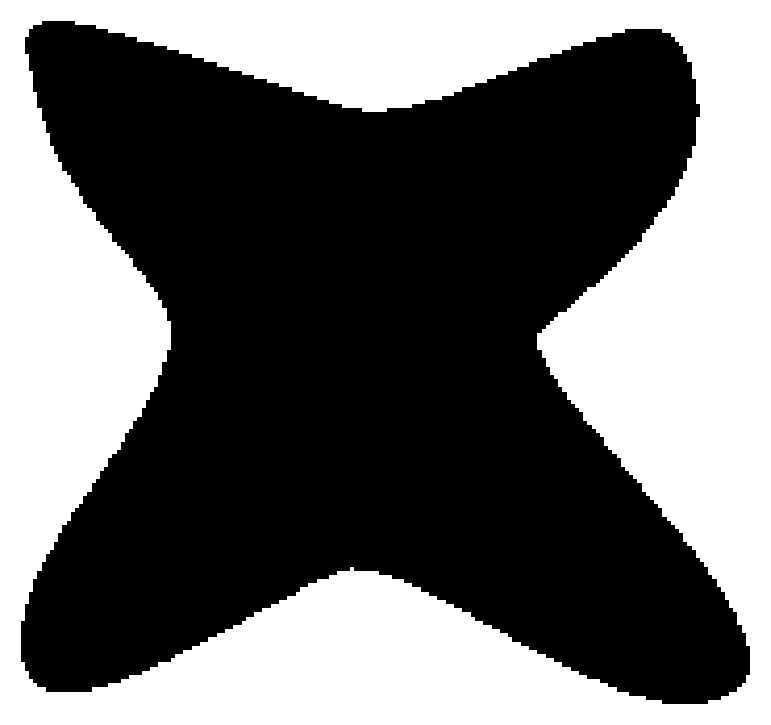,width=0.700000cm}&
\epsfig{figure=PAMI08_png/obj2_6.png,width=0.700000cm}&
\epsfig{figure=PAMI08_png/obj2_2.png,height=0.700000cm}&
\epsfig{figure=PAMI08_png/obj10_2.png,height=0.700000cm}&
\epsfig{figure=PAMI08_png/obj2_1.png,height=0.700000cm}&
\epsfig{figure=PAMI08_png/obj2_5.png,width=0.700000cm}&
\epsfig{figure=PAMI08_png/obj2_4.png,height=0.700000cm}\\
\cline{2-12}
  & \bf{0.997} & \bf{0.995} & \bf{0.894} & \bf{0.889} & \bf{0.887} & 0.691 & 0.687 & 0.670 & 0.660 & 0.643 & 0.642 \\
\hline \epsfig{figure=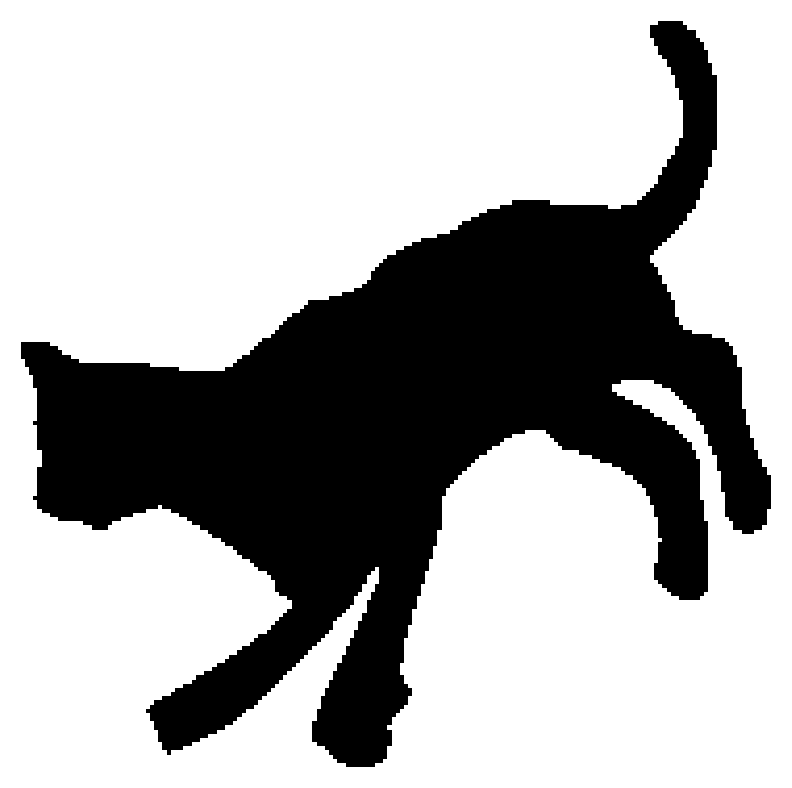,width=0.700000cm}&
\epsfig{figure=PAMI08_png/obj24_2.png,width=0.700000cm}&
\epsfig{figure=PAMI08_png/obj9_2.png,width=0.700000cm}&
\epsfig{figure=PAMI08_png/obj24_3.png,width=0.700000cm}&
\epsfig{figure=PAMI08_png/obj24_4.png,height=0.700000cm}&
\epsfig{figure=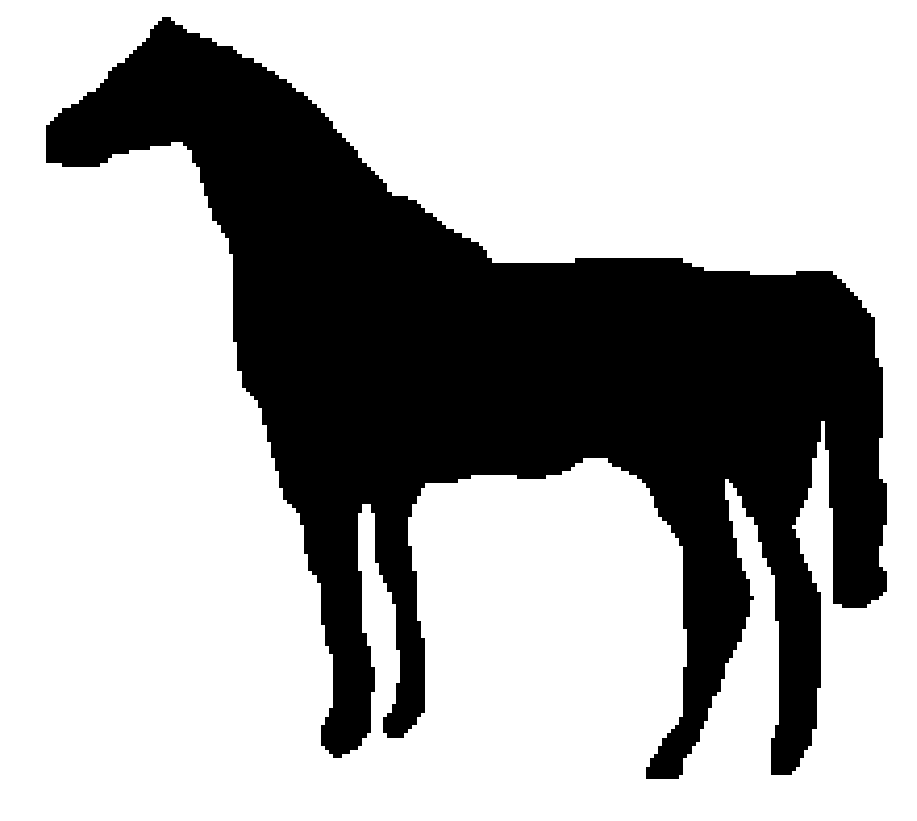,width=0.700000cm}&
\epsfig{figure=PAMI08_png/obj24_5.png,width=0.700000cm}&
\epsfig{figure=PAMI08_png/obj9_6.png,height=0.700000cm}&
\epsfig{figure=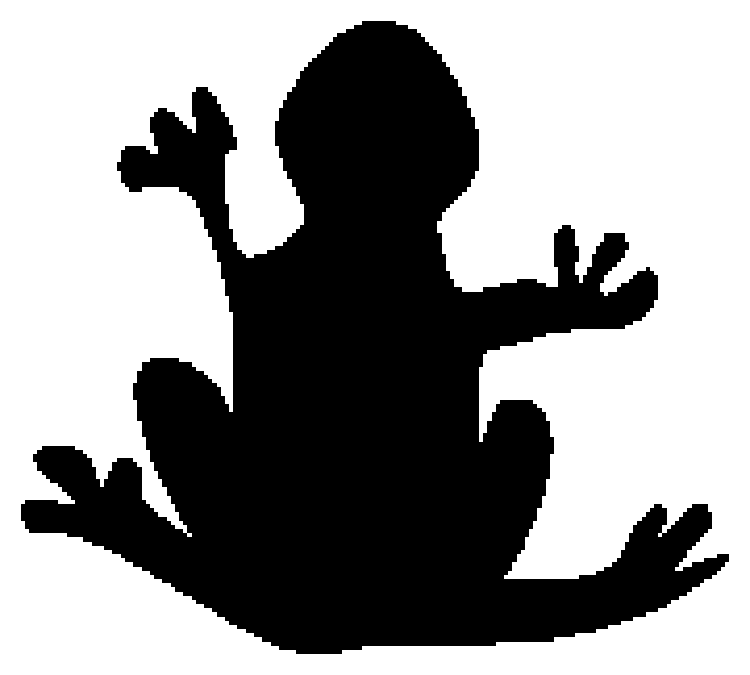,width=0.700000cm}&
\epsfig{figure=PAMI08_png/obj4_4.png,height=0.700000cm}&
\epsfig{figure=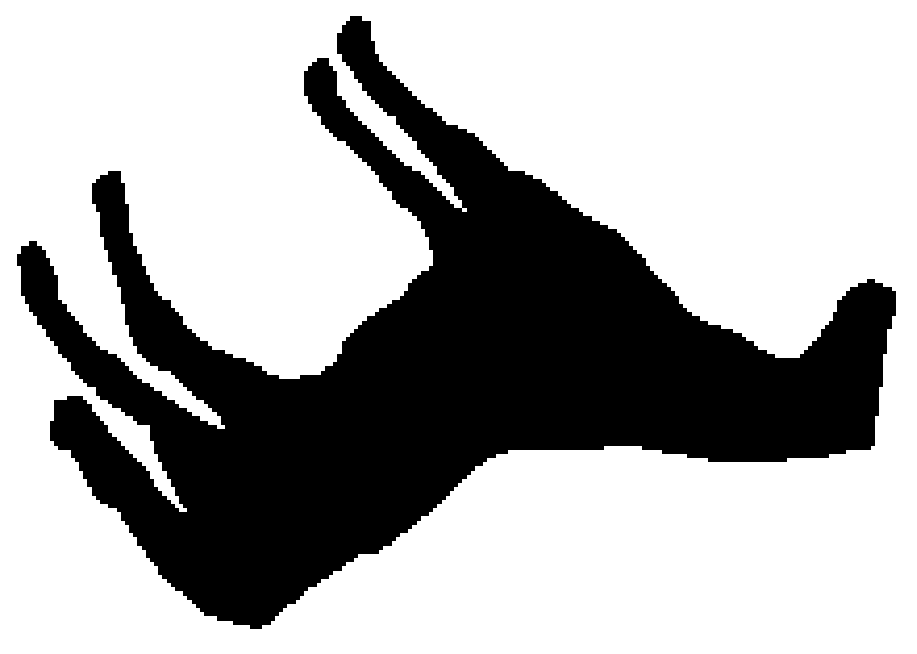,width=0.700000cm}&
\epsfig{figure=PAMI08_png/obj7_4.png,width=0.700000cm}\\
\cline{2-12}
  & 0.897 & \bf{0.849} & 0.803 & 0.776 & 0.752 & 0.745 & \bf{0.692} & 0.647 & 0.639 & 0.638 & 0.635 \\
\hline \epsfig{figure=PAMI08_png/obj10_3.png,height=0.700000cm}&
\epsfig{figure=PAMI08_png/obj10_5.png,width=0.700000cm}&
\epsfig{figure=PAMI08_png/obj2_3.png,height=0.700000cm}&
\epsfig{figure=PAMI08_png/obj10_1.png,width=0.700000cm}&
\epsfig{figure=PAMI08_png/obj10_4.png,height=0.700000cm}&
\epsfig{figure=PAMI08_png/obj10_2.png,height=0.700000cm}&
\epsfig{figure=PAMI08_png/obj10_6.png,width=0.700000cm}&
\epsfig{figure=PAMI08_png/obj2_6.png,width=0.700000cm}&
\epsfig{figure=PAMI08_png/obj2_1.png,height=0.700000cm}&
\epsfig{figure=PAMI08_png/obj2_2.png,height=0.700000cm}&
\epsfig{figure=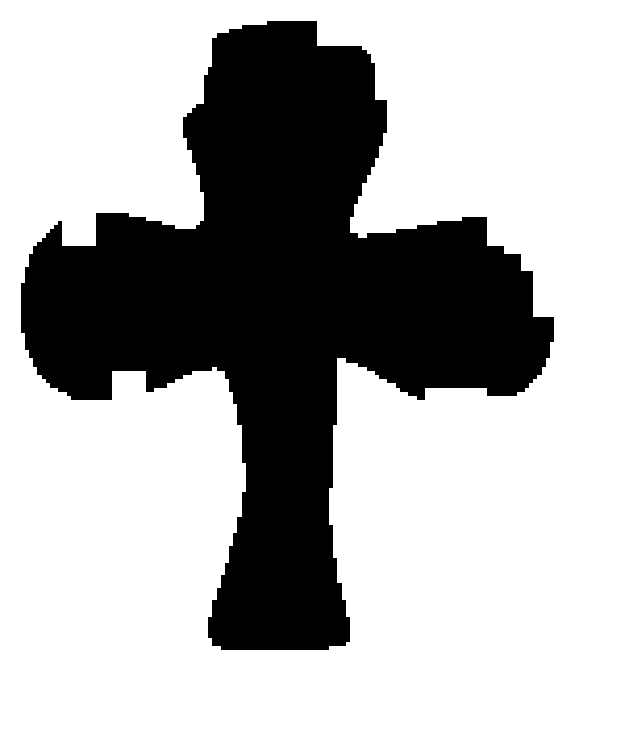,height=0.700000cm}&
\epsfig{figure=PAMI08_png/obj2_4.png,height=0.700000cm}\\
\cline{2-12}
  & \bf{0.914} & 0.869 & \bf{0.836} & \bf{0.835} & \bf{0.807} & \bf{0.802} & 0.790 & 0.772 & 0.729 & 0.697 & 0.696 \\
\hline \epsfig{figure=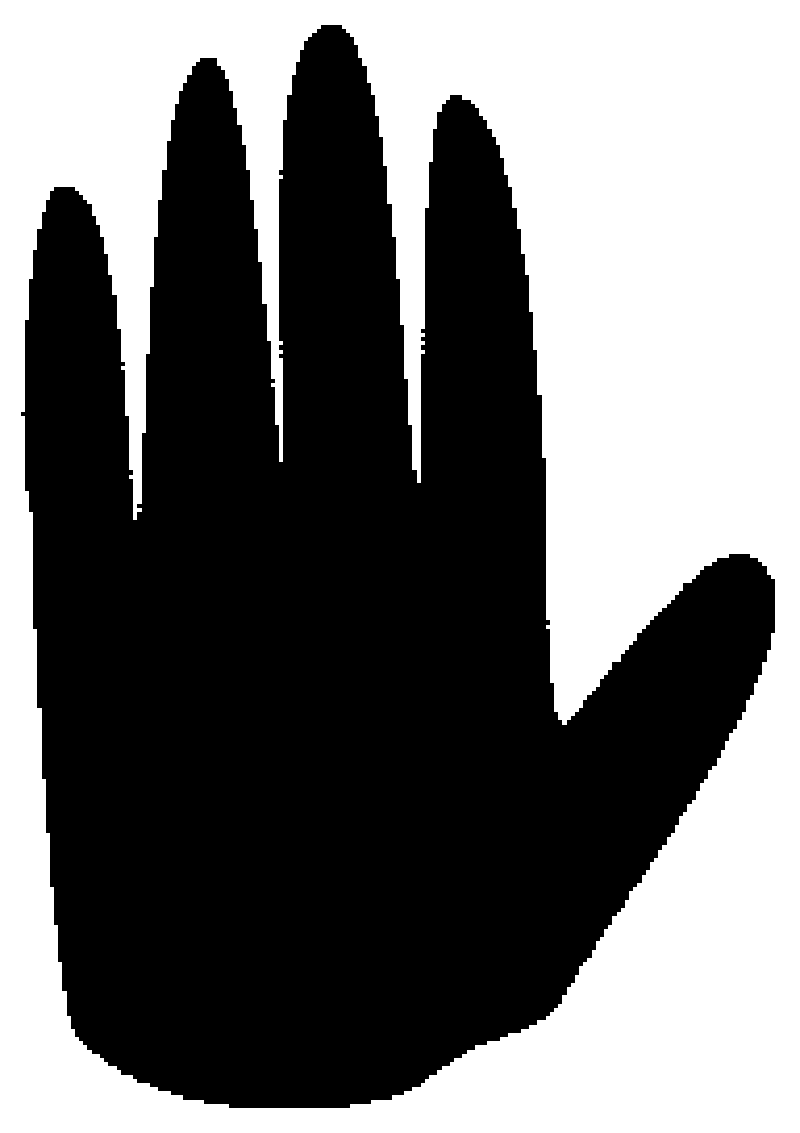,height=0.700000cm}&
\epsfig{figure=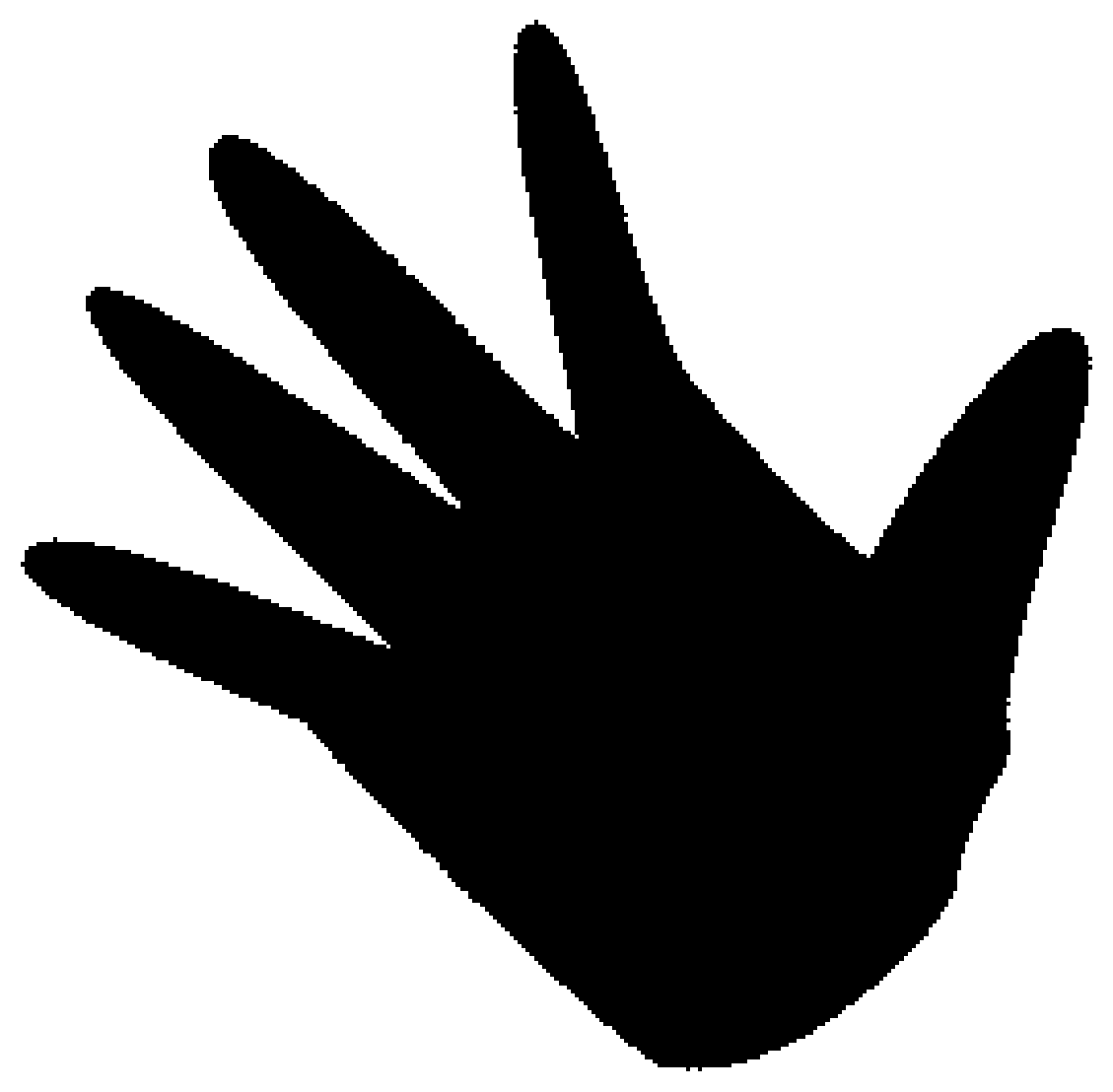,width=0.700000cm}&
\epsfig{figure=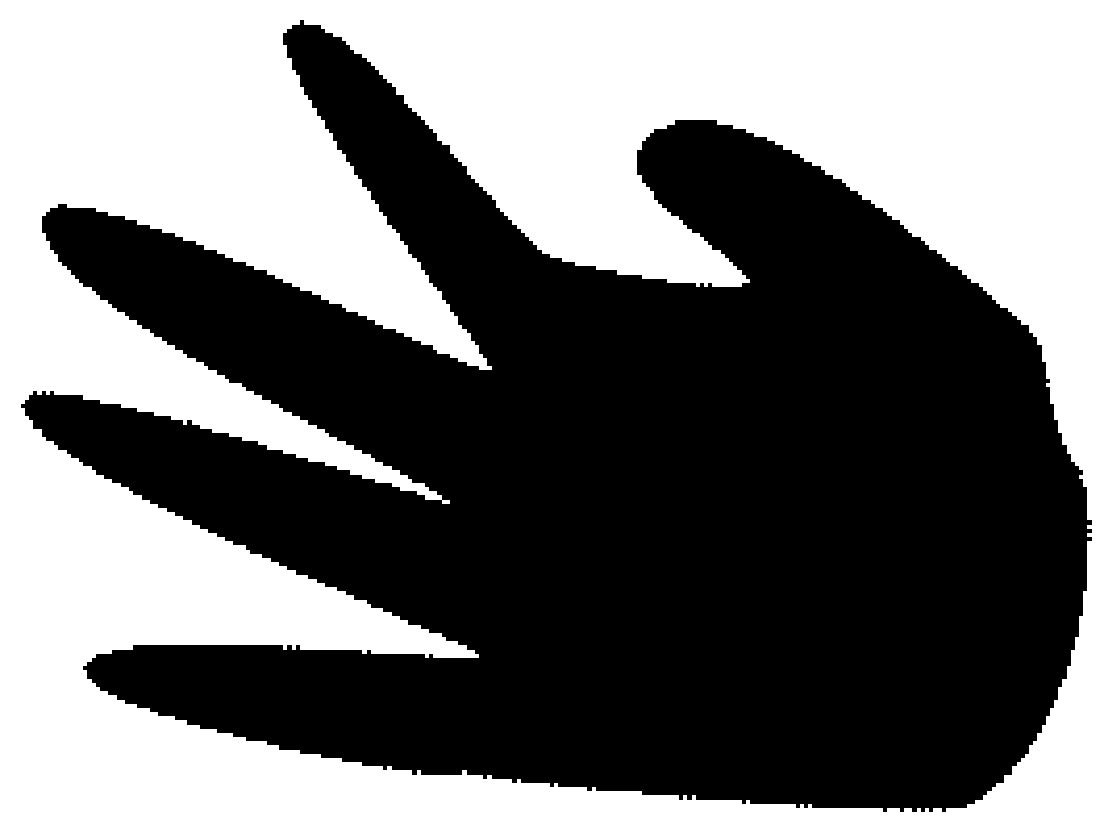,width=0.700000cm}&
\epsfig{figure=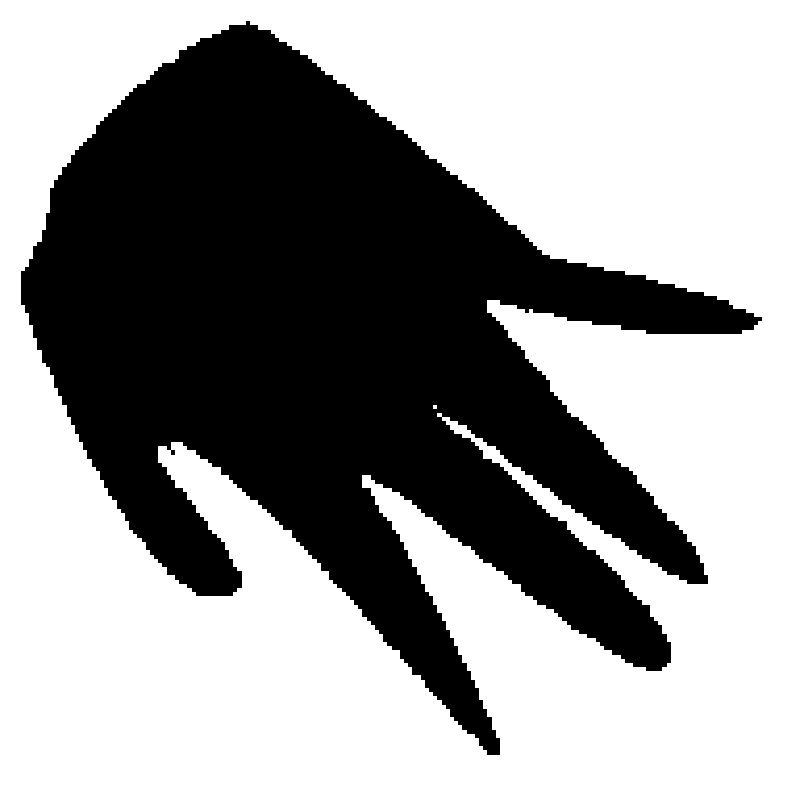,width=0.700000cm}&
\epsfig{figure=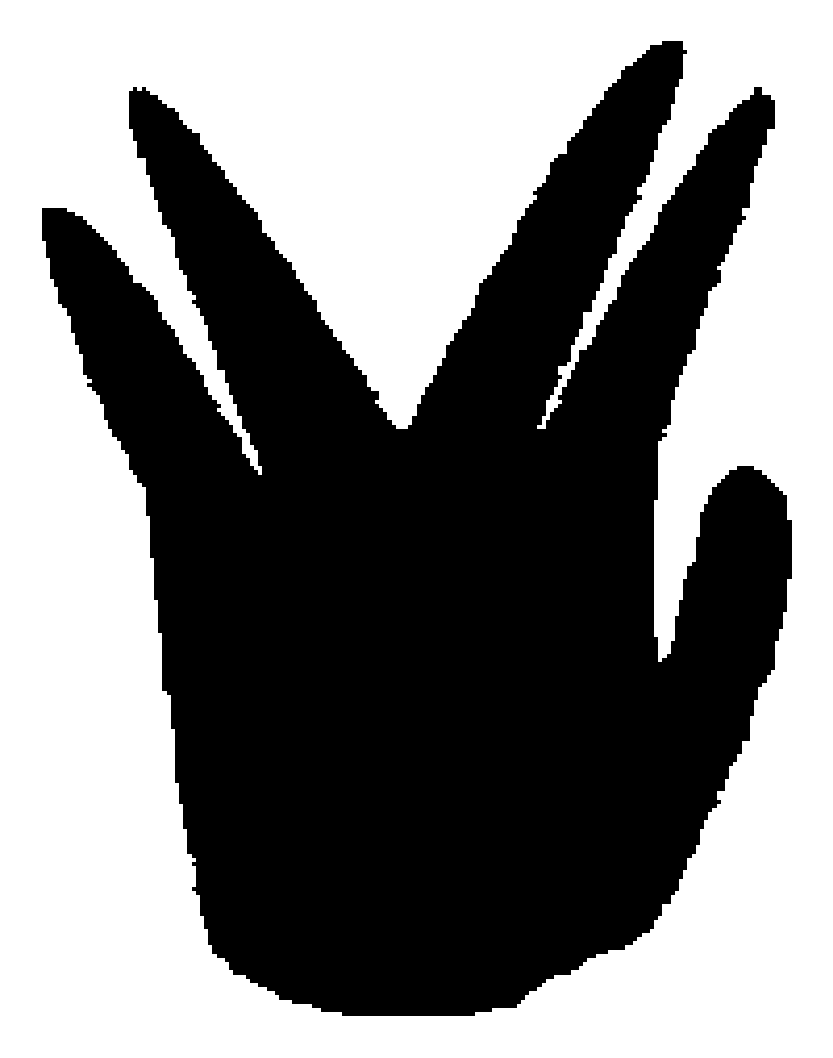,height=0.700000cm}&
\epsfig{figure=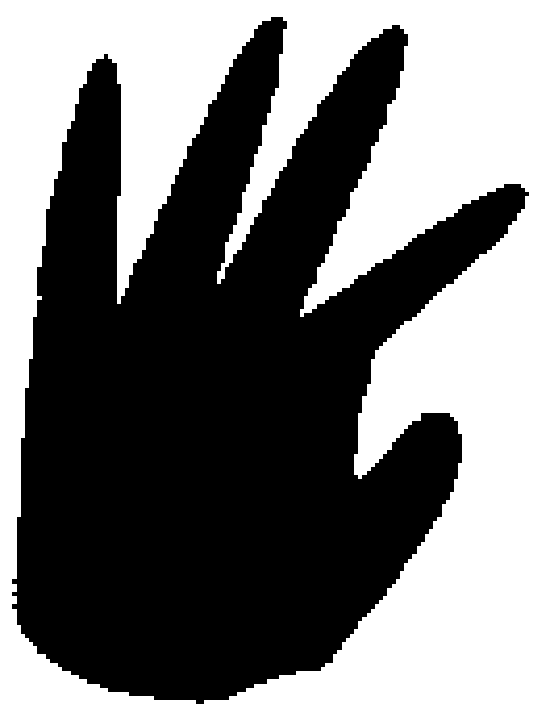,height=0.700000cm}&
\epsfig{figure=PAMI08_png/obj7_4.png,width=0.700000cm}&
\epsfig{figure=PAMI08_png/obj4_4.png,height=0.700000cm}&
\epsfig{figure=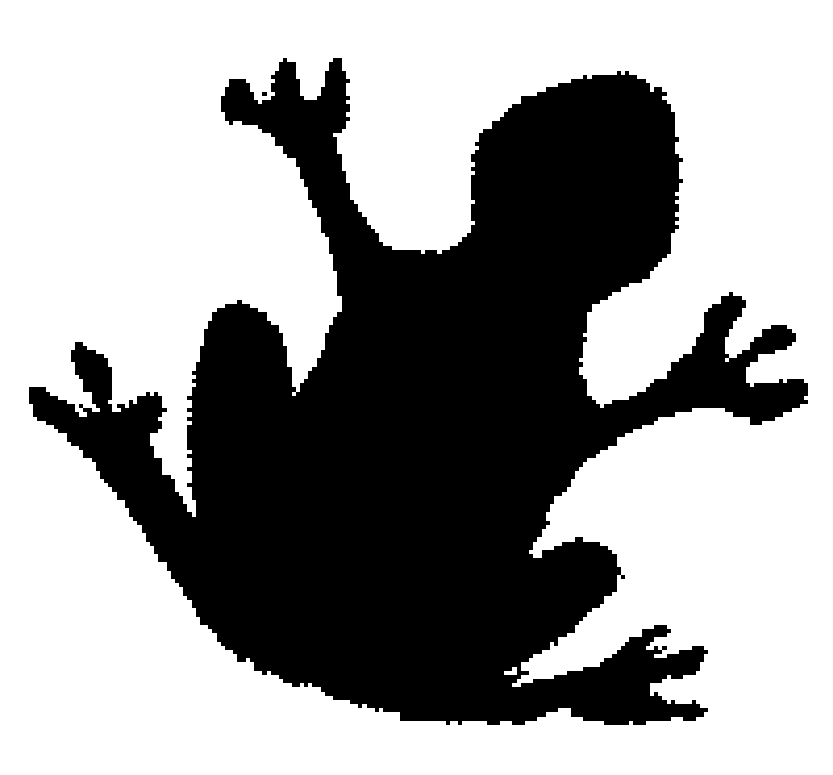,width=0.700000cm}&
\epsfig{figure=PAMI08_png/obj4_1.png,height=0.700000cm}&
\epsfig{figure=PAMI08_png/obj7_1.png,height=0.700000cm}&
\epsfig{figure=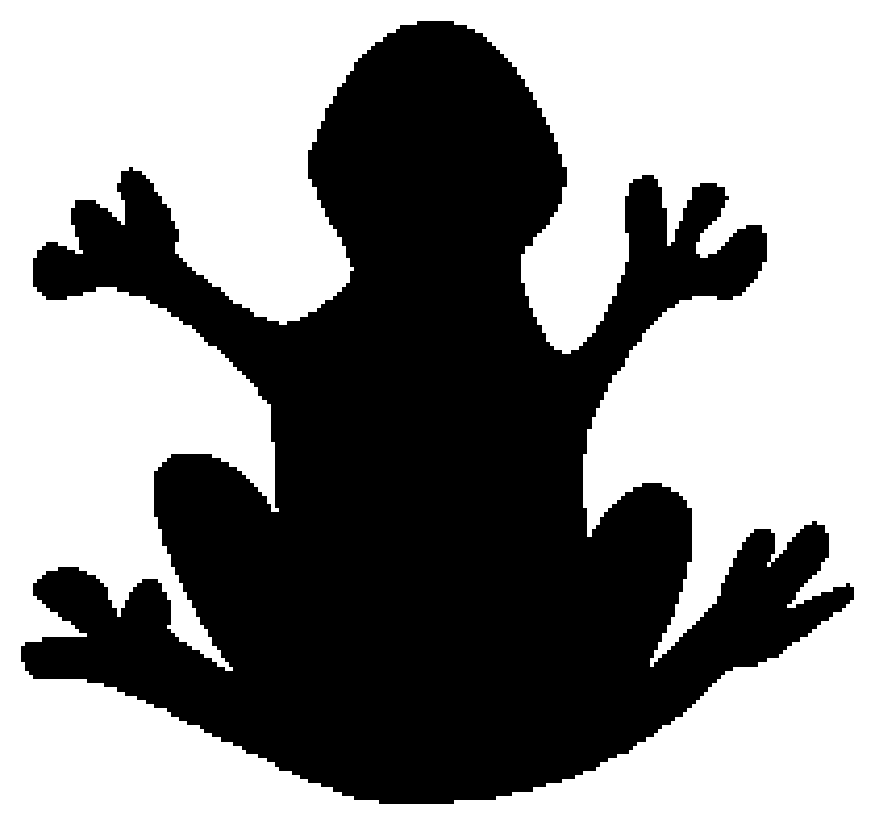,width=0.700000cm}\\
\cline{2-12}
  & \bf{0.966} & \bf{0.922} & \bf{0.890} & \bf{0.862} & \bf{0.843} & 0.578 & 0.572 & 0.568 & 0.551 & 0.542 & 0.533 \\
\hline \epsfig{figure=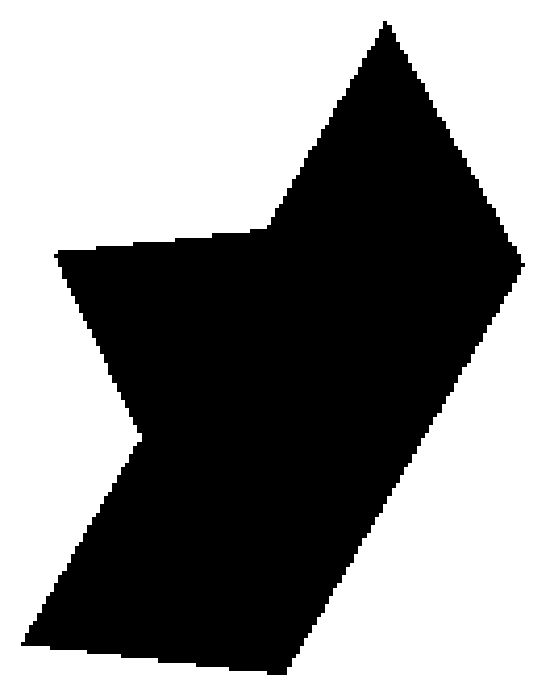,height=0.700000cm}&
\epsfig{figure=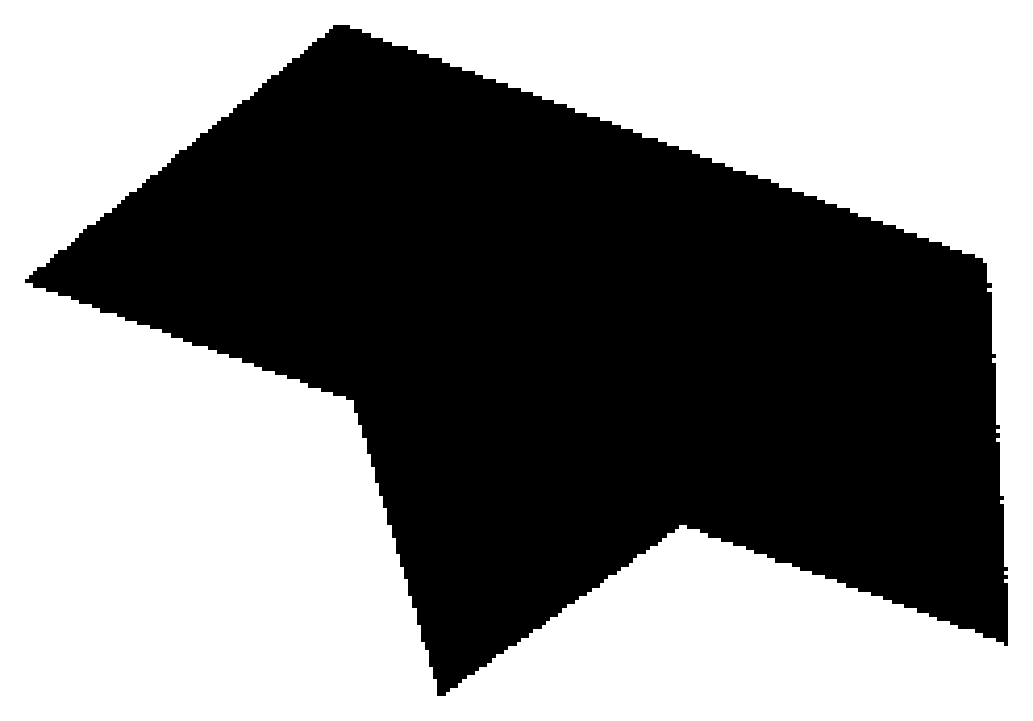,width=0.700000cm}&
\epsfig{figure=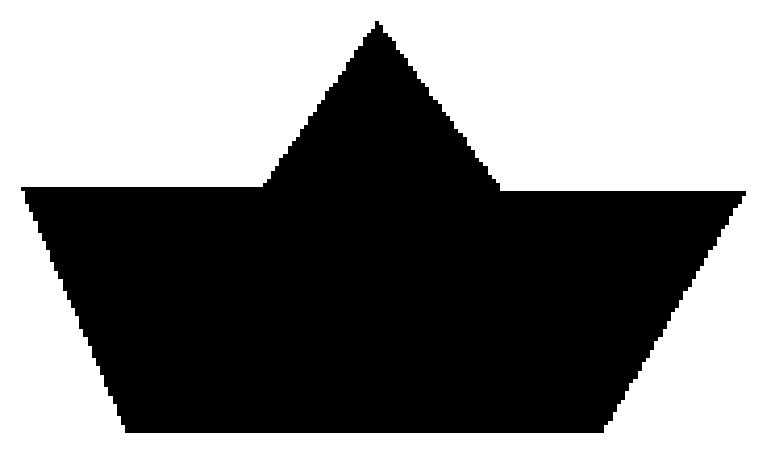,width=0.700000cm}&
\epsfig{figure=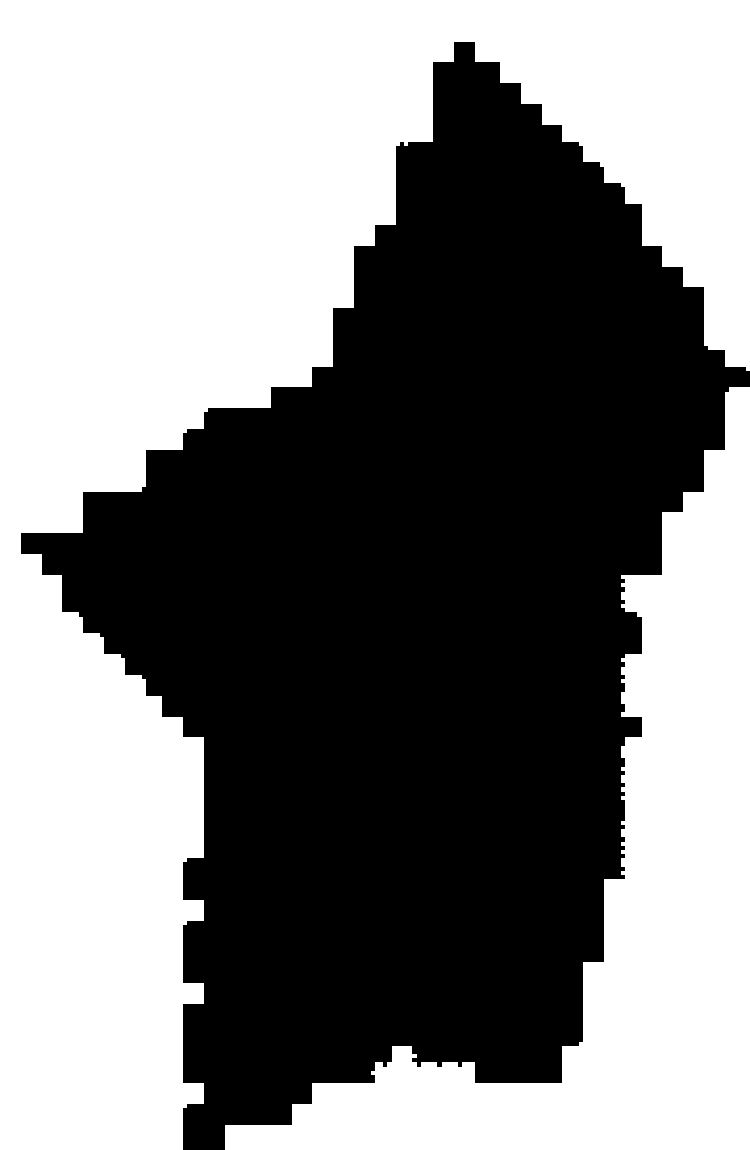,height=0.700000cm}&
\epsfig{figure=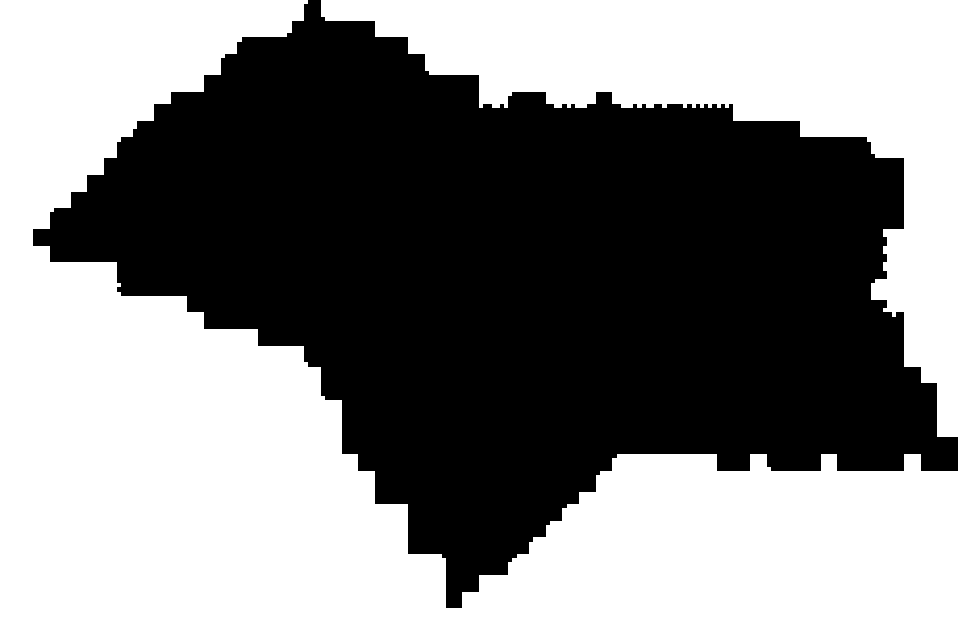,width=0.700000cm}&
\epsfig{figure=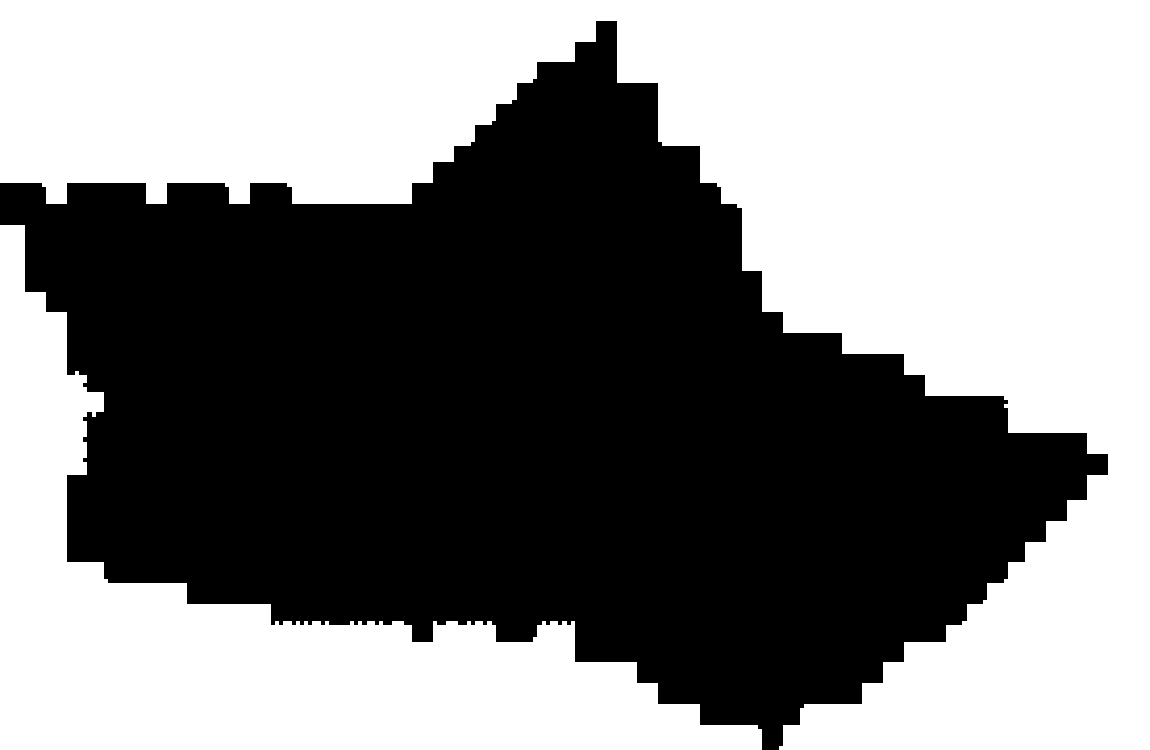,width=0.700000cm}&
\epsfig{figure=PAMI08_png/obj5_6.png,height=0.700000cm}&
\epsfig{figure=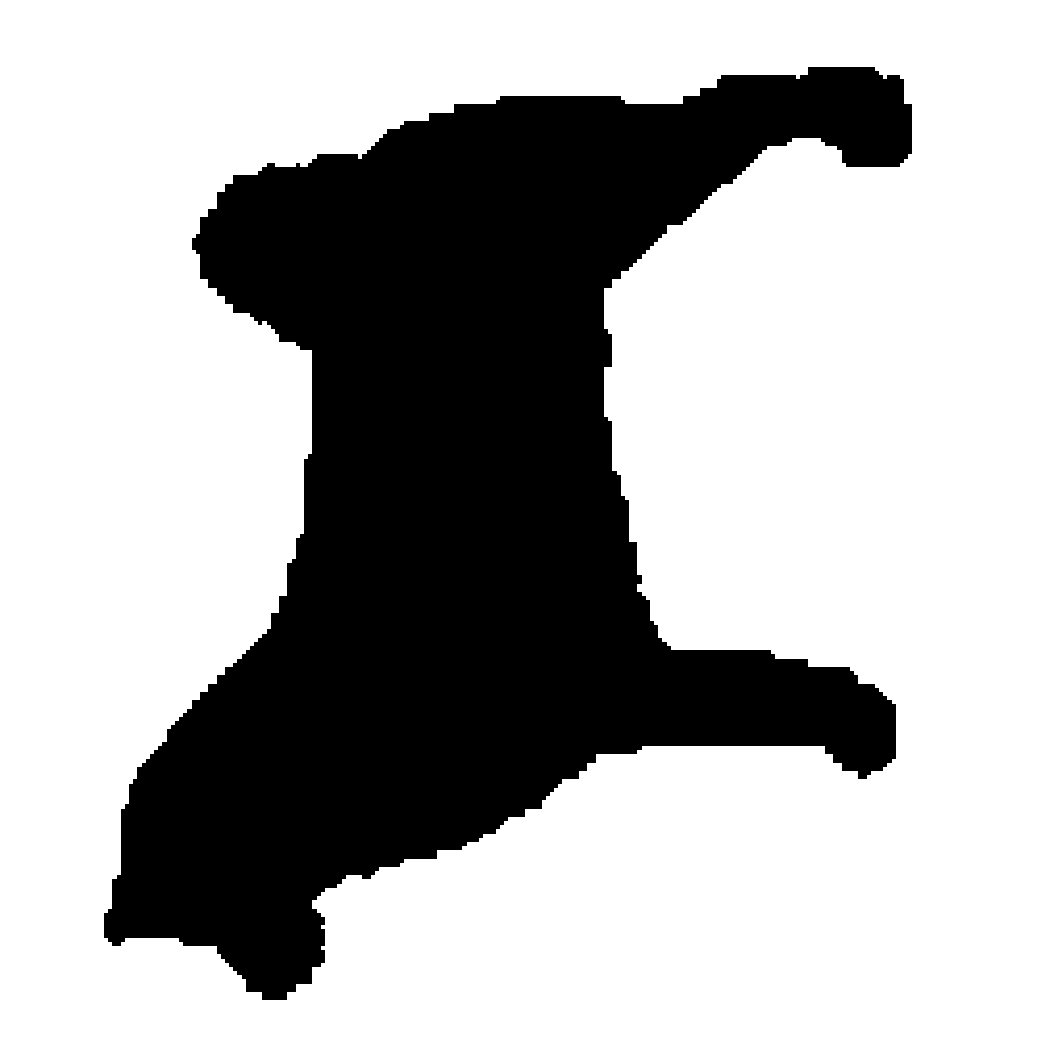,height=0.700000cm}&
\epsfig{figure=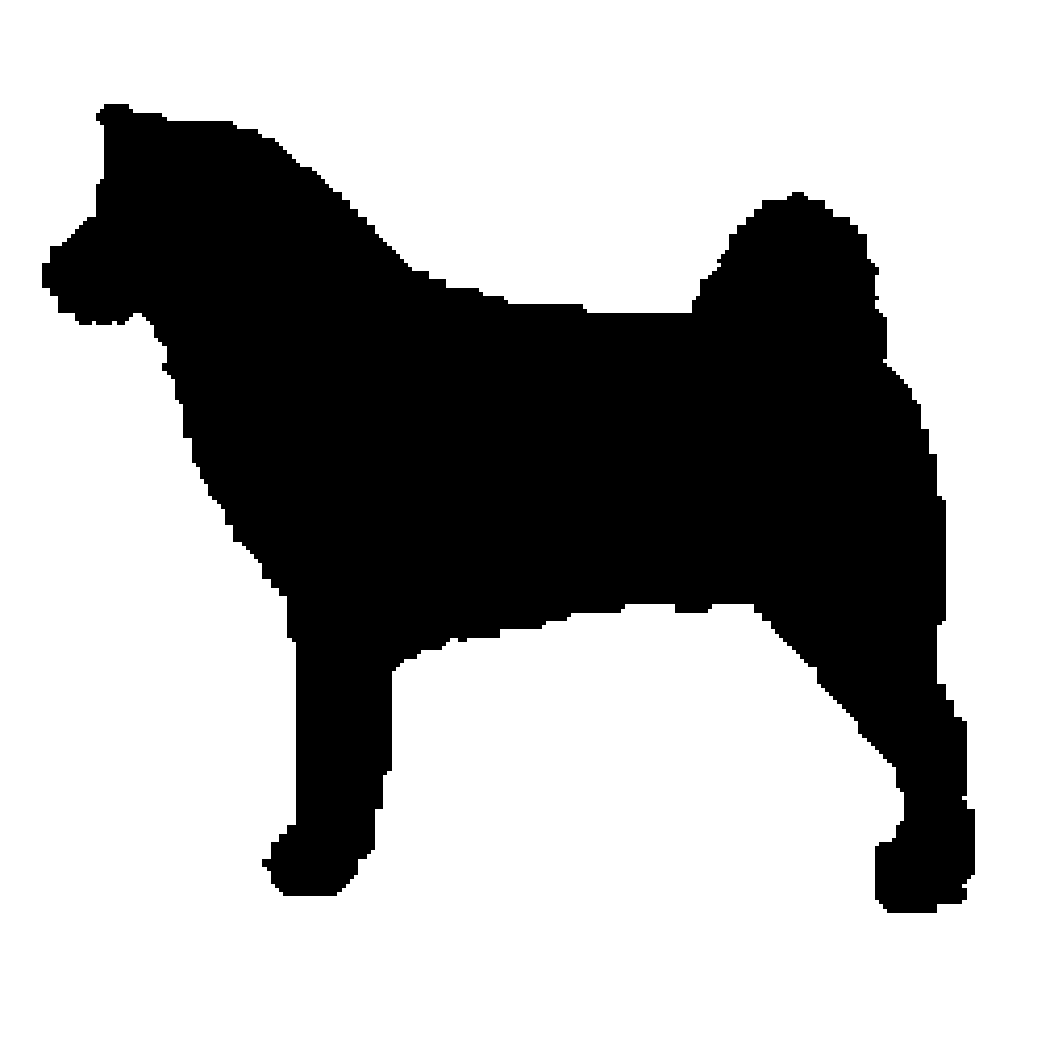,height=0.700000cm}&
\epsfig{figure=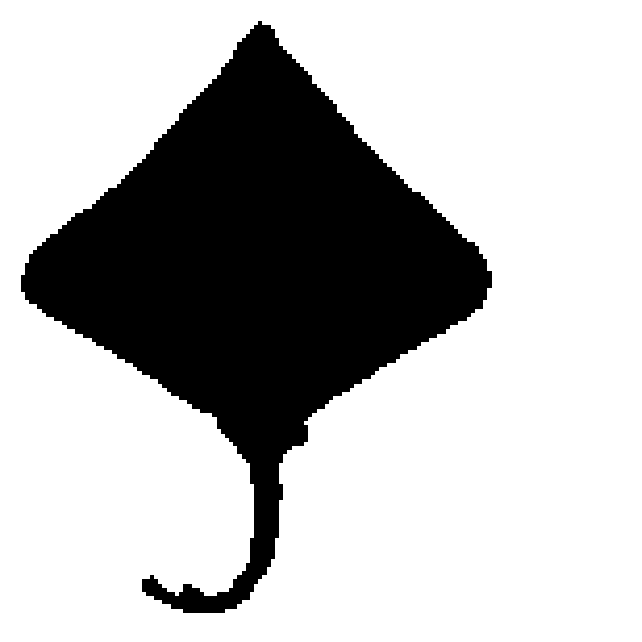,height=0.700000cm}&
\epsfig{figure=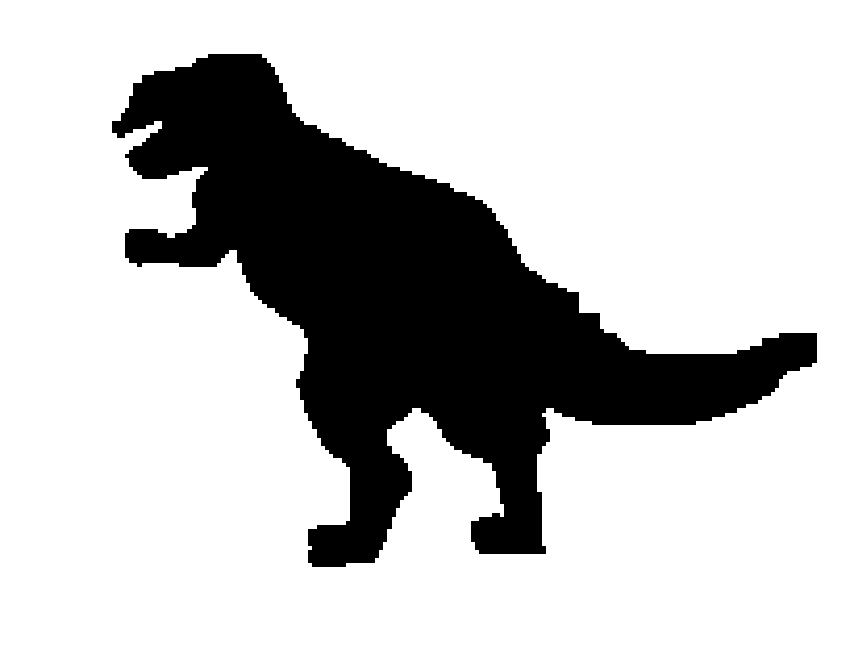,width=0.700000cm}&
\epsfig{figure=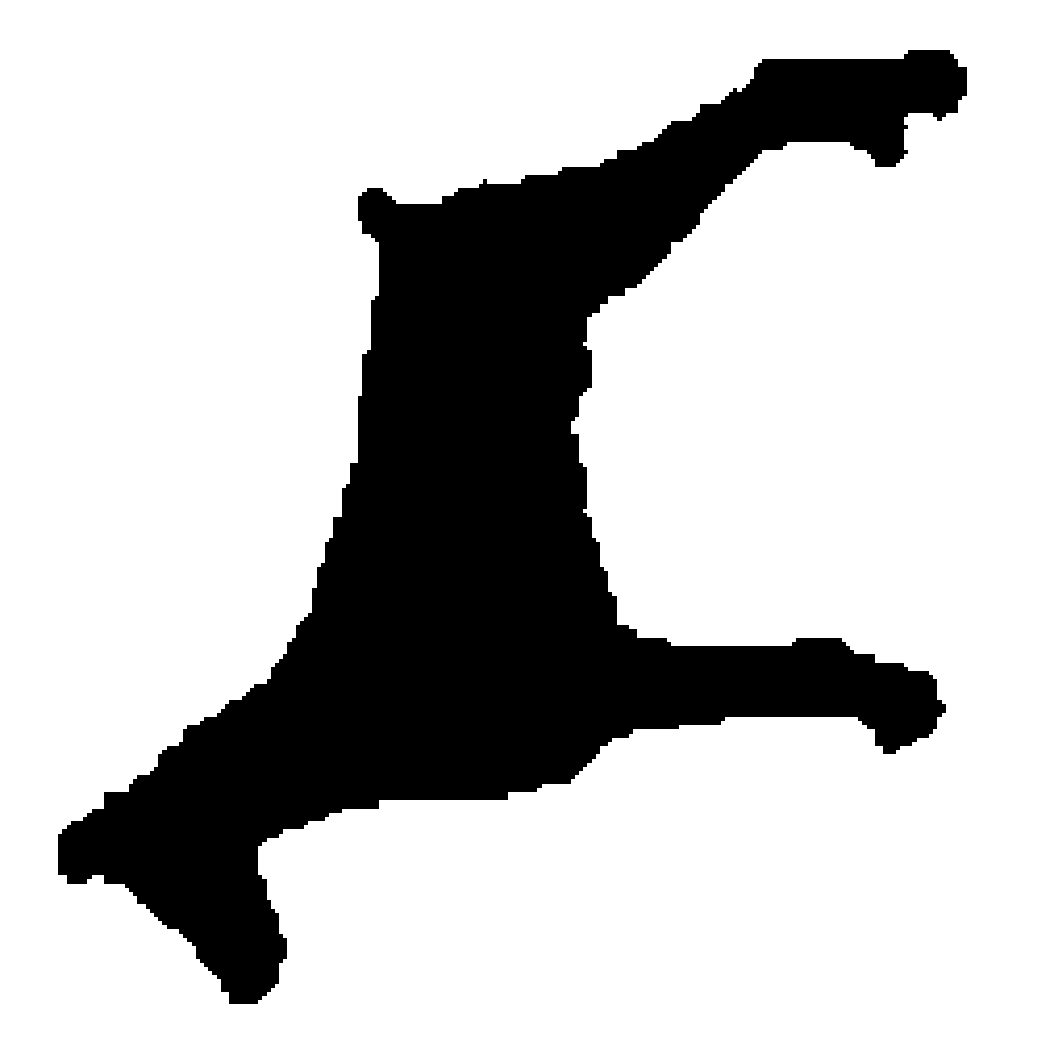,height=0.700000cm}\\
\cline{2-12}
  & \bf{0.898} & \bf{0.893} & \bf{0.821} & \bf{0.813} & \bf{0.803} & 0.442 & 0.398 & 0.395 & 0.384 & 0.377 & 0.362 \\
\hline \epsfig{figure=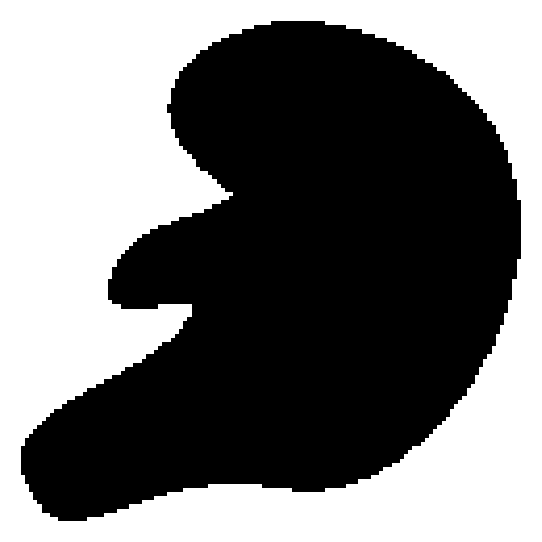,height=0.700000cm}&
\epsfig{figure=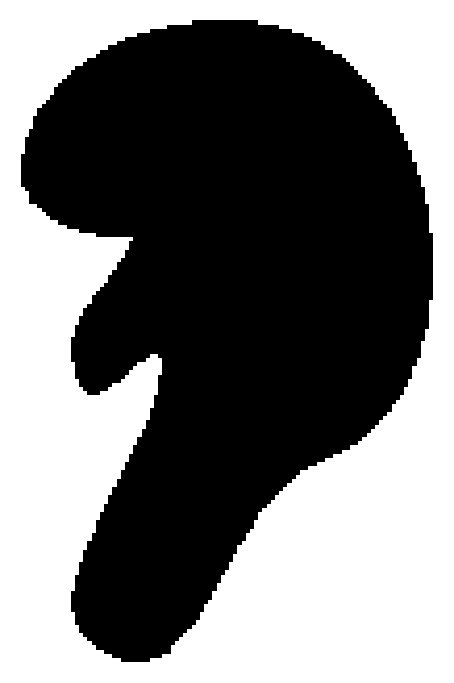,height=0.700000cm}&
\epsfig{figure=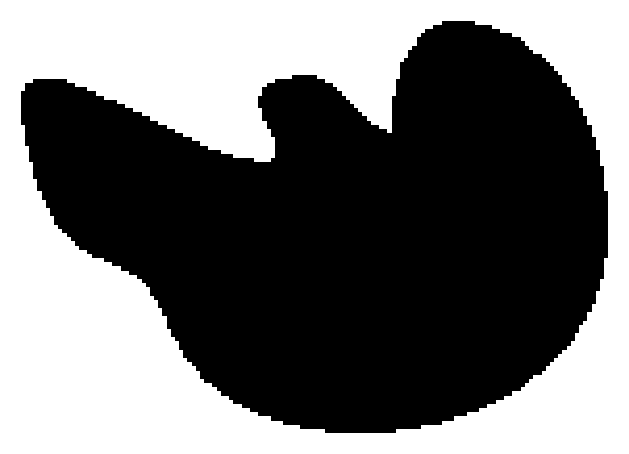,width=0.700000cm}&
\epsfig{figure=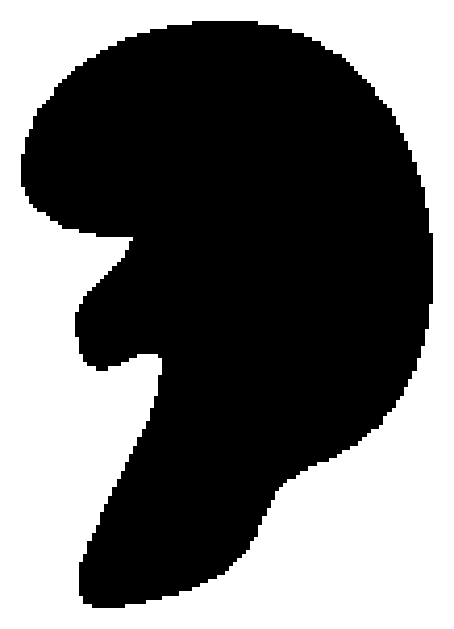,height=0.700000cm}&
\epsfig{figure=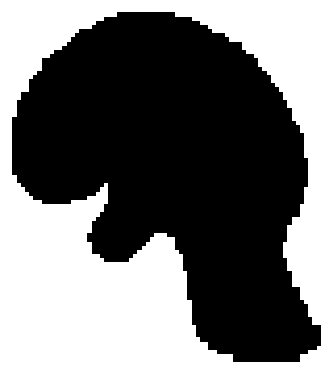,height=0.700000cm}&
\epsfig{figure=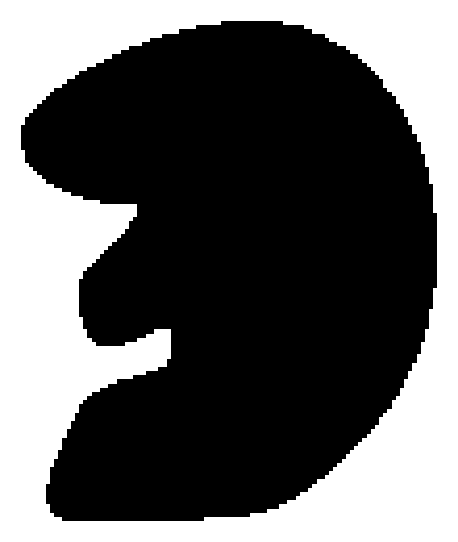,height=0.700000cm}&
\epsfig{figure=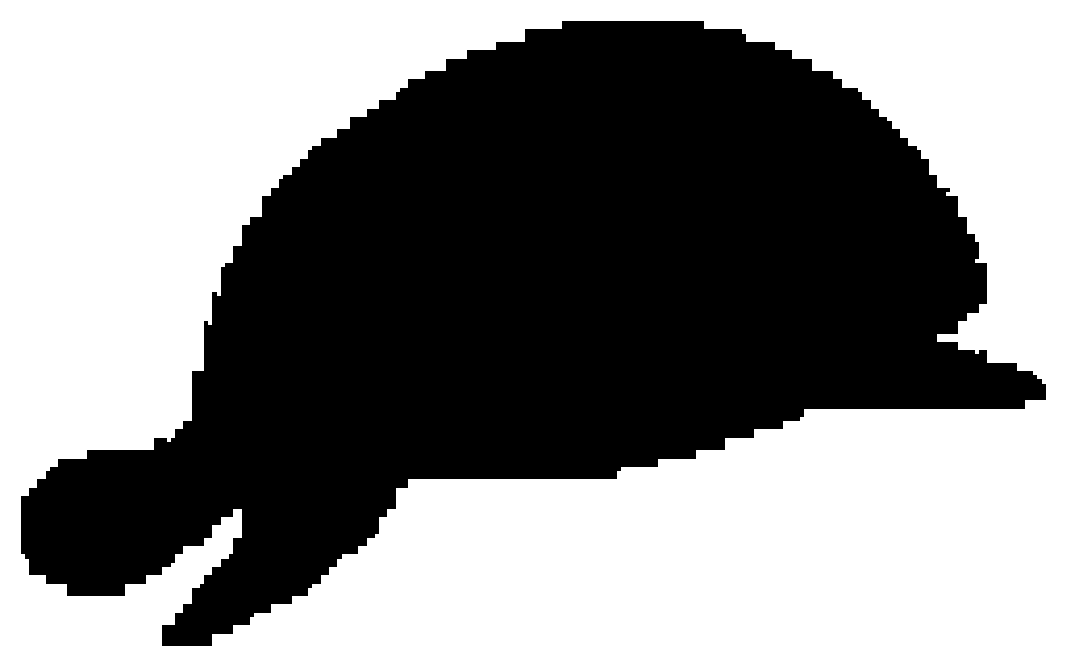,width=0.700000cm}&
\epsfig{figure=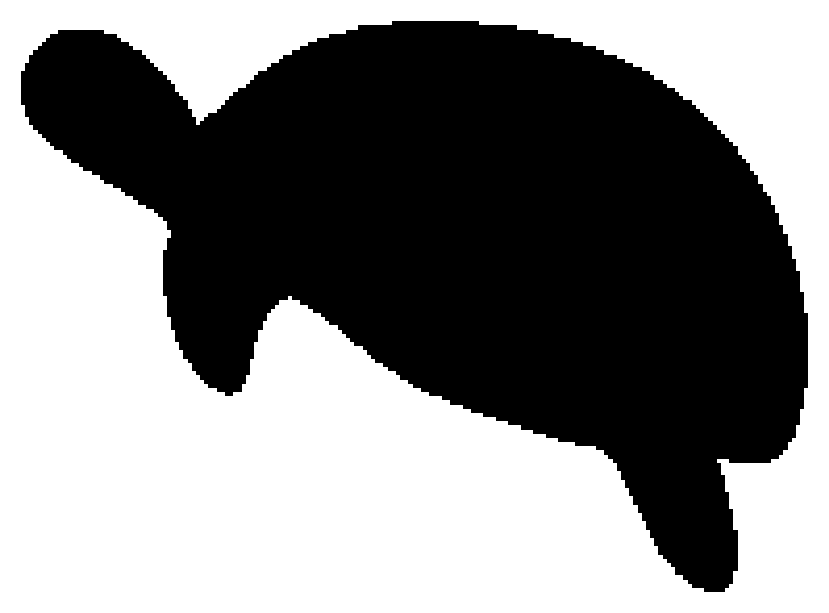,width=0.700000cm}&
\epsfig{figure=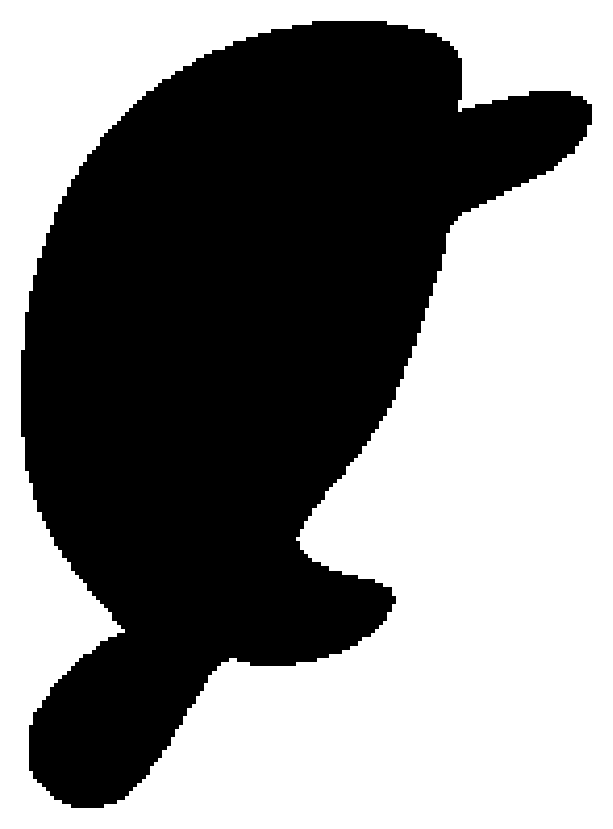,height=0.700000cm}&
\epsfig{figure=PAMI08_png/obj5_6.png,height=0.700000cm}&
\epsfig{figure=PAMI08_png/obj5_4.png,height=0.700000cm}&
\epsfig{figure=PAMI08_png/obj5_5.png,height=0.700000cm}\\
\cline{2-12}
  & \bf{0.839} & \bf{0.788} & \bf{0.785} & \bf{0.634} & \bf{0.611} & 0.391 & 0.302 & 0.302 & 0.293 & 0.281 & 0.273 \\
\hline \epsfig{figure=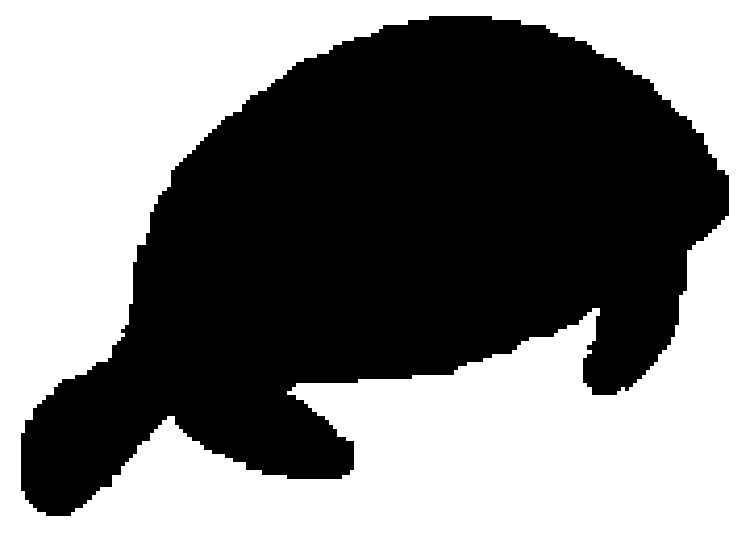,width=0.700000cm}&
\epsfig{figure=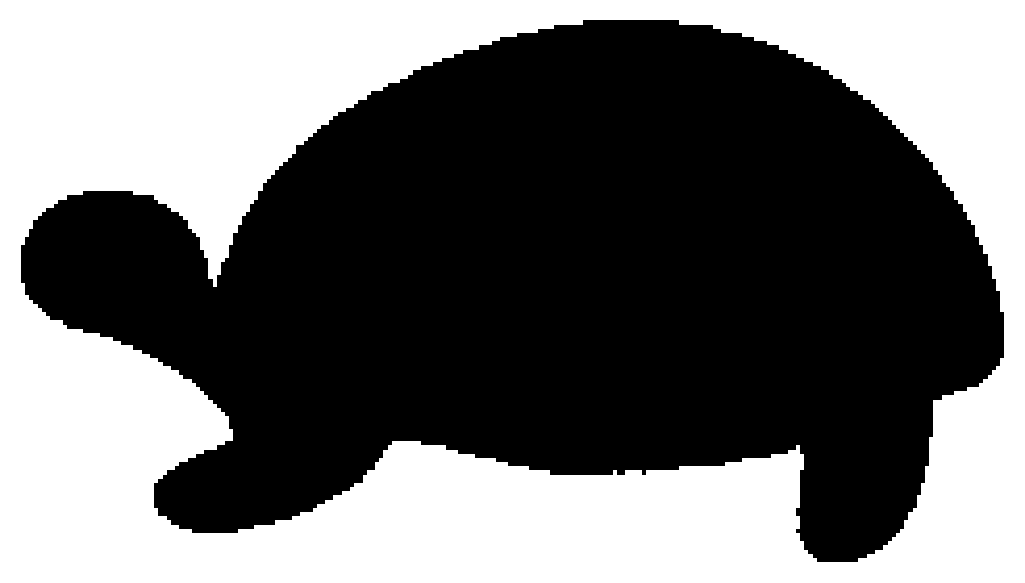,width=0.700000cm}&
\epsfig{figure=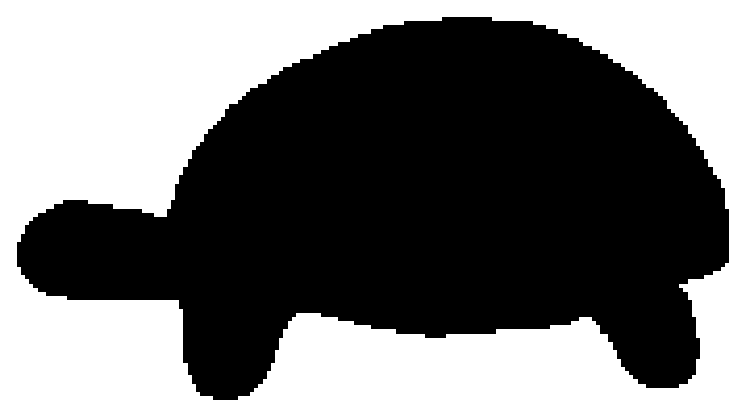,width=0.700000cm}&
\epsfig{figure=PAMI08_png/obj14_6.png,width=0.700000cm}&
\epsfig{figure=PAMI08_png/obj14_2.png,height=0.700000cm}&
\epsfig{figure=PAMI08_png/obj14_4.png,width=0.700000cm}&
\epsfig{figure=PAMI08_png/obj13_2.png,height=0.700000cm}&
\epsfig{figure=PAMI08_png/obj13_6.png,width=0.700000cm}&
\epsfig{figure=PAMI08_png/obj13_3.png,height=0.700000cm}&
\epsfig{figure=PAMI08_png/obj13_1.png,height=0.700000cm}&
\epsfig{figure=PAMI08_png/obj13_4.png,height=0.700000cm}&
\epsfig{figure=PAMI08_png/obj13_5.png,height=0.700000cm}\\
\cline{2-12}
  & \bf{0.935} & \bf{0.883} & \bf{0.533} & \bf{0.464} & \bf{0.456} & 0.274 & 0.272 & 0.269 & 0.251 & 0.229 & 0.221 \\
\hline \epsfig{figure=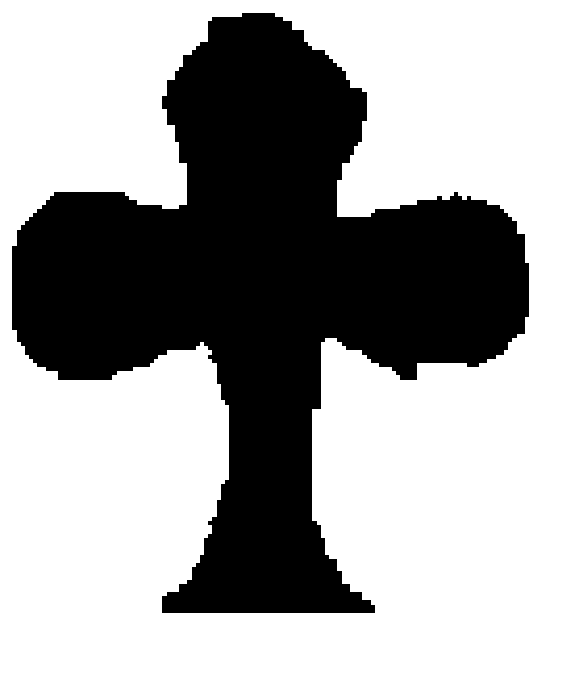,height=0.700000cm}&
\epsfig{figure=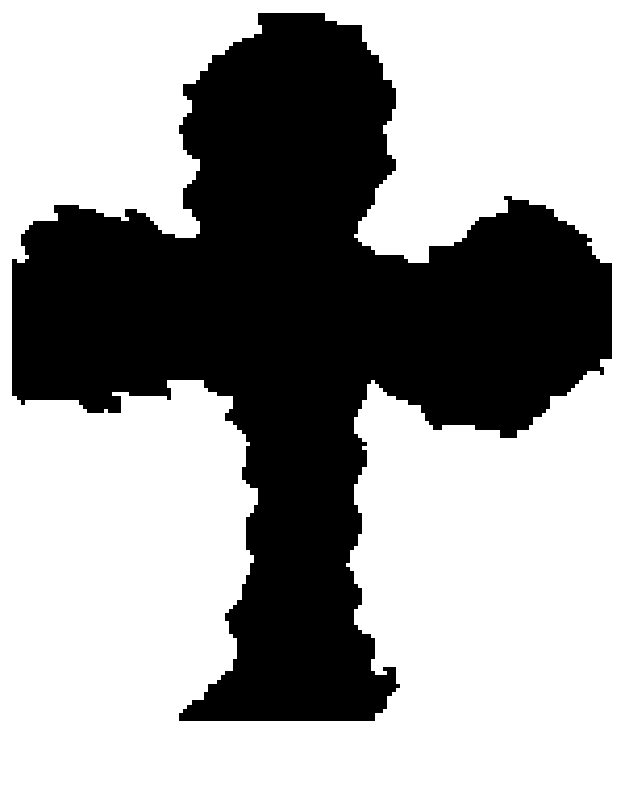,height=0.700000cm}&
\epsfig{figure=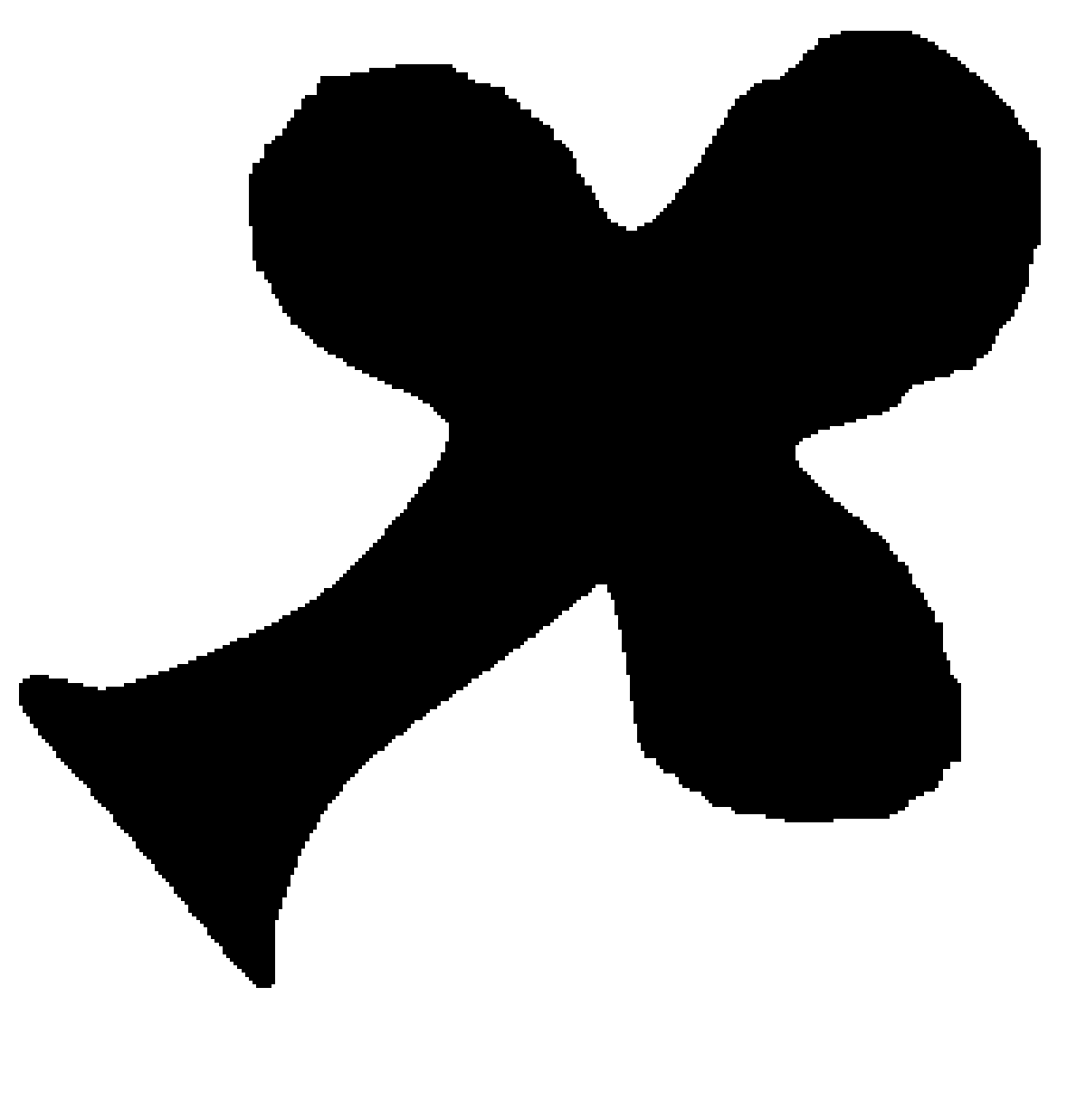,height=0.700000cm}&
\epsfig{figure=PAMI08_png/obj15_6.png,height=0.700000cm}&
\epsfig{figure=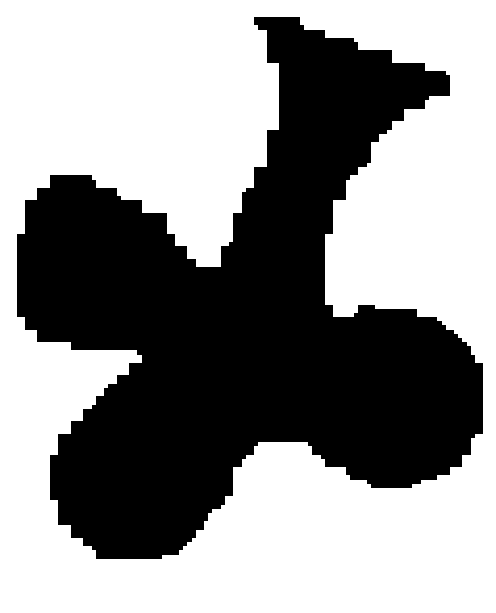,height=0.700000cm}&
\epsfig{figure=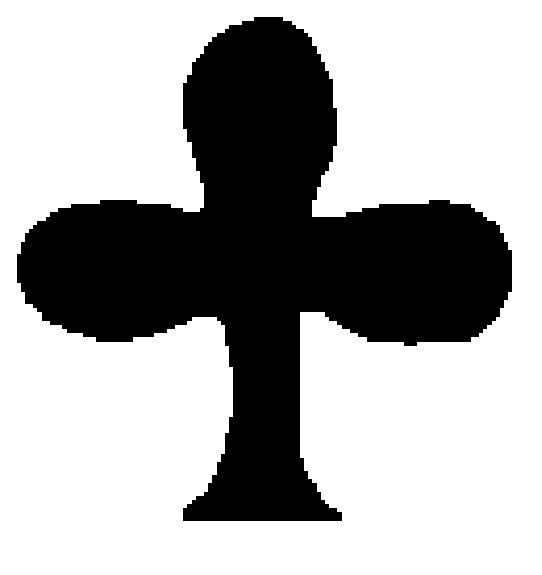,height=0.700000cm}&
\epsfig{figure=PAMI08_png/obj10_3.png,height=0.700000cm}&
\epsfig{figure=PAMI08_png/obj10_4.png,height=0.700000cm}&
\epsfig{figure=PAMI08_png/obj10_1.png,width=0.700000cm}&
\epsfig{figure=PAMI08_png/obj2_5.png,width=0.700000cm}&
\epsfig{figure=PAMI08_png/obj10_5.png,width=0.700000cm}&
\epsfig{figure=PAMI08_png/obj28_4.png,height=0.700000cm}\\
\cline{2-12}
  & \bf{0.957} & \bf{0.955} & \bf{0.859} & \bf{0.843} & \bf{0.791} & 0.644 & 0.606 & 0.584 & 0.575 & 0.571 & 0.563 \\
\hline
\end{tabular}
\label{table:results}
\end{table}

\clearpage
\begin{table}[!h]
\caption{Some query results.}
\centering
\begin {tabular} {|c|c|c|c|c|c||c|c|c|c|c|c|}
\hline \epsfig{figure=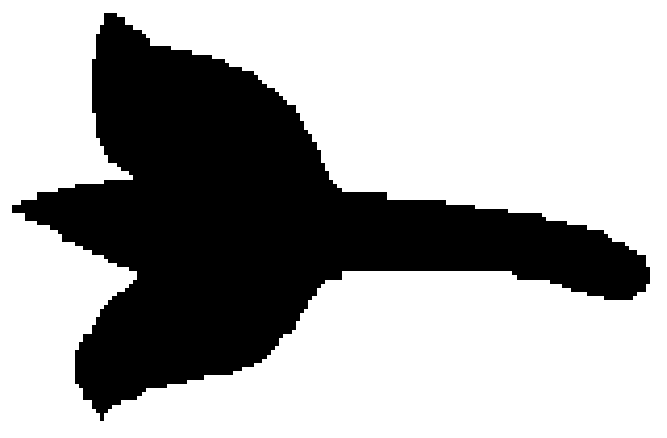,width=0.700000cm}&
\epsfig{figure=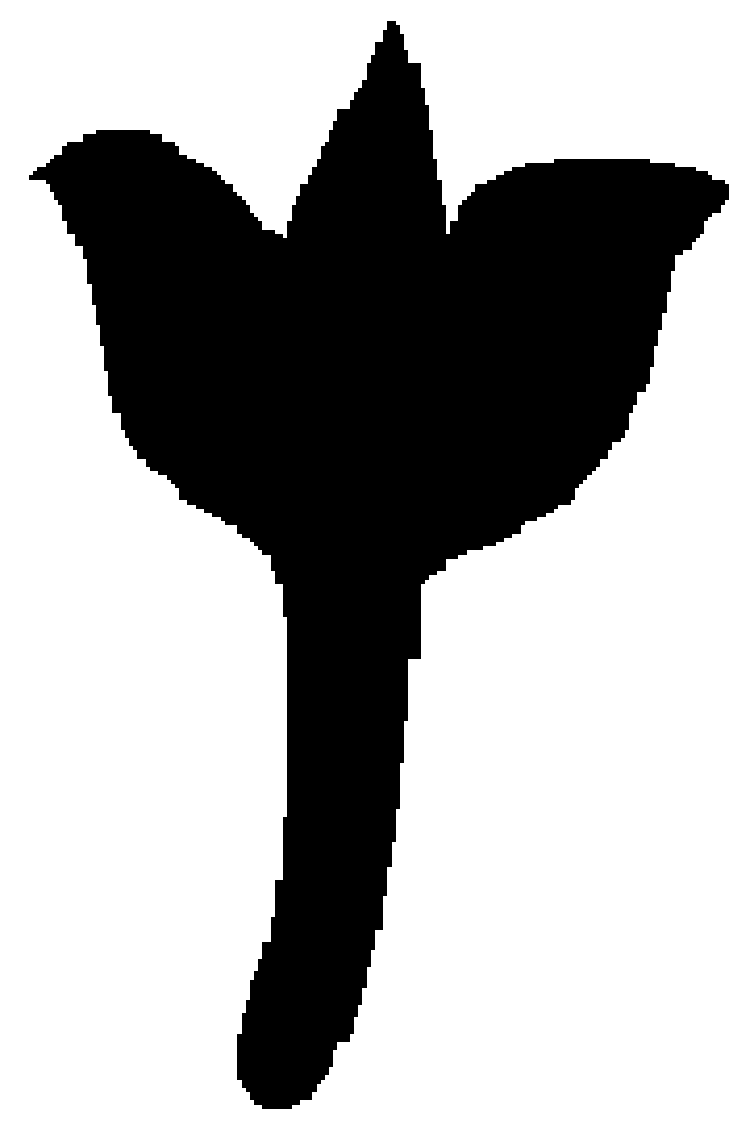,height=0.700000cm}&
\epsfig{figure=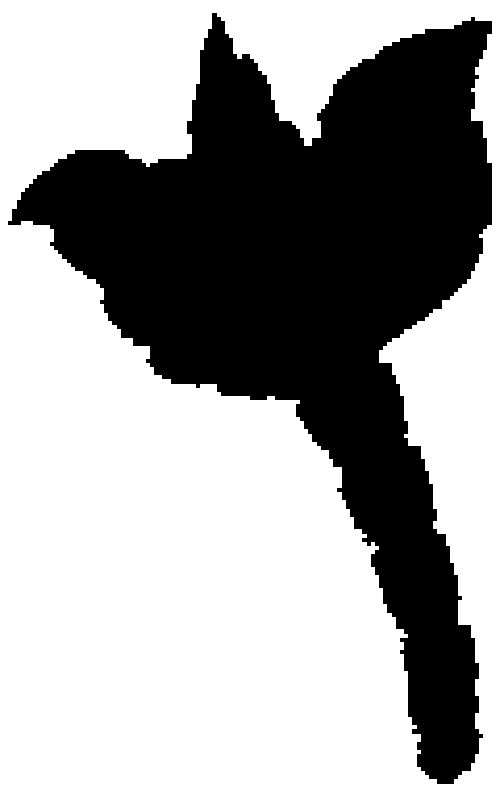,height=0.700000cm}&
\epsfig{figure=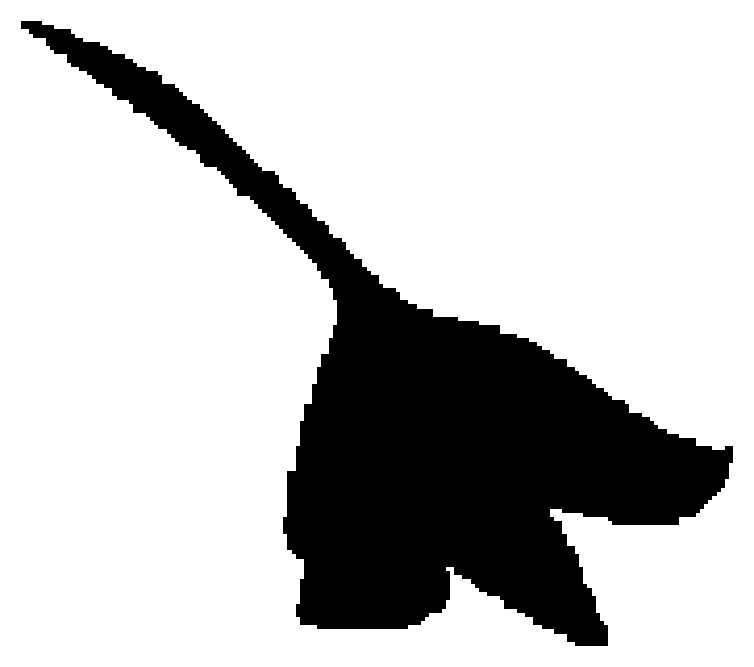,width=0.700000cm}&
\epsfig{figure=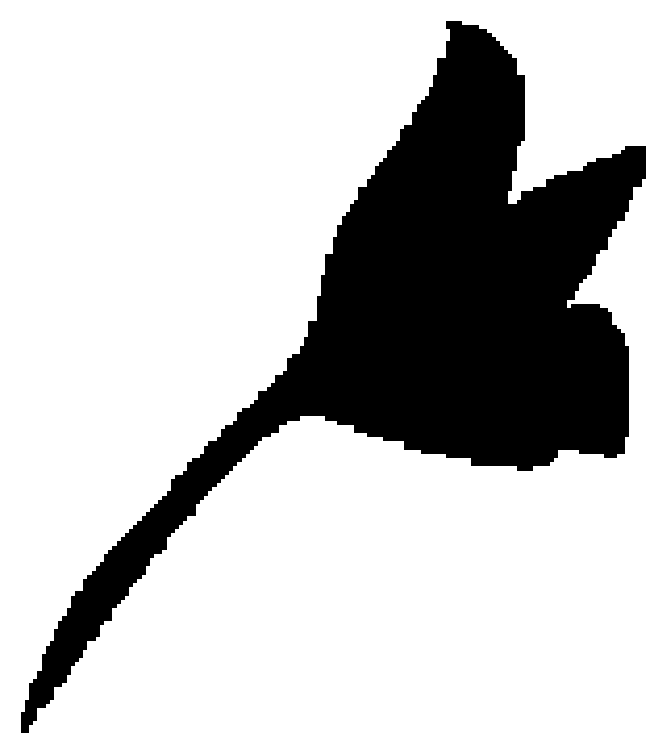,height=0.700000cm}&
\epsfig{figure=PAMI08_png/obj28_4.png,height=0.700000cm}&
\epsfig{figure=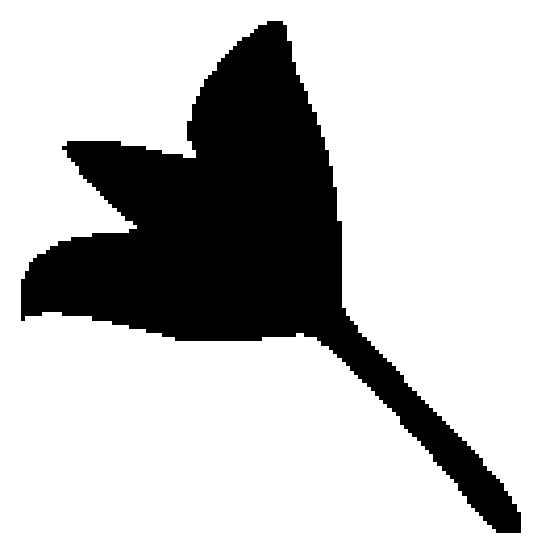,height=0.700000cm}&
\epsfig{figure=PAMI08_png/obj15_4.png,height=0.700000cm}&
\epsfig{figure=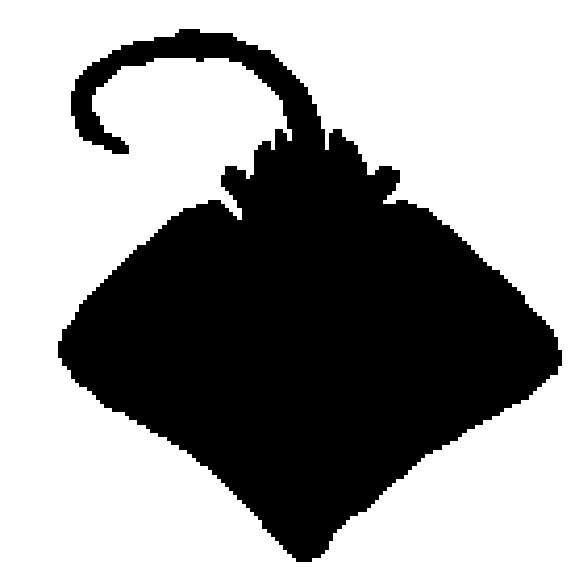,height=0.700000cm}&
\epsfig{figure=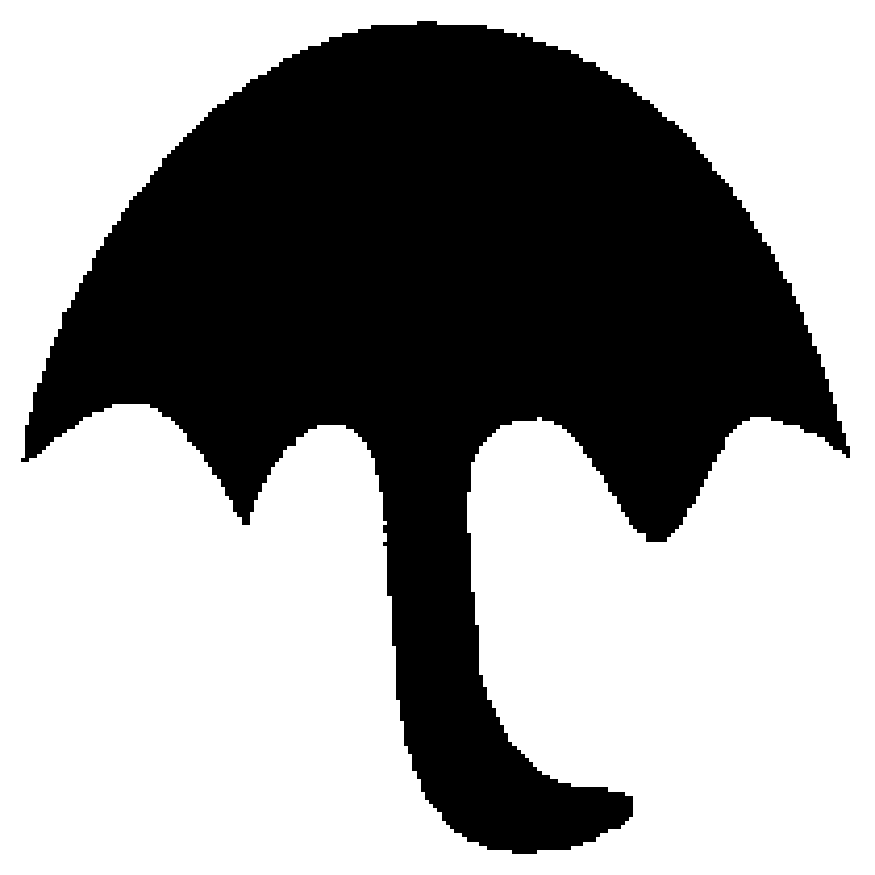,height=0.700000cm}&
\epsfig{figure=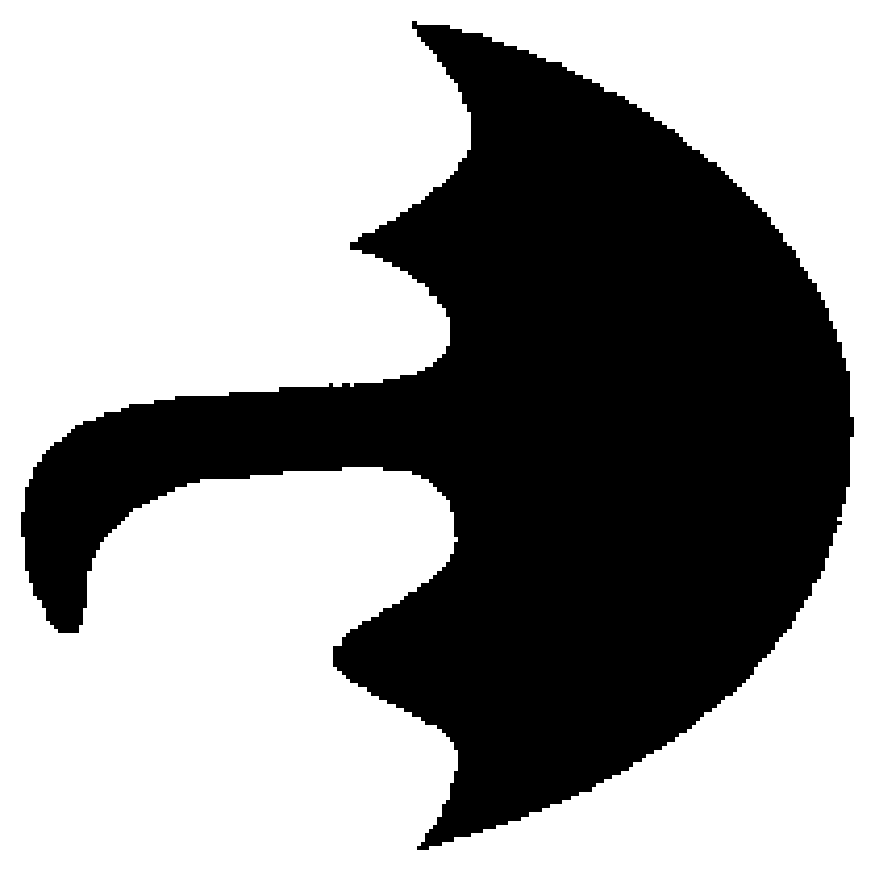,height=0.700000cm}&
\epsfig{figure=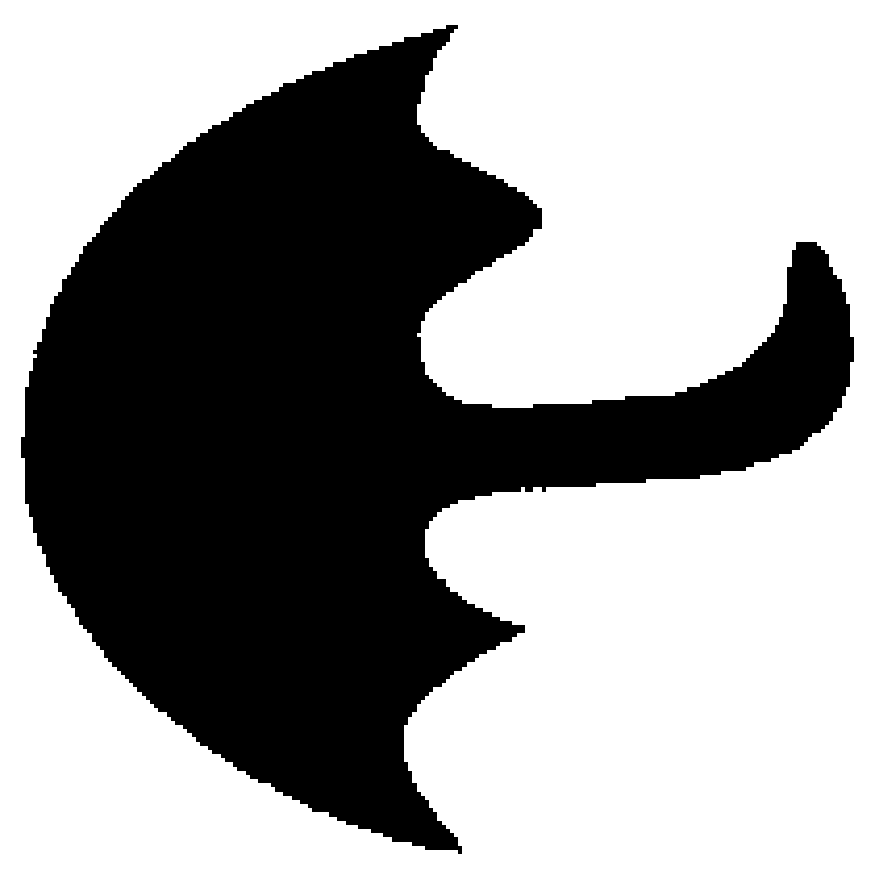,height=0.700000cm}\\
\cline{2-12}
  & \bf{0.992} & \bf{0.882} & \bf{0.781} & \bf{0.692} & 0.638 & \bf{0.627} & 0.582 & 0.574 & 0.548 & 0.546 & 0.536 \\
\hline \epsfig{figure=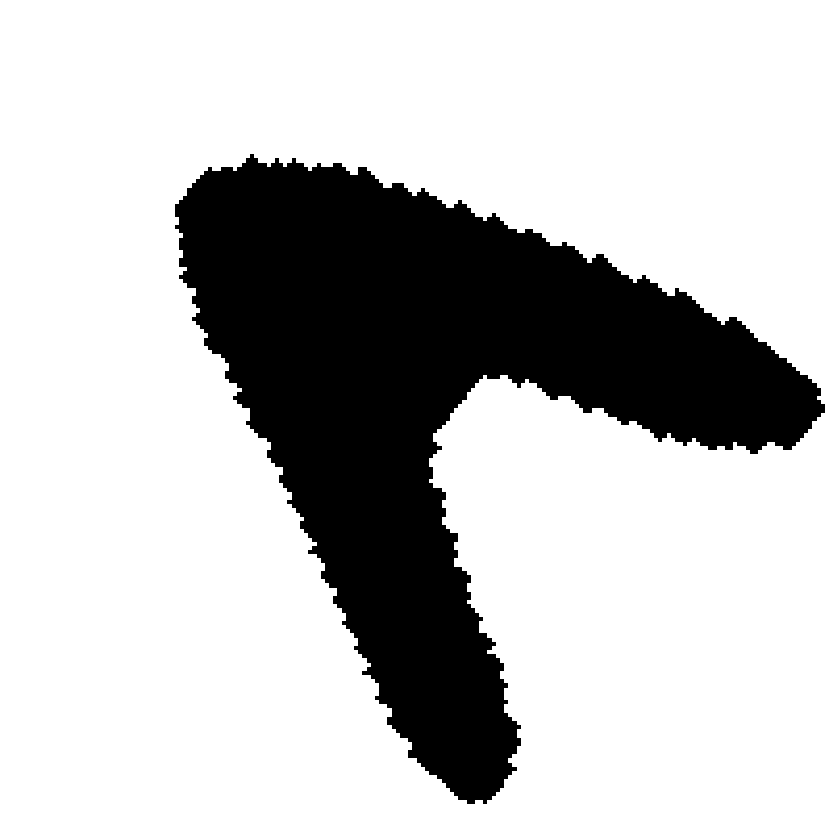,width=0.700000cm}&
\epsfig{figure=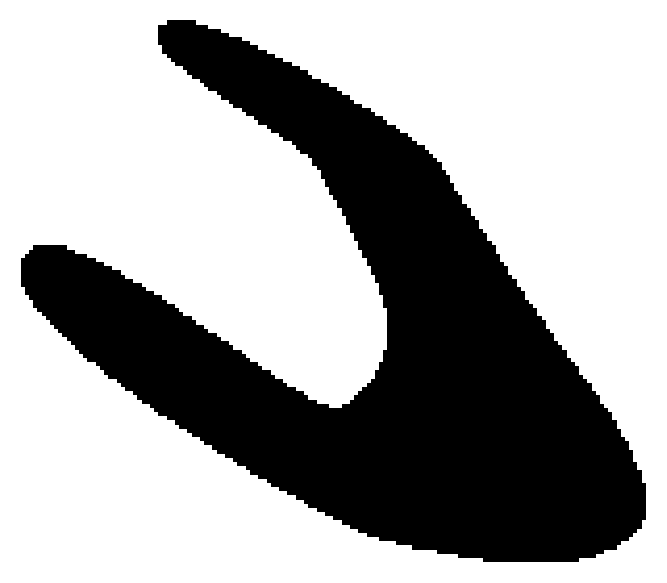,width=0.700000cm}&
\epsfig{figure=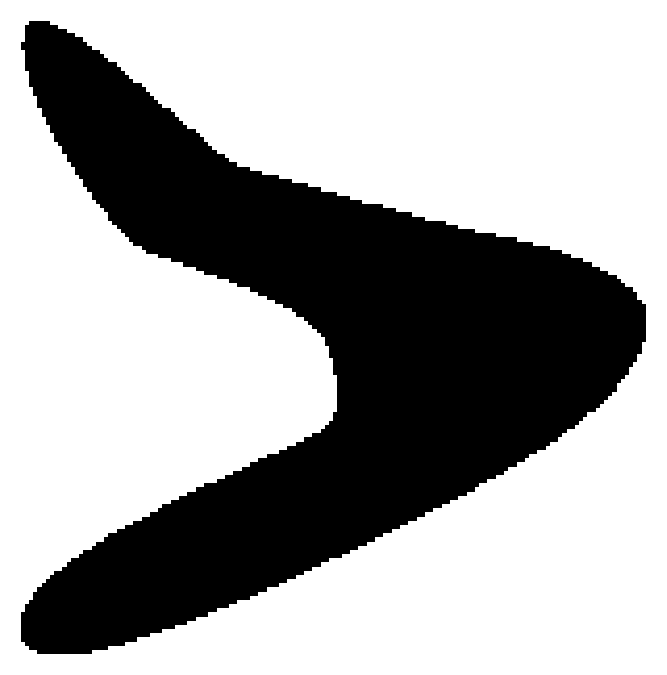,height=0.700000cm}&
\epsfig{figure=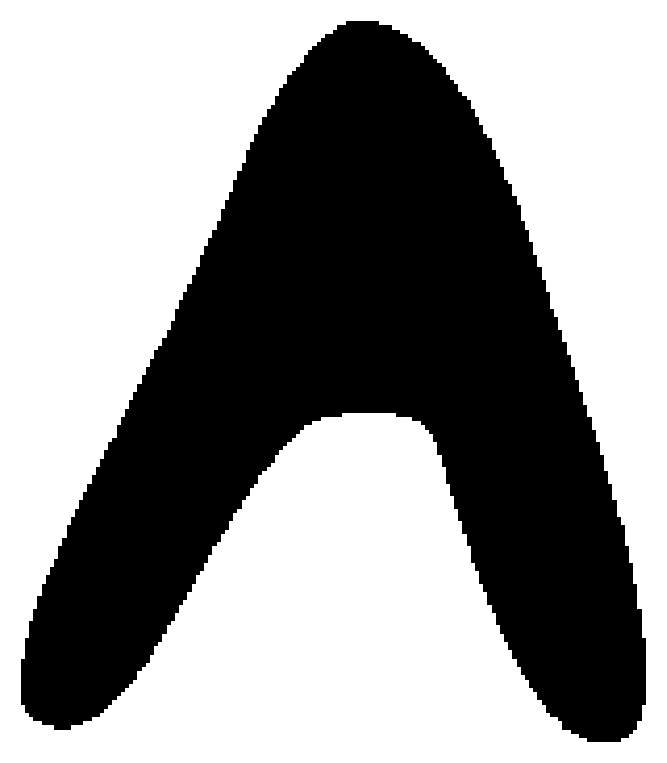,height=0.700000cm}&
\epsfig{figure=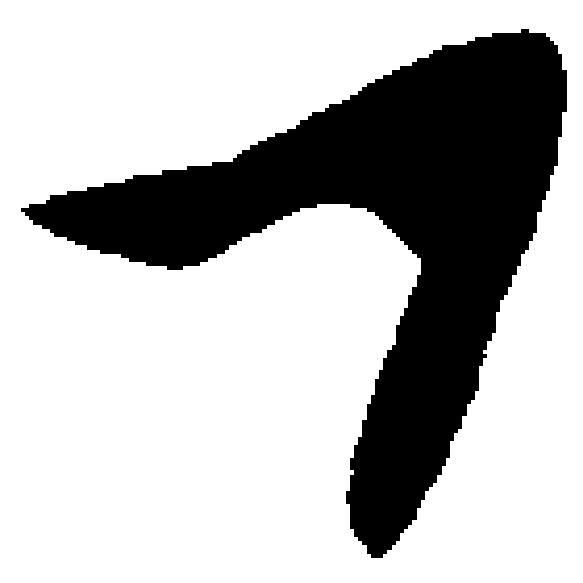,height=0.700000cm}&
\epsfig{figure=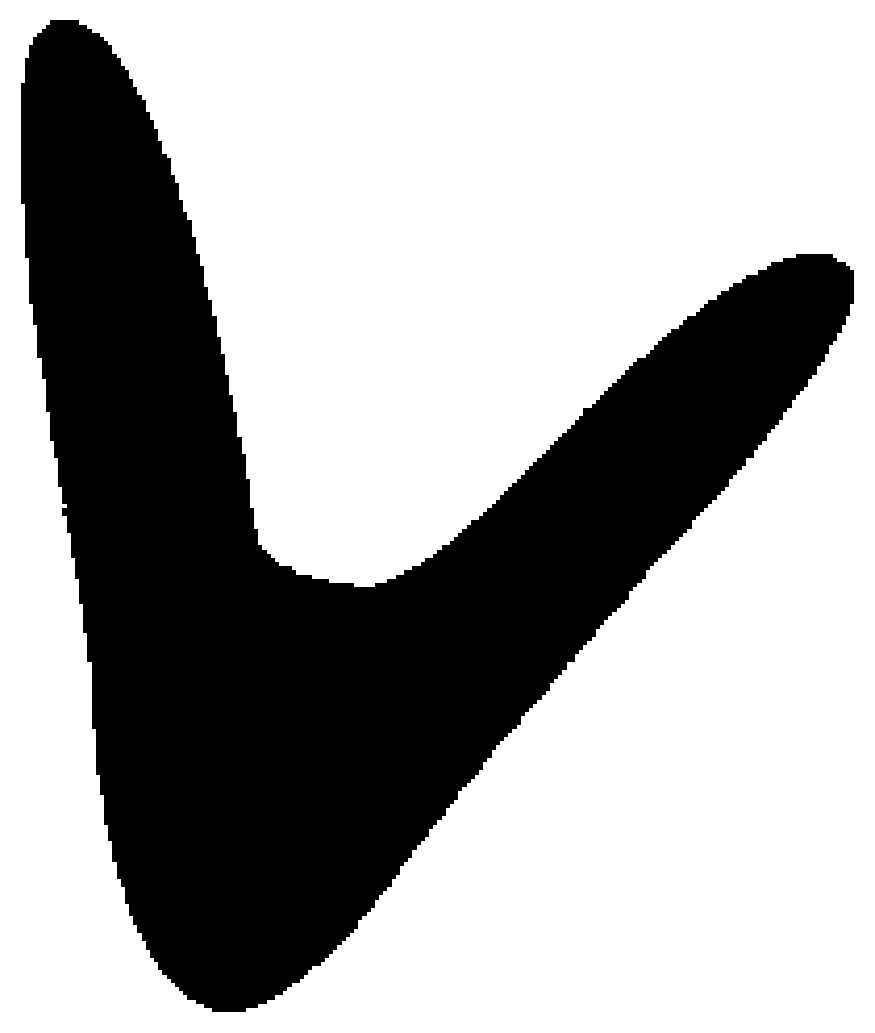,height=0.700000cm}&
\epsfig{figure=PAMI08_png/obj16_5.png,width=0.700000cm}&
\epsfig{figure=PAMI08_png/obj16_2.png,height=0.700000cm}&
\epsfig{figure=PAMI08_png/obj16_3.png,height=0.700000cm}&
\epsfig{figure=PAMI08_png/obj16_6.png,width=0.700000cm}&
\epsfig{figure=PAMI08_png/obj16_4.png,height=0.700000cm}&
\epsfig{figure=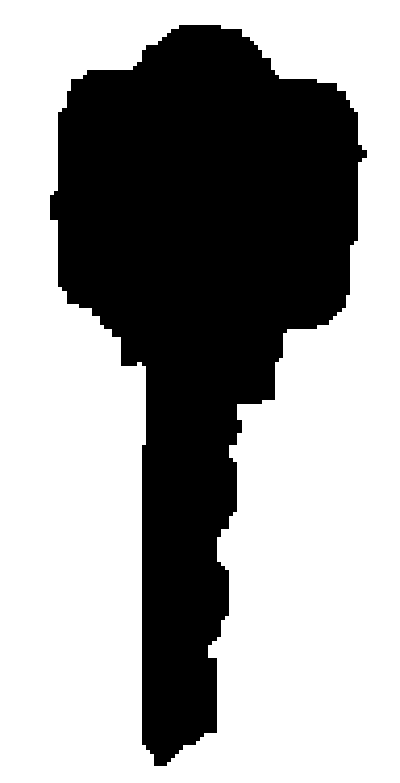,height=0.700000cm}\\
\cline{2-12}
  & \bf{0.936} & \bf{0.925} & \bf{0.855} & \bf{0.854} & \bf{0.829} & 0.571 & 0.563 & 0.501 & 0.489 & 0.440 & 0.410 \\
\hline \epsfig{figure=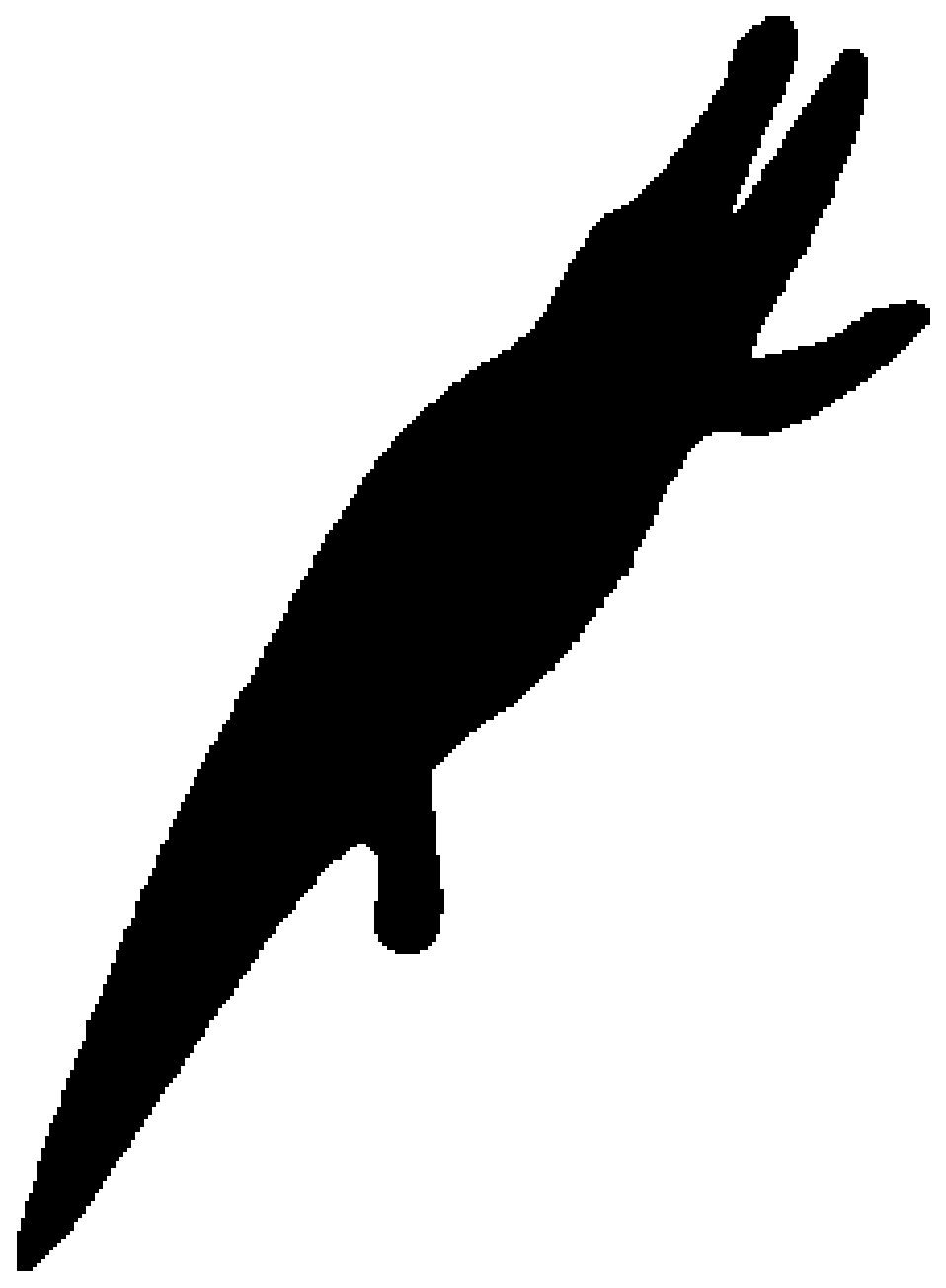,height=0.700000cm}&
\epsfig{figure=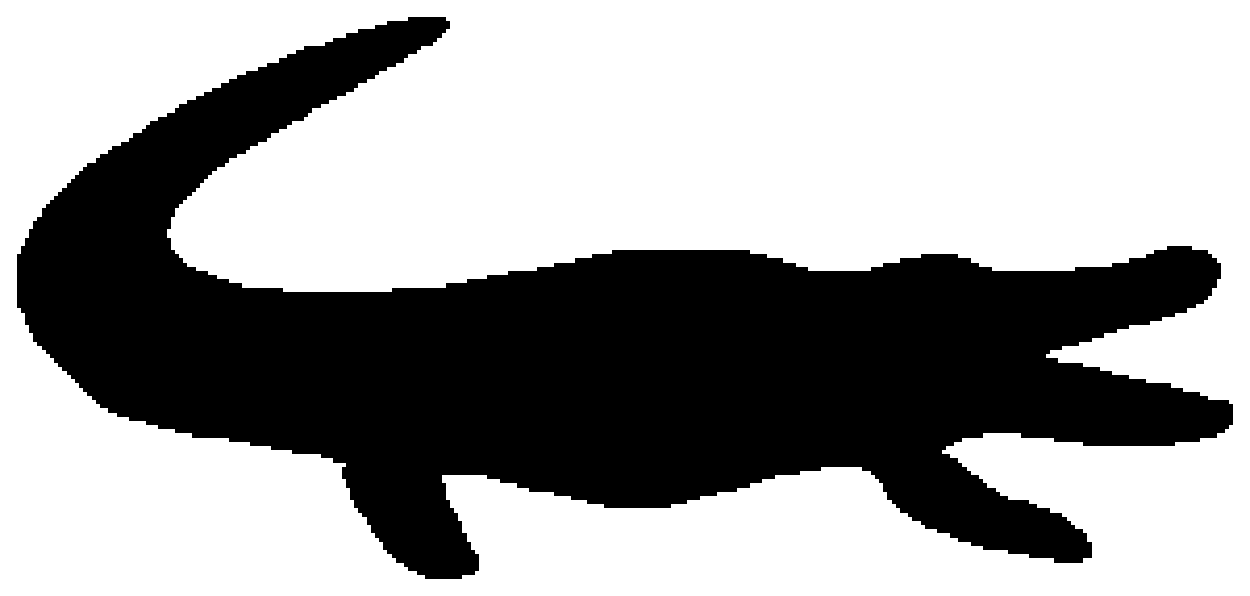,width=0.700000cm}&
\epsfig{figure=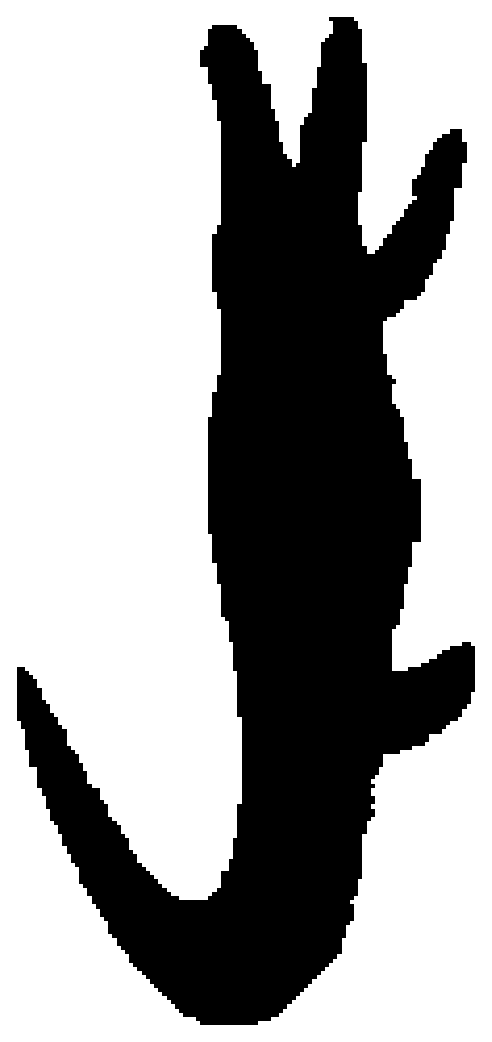,height=0.700000cm}&
\epsfig{figure=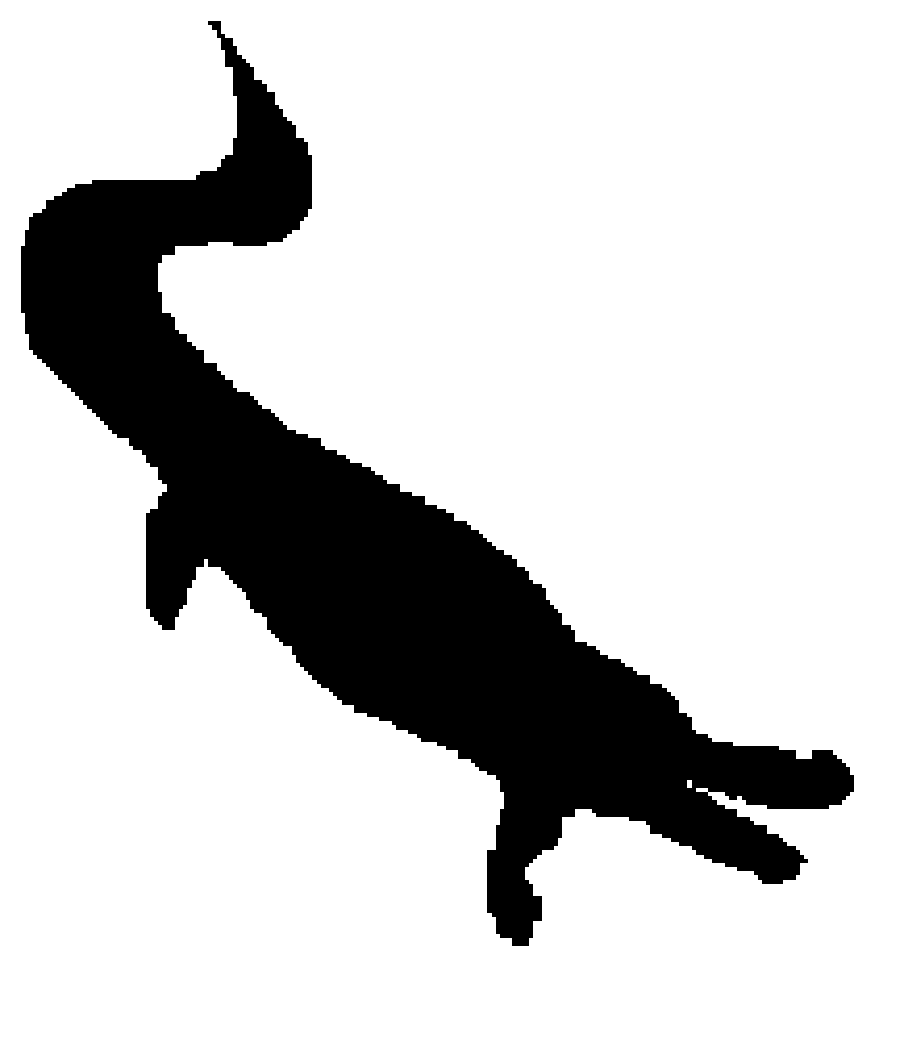,height=0.700000cm}&
\epsfig{figure=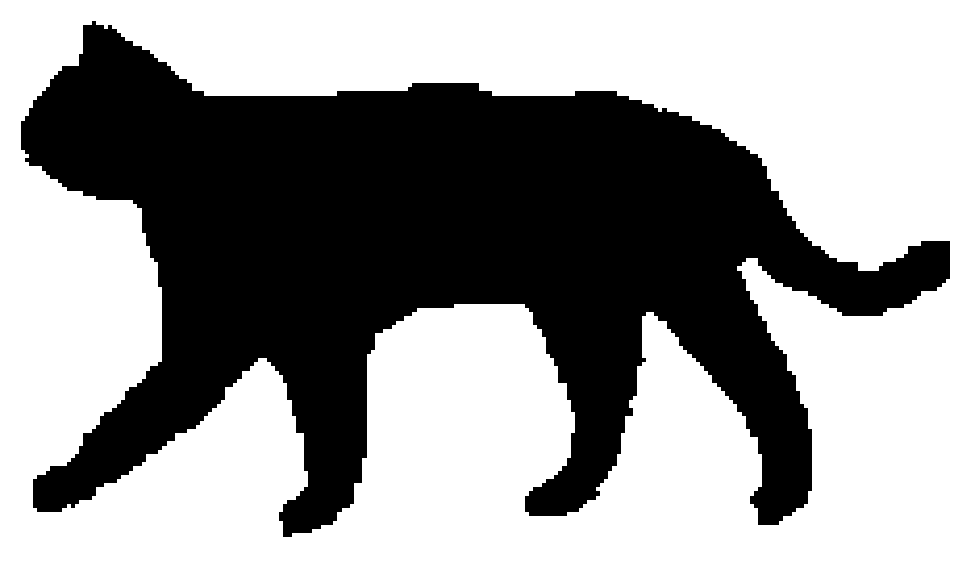,width=0.700000cm}&
\epsfig{figure=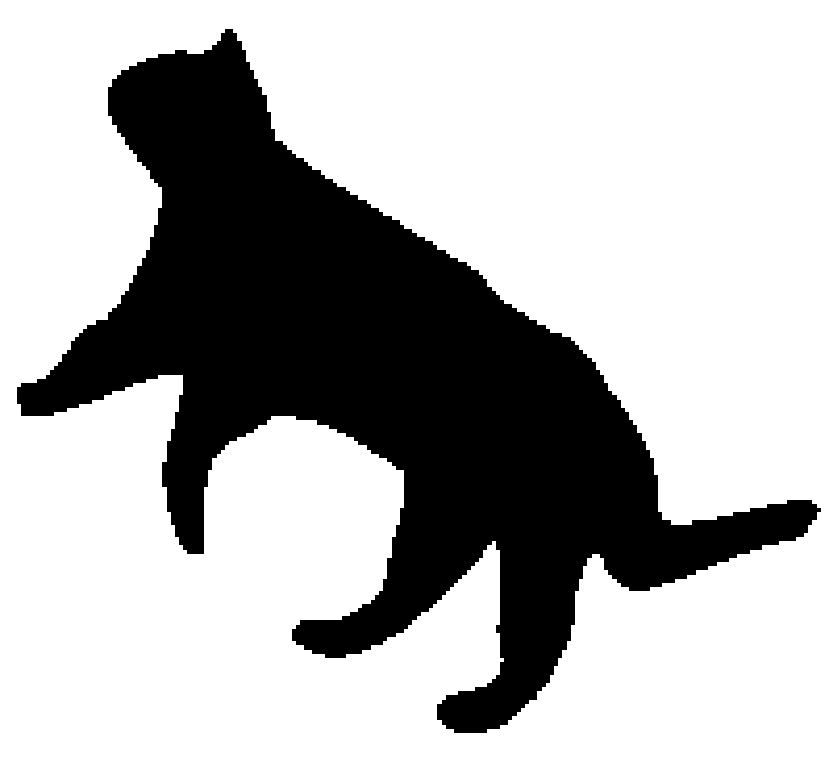,width=0.700000cm}&
\epsfig{figure=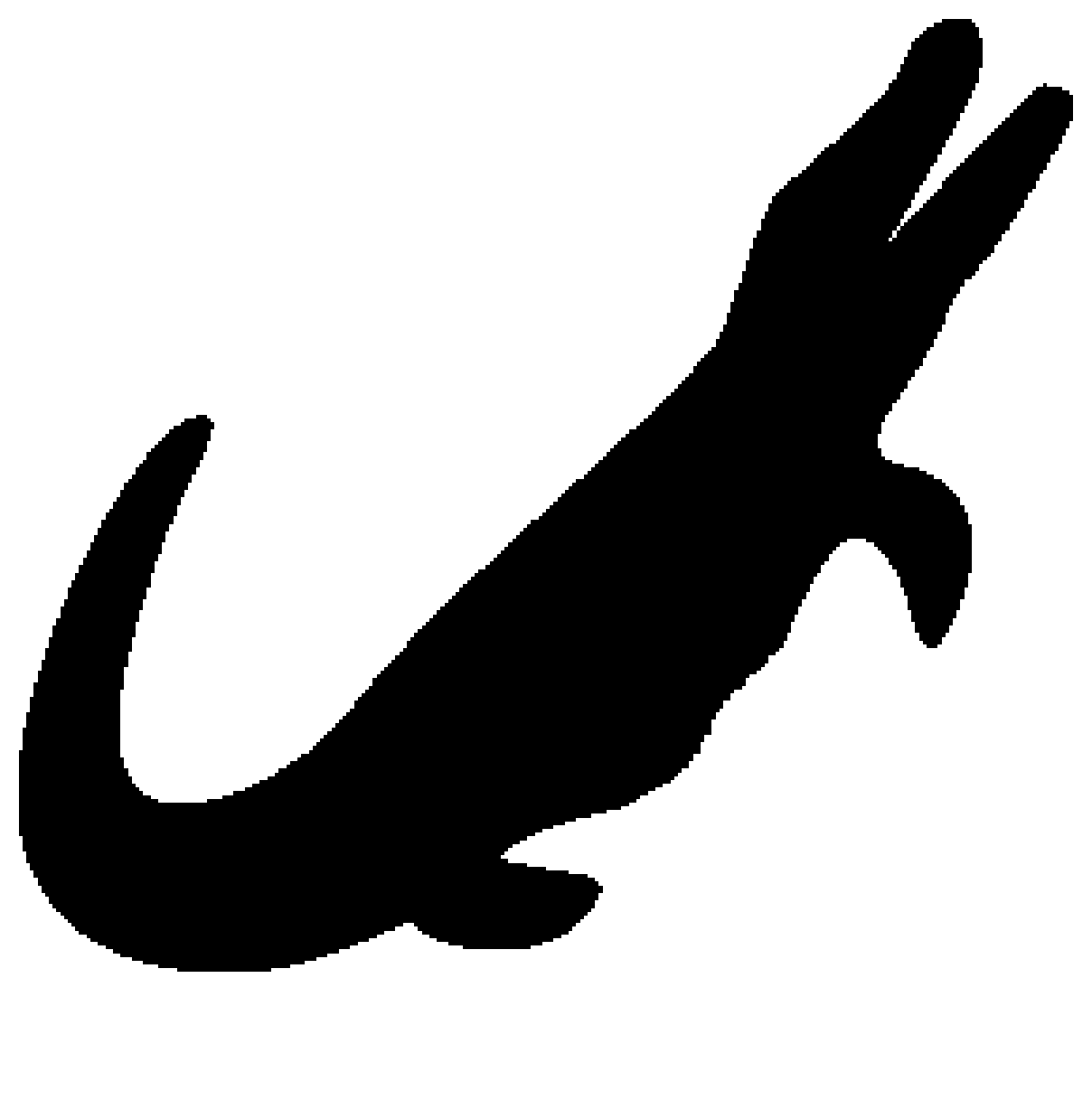,height=0.700000cm}&
\epsfig{figure=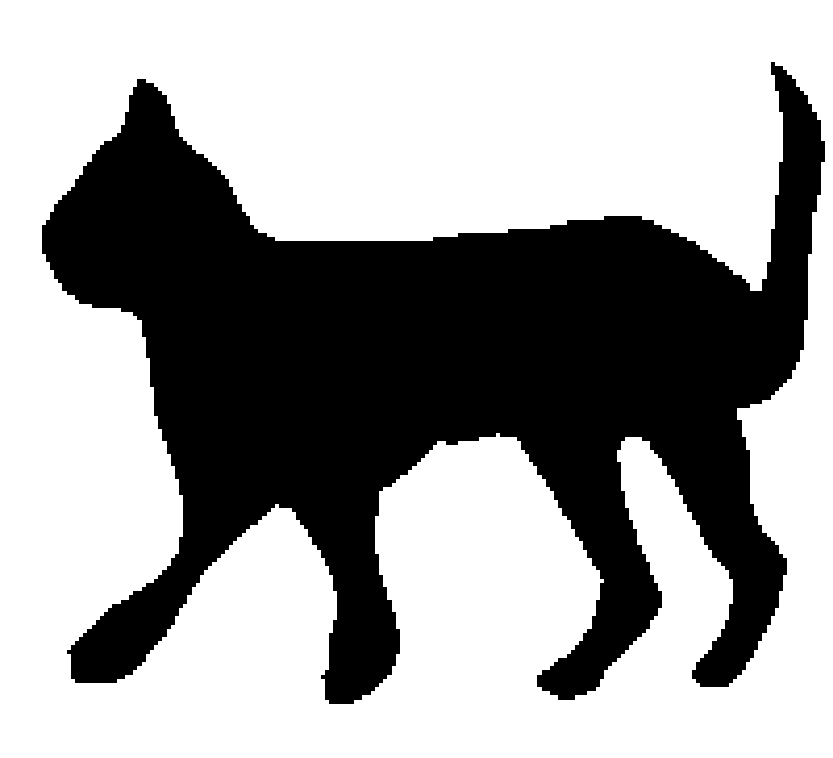,width=0.700000cm}&
\epsfig{figure=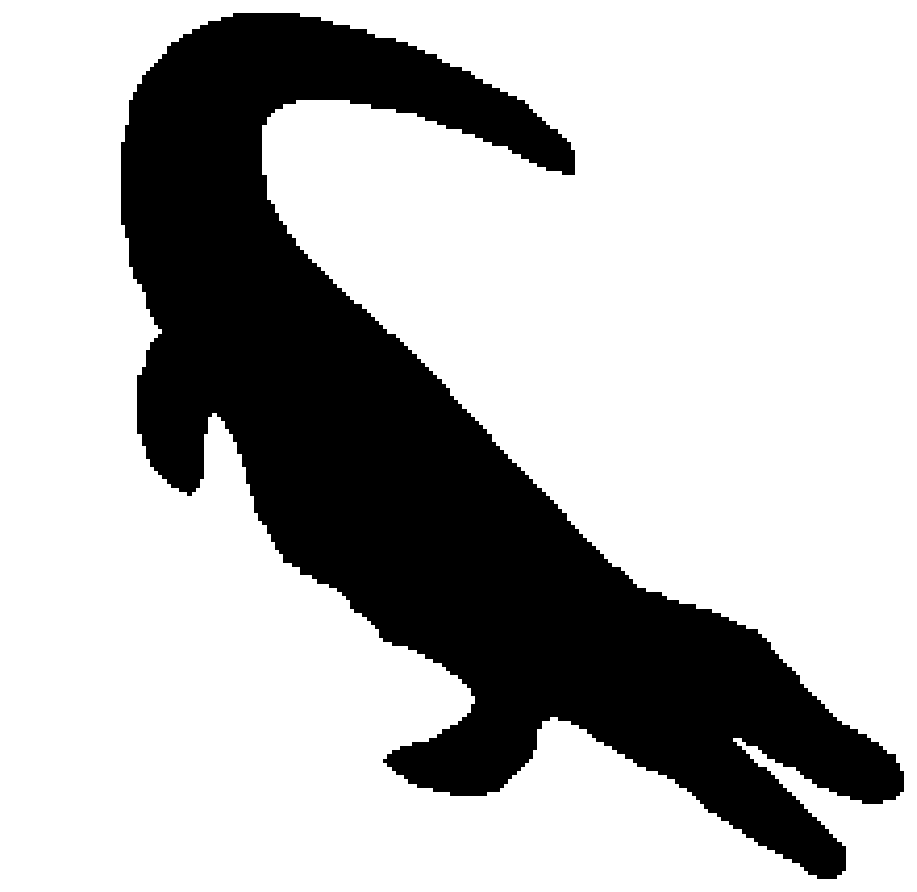,width=0.700000cm}&
\epsfig{figure=PAMI08_png/obj30_2.png,height=0.700000cm}&
\epsfig{figure=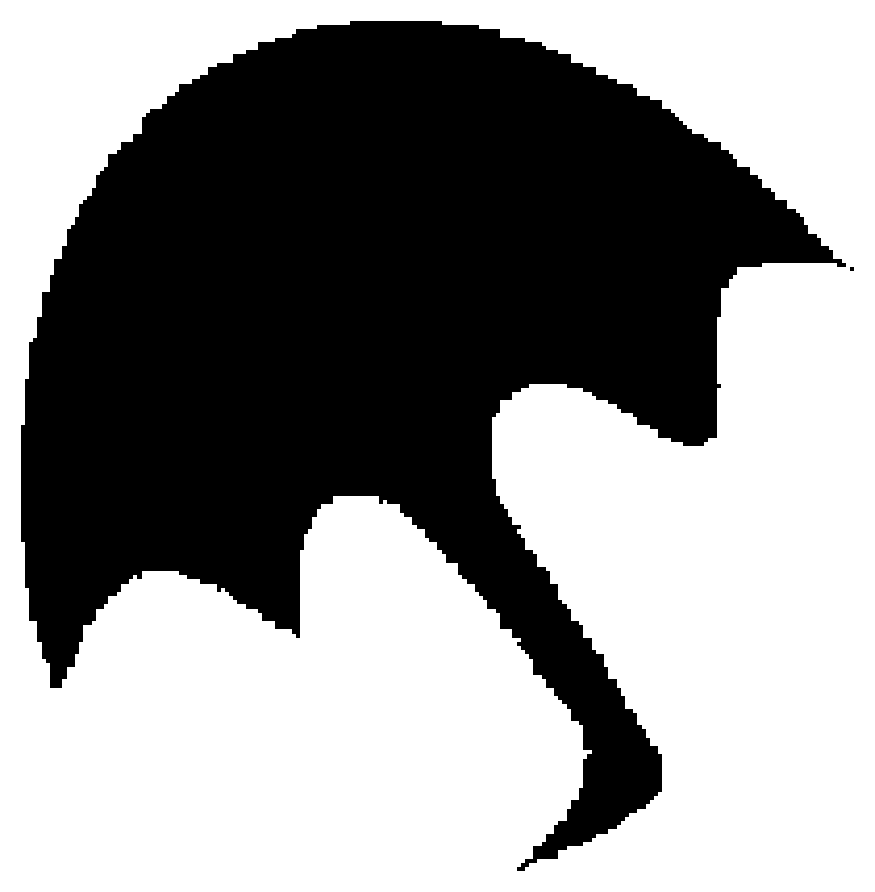,height=0.700000cm}&
\epsfig{figure=PAMI08_png/obj30_4.png,height=0.700000cm}\\
\cline{2-12}
  & \bf{0.560} & \bf{0.529} & \bf{0.483} & 0.443 & 0.431 & \bf{0.414} & 0.403 & \bf{0.330} & 0.323 & 0.313 & 0.304 \\
\hline \epsfig{figure=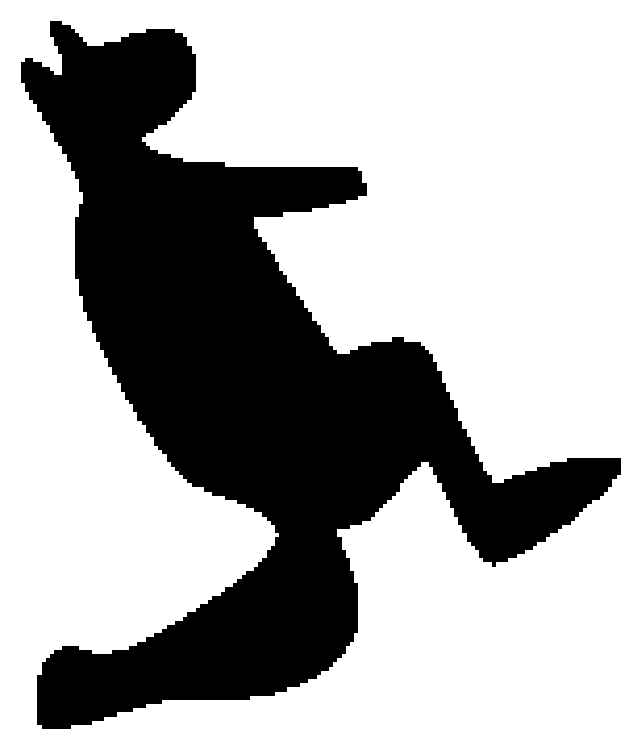,height=0.700000cm}&
\epsfig{figure=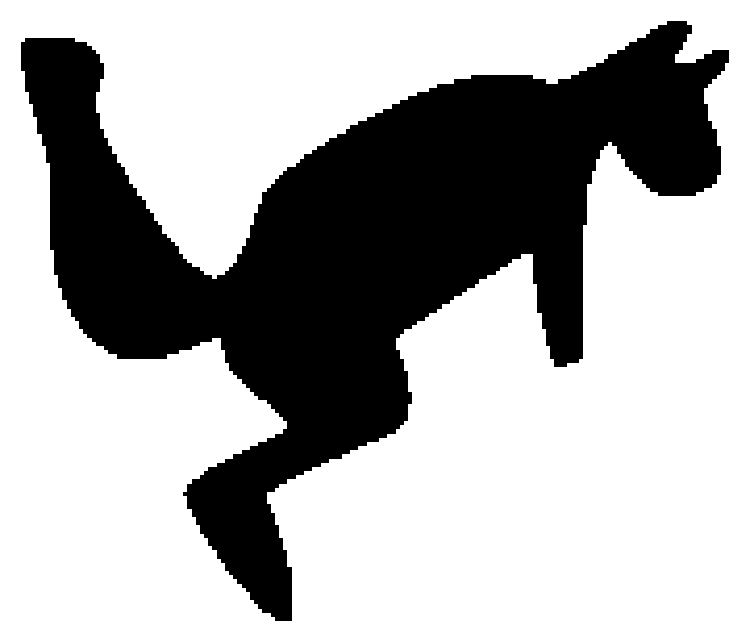,width=0.700000cm}&
\epsfig{figure=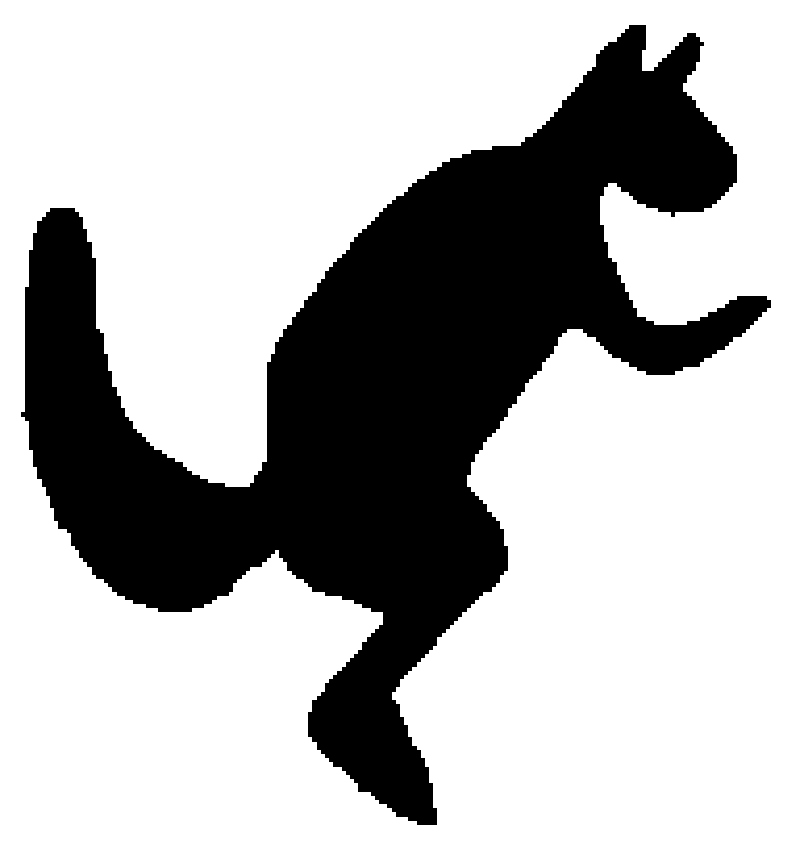,height=0.700000cm}&
\epsfig{figure=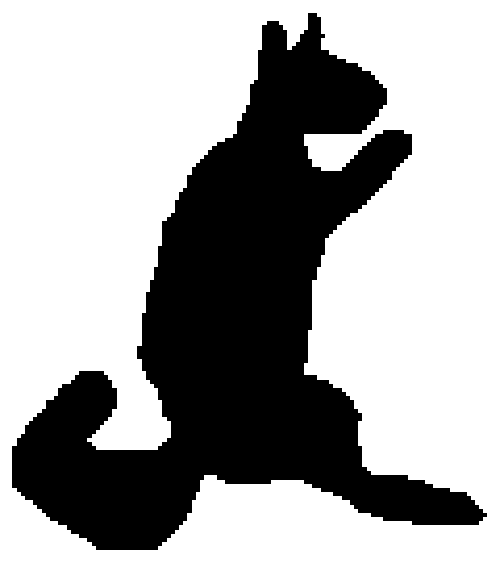,height=0.700000cm}&
\epsfig{figure=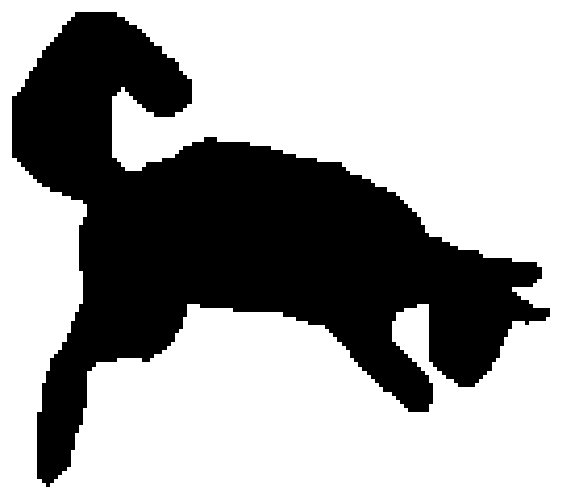,width=0.700000cm}&
\epsfig{figure=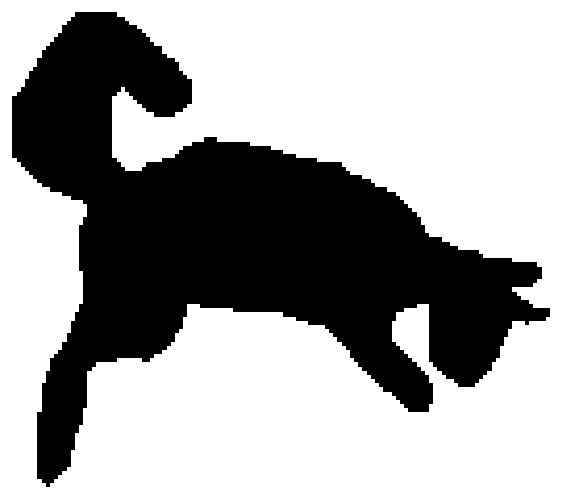,width=0.700000cm}&
\epsfig{figure=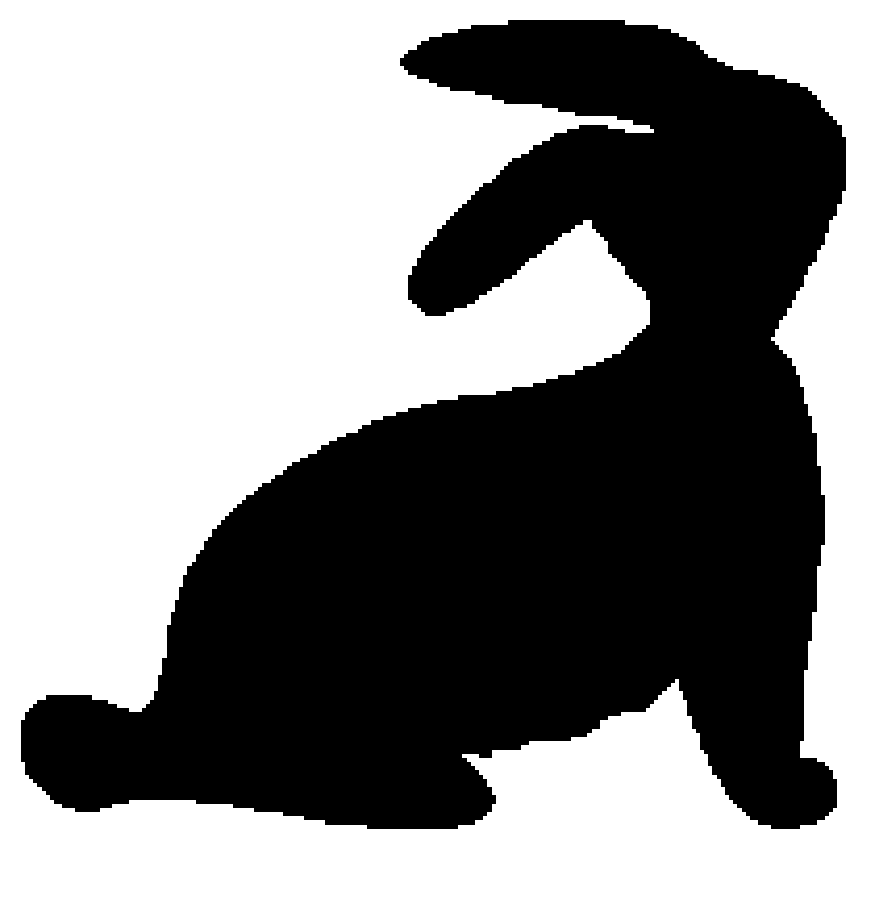,height=0.700000cm}&
\epsfig{figure=PAMI08_png/obj3_2.png,width=0.700000cm}&
\epsfig{figure=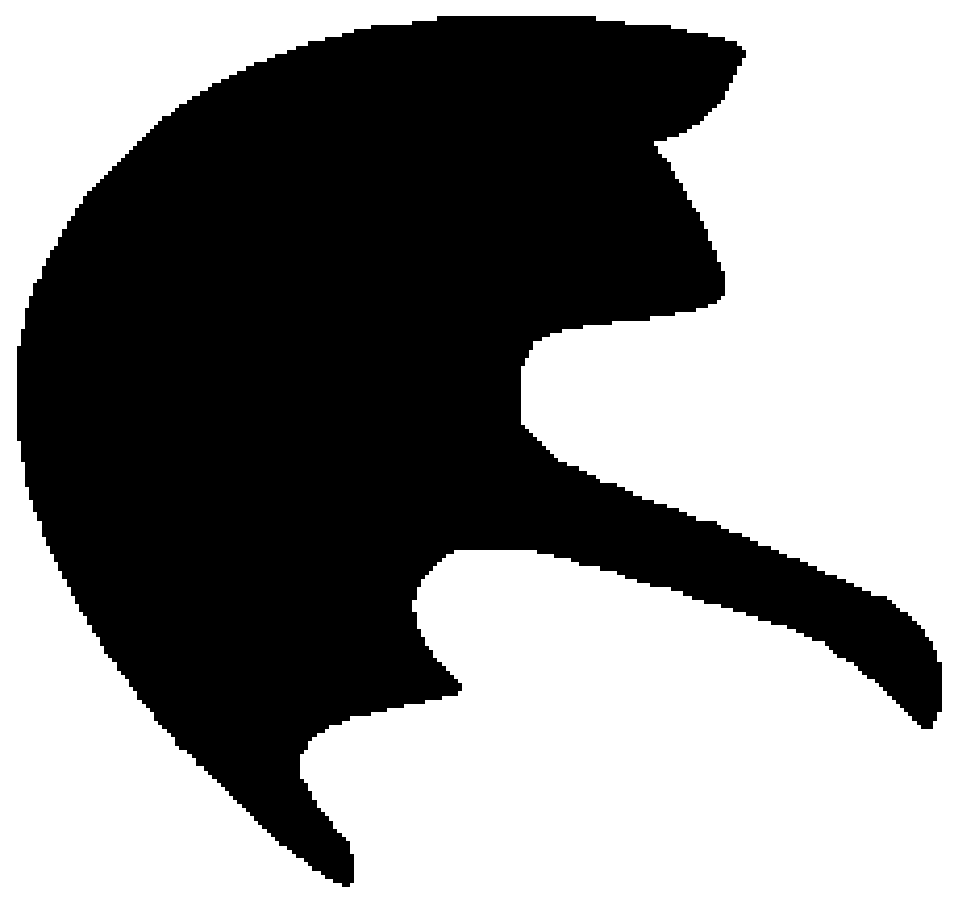,width=0.700000cm}&
\epsfig{figure=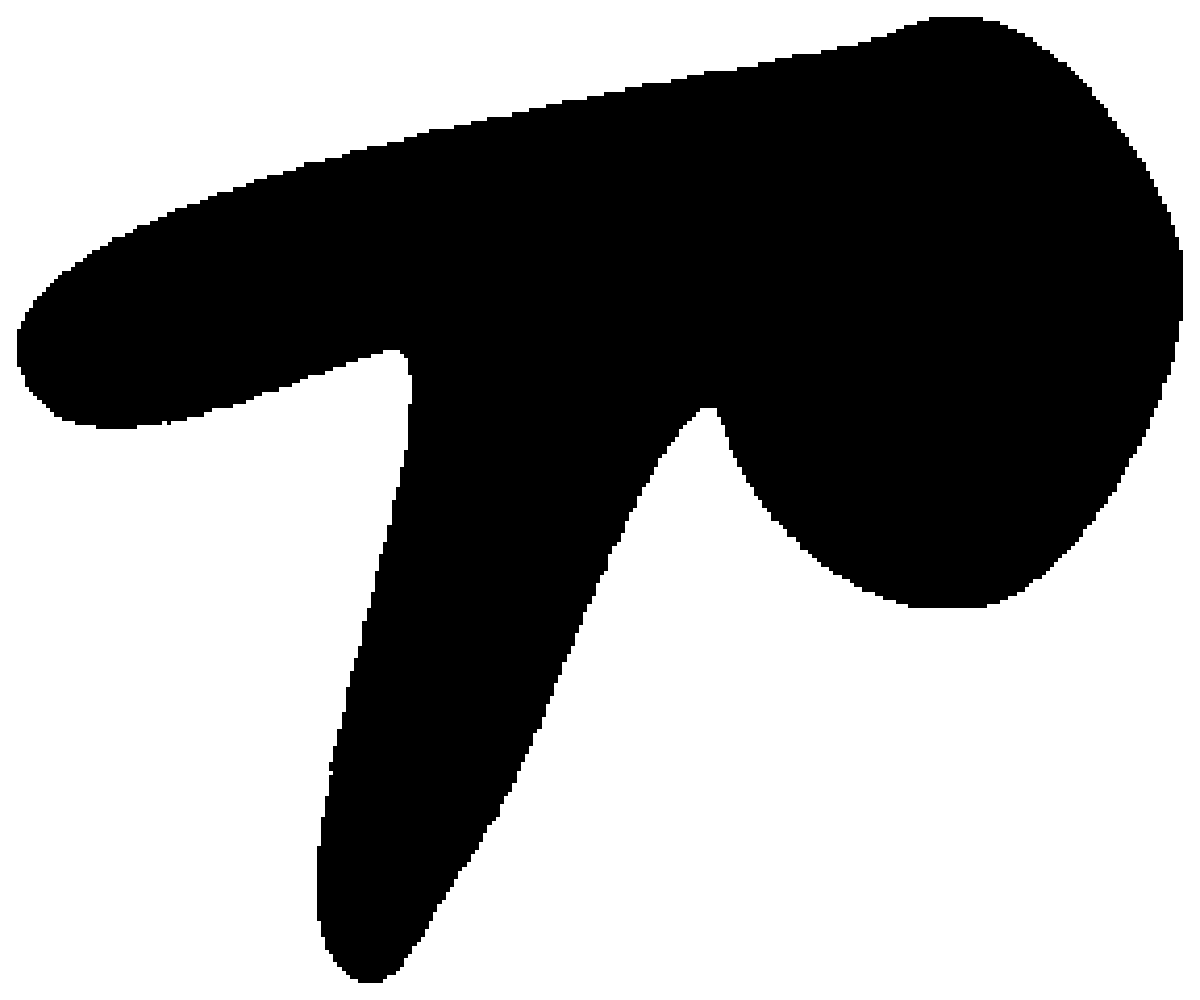,width=0.700000cm}&
\epsfig{figure=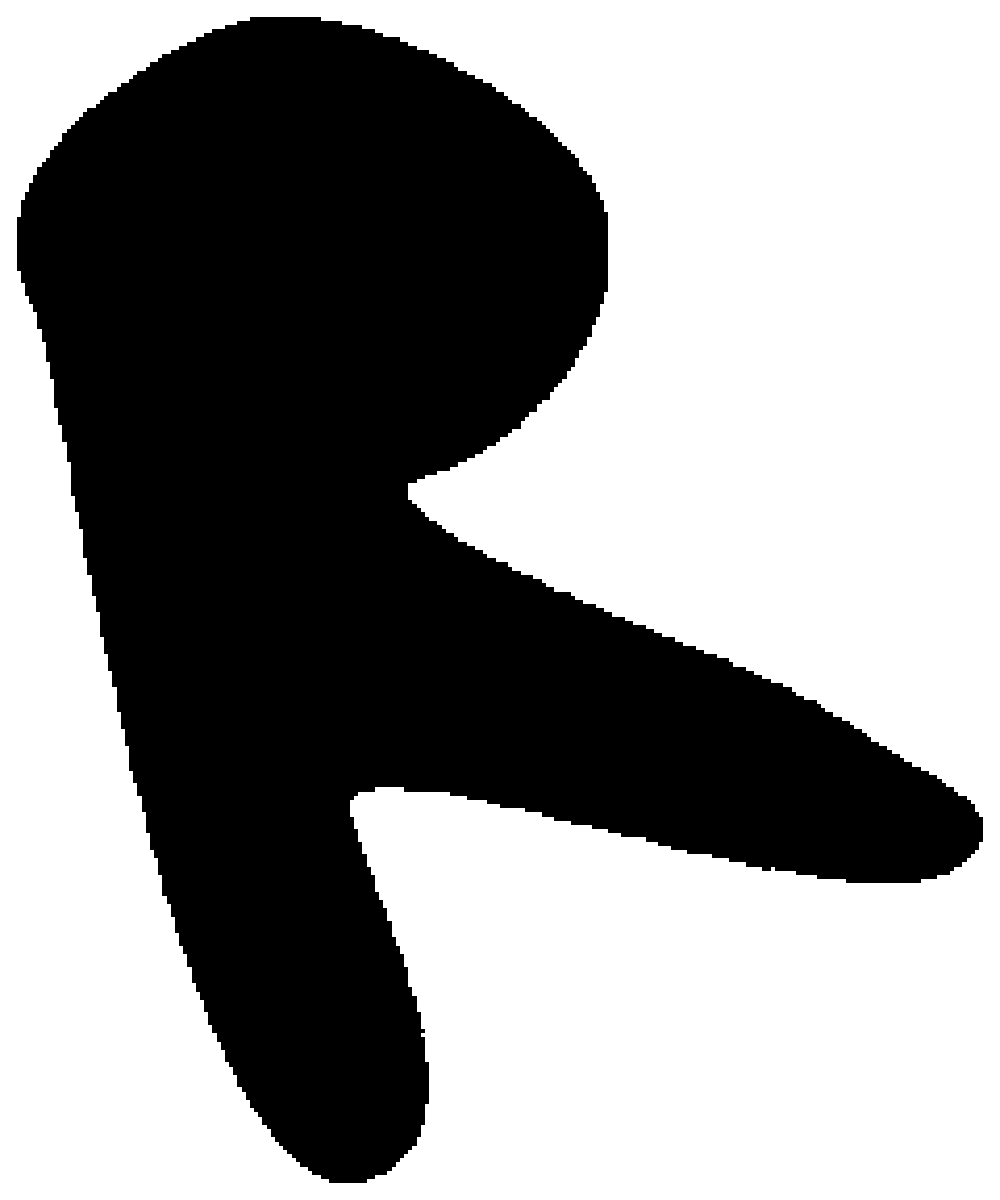,height=0.700000cm}&
\epsfig{figure=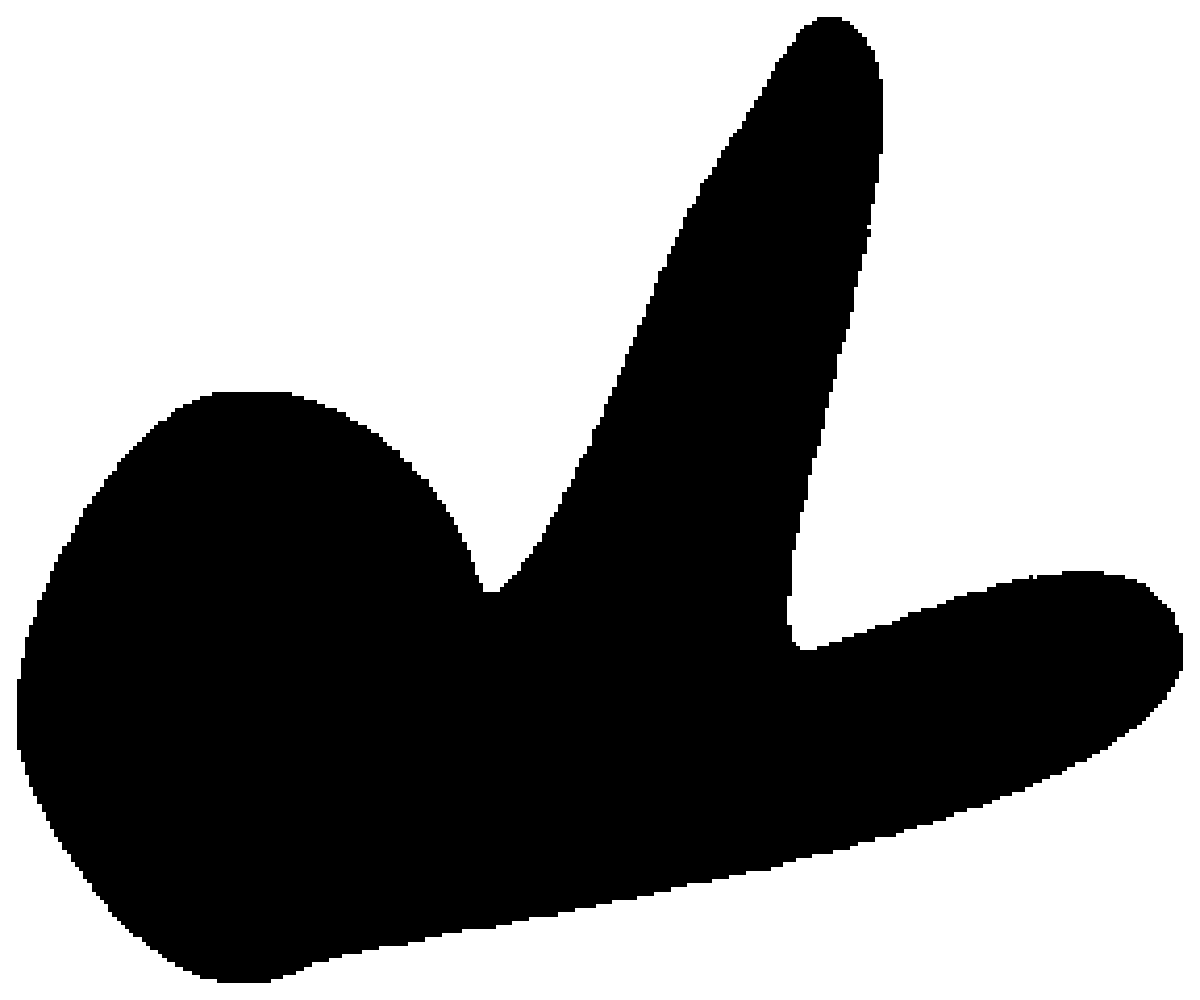,width=0.700000cm}\\
\cline{2-12}
  & \bf{0.967} & \bf{0.904} & \bf{0.599} & \bf{0.585} & \bf{0.585} & 0.446 & 0.429 & 0.404 & 0.396 & 0.391 & 0.389 \\
\hline \epsfig{figure=PAMI08_png/obj20_1.png,width=0.700000cm}&
\epsfig{figure=PAMI08_png/obj20_3.png,width=0.700000cm}&
\epsfig{figure=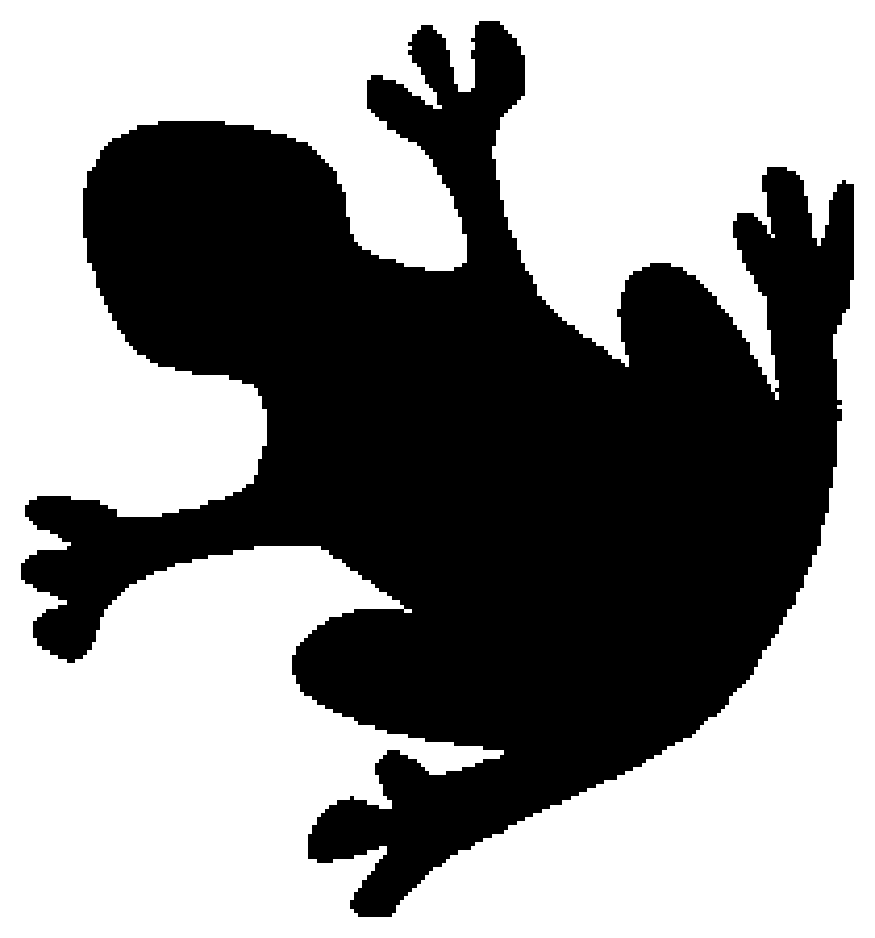,height=0.700000cm}&
\epsfig{figure=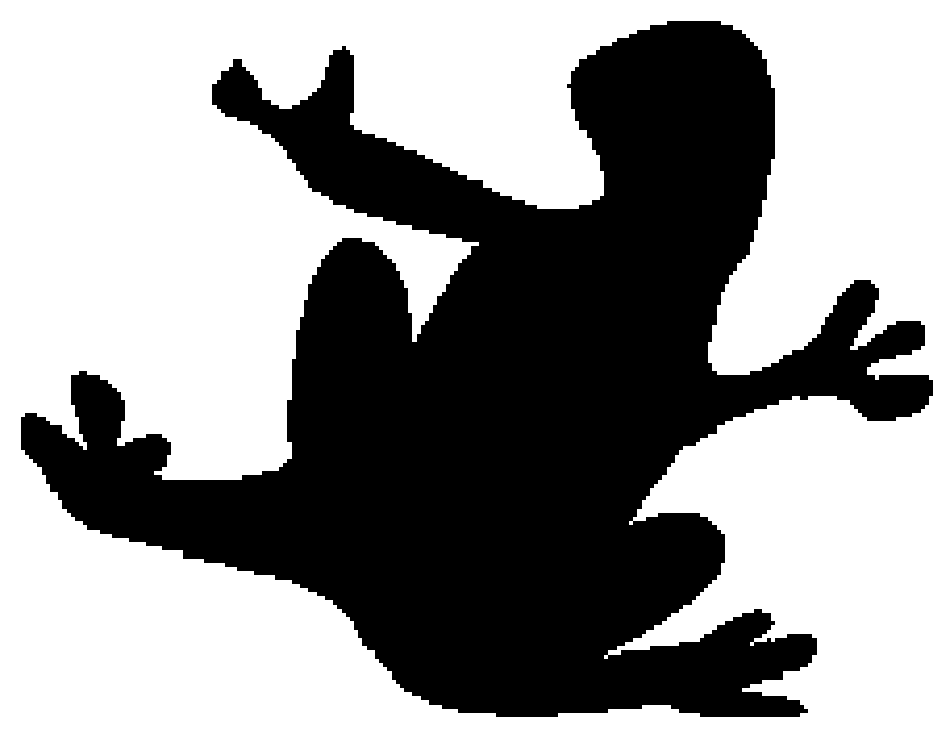,width=0.700000cm}&
\epsfig{figure=PAMI08_png/obj20_5.png,width=0.700000cm}&
\epsfig{figure=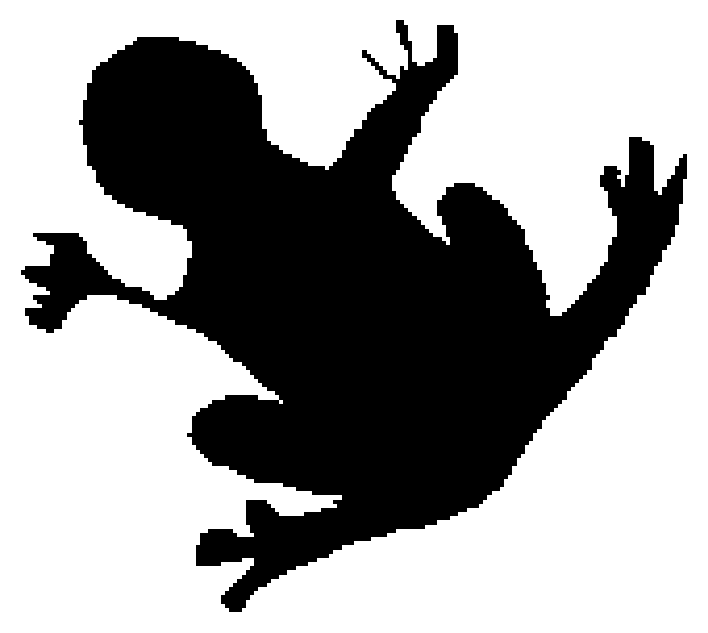,width=0.700000cm}&
\epsfig{figure=PAMI08_png/obj7_4.png,width=0.700000cm}&
\epsfig{figure=PAMI08_png/obj9_5.png,width=0.700000cm}&
\epsfig{figure=PAMI08_png/obj24_2.png,width=0.700000cm}&
\epsfig{figure=PAMI08_png/obj7_1.png,height=0.700000cm}&
\epsfig{figure=PAMI08_png/obj26_3.png,height=0.700000cm}&
\epsfig{figure=PAMI08_png/obj4_3.png,height=0.700000cm}\\
\cline{2-12}
  & \bf{0.977} & \bf{0.969} & \bf{0.962} & \bf{0.901} & \bf{0.815} & 0.670 & 0.635 & 0.597 & 0.585 & 0.571 & 0.569 \\
\hline \epsfig{figure=PAMI08_png/obj21_2.png,height=0.700000cm}&
\epsfig{figure=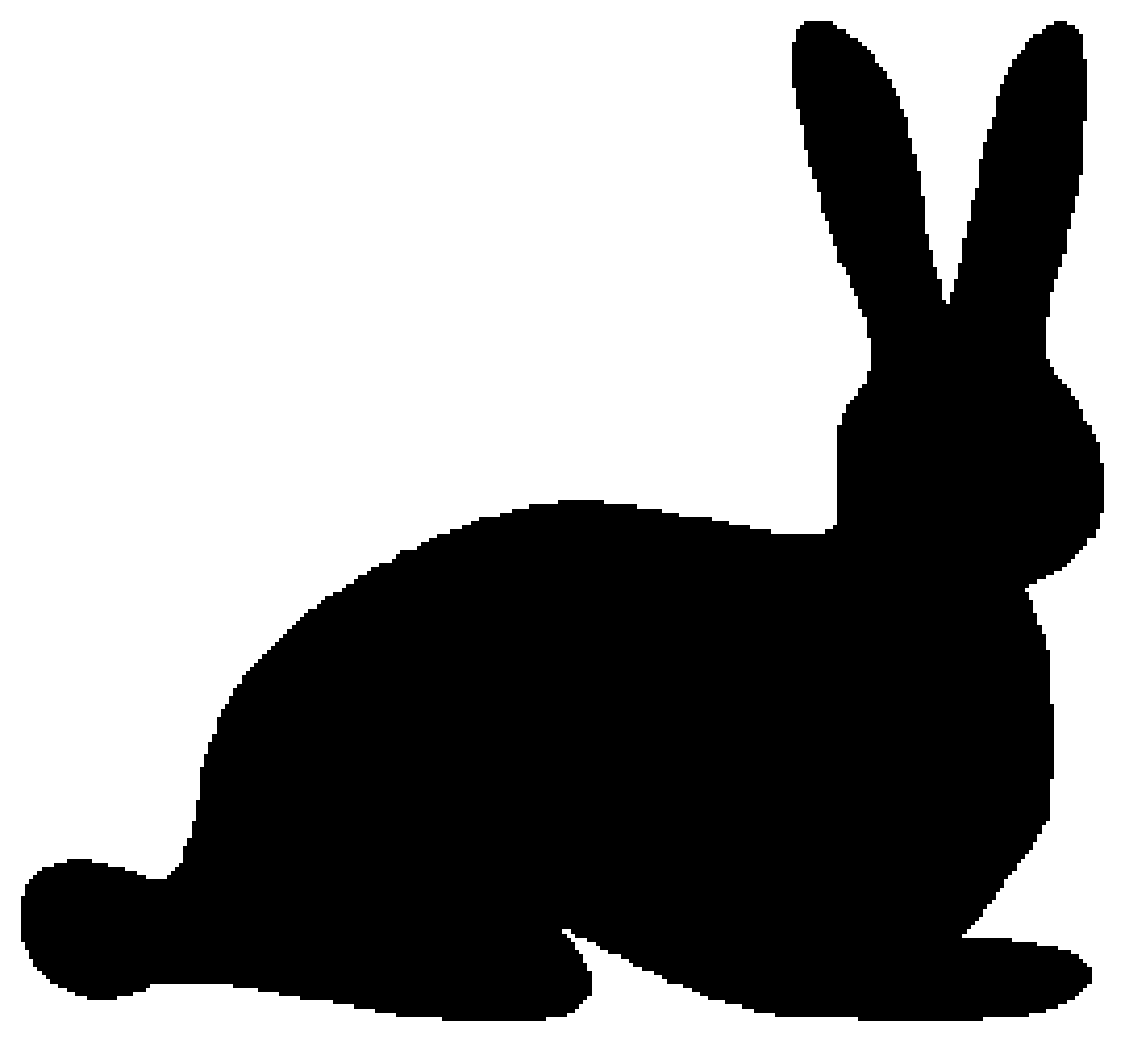,width=0.700000cm}&
\epsfig{figure=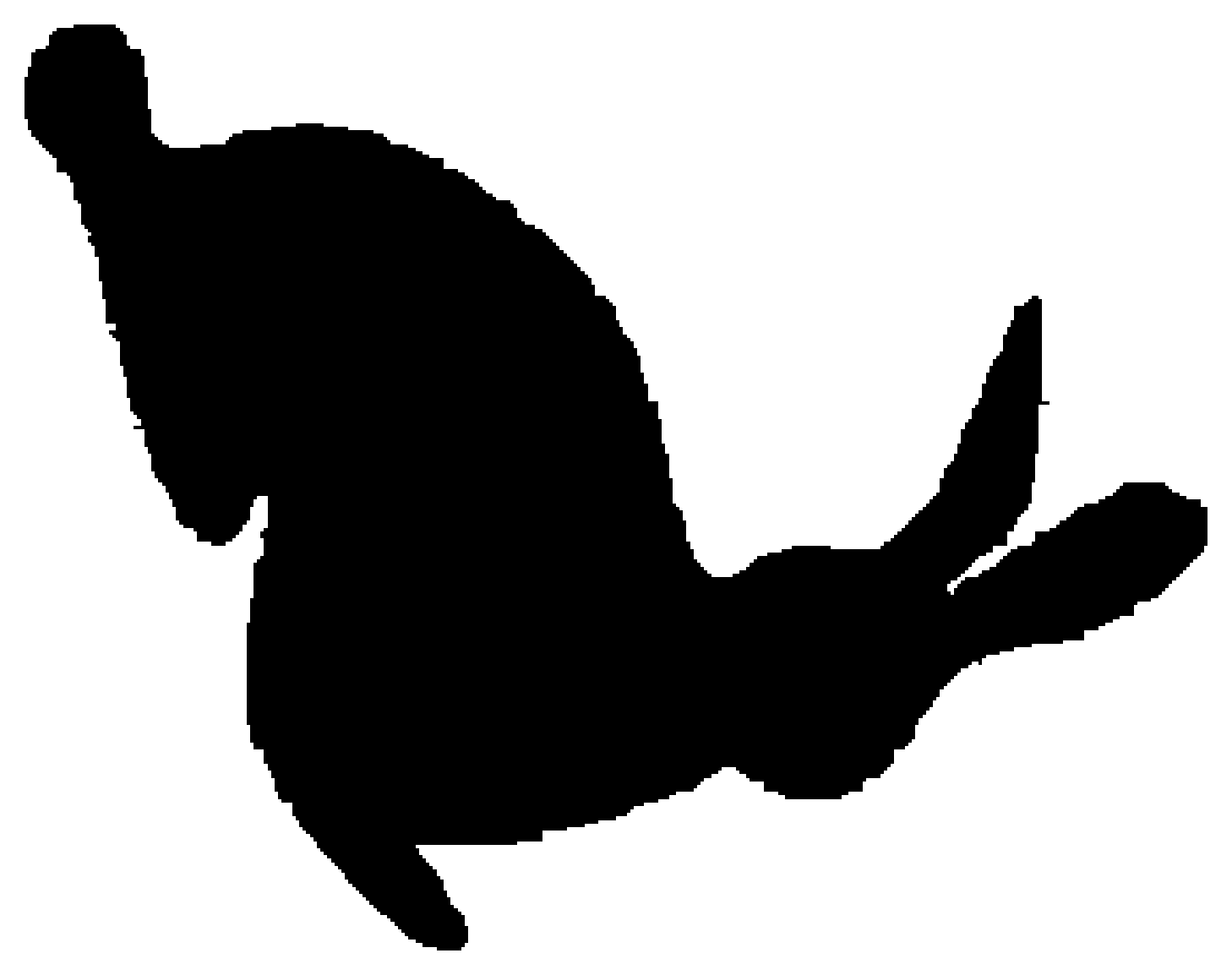,width=0.700000cm}&
\epsfig{figure=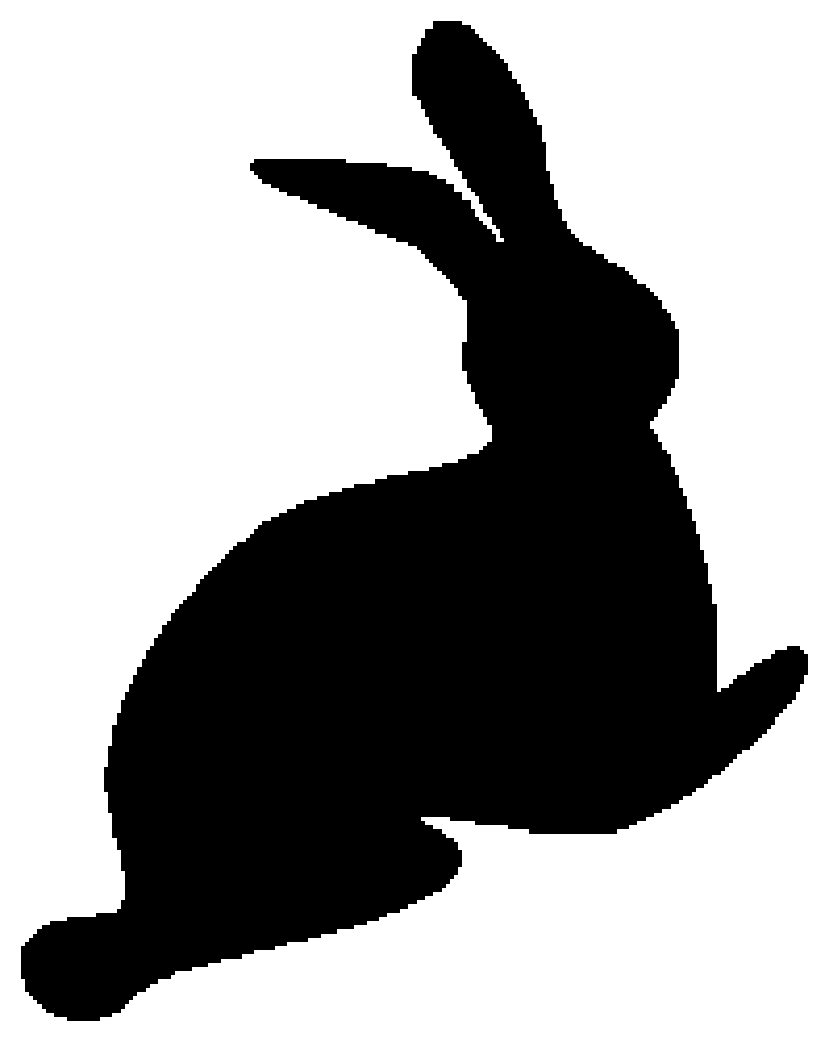,height=0.700000cm}&
\epsfig{figure=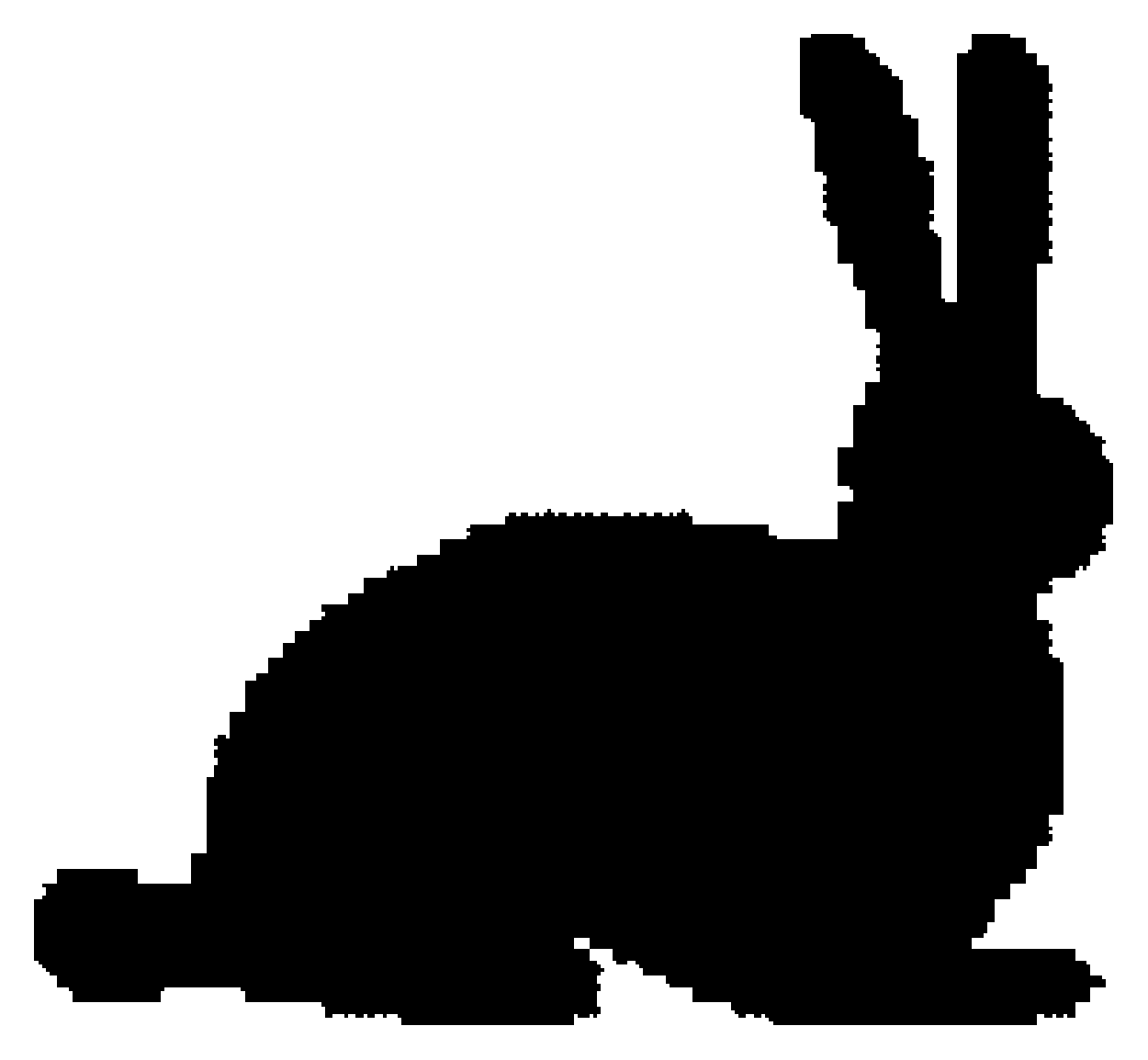,width=0.700000cm}&
\epsfig{figure=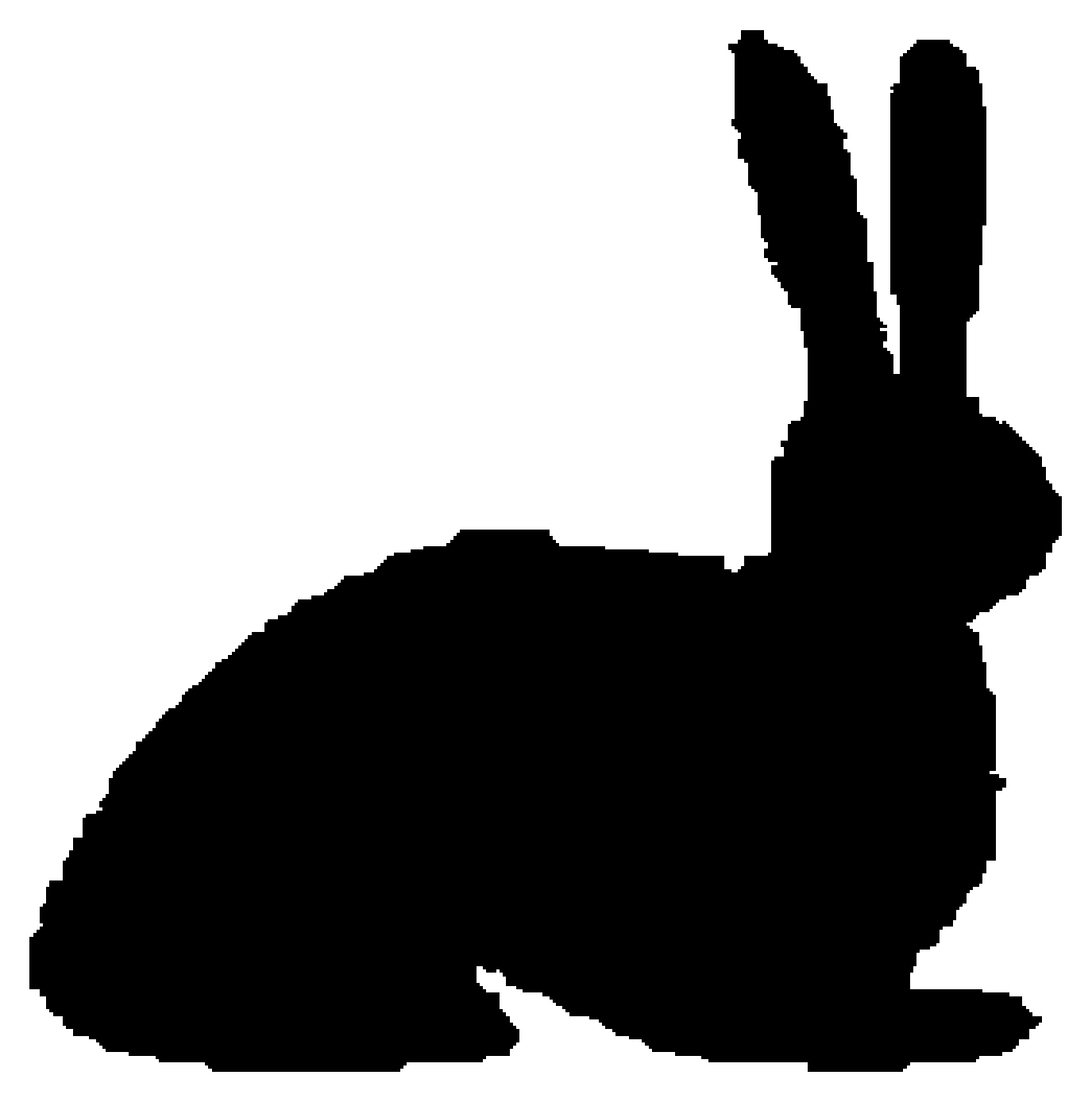,height=0.700000cm}&
\epsfig{figure=PAMI08_png/obj25_3.png,width=0.700000cm}&
\epsfig{figure=PAMI08_png/obj25_5.png,height=0.700000cm}&
\epsfig{figure=PAMI08_png/obj25_4.png,height=0.700000cm}&
\epsfig{figure=PAMI08_png/obj19_4.png,height=0.700000cm}&
\epsfig{figure=PAMI08_png/obj25_1.png,height=0.700000cm}&
\epsfig{figure=PAMI08_png/obj19_2.png,width=0.700000cm}\\
\cline{2-12}
  & \bf{0.838} & \bf{0.837} & \bf{0.817} & \bf{0.815} & \bf{0.637} & 0.514 & 0.488 & 0.475 & 0.446 & 0.442 & 0.426 \\
\hline \epsfig{figure=PAMI08_png/obj22_2.png,height=0.700000cm}&
\epsfig{figure=PAMI08_png/obj22_6.png,width=0.700000cm}&
\epsfig{figure=PAMI08_png/obj22_4.png,width=0.700000cm}&
\epsfig{figure=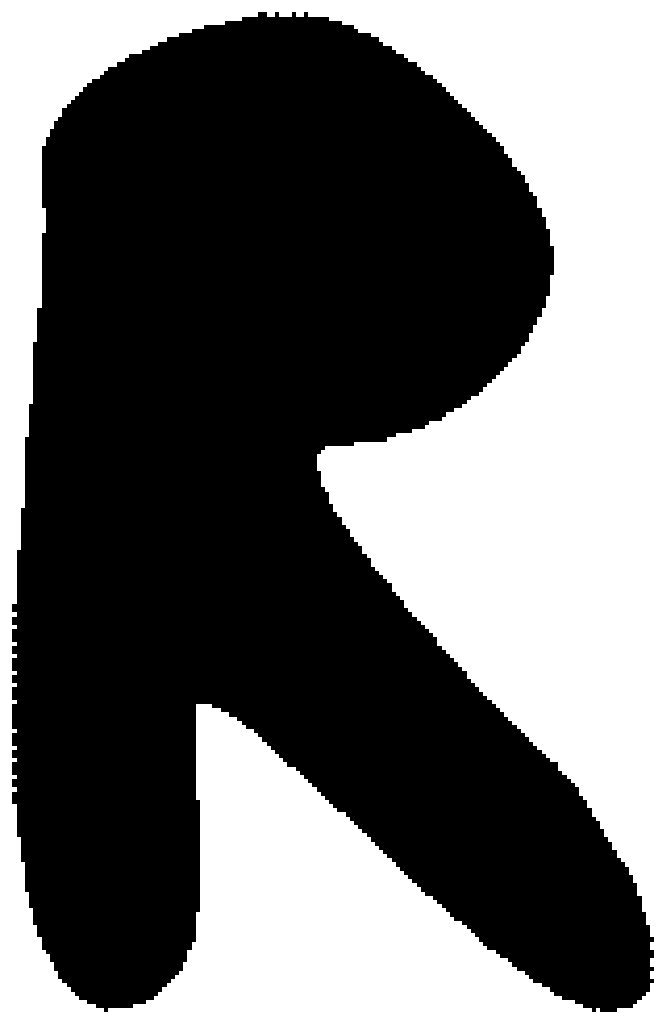,height=0.700000cm}&
\epsfig{figure=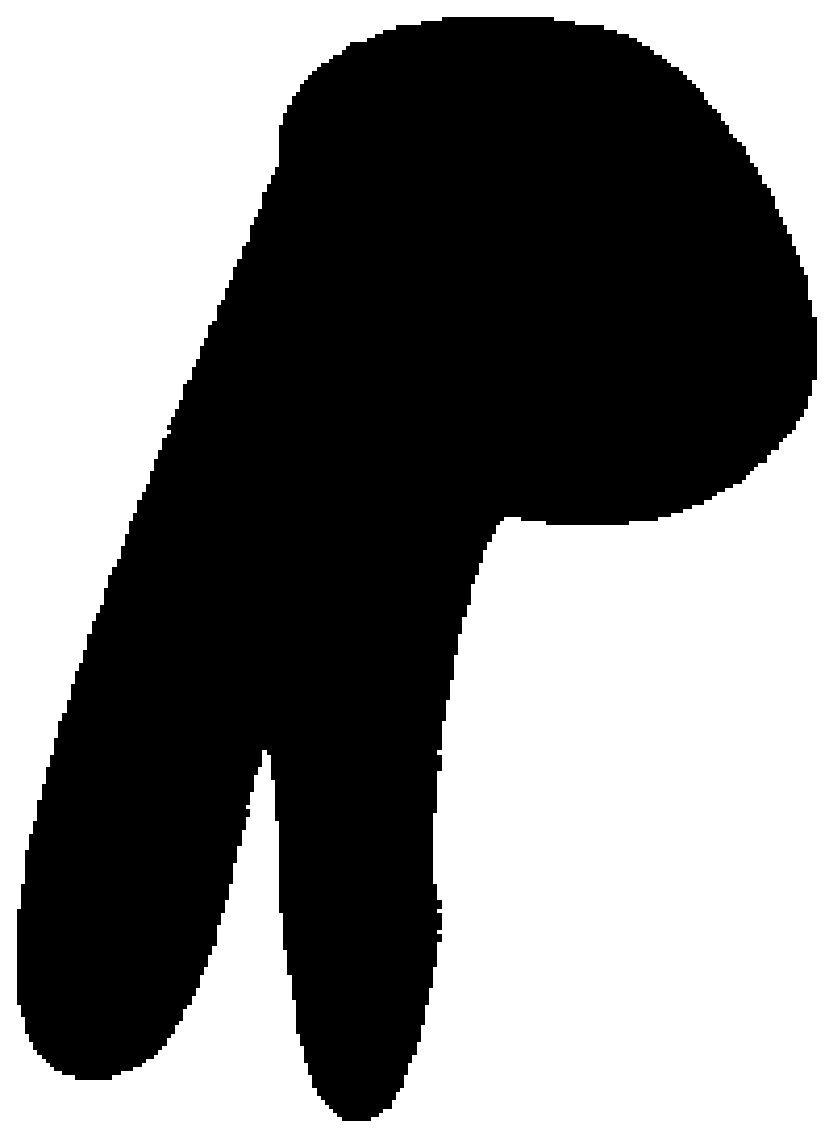,height=0.700000cm}&
\epsfig{figure=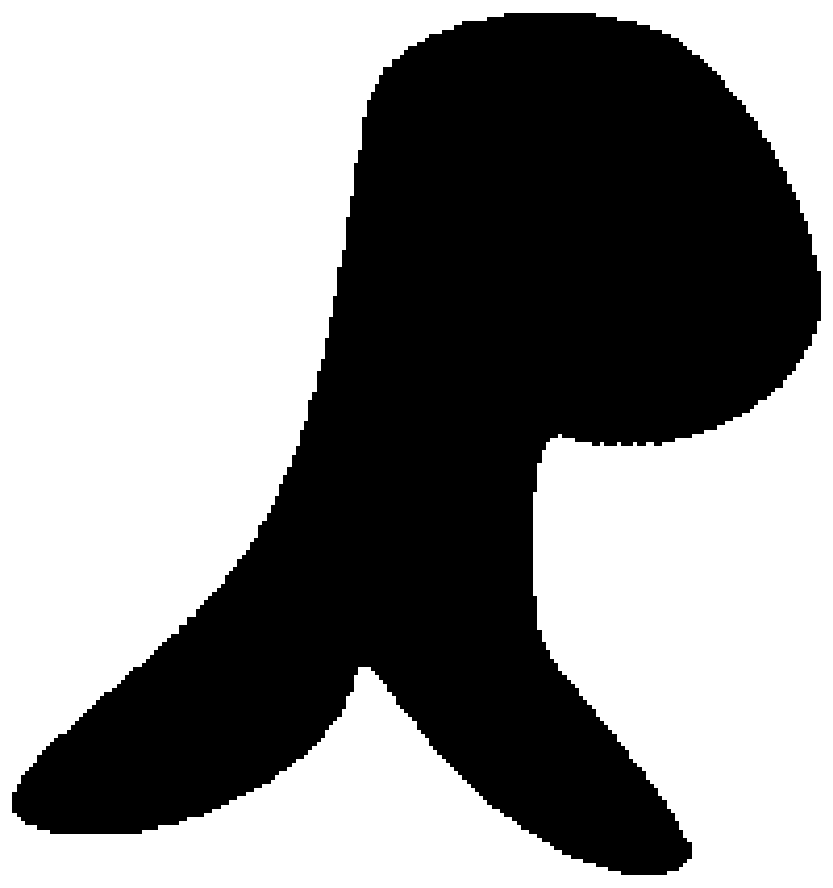,height=0.700000cm}&
\epsfig{figure=PAMI08_png/obj19_2.png,width=0.700000cm}&
\epsfig{figure=PAMI08_png/obj19_4.png,height=0.700000cm}&
\epsfig{figure=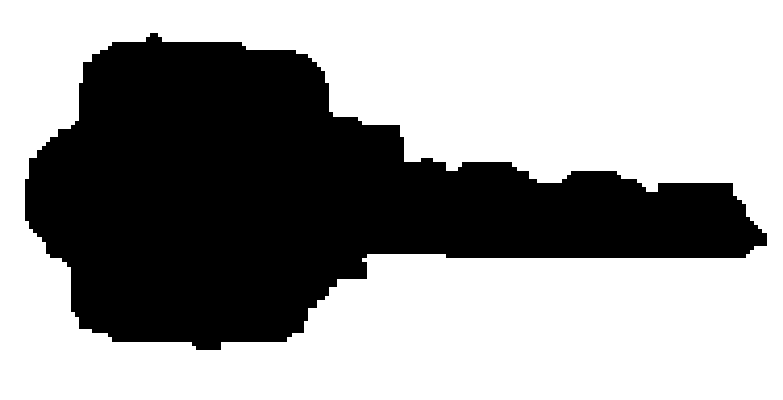,width=0.700000cm}&
\epsfig{figure=PAMI08_png/obj29_5.png,height=0.700000cm}&
\epsfig{figure=PAMI08_png/obj4_1.png,height=0.700000cm}&
\epsfig{figure=PAMI08_png/obj7_4.png,width=0.700000cm}\\
\cline{2-12}
  & \bf{0.995} & \bf{0.993} & \bf{0.897} & \bf{0.670} & \bf{0.486} & 0.418 & 0.391 & 0.378 & 0.370 & 0.352 & 0.348 \\
\hline \epsfig{figure=PAMI08_png/obj23_4.png,height=0.700000cm}&
\epsfig{figure=PAMI08_png/obj23_2.png,height=0.700000cm}&
\epsfig{figure=PAMI08_png/obj23_1.png,width=0.700000cm}&
\epsfig{figure=PAMI08_png/obj23_3.png,height=0.700000cm}&
\epsfig{figure=PAMI08_png/obj23_5.png,width=0.700000cm}&
\epsfig{figure=PAMI08_png/obj23_6.png,width=0.700000cm}&
\epsfig{figure=PAMI08_png/obj6_2.png,height=0.700000cm}&
\epsfig{figure=PAMI08_png/obj6_6.png,width=0.700000cm}&
\epsfig{figure=PAMI08_png/obj6_5.png,width=0.700000cm}&
\epsfig{figure=PAMI08_png/obj6_1.png,height=0.700000cm}&
\epsfig{figure=PAMI08_png/obj6_3.png,width=0.700000cm}&
\epsfig{figure=PAMI08_png/obj6_4.png,width=0.700000cm}\\
\cline{2-12}
  & \bf{0.963} & \bf{0.885} & \bf{0.878} & \bf{0.698} & \bf{0.624} & 0.579 & 0.559 & 0.535 & 0.491 & 0.425 & 0.411 \\
\hline \epsfig{figure=PAMI08_png/obj24_3.png,width=0.700000cm}&
\epsfig{figure=PAMI08_png/obj24_1.png,width=0.700000cm}&
\epsfig{figure=PAMI08_png/obj24_5.png,width=0.700000cm}&
\epsfig{figure=PAMI08_png/obj24_4.png,height=0.700000cm}&
\epsfig{figure=PAMI08_png/obj9_5.png,width=0.700000cm}&
\epsfig{figure=PAMI08_png/obj24_6.png,width=0.700000cm}&
\epsfig{figure=PAMI08_png/obj24_2.png,width=0.700000cm}&
\epsfig{figure=PAMI08_png/obj9_2.png,width=0.700000cm}&
\epsfig{figure=PAMI08_png/obj9_6.png,height=0.700000cm}&
\epsfig{figure=PAMI08_png/obj9_4.png,width=0.700000cm}&
\epsfig{figure=PAMI08_png/obj1_4.png,height=0.700000cm}&
\epsfig{figure=PAMI08_png/obj9_1.png,width=0.700000cm}\\
\cline{2-12}
  & \bf{0.897} & \bf{0.881} & \bf{0.872} & 0.803 & \bf{0.769} & \bf{0.764} & 0.732 & 0.716 & 0.661 & 0.625 & 0.613 \\
\hline \epsfig{figure=PAMI08_png/obj25_4.png,height=0.700000cm}&
\epsfig{figure=PAMI08_png/obj25_1.png,height=0.700000cm}&
\epsfig{figure=PAMI08_png/obj25_5.png,height=0.700000cm}&
\epsfig{figure=PAMI08_png/obj25_3.png,width=0.700000cm}&
\epsfig{figure=PAMI08_png/obj25_2.png,height=0.700000cm}&
\epsfig{figure=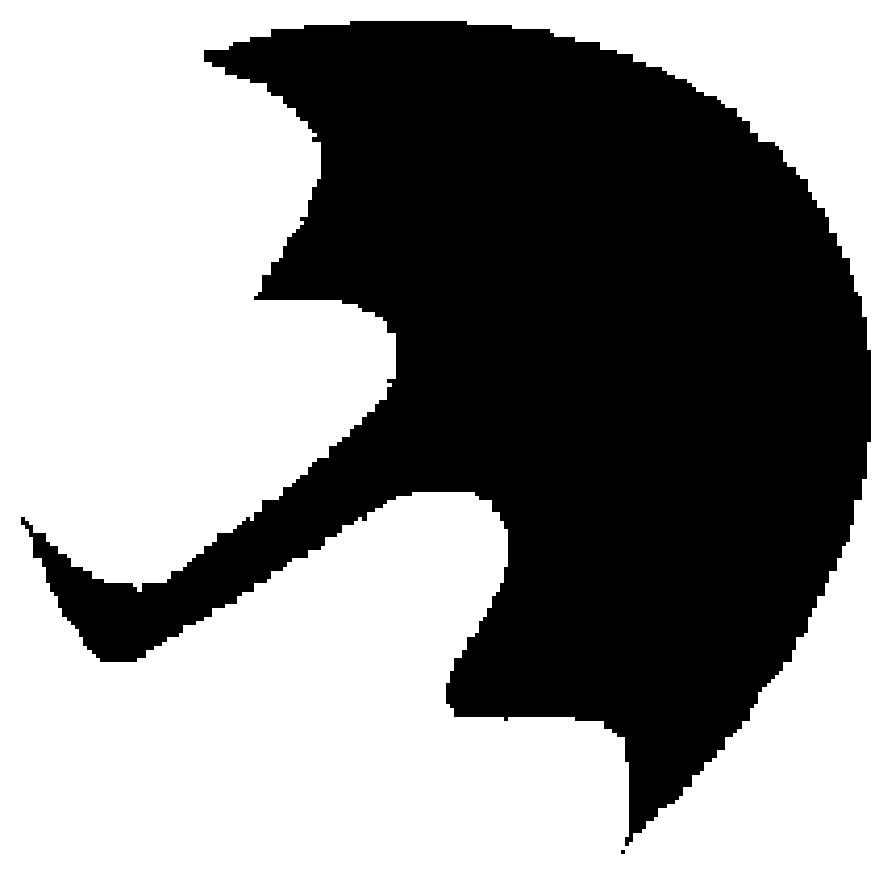,width=0.700000cm}&
\epsfig{figure=PAMI08_png/obj26_4.png,width=0.700000cm}&
\epsfig{figure=PAMI08_png/obj26_5.png,height=0.700000cm}&
\epsfig{figure=PAMI08_png/obj16_6.png,width=0.700000cm}&
\epsfig{figure=PAMI08_png/obj26_6.png,width=0.700000cm}&
\epsfig{figure=PAMI08_png/obj16_3.png,height=0.700000cm}&
\epsfig{figure=PAMI08_png/obj21_1.png,width=0.700000cm}\\
\cline{2-12}
  & \bf{0.993} & \bf{0.991} & \bf{0.859} & \bf{0.838} & \bf{0.776} & 0.573 & 0.550 & 0.546 & 0.542 & 0.518 & 0.506 \\
\hline \epsfig{figure=PAMI08_png/obj26_4.png,width=0.700000cm}&
\epsfig{figure=PAMI08_png/obj26_6.png,width=0.700000cm}&
\epsfig{figure=PAMI08_png/obj26_5.png,height=0.700000cm}&
\epsfig{figure=PAMI08_png/obj26_3.png,height=0.700000cm}&
\epsfig{figure=PAMI08_png/obj26_1.png,width=0.700000cm}&
\epsfig{figure=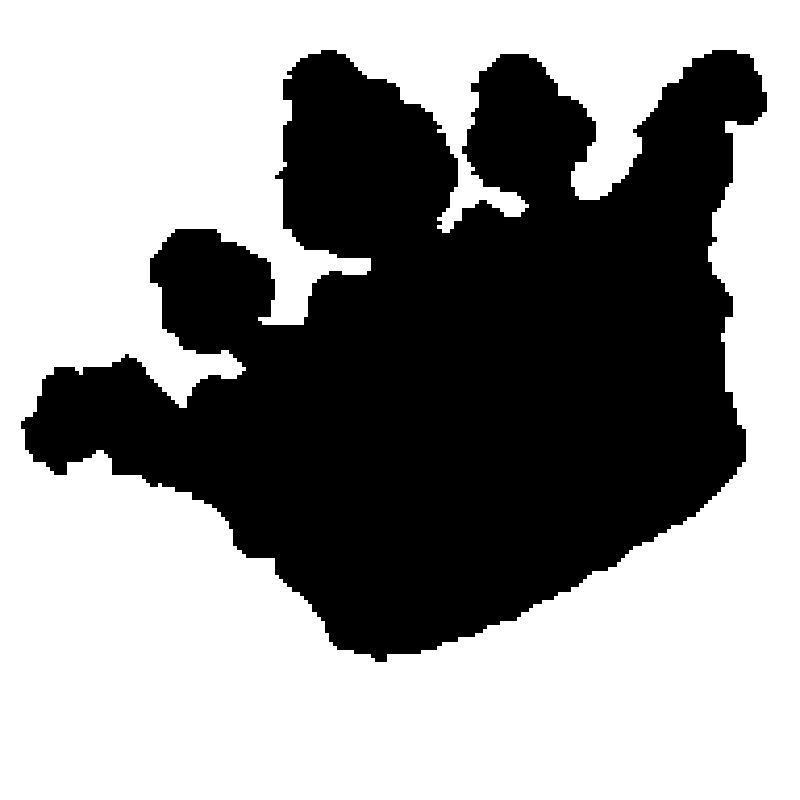,height=0.700000cm}&
\epsfig{figure=PAMI08_png/obj5_4.png,height=0.700000cm}&
\epsfig{figure=PAMI08_png/obj4_1.png,height=0.700000cm}&
\epsfig{figure=PAMI08_png/obj5_5.png,height=0.700000cm}&
\epsfig{figure=PAMI08_png/obj5_1.png,height=0.700000cm}&
\epsfig{figure=PAMI08_png/obj25_1.png,height=0.700000cm}&
\epsfig{figure=PAMI08_png/obj25_5.png,height=0.700000cm}\\
\cline{2-12}
  & \bf{0.933} & \bf{0.931} & \bf{0.786} & \bf{0.784} & \bf{0.628} & 0.623 & 0.614 & 0.582 & 0.580 & 0.574 & 0.573 \\
\hline \epsfig{figure=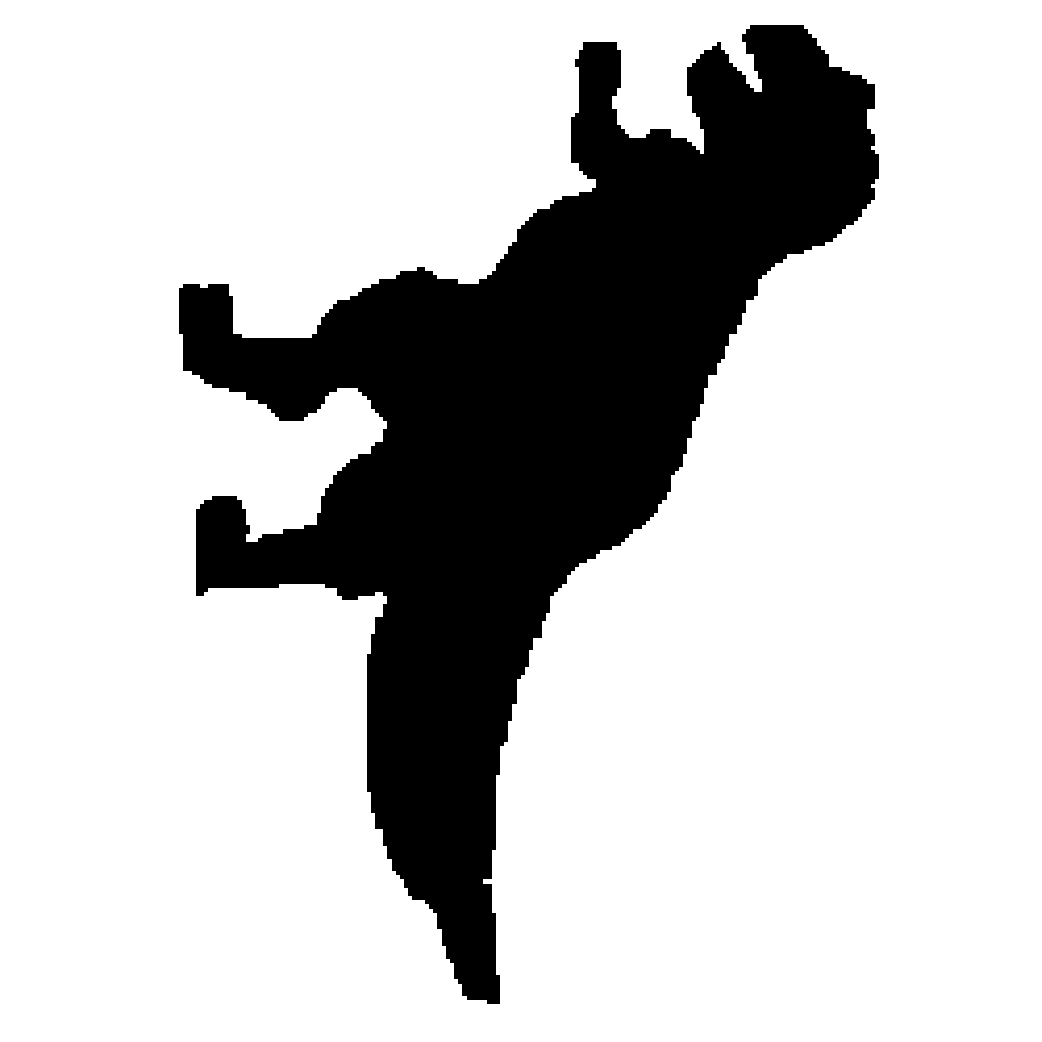,height=0.700000cm}&
\epsfig{figure=PAMI08_png/obj27_4.png,width=0.700000cm}&
\epsfig{figure=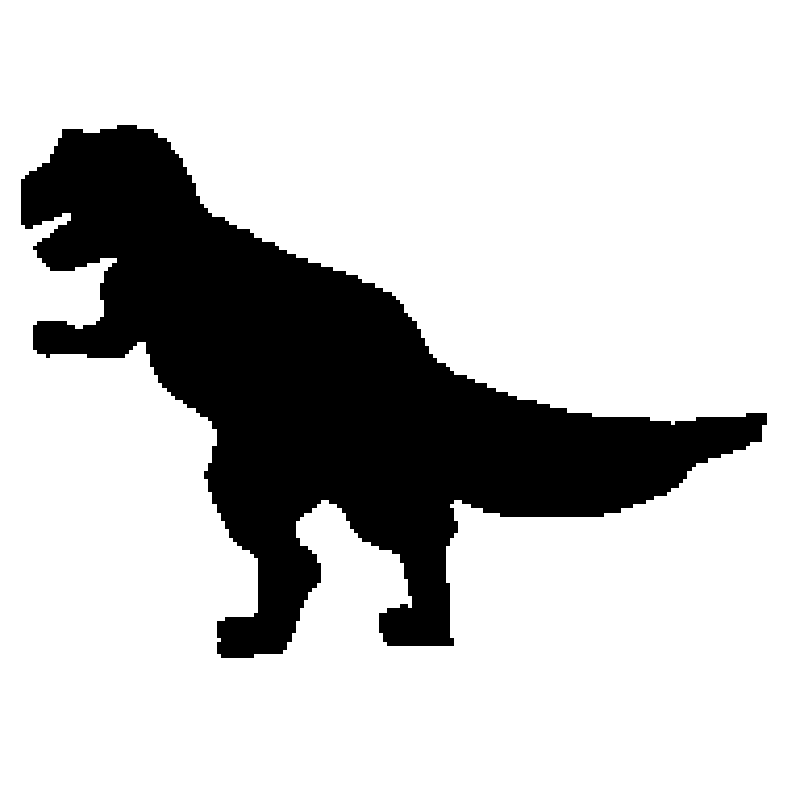,height=0.700000cm}&
\epsfig{figure=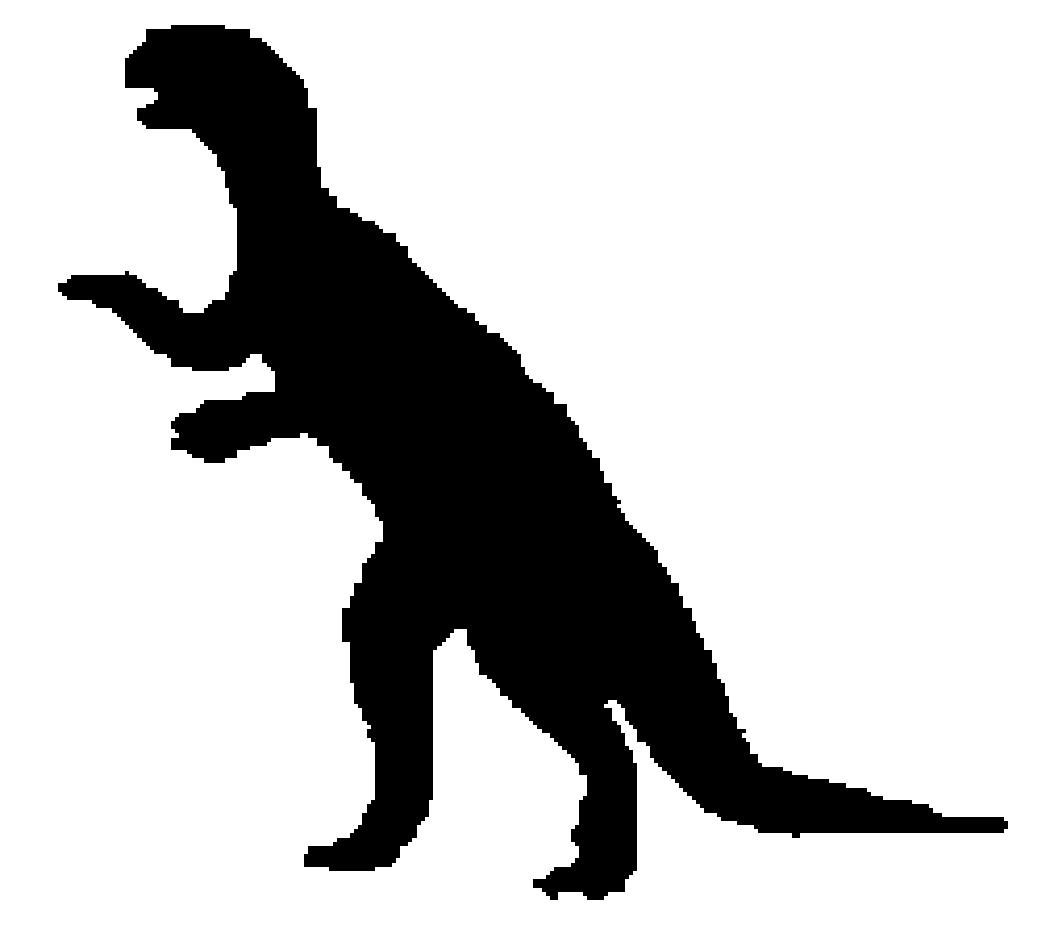,width=0.700000cm}&
\epsfig{figure=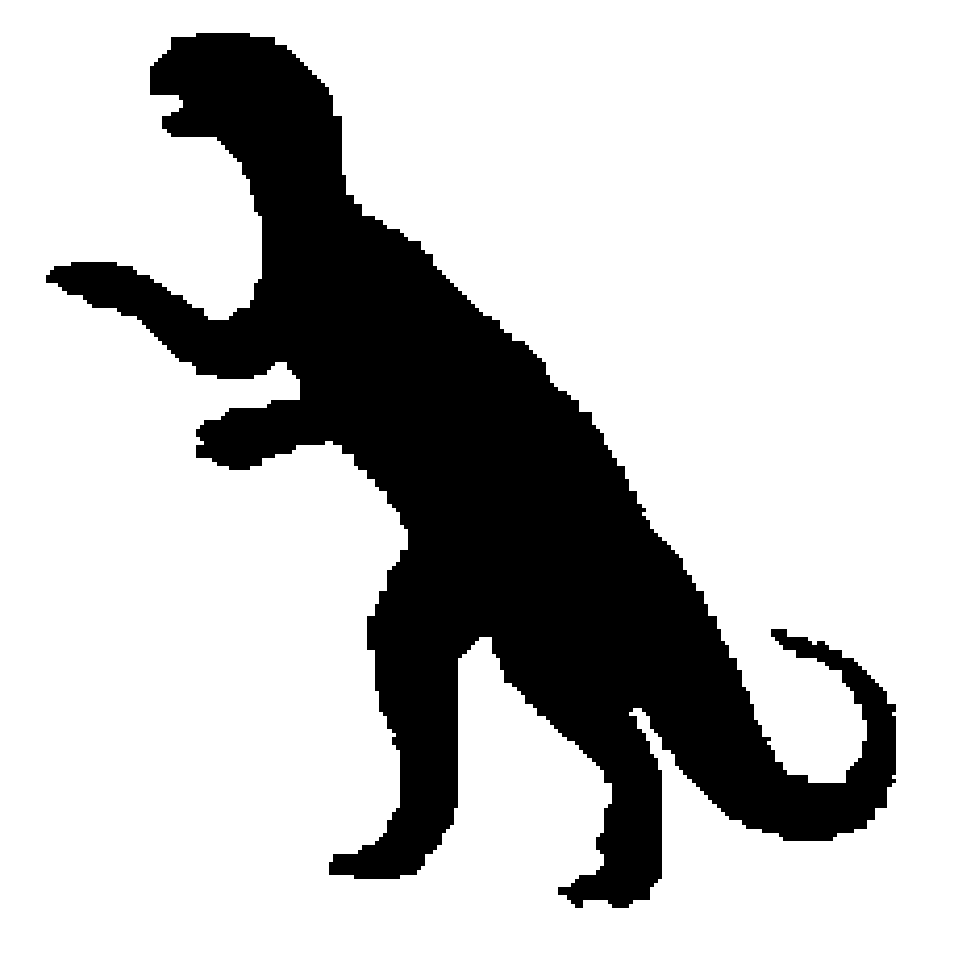,height=0.700000cm}&
\epsfig{figure=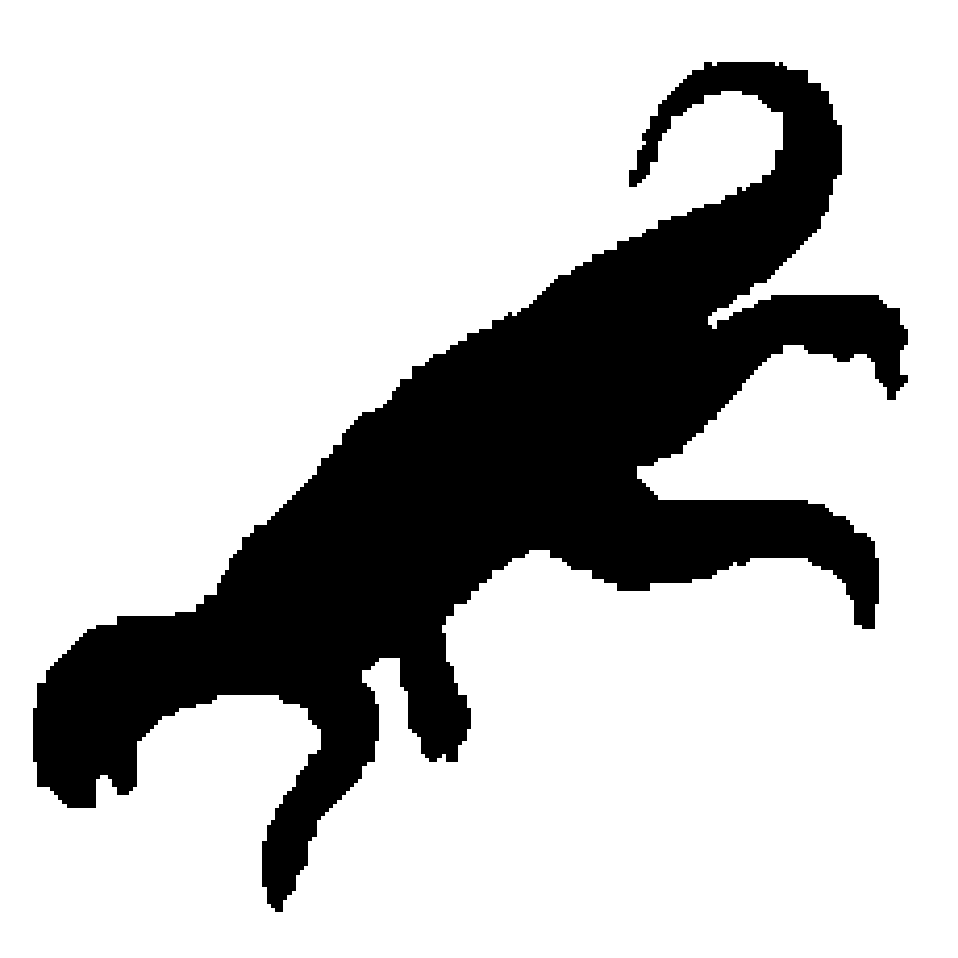,height=0.700000cm}&
\epsfig{figure=PAMI08_png/obj5_3.png,height=0.700000cm}&
\epsfig{figure=PAMI08_png/obj26_3.png,height=0.700000cm}&
\epsfig{figure=PAMI08_png/obj26_1.png,width=0.700000cm}&
\epsfig{figure=PAMI08_png/obj30_6.png,height=0.700000cm}&
\epsfig{figure=PAMI08_png/obj20_3.png,width=0.700000cm}&
\epsfig{figure=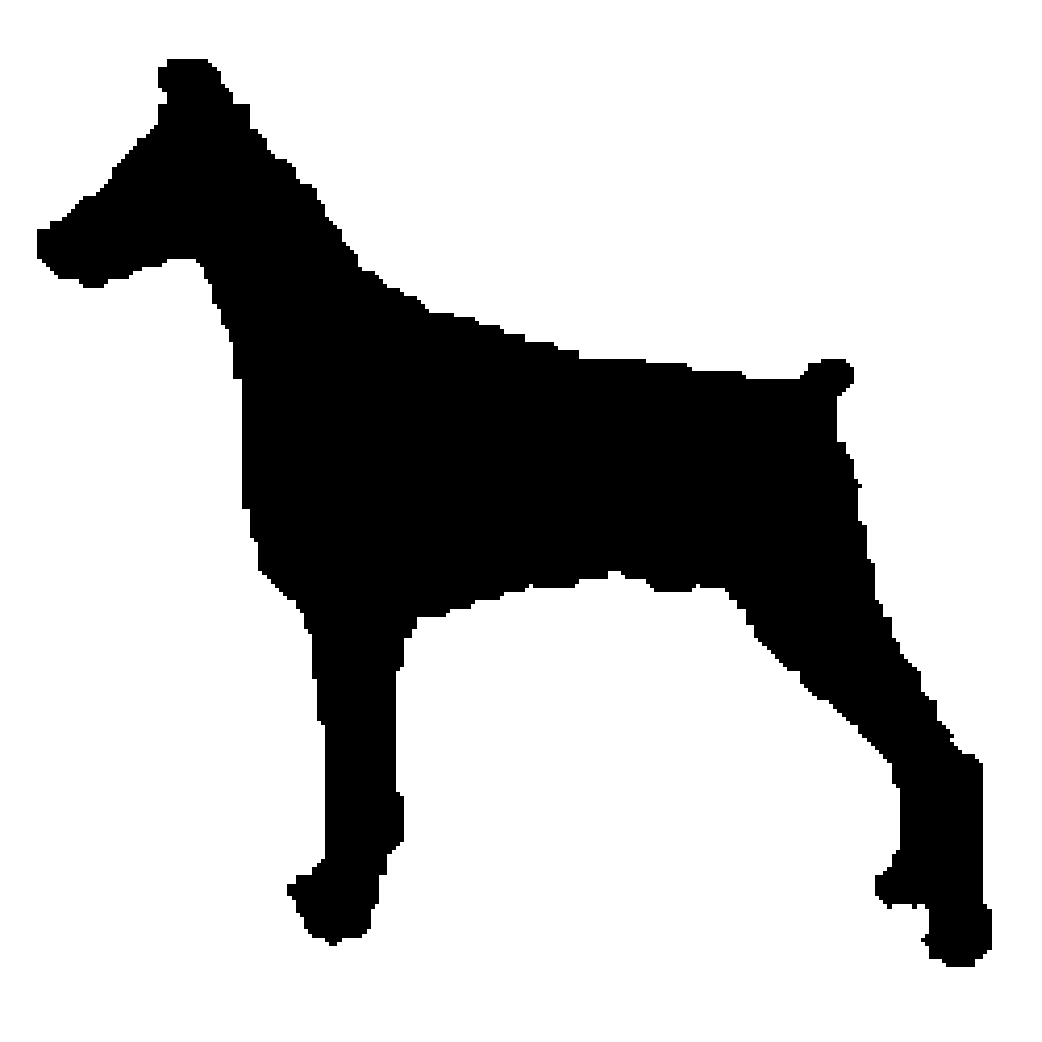,height=0.700000cm}\\
\cline{2-12}
  & \bf{0.961} & \bf{0.923} & \bf{0.842} & \bf{0.820} & \bf{0.812} & 0.617 & 0.516 & 0.502 & 0.490 & 0.483 & 0.481 \\
\hline \epsfig{figure=PAMI08_png/obj28_2.png,height=0.700000cm}&
\epsfig{figure=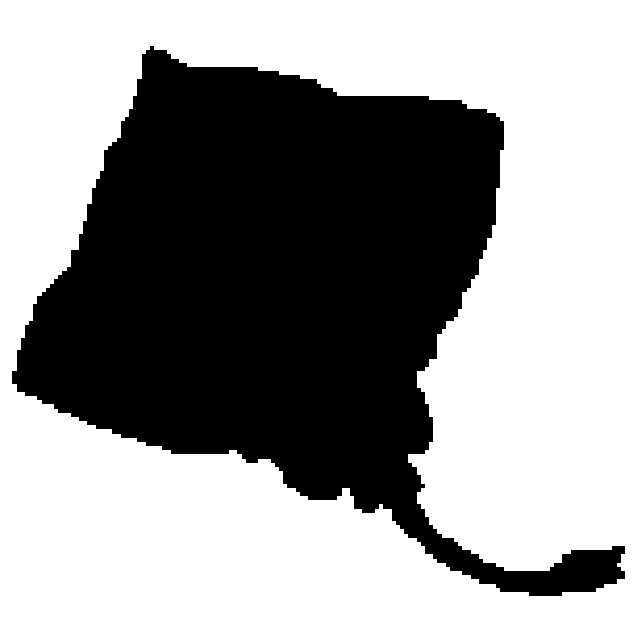,height=0.700000cm}&
\epsfig{figure=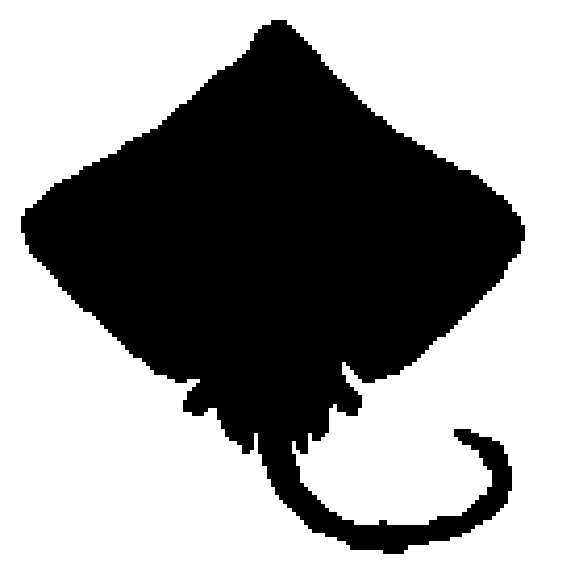,height=0.700000cm}&
\epsfig{figure=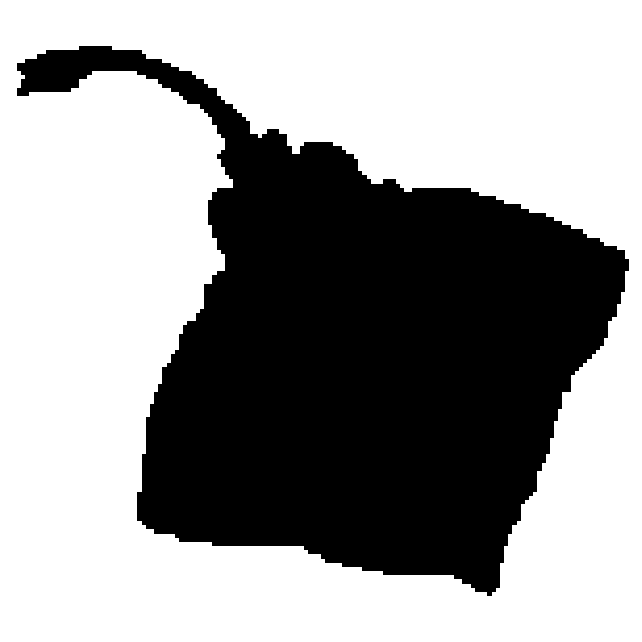,height=0.700000cm}&
\epsfig{figure=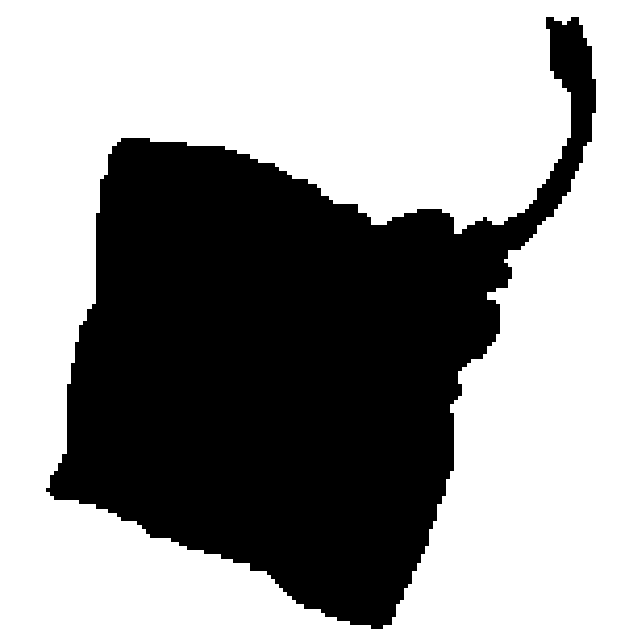,height=0.700000cm}&
\epsfig{figure=PAMI08_png/obj28_4.png,height=0.700000cm}&
\epsfig{figure=PAMI08_png/obj15_4.png,height=0.700000cm}&
\epsfig{figure=PAMI08_png/obj16_6.png,width=0.700000cm}&
\epsfig{figure=PAMI08_png/obj16_3.png,height=0.700000cm}&
\epsfig{figure=PAMI08_png/obj16_4.png,height=0.700000cm}&
\epsfig{figure=PAMI08_png/obj15_2.png,height=0.700000cm}&
\epsfig{figure=PAMI08_png/obj25_1.png,height=0.700000cm}\\
cline{2-12}
  & \bf{0.892} & \bf{0.881} & \bf{0.879} & \bf{0.869} & \bf{0.844} & 0.690 & 0.574 & 0.560 & 0.553 & 0.537 & 0.517 \\
\hline \epsfig{figure=PAMI08_png/obj29_5.png,height=0.700000cm}&
\epsfig{figure=PAMI08_png/obj29_6.png,width=0.700000cm}&
\epsfig{figure=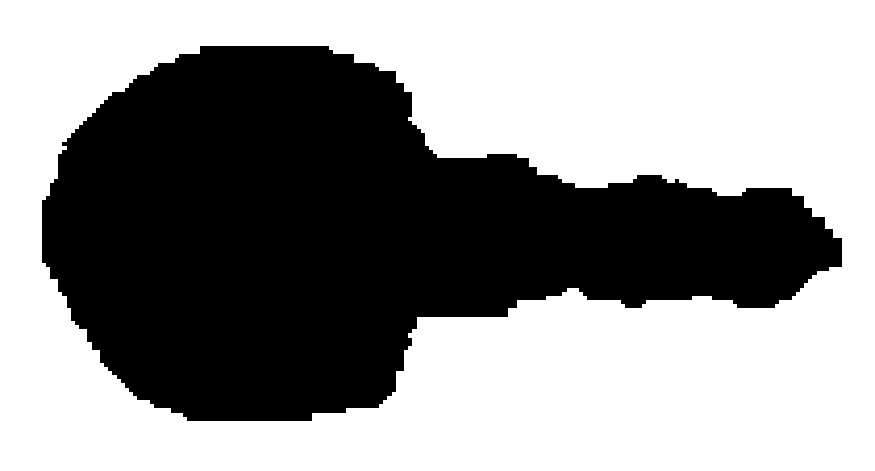,width=0.700000cm}&
\epsfig{figure=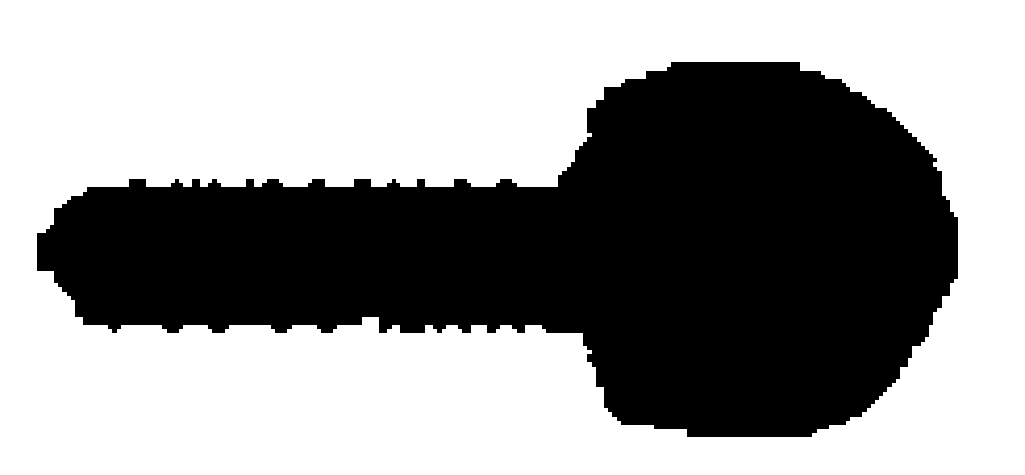,width=0.700000cm}&
\epsfig{figure=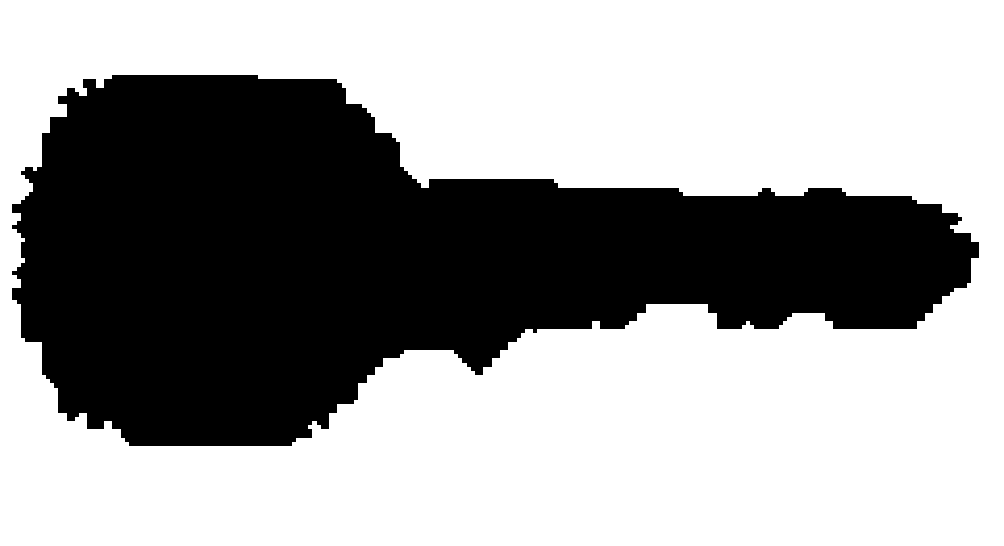,width=0.700000cm}&
\epsfig{figure=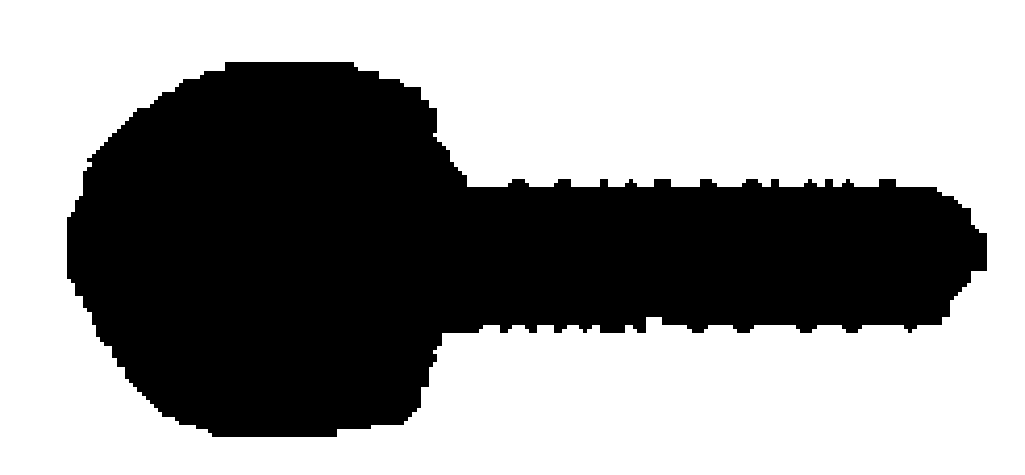,width=0.700000cm}&
\epsfig{figure=PAMI08_png/obj7_4.png,width=0.700000cm}&
\epsfig{figure=PAMI08_png/obj7_6.png,width=0.700000cm}&
\epsfig{figure=PAMI08_png/obj28_2.png,height=0.700000cm}&
\epsfig{figure=PAMI08_png/obj7_3.png,height=0.700000cm}&
\epsfig{figure=PAMI08_png/obj7_1.png,height=0.700000cm}&
\epsfig{figure=PAMI08_png/obj26_2.png,height=0.700000cm}\\
\cline{2-12}
  & \bf{0.995} & \bf{0.747} & \bf{0.686} & \bf{0.683} & \bf{0.674} & 0.477 & 0.475 & 0.468 & 0.466 & 0.465 & 0.455 \\
\hline \epsfig{figure=PAMI08_png/obj30_4.png,height=0.700000cm}&
\epsfig{figure=PAMI08_png/obj30_2.png,height=0.700000cm}&
\epsfig{figure=PAMI08_png/obj30_1.png,height=0.700000cm}&
\epsfig{figure=PAMI08_png/obj30_6.png,height=0.700000cm}&
\epsfig{figure=PAMI08_png/obj24_6.png,width=0.700000cm}&
\epsfig{figure=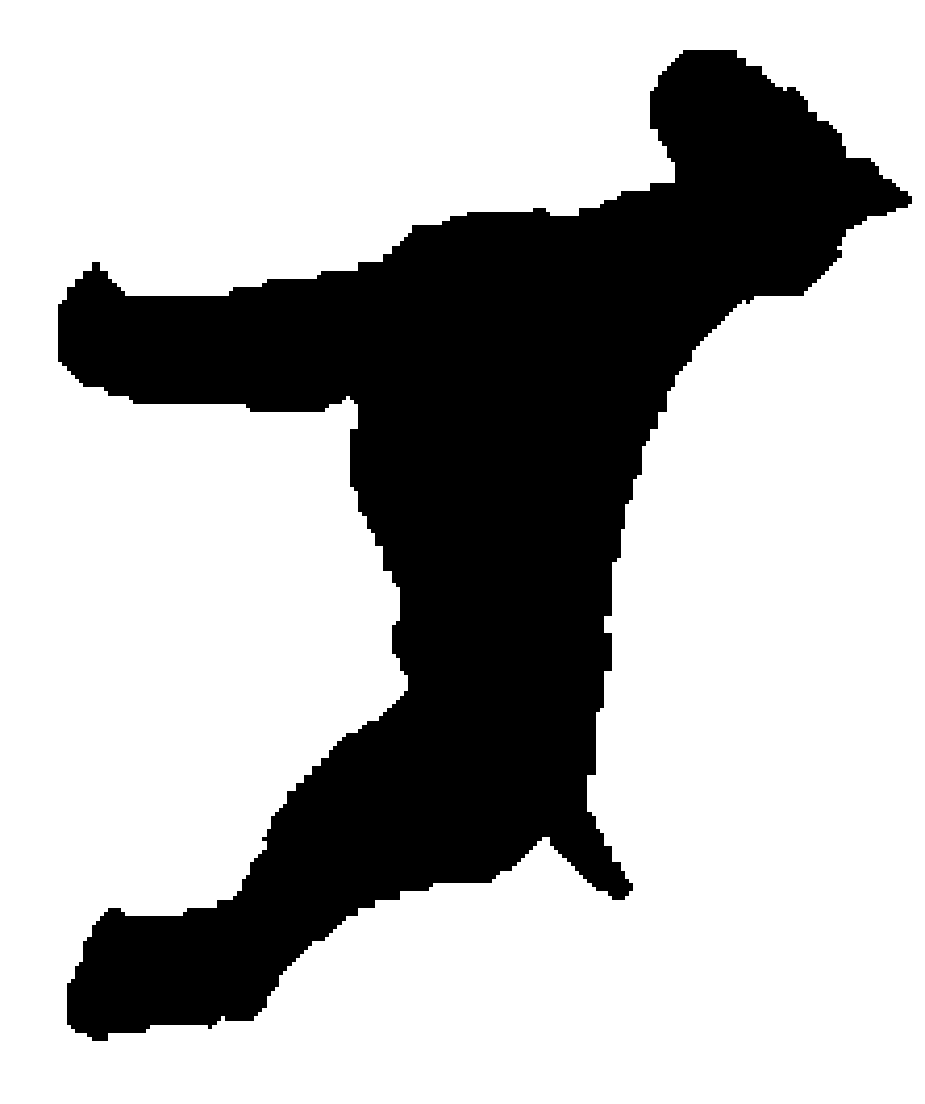,height=0.700000cm}&
\epsfig{figure=PAMI08_png/obj12_1.png,width=0.700000cm}&
\epsfig{figure=PAMI08_png/obj1_6.png,width=0.700000cm}&
\epsfig{figure=PAMI08_png/obj24_1.png,width=0.700000cm}&
\epsfig{figure=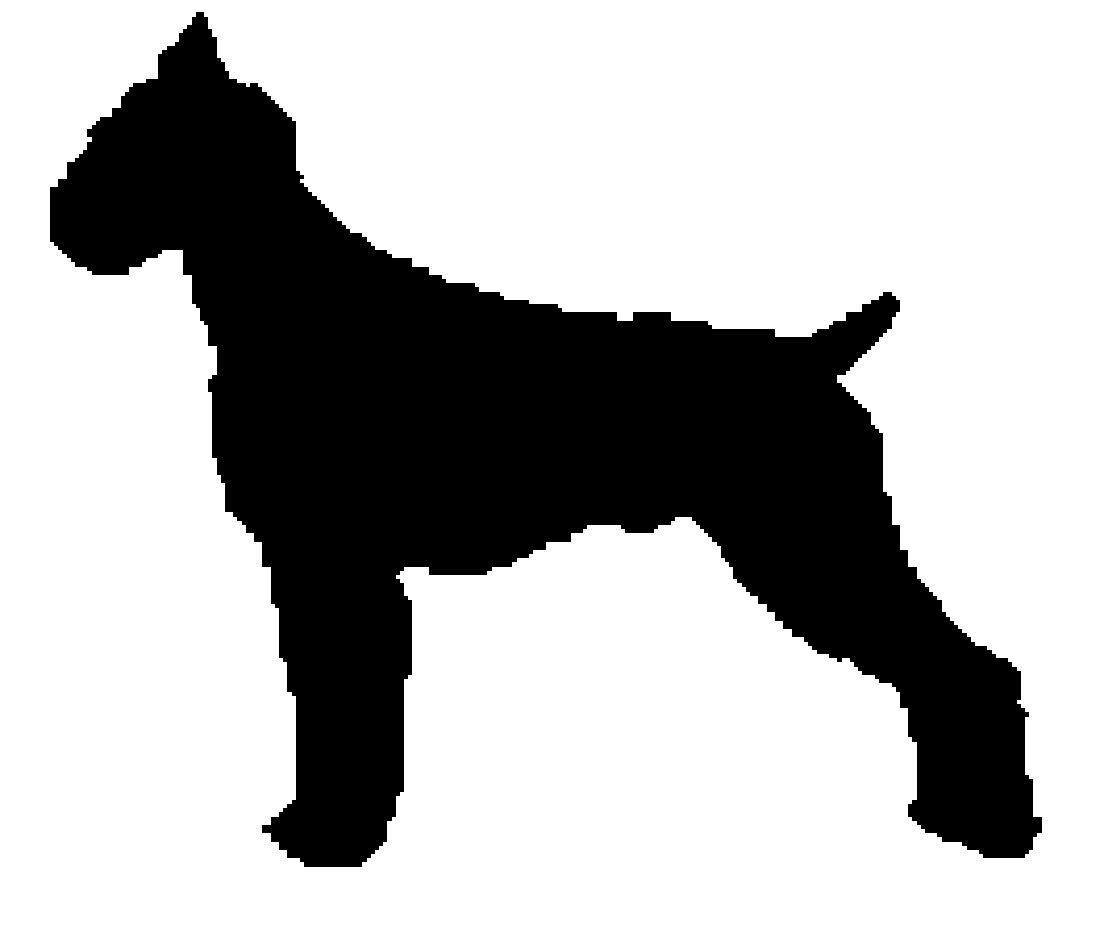,width=0.700000cm}&
\epsfig{figure=PAMI08_png/obj12_6.png,width=0.700000cm}&
\epsfig{figure=PAMI08_png/obj27_4.png,width=0.700000cm}\\
\cline{2-12}
  & \bf{0.994} & \bf{0.679} & \bf{0.675} & 0.512 & \bf{0.488} & 0.484 & 0.476 & 0.473 & \bf{0.469} & 0.452 & 0.440 \\
\hline
\end{tabular}
\label{table:results2}
\end{table}
\clearpage

\section{Semi-Local Coordinate Frame}
\label{sec:slcf}The disconnected skeleton representation, as
described up to this point, is  articulation insensitive. However,
{a certain context} may require articulation sensitivity  as
previously  illustrated in Fig.~\ref{fig:goal}. {In this section, we
define a part-centered coordinate frame which we call the semi-local
frame which can be used
 to introduce sensitivity to articulations.  This frame, together with the global Euclidean frame
 (Section~\ref{sec:smas}) and
deformations  discussed below (Section~\ref{sec:morph}), provides a
means of  both individual and joint handling of visual
transformations corresponding to rigid, piecewise rigid and
infinitesimal.}

\begin{figure}[!h]
\centering
\includegraphics[height=4cm]{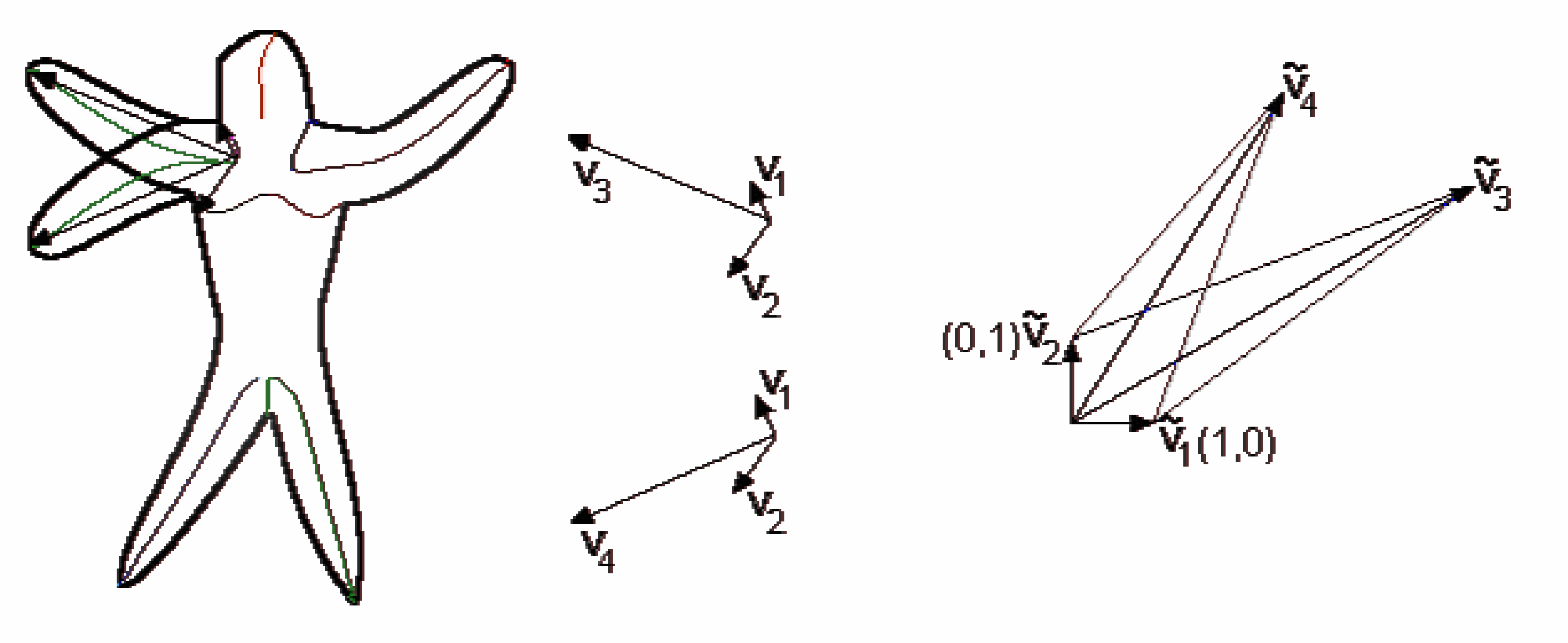}
\caption{Articulation of a section can be described by a single
point in the semi-local coordinate frame.} \label{fig:quadrep}
\end{figure}

\begin{figure}[!h]
\begin{center}
\begin{tabular}{ccc}
\includegraphics[height=2cm]{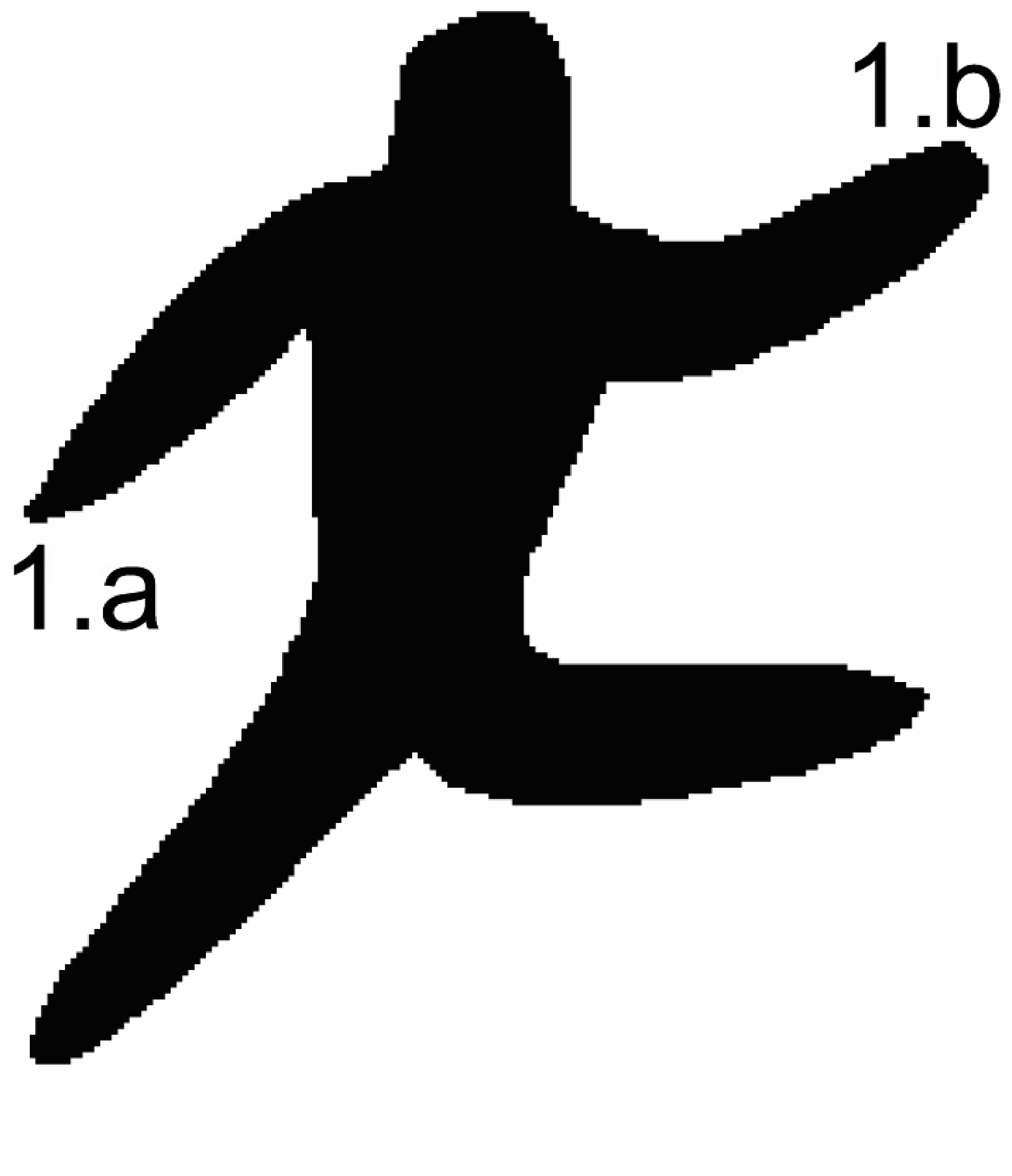}
\includegraphics[height=2cm]{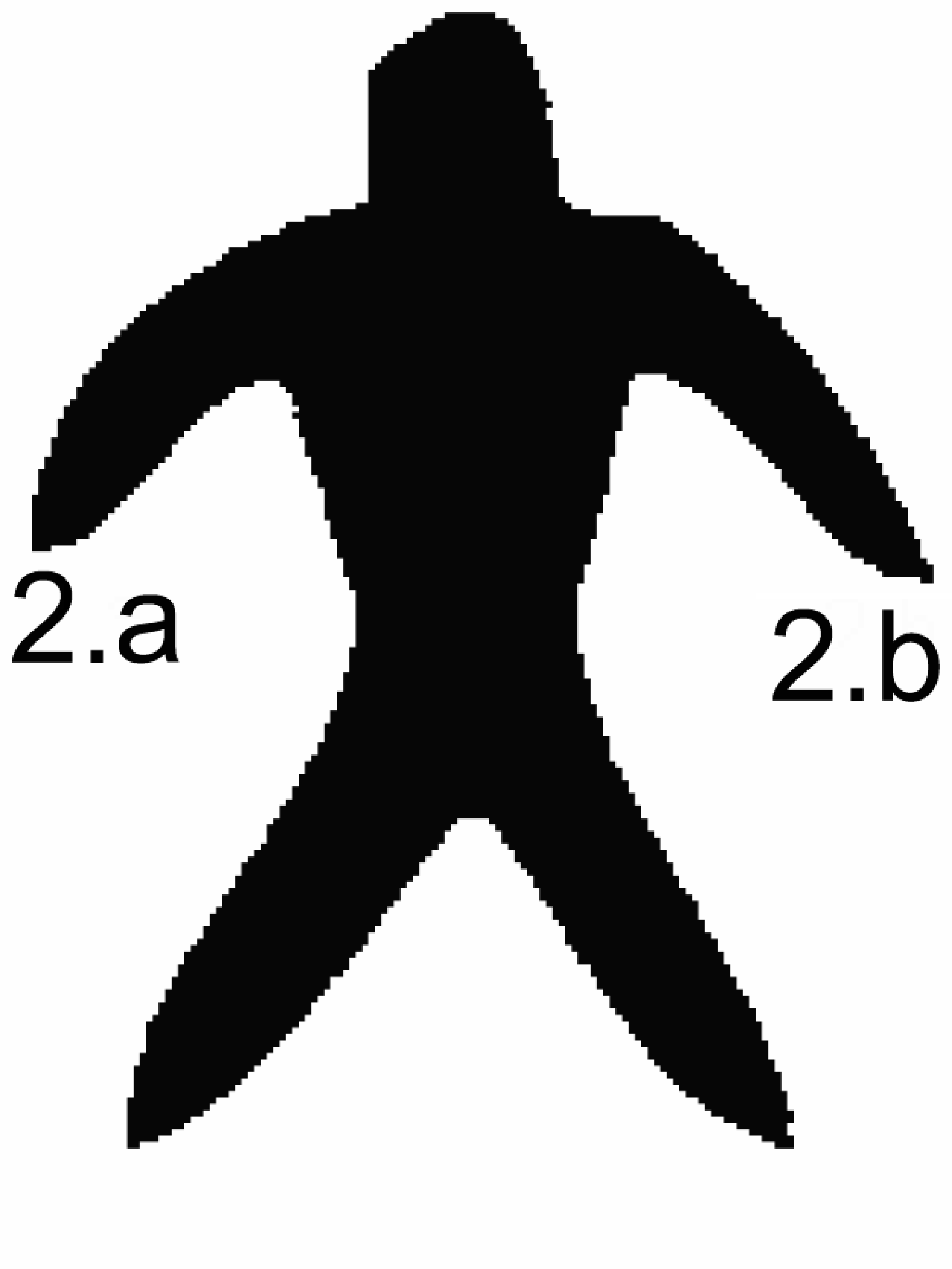} &
\includegraphics[height=3.5cm]{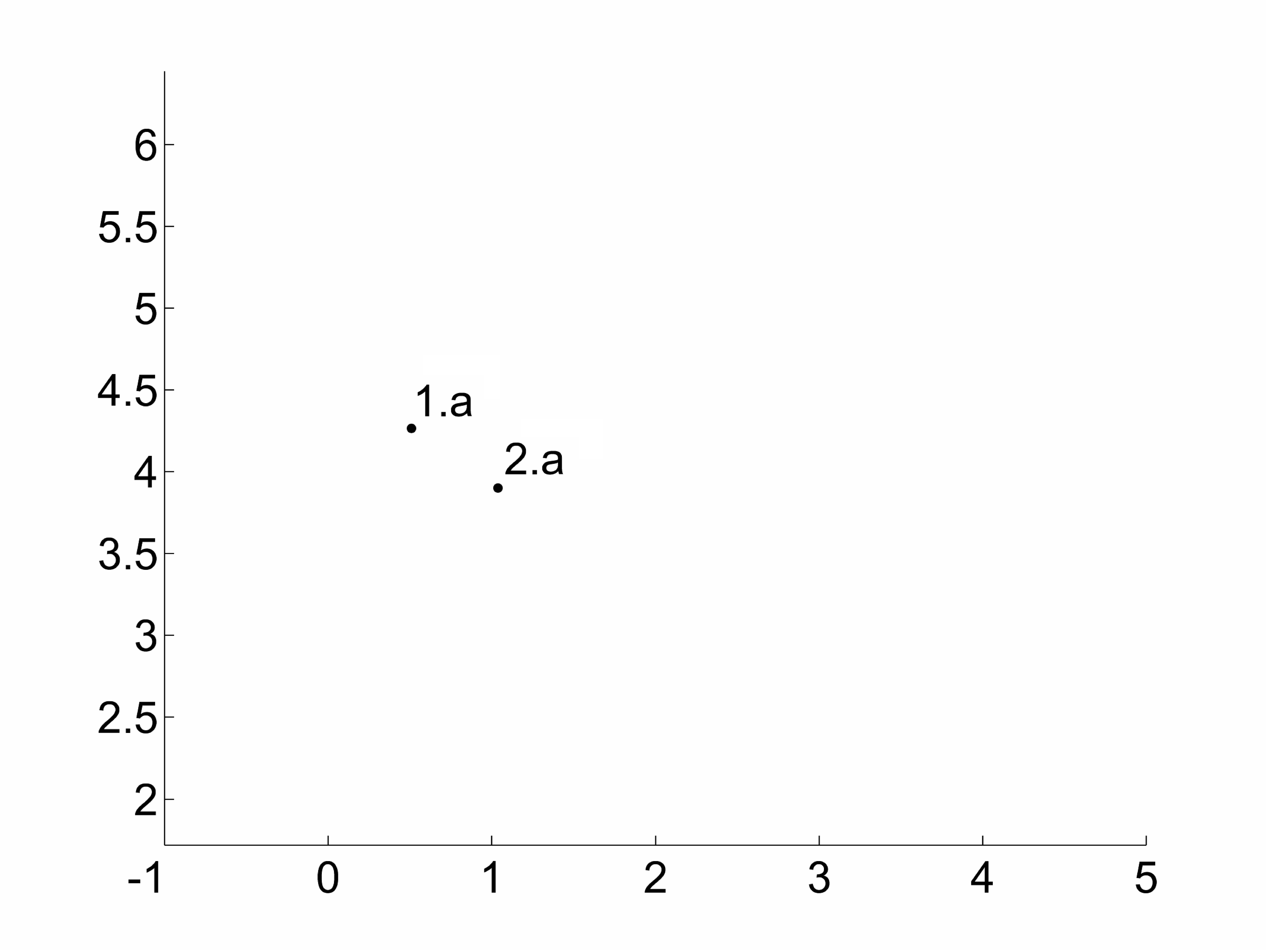} &
\includegraphics[height=3.5cm]{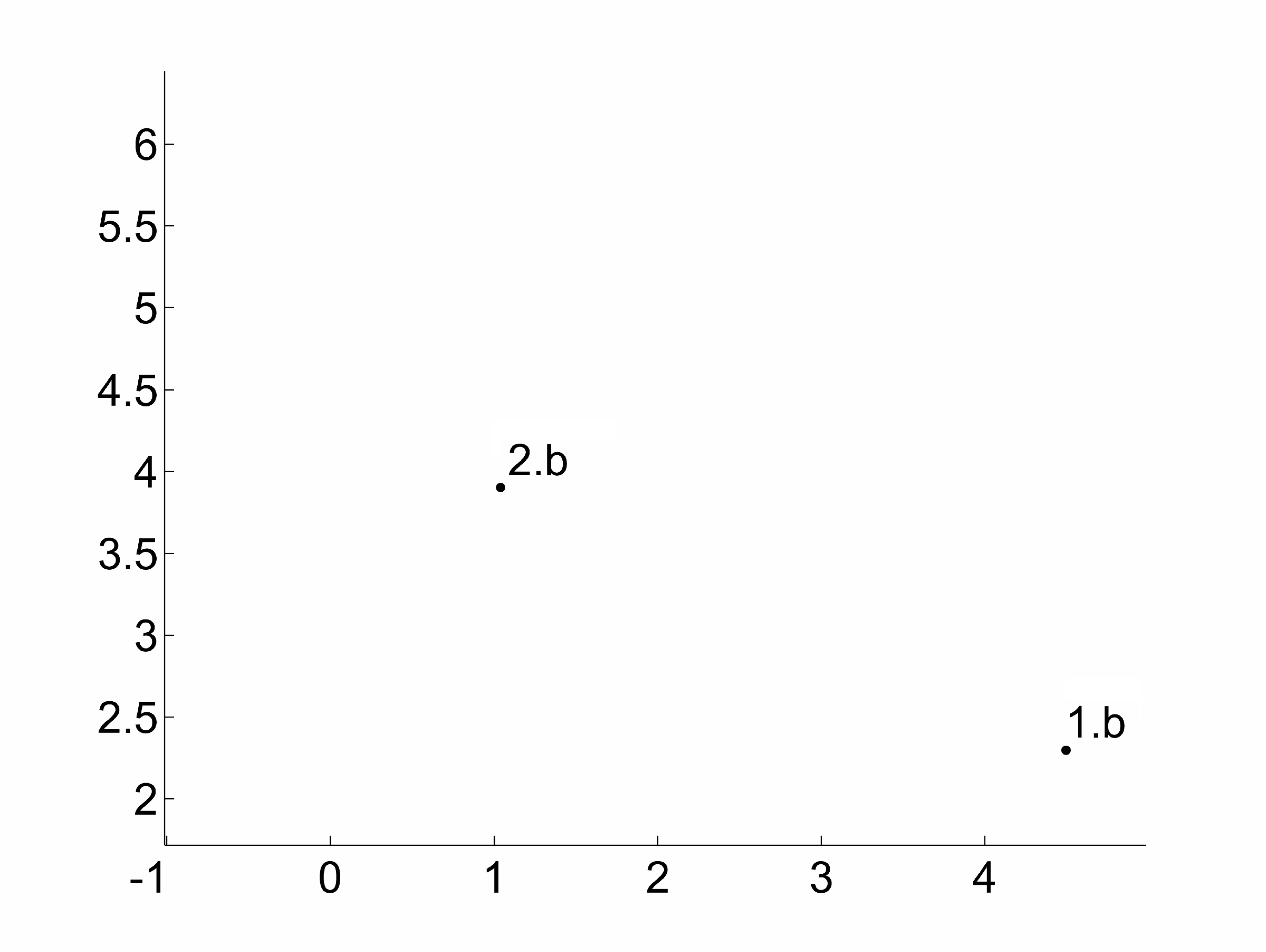} \\
(a) & (b) & (c)
\end{tabular}
\caption{The measurements in the semi-local coordinate frame can be
used to make an inference related to pose. Notice that while the
coordinates of similar postures are close (b), the coordinates of
different postures are far away (c). As shown in (a), 1.a, 2.a, 2.b
are all arm down positions whereas 1.b is an arm up position.}
\label{fig:arms}
\end{center}
\end{figure}

Recall that in the disconnected skeleton, a positive branch is
neighbored by two negative branches. Typically, the start points of
the negative branches as well as all the disconnection locations,
are quite stable under bendings and articulations. It is the tip of
the positive branch that moves freely if the branch is denoting a
deformable section. Consider the disconnected branches of a human
shape as displayed in Fig.~\ref{fig:quadrep}. Four points define
three vectors, starting from the disconnection point of the
protrusion branch and ending respectively at the starts of the two
indentations and the protrusion. The third vector can be represented
as a linear combination of the remaining two. When these vectors are
transformed to standard bases, each configuration can be represented
by only a single point which denotes the {\sl local pose} of a shape
section, which may or may not articulate or bend. This coordinate
measurement is deprived of any shape information as well as
Euclidean transformations, as illustrated in Fig.~\ref{fig:arms}.

{We compute the similarity between two deformable shape sections
denoted by the vectors $\vec{x}$ and $\vec{y}$, by using
\begin{equation}
sim(\vec{x},\vec{y})=\frac{1}{1+d^2(\vec{x},\vec{y})/\epsilon^2}
\label{eq:posesim}
\end{equation}
\noindent where $\epsilon$ is a soft threshold and $d$ is a distance
measure defined in the semi-local coordinate frame. The similar
configurations have high similarity scores.}
Recall that (Section \ref{sec:smas}) the total similarity of two
shapes is determined by the weighted sum of matched branch pairs
where the weights are the normalized lengths of the branches. This
matching score may be reevaluated by integrating the measurements in
the semi-local coordinate frame. For a deformable section, the tip
of the protrusion branch moves freely whereas the indentations are
more or less fixed. Therefore, we find it sufficient to revise the
weights of the positive symmetry branches.

\begin{figure}[!h]
\centering
\includegraphics[height=3.25cm]{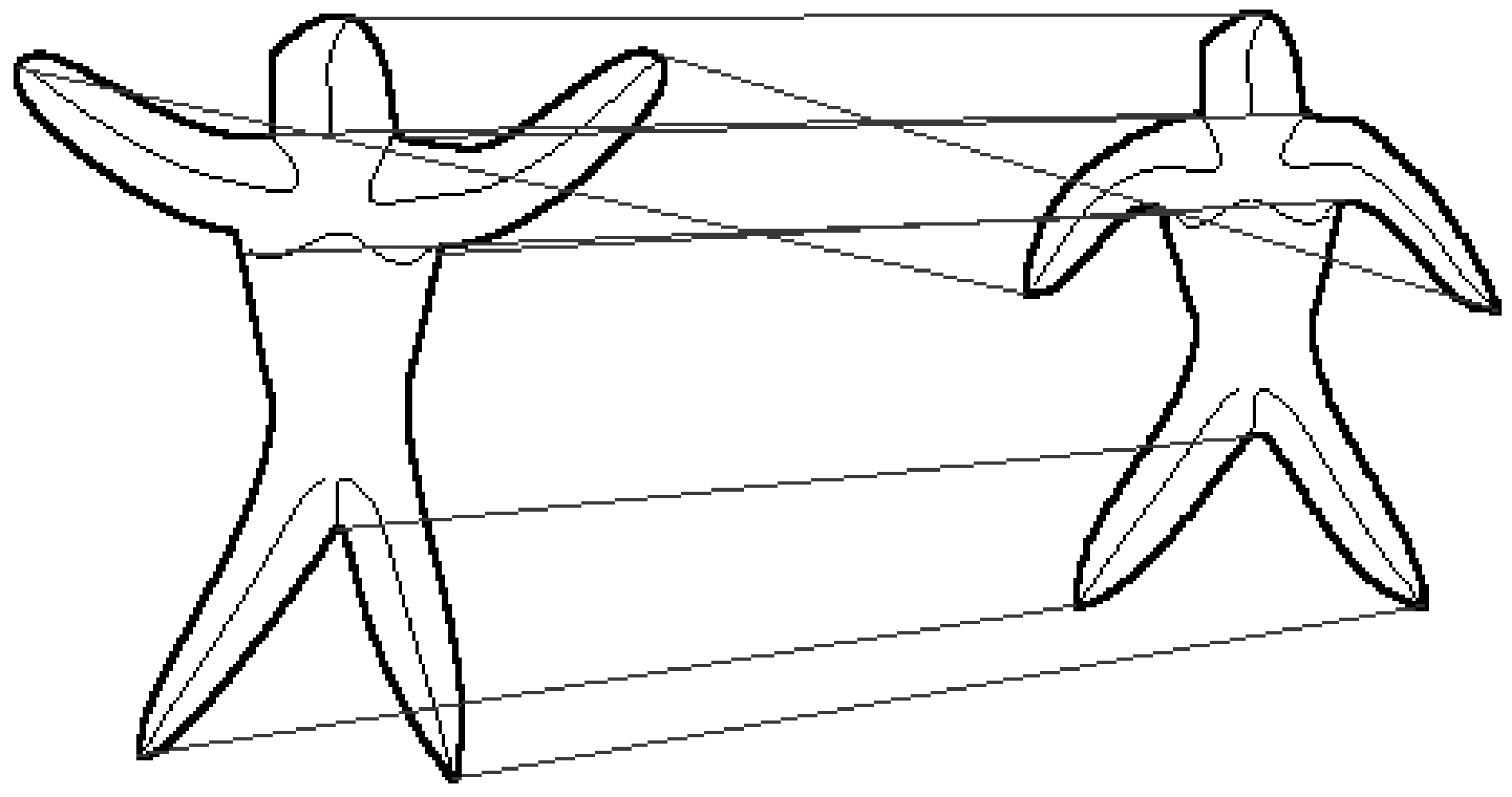}
\caption{Articulation sensitivity can be achieved in the matching
process by using the semi-local coordinate frame. For example, when
the Euclidean differences in the coordinates of matched parts are
incorporated to the skeletal matching with $\epsilon=4$, the
matching score between two human shapes is decreased from $0.9621$
to $0.7191$.} \label{fig:matching}
\end{figure}

The effect of integrating pose similarity into the similarity
measure is demonstrated in Fig. \ref{fig:matching}.
 When the
Euclidean distance is used in (\ref{eq:posesim}), the shape matching
becomes articulation sensitive. As demonstrated in a  companion
work~\cite{Erdem06}, one can consider the measurements in the
semi-local frame as  statistical data and replace the Euclidian
distance in (\ref{eq:posesim}) with the Mahalanobis distance in
order to introduce context-dependent sensitivity to articulations.

\section{{Deformations}}
\label{sec:morph}
 Deforming a given shape to a target shape has been a topic of interest in
 shape analysis, \cite{Mio06,Chui03,Peter06,Shah06,Sebastian04}.
 In a skeleton based representation, one can obtain new shapes simply by modifying a skeletal
representation. Note that even though small changes in the shape
boundary may lead to significant changes in the shape skeleton, the
opposite is not true~\cite{Leonard06}.

Suppose we edit our skeletal representation by changing attributes
or deleting/inserting a primitive. Since we have lost information as
a result of excessive regularization, we can not construct the new
shape boundary simply by propagating the edited skeleton, as
in~\cite{Blum73,Sebastian04}. Therefore, we suggest an alternative
solution by considering the transformations or diffeomorphisms,
\cite{Camion01,Bookstein91,Shah06} that  transform one disconnection
point set into another. Note that when such a transformation is
found, it can be used to form a dense correspondence between two
shape domains. The new shape boundary after the skeletal edit
operations can be formed by applying the computed transformation to
the original boundary points.

\begin{figure}[!h]
\centering
\begin{tabular}{cccccc}
\includegraphics[height=1.125cm]{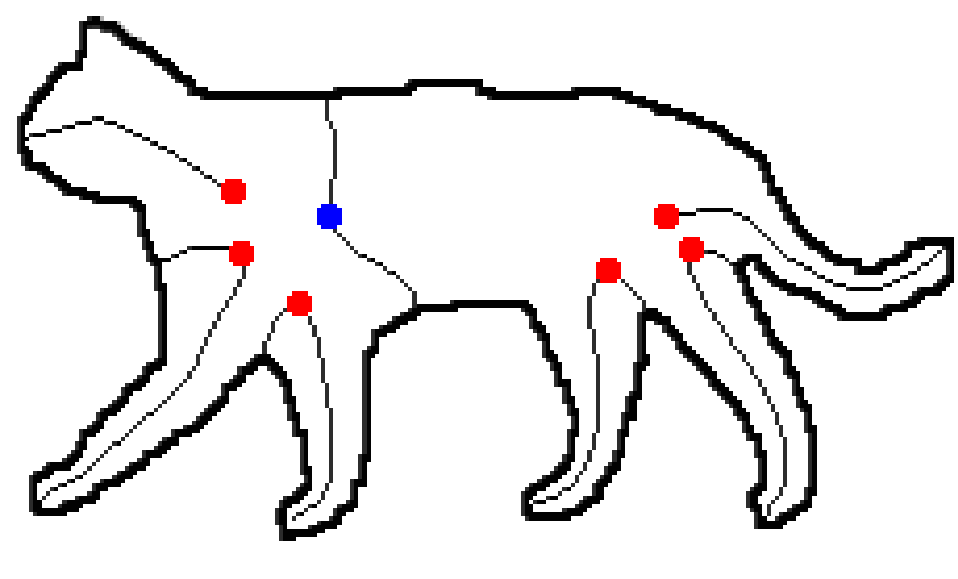} &
\includegraphics[height=1.3125cm]{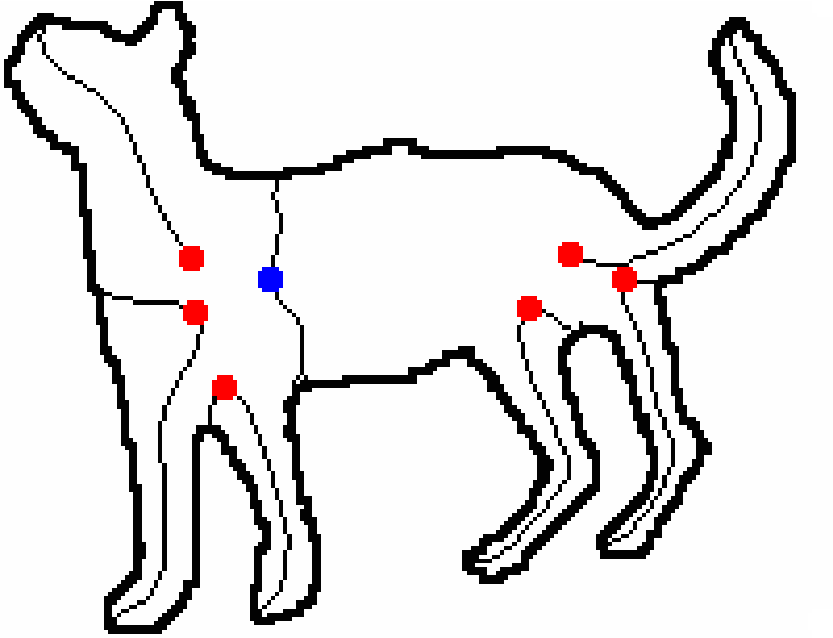} &
\includegraphics[height=1.5cm]{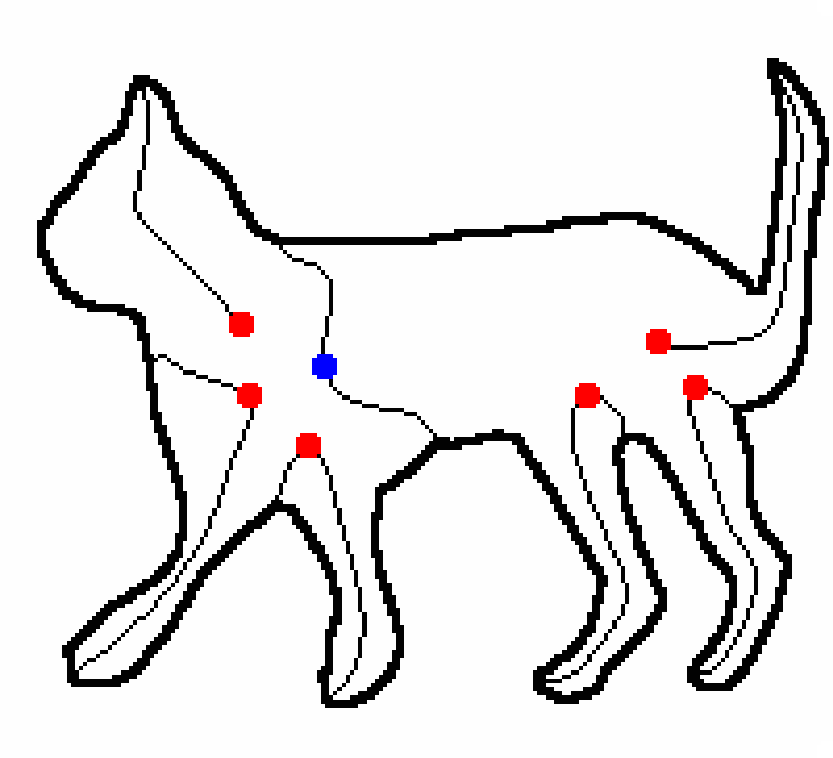} &
\includegraphics[height=1.5cm]{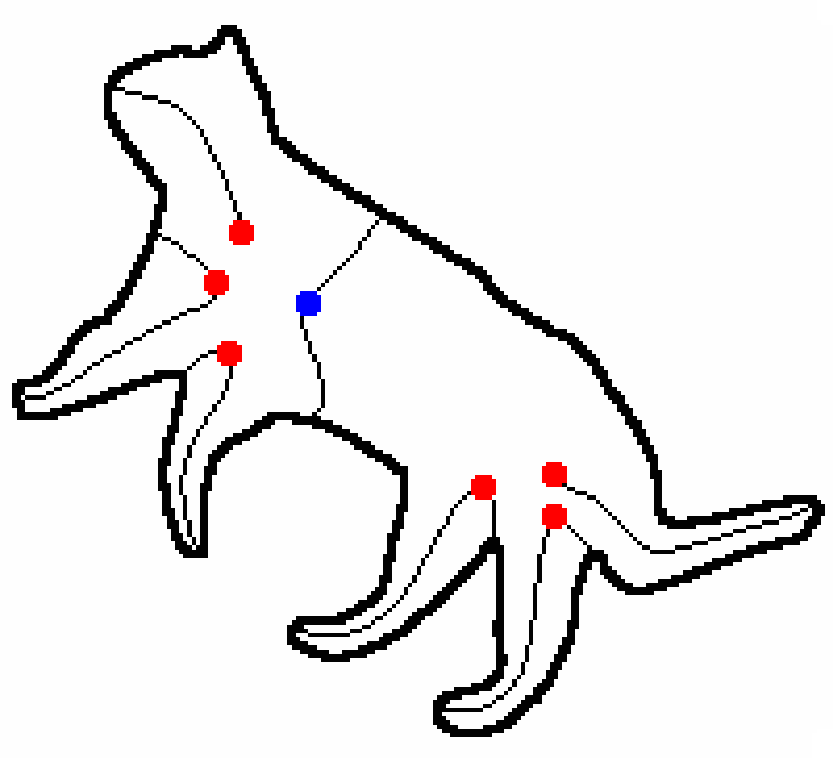} &
\includegraphics[height=1.5cm]{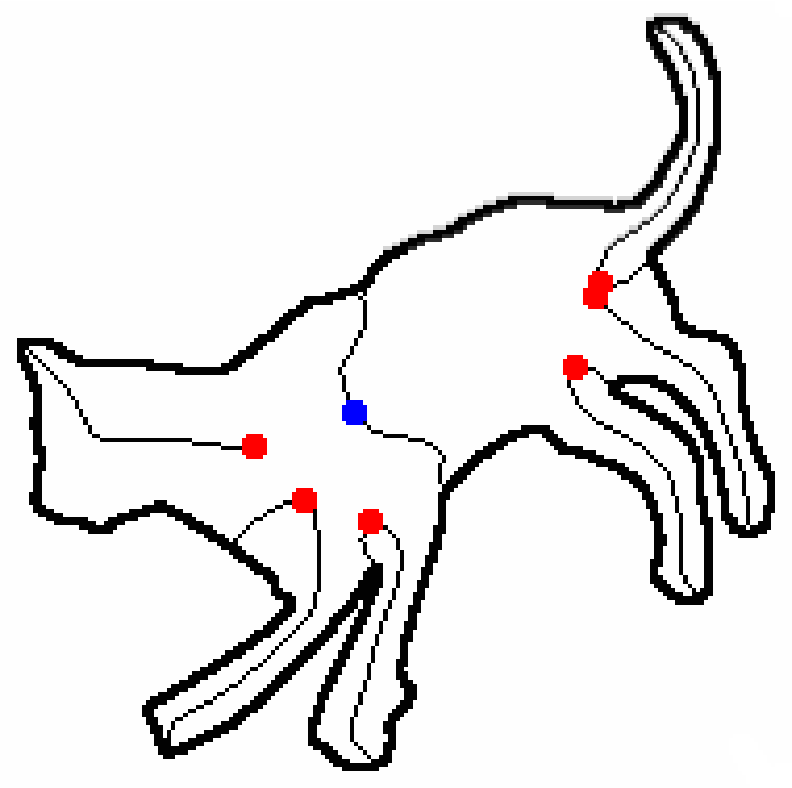} &
\includegraphics[height=1.5cm]{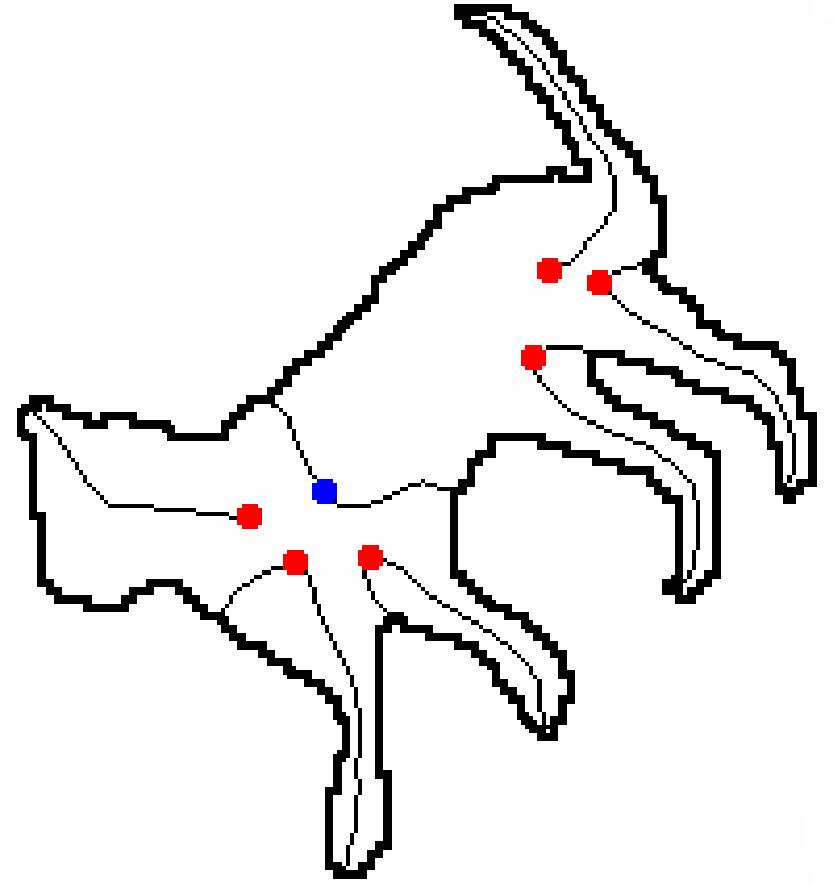} \\
\multicolumn{6}{c}{(a)} \\
\includegraphics[height=1.5cm]{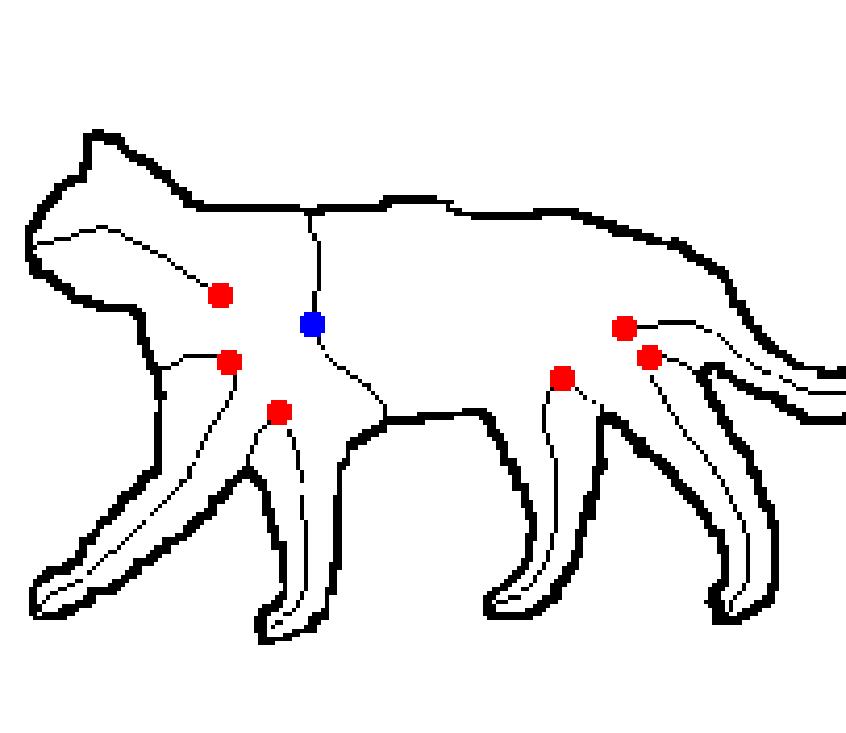} &
\includegraphics[height=1.5cm]{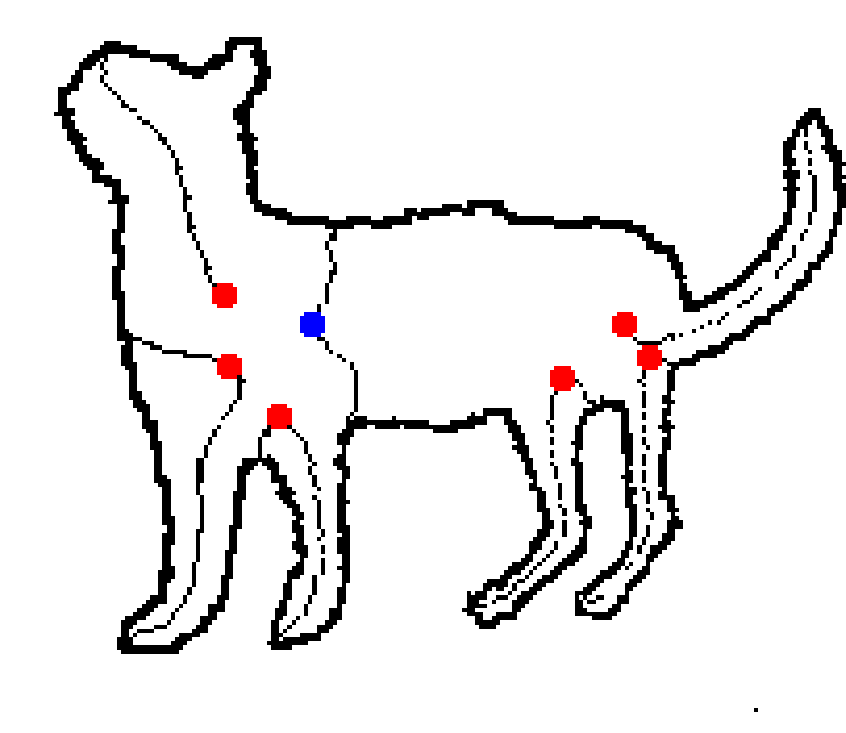} &
\includegraphics[height=1.5cm]{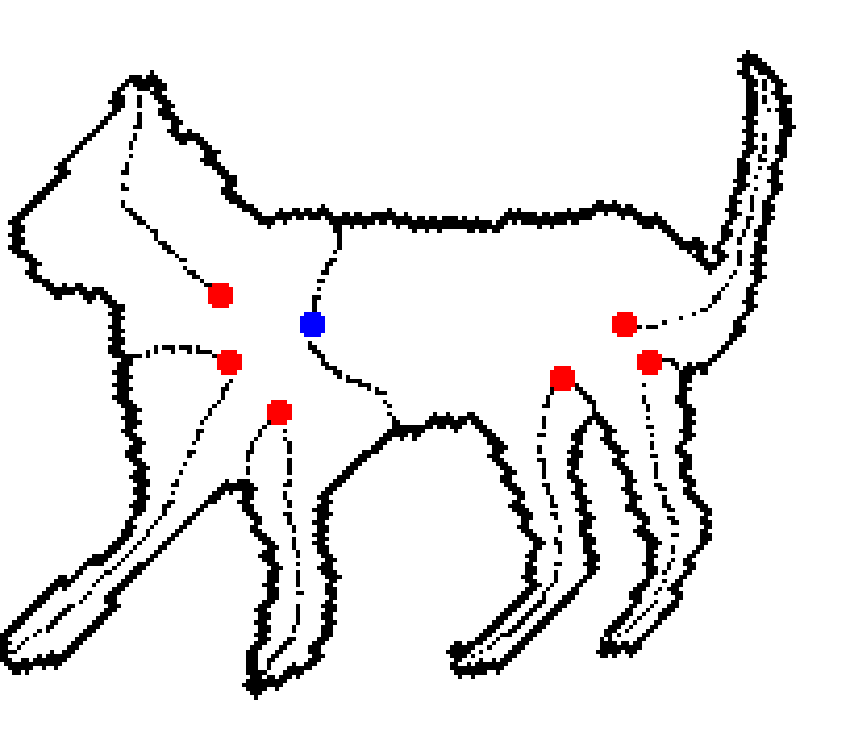} &
\includegraphics[height=1.5cm]{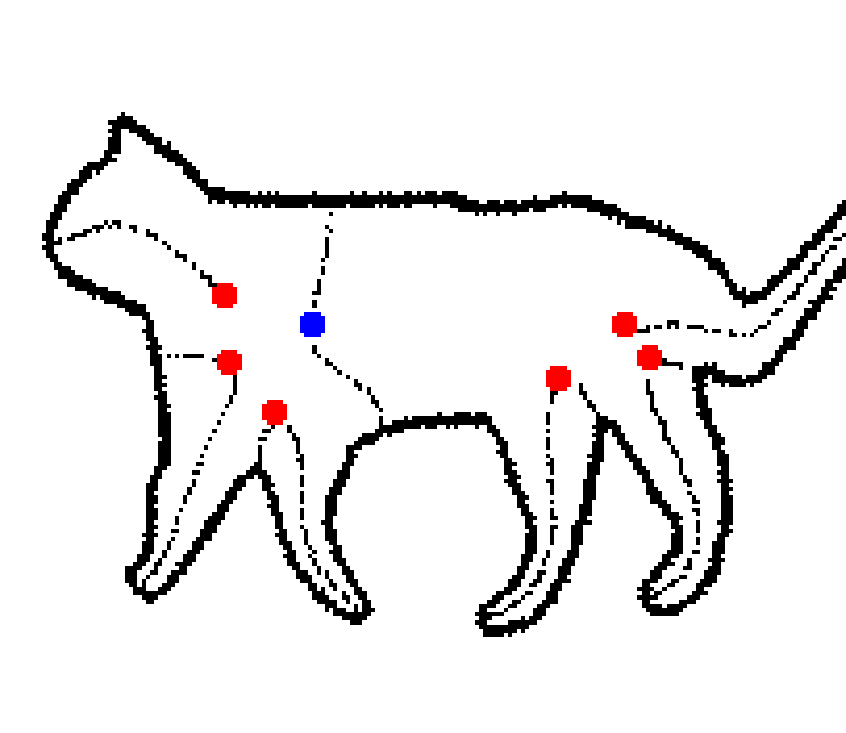} &
\includegraphics[height=1.5cm]{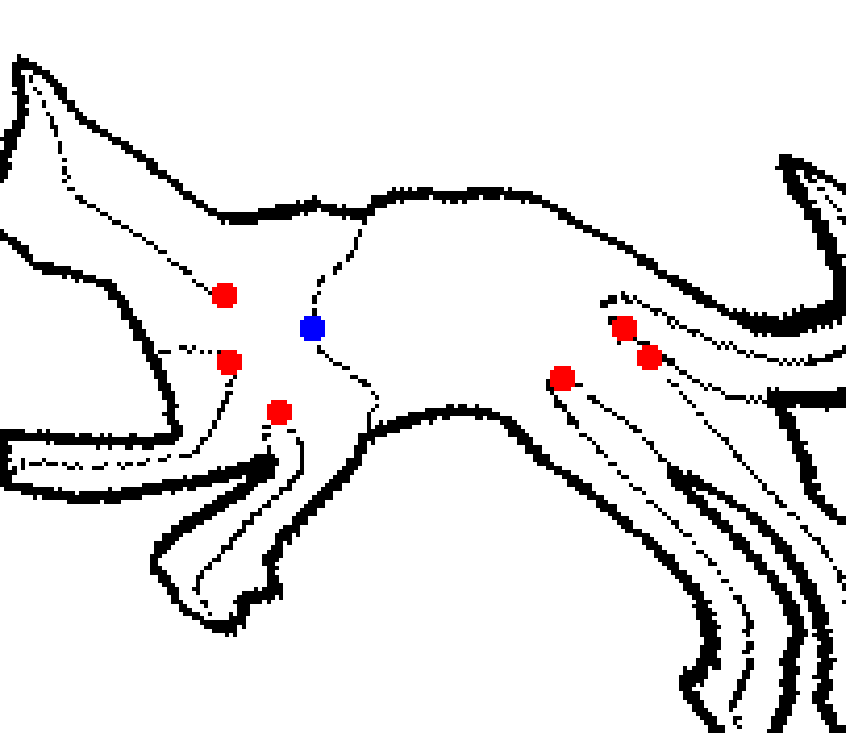} &
\includegraphics[height=1.5cm]{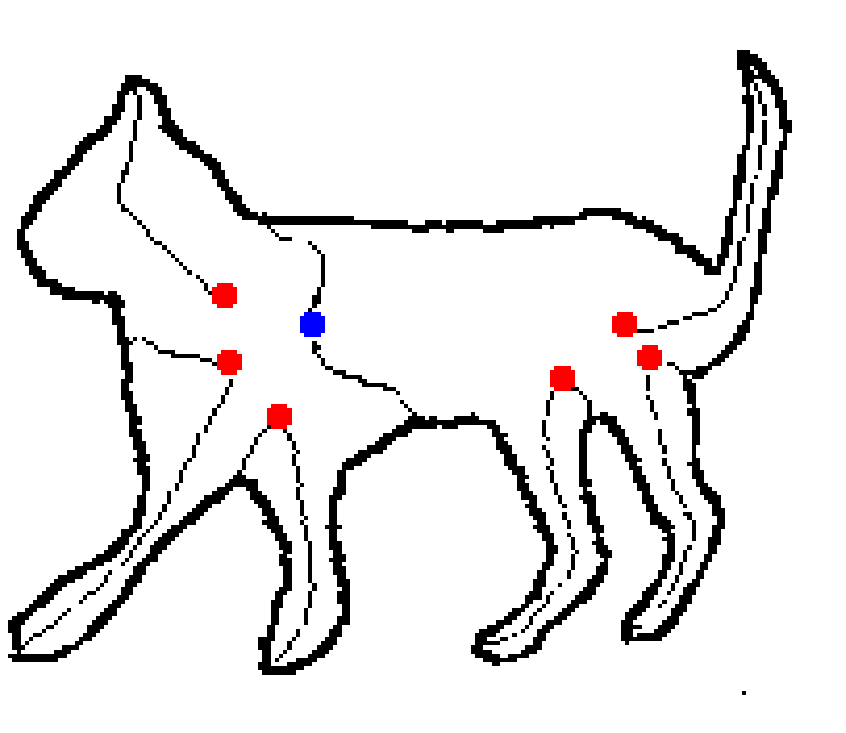}\\
\multicolumn{6}{c}{(b)} \\
\includegraphics[height=1.5cm]{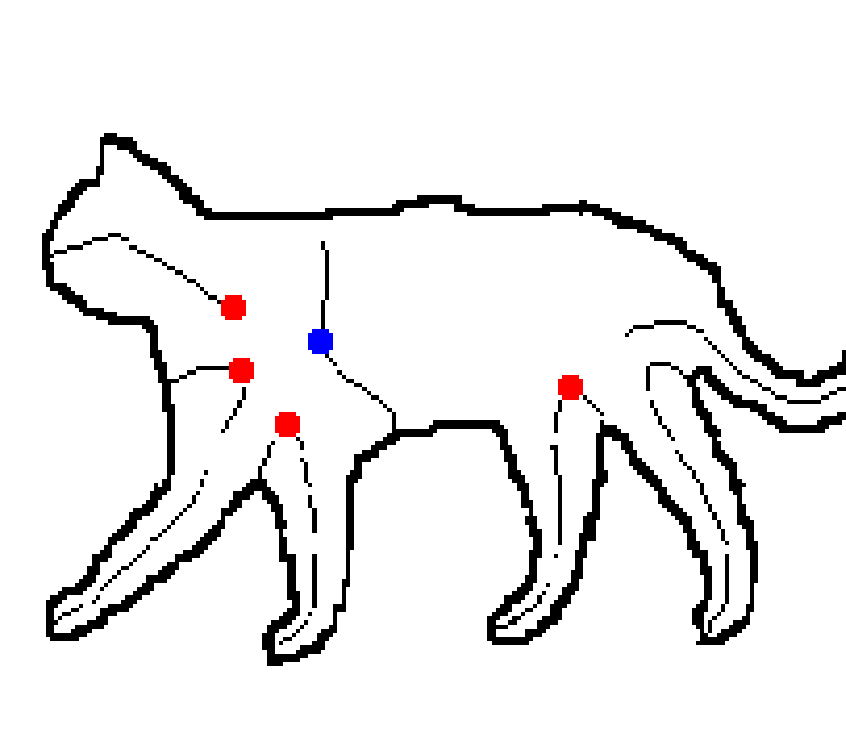} &
\includegraphics[height=1.5cm]{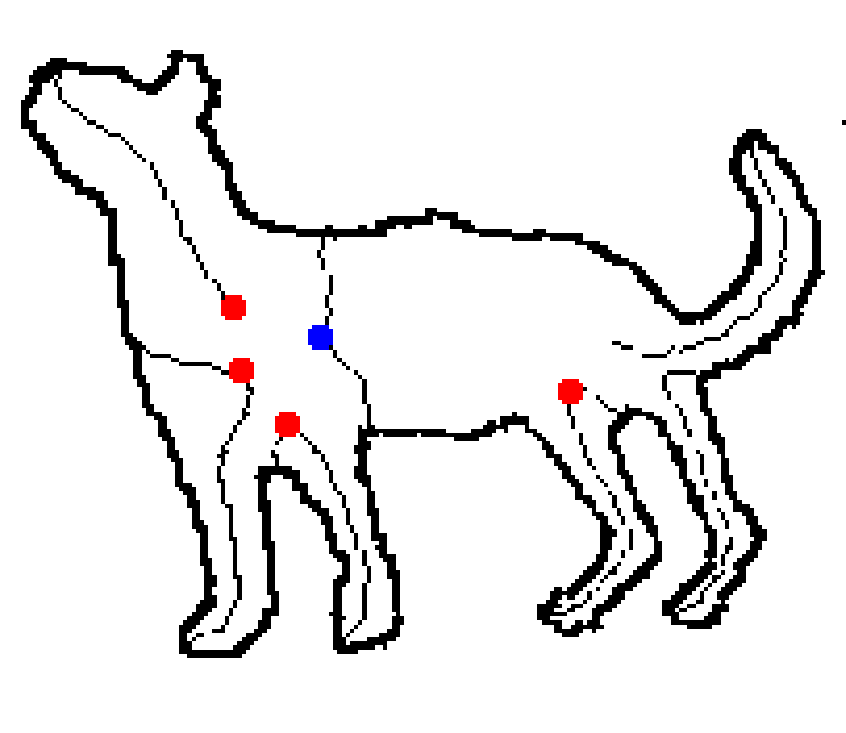} &
\includegraphics[height=1.5cm]{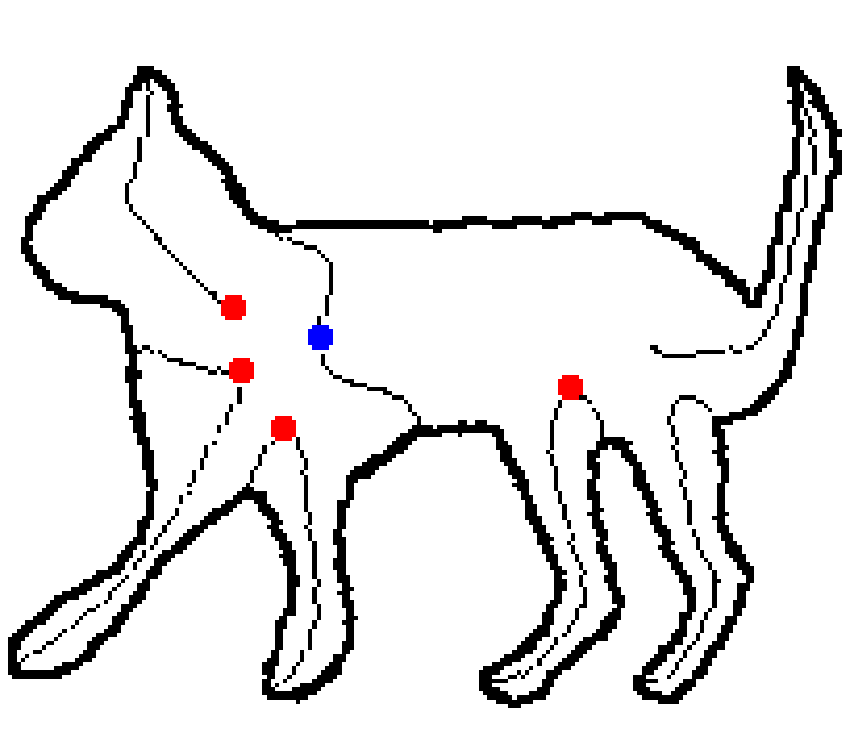} &
\includegraphics[height=1.5cm]{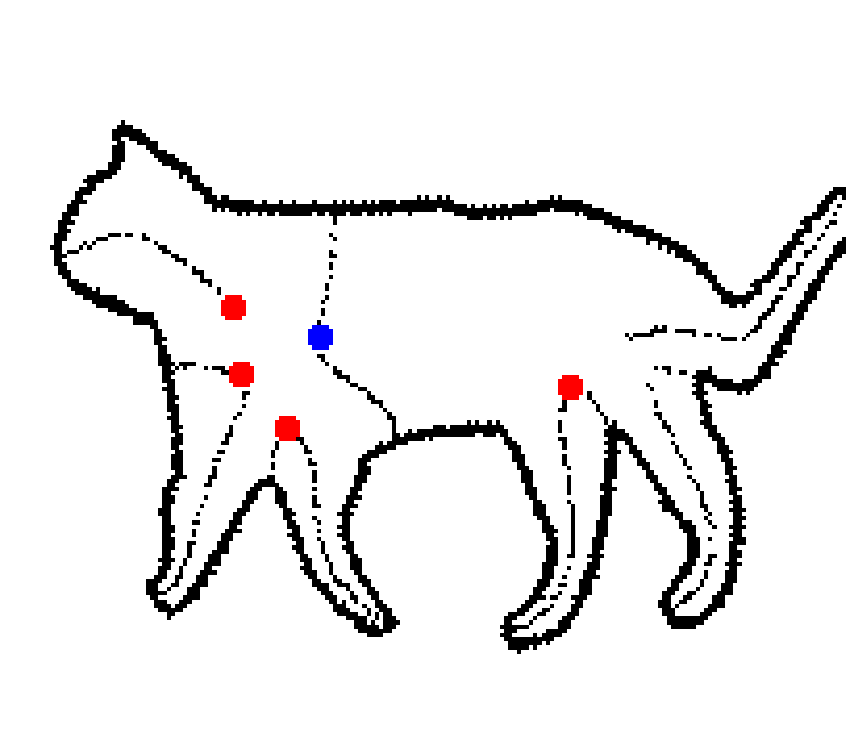} &
\includegraphics[height=1.5cm]{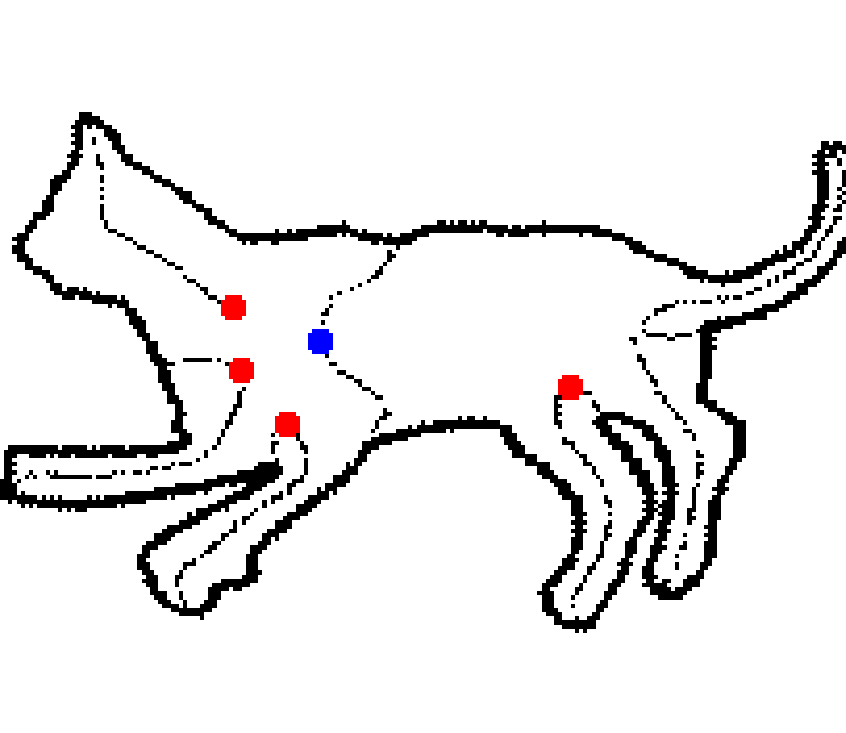} &
\includegraphics[height=1.5cm]{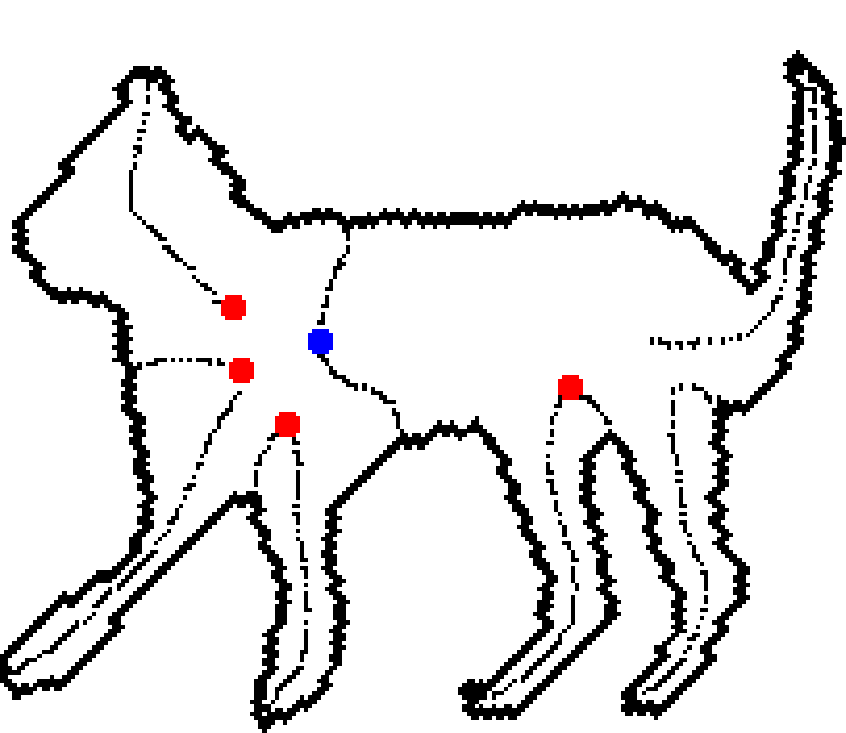}\\
\multicolumn{6}{c}{(c)} \\
\end{tabular}
\caption{Thin Plate Spline deformation applied to disconnection
points (see text).} \label{fig:cat_unwarp}
\end{figure}

 We illustrate this concept in
Fig.~\ref{fig:cat_unwarp}. The shapes in the cat category and the
disconnection points of the primitives are depicted in the top row.
For each of the six shapes we form a point set which includes all
the disconnection points and the shape center. We compute the mean
shape by computing the \textsl{Procrustes mean}~\cite{Small96} of
all the six point sets. For each cat shape we compute the
\textsl{Thin Plate Spline}~\cite{Bookstein91} transformation that
aligns the point set of the cat with the mean point set. The
boundary points for each cat are transformed using the computed
spline transformation. The deformed boundaries are displayed in the
second row. Notice that the fifth cat shape went through a major
deformation. This result is in agreement with the unintuitive
similarity scores reported in Query $9$ of
Table~\ref{table:results}. The fifth cat differs from the other cats
significantly in terms of the disconnection points arising from the
tail and the rightmost leg. That is within category variability of
these disconnection points is significantly higher than the within
category variability of the other disconnection points. The bottom
row in Fig.~\ref{fig:cat_unwarp} displays the cats aligned with the
mean shape when the Procrustes mean is computed by excluding the
disconnection points corresponding to the tail and the rightmost
leg. The point set of the fifth cat no longer deviates too much from
the  mean set. Observe the difference in the fifth column. We
believe that the unintuitive similarity scores in Query $9$ do not
mainly arise from the disconnected skeleton representation but they
are a consequence of the current shape matching algorithm which
treats every primitive equally  and does not take the deformation
into account. This weakness can be eliminated by using some gating
ideas over matched branches or by defining context dependent
saliency values for each primitive~\cite{BaseskiTez}. Furthermore,
it is worth exploring the algorithms that jointly determine point
correspondences and estimate local deformation,
\cite{Chui03,Glaunes04}.

\section{Computational Requirements}
\label{sec:cr} {The experiments are} carried out  on a Pentium4 with
$2$ GHz CPU. The time required to compute the descriptions from
binary images depends mainly on the diffusion step. The size of the
image significantly affects the duration of the diffusion step. The
axis detection step is also affected because as the size of the
image increases, the number of pixels representing the symmetry
points increases. However, scaling does not increase the matching
cost. As an example, when the size of the tulip-like  shape (Fig.
\ref{fig:db}, fifth row) is $100\times113$ pixels, the diffusion
takes one second and the skeleton computation takes another one
second. When the image size is changed to $198\times225$, the
diffusion takes six seconds and the complete computation takes eight
seconds. Notice that the new surface computation is even faster than
the TSP {method}~\cite{Tari96}.

%\begin{figure}[!b]
%\begin{center}
%\includegraphics[height=3cm]{PAMI08_png/performance}
%\caption{The effect of the size of the image on the time it takes to
%compute the descriptions. When the image size is $100\times113$:
%Diffusion length: 1 sec. Axis Detection Step: 1 sec. When the image
%size is $198\times225$: Diffusion Length: 6 sec Axis Detection Step:
%2 sec.} \label{fig:perf}
%\end{center}
%\end{figure}

Since the representation scheme produces coarse descriptions of
shapes, the number of branches and the number of descriptions are
small. Therefore, even the matching of the most complex shapes in
the database takes approximately one second. However, when a shape
is compared to all the shapes in the database, the number of
descriptions of the query shape affects the computation time. For
instance, while it takes fifteen seconds to classify a shape with
two descriptions, it takes twenty-five  seconds to classify a shape
with six descriptions. Because of the fixed complexity of retrieving
shapes from storage, the change in computation time is not as high
as expected.

\section{Summary and Discussion}
\label{sec:summary} We have presented an unconventional approach to
shape representation and recognition by using skeletons. Unlike
common skeletal representations, our branches are disconnected as a
result of excessive regularization. Hence, the representation is a
point set representation. A key difference in our framework from the
conventional approach is the construction of separate mechanisms to
handle visual transformations. The main focus of this paper is the
global frame that is constructed to handle scale, rotation and
translation. This frame, alone,  is sufficient for articulation
insensitive similarity measurement.

We have demonstrated the potential of the representation on shape
matching and similarity computation. On a diverse shape database of
$180$ shapes with $30$ categories, even for a 100\% recall rate, the
precision is over 88\%. The Bull's eye~\cite{Ling05,Latecki00} test
score is 98\%. Even though the shapes are represented at attainable
coarseness, the matching results are comparable to skeletal
representations with complete detail. We offer the following
explanation. Very large numbers of explanatory features are harmful
for  recognition and categorization tasks. Secondary details which
are weakly informative contain errors due to numerics and lower the
recognition performance. Hence, relying on dominant features that
can be extracted with high numerical accuracy alleviates the
problem.

The presented method is easy to implement and fast to compute. The
disconnected nature of our skeletal representation not only
introduces robustness, but also makes the necessary constructions
almost trivial.

%%%%%Sibel: yeni%%%%%%
An important weakness of the method is that it is limited to shapes
with closed boundaries. Even though we can compute symmetry axes for
the shapes with holes, construction of the complete representation
along with the coordinate frame constructions becomes tricky. A
major limitation comes from the fact that shapes with holes are not
easily shrunk to single centers. A simple surface minimum indicating
the center is replaced with a sequence of parabolic points whose
centroid is typically outside the shape. Currently, we are working
on extending the construction of disconnected skeleton to shapes
with holes.
%%%%%Sibel:%%%%%
%
\bibliographystyle{plain}
\bibliography{Aslan08}
\section*{Acknowledgments}
We thank J. Shah, D. Sezer, J. Scheideman, K. Hulten,
and anonymous reviewers for providing feedback. 
Preliminary version by C. Aslan and S. Tari has appeared in \cite{Aslan05}. 

\end{document}